\documentclass{article} 
\usepackage{collas2026_conference,times}

\usepackage{amsmath,amsfonts,bm}

\def\eqref#1{equation~\ref{#1}}

\def\1{\bm{1}}

\DeclareMathAlphabet{\mathsfit}{\encodingdefault}{\sfdefault}{m}{sl}
\SetMathAlphabet{\mathsfit}{bold}{\encodingdefault}{\sfdefault}{bx}{n}

\usepackage{hyperref}
\hypersetup{
    colorlinks=true,
    linkcolor=red,
    filecolor=magenta,
    urlcolor=blue,
    citecolor=purple,
    pdftitle={On the Interaction Between Model Compression and Test-Time Adaptation},
    pdfpagemode=FullScreen,
    }

\usepackage{microtype}
\usepackage{graphicx}
\usepackage{subcaption}
\usepackage{booktabs} 
\usepackage{arydshln} 
\usepackage{hyperref}
\usepackage{url}
\usepackage{multirow}
\usepackage{enumitem}
\usepackage{easyReview}

\usepackage{xspace}

\newcommand{\eg}{\emph{e.g.},\xspace}
\newcommand{\ie}{\emph{i.e.},\xspace}

\newcommand{\dnns}{DNNs\xspace}

\newcommand\cka{CKA\xspace}

\newcommand\tta{TTA\xspace}
\newcommand\ame{AME\xspace}

\title{On the Interaction Between Model Compression and Test-Time Adaptation}

\author{Francesco Corti \\
Graz University of Technology \\
Graz, Austria \\
\texttt{francesco.corti@tugraz.at} \\
\And 
Dong Wang \\
Graz University of Technology \\
Graz, Austria \\
\texttt{dong.wang@tugraz.at} \\
\And
Young D. Kwon \\
Samsung AI Center-Cambridge \\
University of Cambridge \\
Cambridge, United Kingdom \\
\texttt{yd.kwon@samsung.com} \\
\AND 
Cecilia Mascolo \\
University of Cambridge \\
Cambridge, United Kingdom \\
\texttt{cm542@cam.ac.uk} \\
\And
Olga Saukh \\
Graz University of Technology \\ 
Complexity Science Hub \\
Graz, Vienna, Austria \\
\texttt{saukh@tugraz.at}
}

\collasfinalcopy 

\begin{document}
\maketitle

\begin{abstract}

Deep neural networks deployed in the wild must be both efficient and adaptable, requiring model compression and test-time adaptation (TTA). While both are well studied in isolation, their interaction remains poorly understood. We systematically analyze how structured compression affects a model’s ability to adapt under distribution shift. Using ResNet-18 and ViT-Base on CIFAR-10-C and ImageNet-C, we evaluate multiple compression methods combined with standard TTA techniques. We introduce a diagnostic framework that examines representational expressivity and adaptation subspace compatibility. Our results reveal a consistent gap: although compressed models retain high accuracy under supervised adaptation, their TTA performance degrades significantly with increasing compression. We show that this stems from reduced representational diversity and structural constraints that limit recoverability. These effects strongly depend on the compression method, highlighting the need to design compression strategies that preserve adaptability.
\end{abstract}

\section{Introduction}
\label{sec:intro}

Deployed deep neural networks (\dnns) often operate under resource constraints and distribution shifts, necessitating both compression to meet memory and latency budgets and adaptation to cope with changing environments~\citep{han2015deep, sun2020test, wang2021tent}. While compression and model adaptation have been extensively studied in isolation, their interaction remains largely unexplored~\citep{liang2025comprehensive, choudhary2020comprehensive}. In particular, it is unclear whether compression strategies can preserve the capacity of a model to adapt to distribution shifts or task variations~\citep{liebenwein2021lost}.

This interaction is especially critical in edge settings, where adaptation incurs substantial memory overhead due to gradient computation and the need to store intermediate activations~\citep{dong2025bats, gomez2017reversible, xiao2026efficient}. For example, on resource-constrained devices such as the Raspberry Pi Zero, adaptation via entropy minimization leads to out-of-memory errors even at small batch sizes~\citep{jia2024tinytta}. Consequently, compression is not only desirable for efficient inference but also a prerequisite for enabling on-device adaptation~\citep{lin2022ondevice}.

Structured compression methods reduce model size by removing or merging entire computational units~\citep{li2017_pruning_filters, wang2025forget}. These methods can be divided into data-dependent methods, which leverage calibration data to estimate unit importance~\citep{sun2023simple, molchanov2019importance, lecun1989optimal}, and data-free methods, which rely on weight statistics~\citep{li2017_pruning_filters} or structural redundancies~\citep{wang2025forget}. 

Test-time adaptation (TTA) has emerged as a promising paradigm for adapting deep models using only unlabeled incoming data~\citep{liang2025comprehensive, liang2020we, wang2021tent}. Existing approaches can be broadly categorized into backpropagation-based (BP) and backpropagation-free (BP-free) methods. BP-based methods update model parameters via gradient-based optimization, typically fine-tuning a subset of parameters~\citep{wang2021tent, niu2023towards, gong2023sotta}. While achieving strong adaptation performance, they incur substantial memory overhead due to gradient computation and activation storage, limiting their applicability on resource-constrained devices~\citep{jia2024tinytta}. 

In contrast, BP-free methods avoid gradient computation by adapting parameters during the forward pass~\citep{mirza2022norm}, modifying lightweight input prompts~\citep{niu2024test} or correcting embedding distortions via layer-wise alignment~\citep{xiao2026architectureagnostic}, making them more suitable for edge deployment. Despite this progress, existing studies predominantly assume dense models and do not account for compression~\citep{mirza2022norm, wang2021tent, niu2023towards, wang2022continual, niu2022efficient, sun2020test}. This assumption is misaligned with resource-constrained deployment scenarios, where models are routinely compressed to satisfy hardware and efficiency constraints~\citep{fang2023depgraph, ashkboos2024slicegpt, ma2023llm}.

Recent studies have shown that compression degrades learned representation quality~\citep{corti2022studying}, and erodes robustness under distribution shift~\citep{liebenwein2021lost}. Yet existing work that explicitly considers the compression and adaptation interaction remains limited to quantization~\citep{xiao2024ttaq,deng2025test}. Whether structurally pruned or folded models preserve the representational and geometrical properties required for test-time adaptation remains an open question. In this work, we address the following question:
\begin{quote}
\textit{Does structured compression degrade a model's ability to adapt at test time, or can suitable compression strategies preserve adaptability under distribution shift and task variation?}
\end{quote}

\begin{figure*}[t]
\centering
\begin{subfigure}[t]{0.32\textwidth}
\centering
\includegraphics[width=\textwidth]{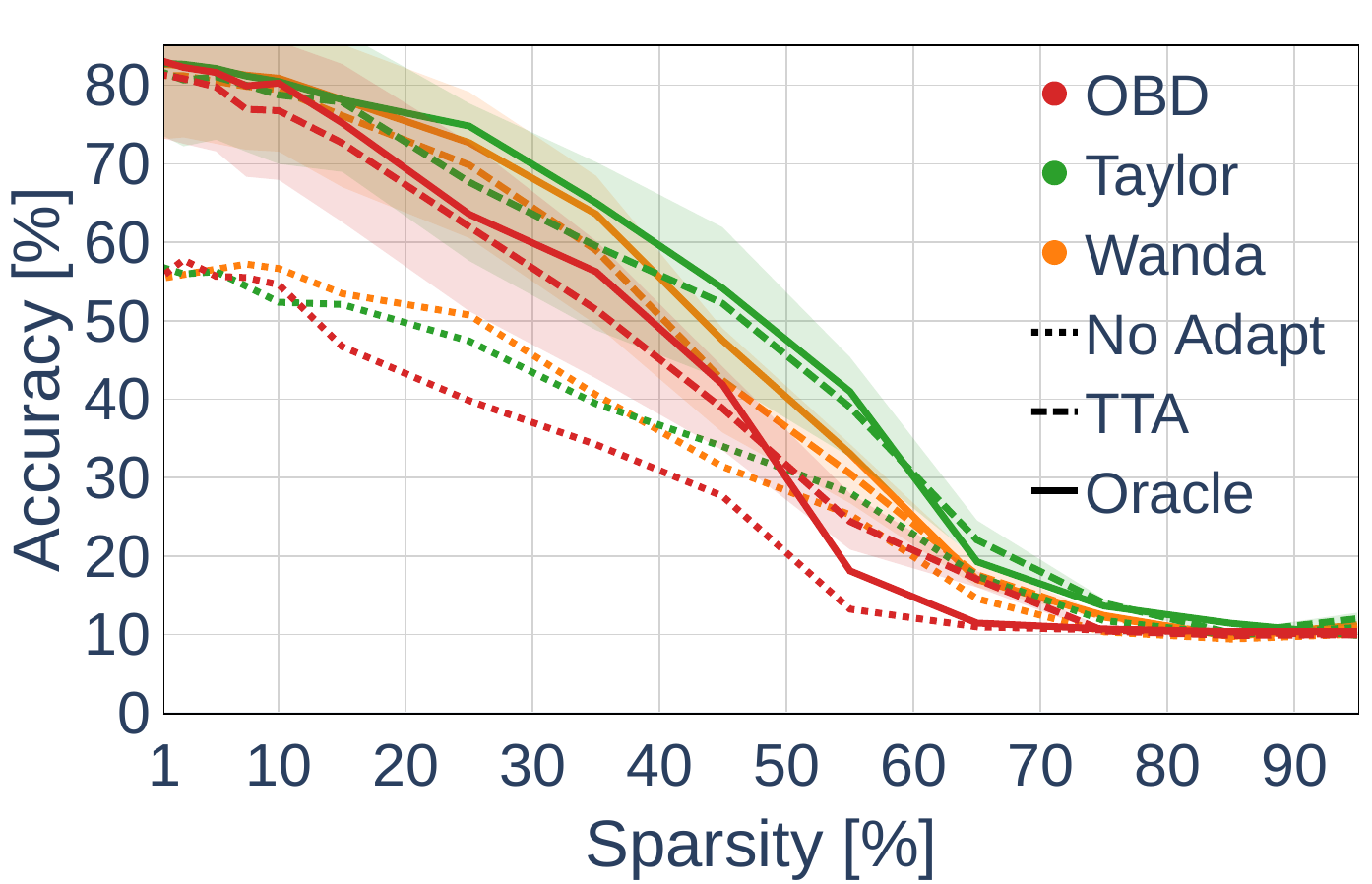}
\caption{ResNet-18 on CIFAR-10-C}
\label{fig:hero_data_cifar10}
\end{subfigure}
\hfill
\begin{subfigure}[t]{0.32\textwidth}
\centering
\includegraphics[width=\textwidth]{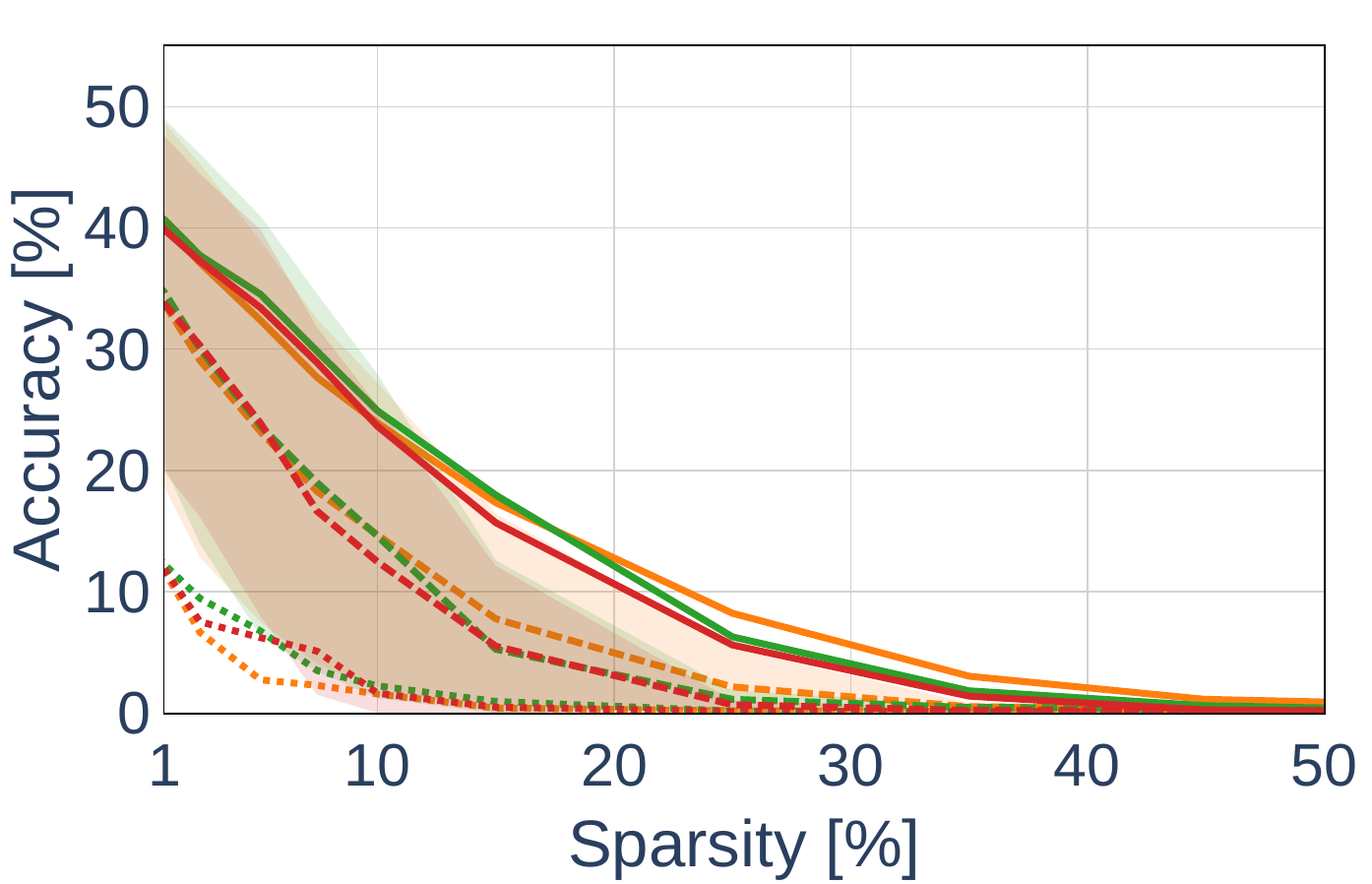}
\caption{ResNet-18 on ImageNet-C}
\label{fig:hero_data_imagenet_resnet}
\end{subfigure}
\hfill
\begin{subfigure}[t]{0.32\textwidth}
\centering
\includegraphics[width=\textwidth]{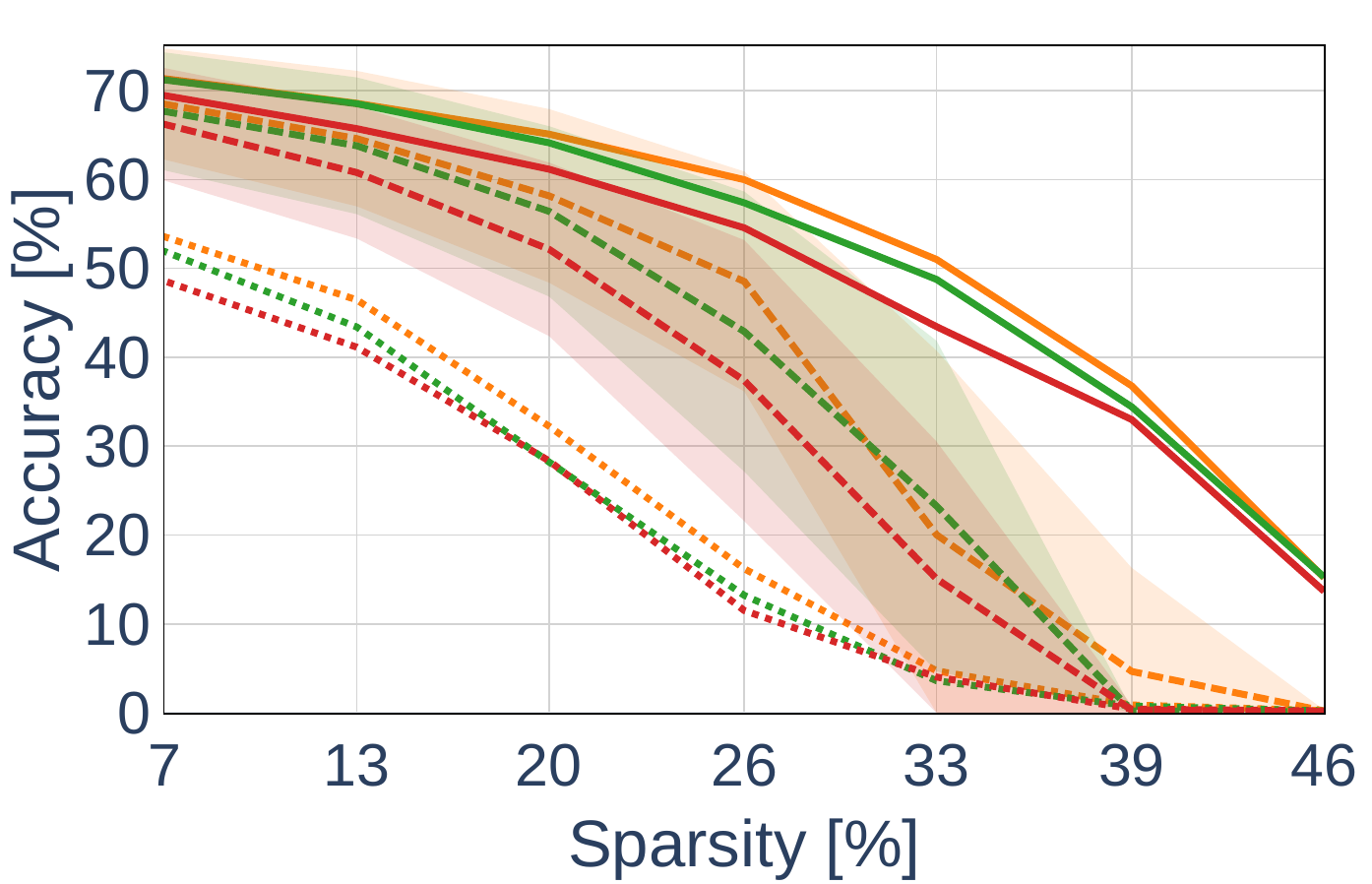}
\caption{ViT-Base on ImageNet-C}
\label{fig:hero_data_imagenet_vit}
\end{subfigure}

\vspace{0.3em}

\begin{subfigure}[t]{0.32\textwidth}
\centering
\includegraphics[width=\textwidth]{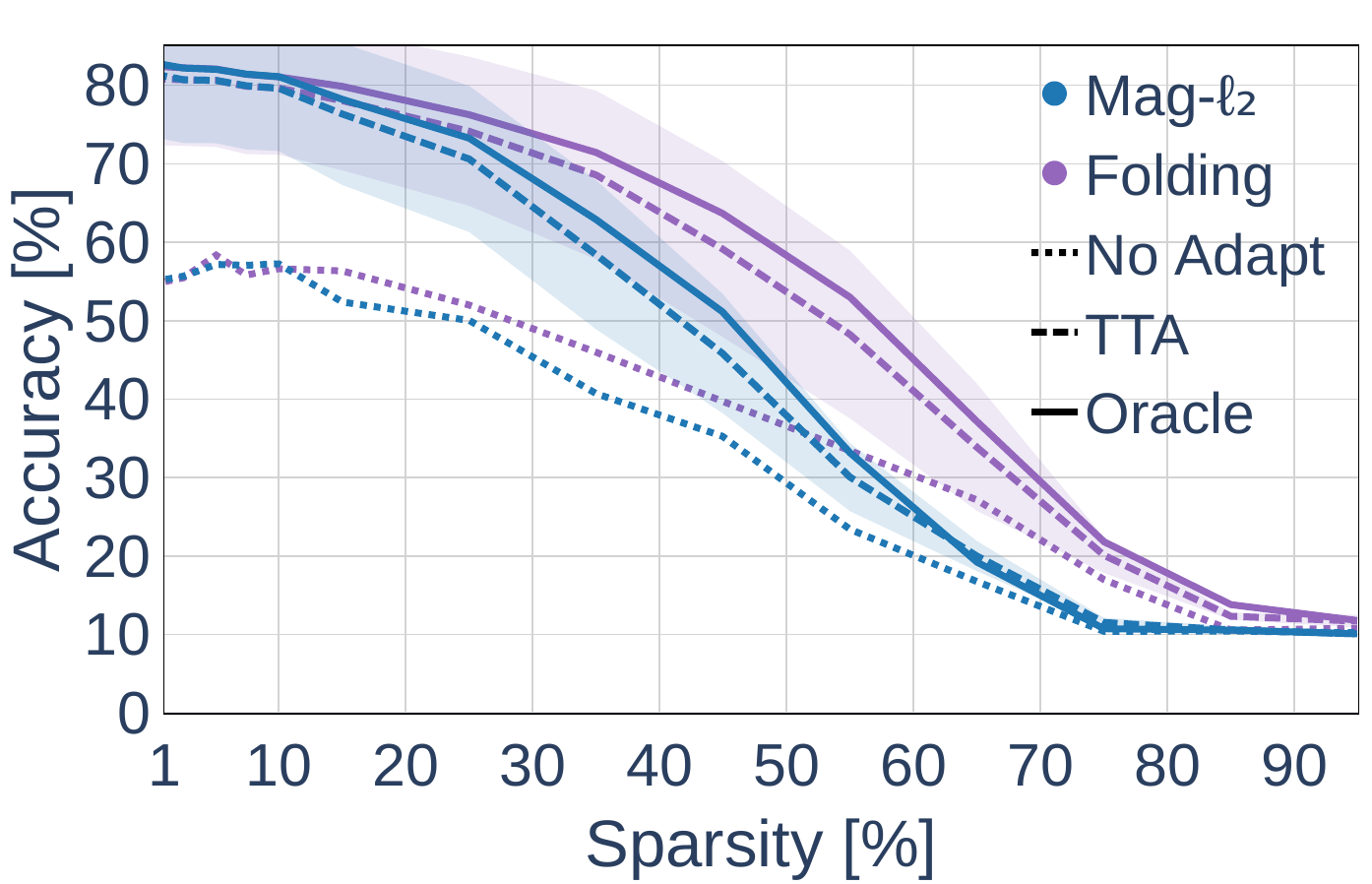}
\caption{ResNet-18 on CIFAR-10-C}
\label{fig:hero_free_cifar10}
\end{subfigure}
\hfill
\begin{subfigure}[t]{0.32\textwidth}
\centering
\includegraphics[width=\textwidth]{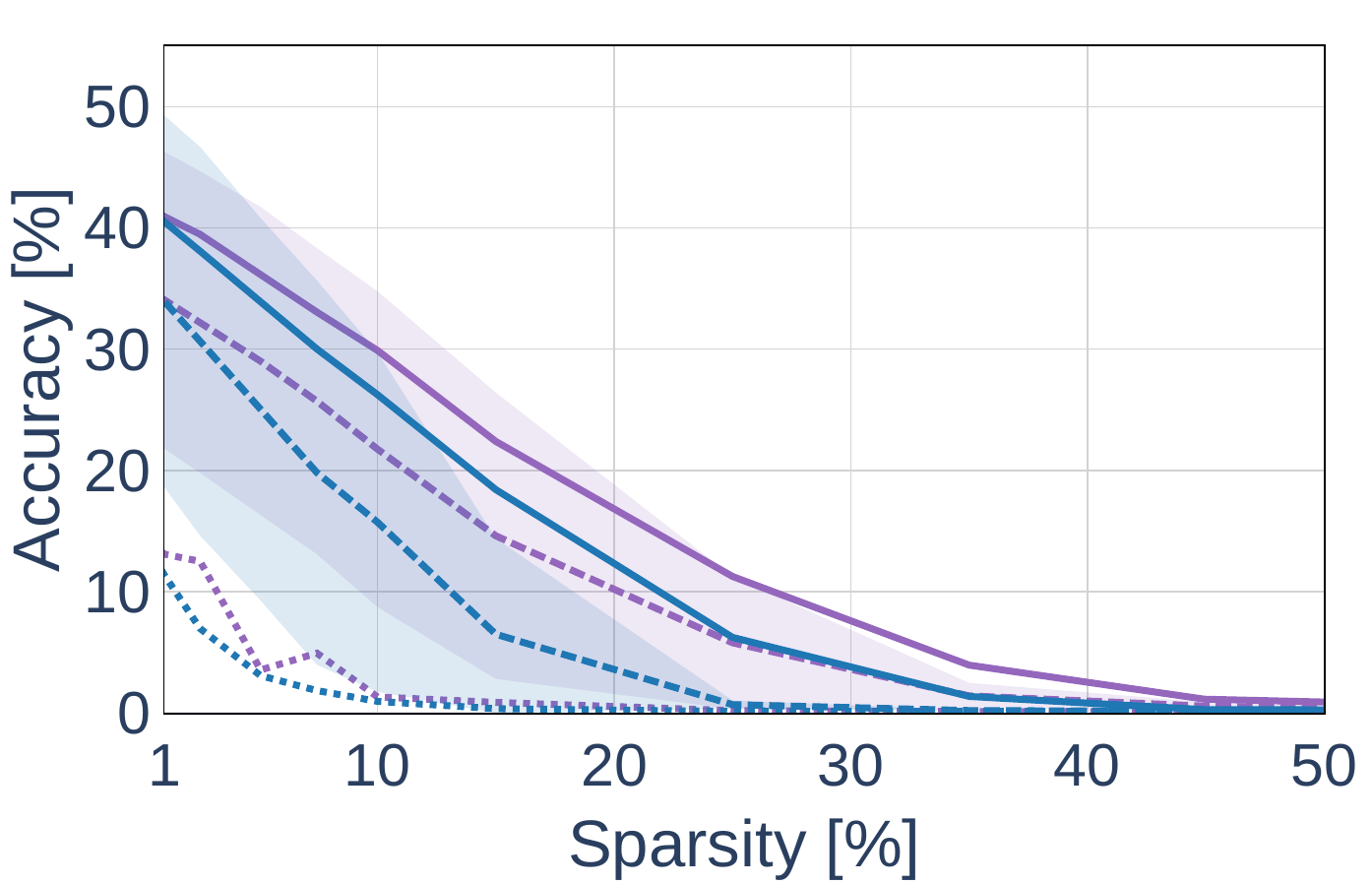}
\caption{ResNet-18 on ImageNet-C}
\label{fig:hero_free_imagenet_resnet}
\end{subfigure}
\hfill
\begin{subfigure}[t]{0.32\textwidth}
\centering
\includegraphics[width=\textwidth]{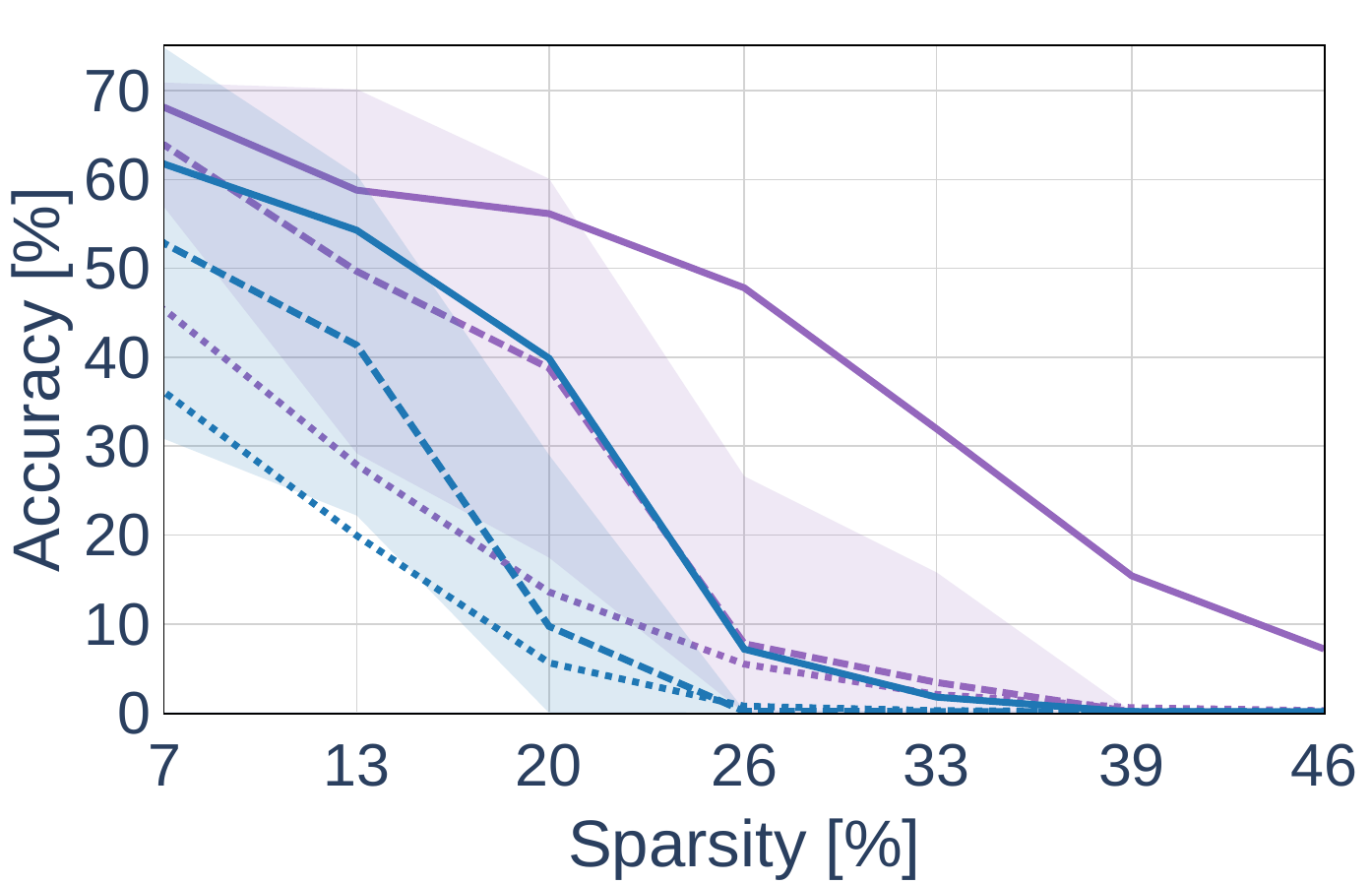}
\caption{ViT-Base on ImageNet-C}
\label{fig:hero_free_imagenet_vit}
\end{subfigure}
\caption{\textbf{The gap between test-time adaptation (\tta) and supervised fine-tuning (Oracle) widens with increasing compression, particularly on ImageNet-C.} We compare post-compression \tta against supervised fine-tuning across architectures and datasets, averaged across all corruptions. At low sparsity levels, \tta remains competitive, suggesting that adaptation signals are largely preserved. However, as compression increases, the gap between \tta and supervised fine-tuning grows, suggesting that compressed architectures increasingly hinder effective adaptation with \tta. This effect is especially pronounced on ImageNet-C. Shaded regions indicate $\pm\sigma$ across corruption types. \textbf{Top row} (a--c): data-dependent pruning (Wanda, Taylor, OBD). \textbf{Bottom row} (d--f): data-free compression (Folding, Mag-$\ell_2$). \tta is performed using SAR~\citep{niu2023towards} for ResNet-18 and SPA~\citep{niu2025self} for ViT-Base.}
\label{fig:hero}
\end{figure*}

This question is closely related to continual learning: models deployed at the edge must retain \emph{plasticity}, \ie the ability to adapt to incoming data over time~\citep{wang2022continual}. However, compression applied to meet deployment constraints may inadvertently erode this capacity. If compression reduces feature diversity, misaligns the unsupervised adaptation signal from the supervised gradient direction, or restricts the useful update subspace, the standard compress-and-deploy pipeline can yield models that maintain high accuracy on source data but lose the ability to adapt. We refer to this phenomenon as \emph{silent plasticity loss}. 

Figure~\ref{fig:hero} provides empirical evidence of this effect: while compressed models retain strong performance under supervised adaptation, their test-time adaptation performance degrades substantially as compression increases, leading to a widening gap between the two. This gap indicates that, despite preserved predictive performance, compression can impair the model’s ability to effectively utilize unlabeled test-time signals. A per-corruption decomposition of this gap (Fig.~\ref{fig:gap_scatter}) shows that the widening trend generalizes across most corruption types rather than being driven by a subset. To isolate the compression-induced component of this widening from the supervised-unsupervised signal-quality difference already present for the dense model, Fig.~\ref{fig:compression_induced_gap} reports the Oracle-\tta gap after subtracting its zero-compression value.  

We introduce a framework to diagnose how compression interacts with \tta along two complementary dimensions (Sec.~\ref{sec:framework}). First, \textit{diversity/expressivity}: compression may reduce the variability and richness of internal representations, limiting the range of functions the model can realize during \tta. We quantify this using representation similarity between dense and compressed models~\citep{kornblith2019similarity}, as well as activation entropy across feature maps and tokens~\citep{skean2025layer}. Second, \textit{subspace compatibility}: \tta updates are restricted to a small parameter subspace (\eg normalization layers), and compression may misalign this subspace with useful update directions. We quantify this via gradient cosine similarity between unsupervised and supervised updates, and provide a theoretical analysis of objective-induced gradient degeneracy in both entropy-based and consistency-based \tta objectives.

Our empirical analysis across convolutional and transformer architectures reveals the following insights:
\begin{itemize}
    \item The gap between \tta and supervised adaptation (Oracle) widens with increasing compression, suggesting that \tta depends on structural properties of dense models that are not preserved under compression.
    \item \tta degrades representations already at low compression levels and fails to recover them at higher sparsity, whereas supervised adaptation consistently restores them, highlighting \tta’s sensitivity to compression-induced representational distortions.
    \item Compression induces two distinct failure modes of gradient alignment between \tta and supervised updates: \emph{gradient degeneracy}, where near-uniform predictions attenuate the \tta signal, and \emph{active divergence}, where the \tta gradient opposes the supervised gradient direction even at minimal sparsity. We show theoretically that both entropy- and consistency-based \tta objectives exhibit gradient collapse under compression, while the supervised cross-entropy gradient can remain more informative at the same sparsity. 
    \item The severity of these effects depends on the compression method: Wanda and Taylor better preserve recoverable structure, while OBD and Mag-$\ell_2$ lead to pronounced degradation; folding achieves more stable recovery.
    \item These effects are substantially more pronounced on ImageNet-C than on CIFAR-10-C, highlighting the increased difficulty of adaptation under complex tasks.
\end{itemize}

\section{Related Work}
\label{sec:related_work}

\textbf{Structured Model Compression.} 
Structured pruning removes entire computational units (\eg convolutional filters~\citep{li2017_pruning_filters} or attention heads~\citep{michel2019sixteen}), enabling efficient inference on commodity hardware~\citep{li2017_pruning_filters}. Selection criteria range from magnitude-based heuristics~\citep{li2017_pruning_filters} to activation-aware scores~\citep{sun2023simple}, first-order Taylor approximations~\citep{molchanov2019importance}, and second-order Hessian-based methods~\citep{lecun1989optimal}. Beyond pruning, model folding clusters similar channels and replaces them with shared representations~\citep{wang2025forget, saukh2026cut, son2018clustering}. 
Existing benchmarks primarily evaluate compressed models by their accuracy on the source distribution~\citep{liebenwein2021lost, hooker2020characterising}. This protocol does not capture whether compressed models retain the capacity to adapt under distribution shift. As reported in Fig.~\ref{fig:hero}, these properties can diverge, indicating that standard compression benchmarks provide an incomplete view of model quality.

\textbf{Test-Time Adaptation.}
Test-time adaptation (\tta) adapts deployed models to distribution shifts using only unlabeled data. TENT~\citep{wang2021tent} updates batch-normalization parameters via entropy minimization; SAR~\citep{niu2023towards} extends this with sharpness-aware optimization and reliable sample filtering; EATA~\citep{niu2022efficient} incorporates an anti-forgetting regularizer to preserve important weights. For vision transformers, SPA~\citep{niu2025self} enforces consistency between clean and augmented inputs. In contrast, BP-free methods eliminate gradient computation at test time: FOA~\citep{niu2024test} uses a derivative-free covariance matrix adaptation of new prompt embeddings; PEA~\citep{xiao2026architectureagnostic} corrects the embedding distortions at each intermediate layer through a covariance alignment procedure. 

\textbf{Loss Landscape Geometry and Mode Connectivity.}
Solutions found by SGD in overparameterized networks are often connected by low-loss curves in weight space~\citep{garipov2018loss, draxler2018essentially}, and become linearly connected when accounting for permutation symmetries~\citep{entezari2022role, ainsworth2022git}. \citet{frankle2020linear} formalize \emph{linear mode connectivity} (LMC) and show that pruned subnetworks recover full accuracy only when they are stable to SGD noise. Recent work by~\citet{saukh2026cut} extends this perspective to model folding, showing how clustering-based compression reshapes the loss landscape geometry. While these works characterize how compression alters the loss landscape, they do not address whether adaptation can compensate for these changes.

\textbf{Compression-Adaptation Interaction.}
Plasticity loss in deep networks has been linked to loss landscape properties even without compression~\citep{lyle2023understanding, dohare2024loss}. Compression can further amplify representational biases, \eg toward low-frequency subgroups~\citep{hooker2020characterising}. \citet{liebenwein2021lost} suggest that overparameterization may be necessary to preserve robustness under distribution shift. 

Despite these insights, the mechanistic question remains open: \emph{which structural properties must a compression method preserve to enable effective adaptation?}

\section{Diagnostic Framework}
\label{sec:framework}

Given a classifier $f(x;\theta)$ with pretrained parameters $\theta_0$, trained on a source distribution $\mathcal{P}$, we consider deployment under a shifted target distribution $\mathcal{Q} \neq \mathcal{P}$. To improve performance on $\mathcal{Q}$, test-time adaptation (\tta) updates the model parameters as:
\begin{equation}
   \theta^{*} = \theta_0 + \Delta, \quad \Delta \in \mathcal{U},
\label{eq:tta}
\end{equation}
where $\mathcal{U}$ denotes the set of admissible update directions (\eg affine normalization parameters). In contrast, we refer to \emph{Oracle} as supervised adaptation over the same subspace $\mathcal{U}$, using labeled target data.

In practice, models are often compressed prior to deployment. We formalize structured compression as a projection $\Pi: \mathbb{R}^D \rightarrow \mathbb{R}^d$ with $d \ll D$, mapping the original parameters to a compressed representation $\phi_0 = \Pi(\theta_0)$. This formulation captures two common compression strategies:
\begin{itemize}
    \item \textbf{Structured Pruning:} $\Pi$ acts as a coordinate selection operator, applying binary masks to channels or attention heads and retaining only a subset of parameters indexed by $S \subset \{1, \dots, D\}$.
    
    \item \textbf{Folding:} $\Pi$ performs cluster-based aggregation~\citep{wang2025forget, son2018clustering}, grouping similar channels and replacing each cluster with its centroid. This reduces dimensionality via linear combinations rather than explicit removal.
\end{itemize}

To enable a controlled comparison, we use equal per-layer compression ratios, resulting in matched parameter budgets across methods. The central question is how structured compression affects the model’s ability to adapt to $\mathcal{Q}$. While compression reduces model capacity, its interaction with adaptation is governed by more subtle mechanisms, which we analyze next.

\textbf{Models and Datasets.}
We evaluate the diagnostic framework on both convolutional and self-attention architectures. Specifically, we use ResNet-18~\citep{he2016deep} checkpoints trained on CIFAR-10 (11.1M parameters) and adapted to CIFAR-10-C~\citep{hendrycks2019benchmarking}, as well as ResNet-18 (11.7M parameters) and ViT-Base~\citep{dosovitskiy2020image} (86M parameters) checkpoints trained on ImageNet and adapted to ImageNet-C~\citep{hendrycks2019benchmarking}. All experiments are conducted at severity level 5 across the 15 corruption types defined in CIFAR-10-C and ImageNet-C. Multi-seed and severity-3 experiments are reported in Appendix~\ref{sec:appendix}.

\textbf{Compression Methods.}
We analyze five structured compression strategies: magnitude-based $\ell_2$ pruning~\citep{li2017_pruning_filters}, Wanda~\citep{sun2023simple} (combining weight magnitudes with activation norms), Taylor-based pruning~\citep{molchanov2019importance}, Optimal Brain Damage (OBD)~\citep{lecun1989optimal} using Hessian-based importance scores, and model folding via $k$-means channel clustering~\citep{wang2025forget}.

The \emph{compression ratio} $r$ denotes the uniform fraction of channels removed per layer. For ResNet-18, $r$ corresponds to the fraction of convolutional filters (output channels) removed. For ViT-Base, $r$ specifies the fraction of hidden units removed from the feed-forward network layers in each transformer block. We evaluate $r \in \{0.0, 0.01, \ldots, 0.95\}$ for ResNet-18 (sparsity $s \in [0\%, 94.9\%]$) and $r \in \{0.0, 0.1, \ldots, 0.7\}$ for ViT-Base (sparsity $s \in [0\%, 45.8\%]$). An analysis of the memory use, latency, FLOPs and adaptation time of the compressed models is reported in Fig.~\ref{fig:compute_overheads_appendix}.

For compressed ResNet-18 models, we apply REPAIR~\citep{jordan2022repair} to re-estimate batch normalization statistics using four batches of training data, following~\citet{wang2025forget}. In contrast, ViT-Base uses layer normalization~\citep{ba2016layer}, which computes statistics per sample and does not require post-compression re-estimation.

\textbf{Adaptation Methods.}
We focus on backpropagation-based \tta methods, as our diagnostic framework relies on gradient-based analysis (\eg gradient cosine similarity and objective-induced degeneracy), which requires backpropagation gradients to exist. For ResNet-18, we use Sharpness-Aware and Reliable (SAR) entropy minimization~\citep{niu2023towards} with the original hyperparameters. For ViT-Base, we employ Self-Bootstrapping Adaptation (SPA)~\citep{niu2025self}, which optimizes a consistency objective between clean and augmented inputs. Together, these two methods cover \tta entropy-based and consistency-based objectives; whether backpropagation-free methods~\citep{niu2024test, xiao2026architectureagnostic} exhibit different sensitivity to compression is an important direction for future work.

To isolate the effect of structured compression from the adaptation objective, we introduce supervised Oracle variants that share the \emph{same} optimization settings (optimizer, trainable parameters, learning rate, and number of adaptation steps) as their analyzed \tta counterparts. For ResNet-18, Oracle-SAR replaces the entropy-based objectives in both inner and outer loops with supervised cross-entropy. For ViT-Base, Oracle-SPA replaces all self-generated signals with supervised cross-entropy losses across all views. We emphasize that the Oracle is not intended as a fair comparison with \tta, but as a diagnostic upper bound on the recoverability of the adaptation subspace $\mathcal{U}$.

\begin{figure*}[t]
\centering
\begin{subfigure}[t]{0.33\textwidth}
\centering
\includegraphics[width=\textwidth]{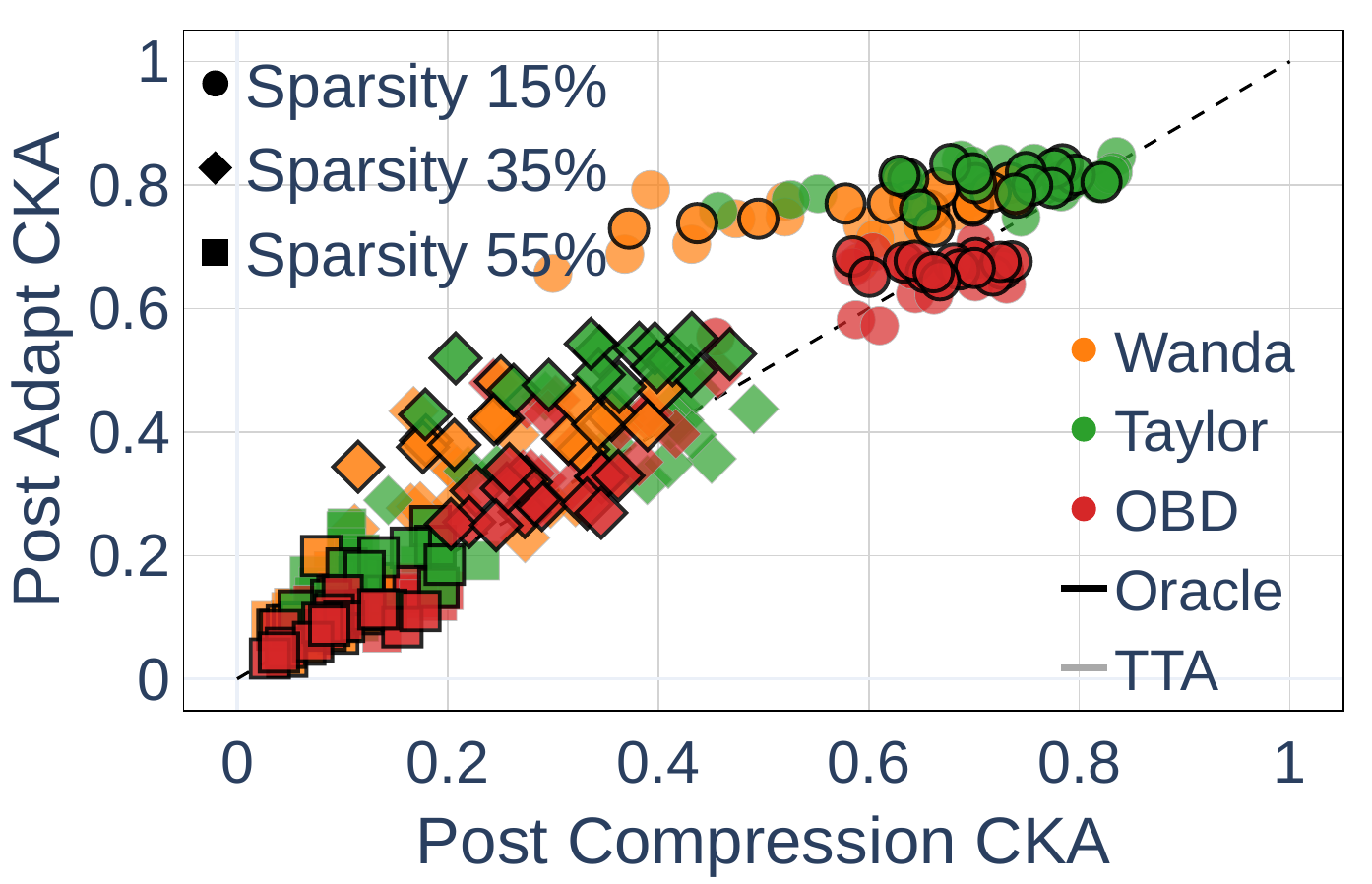}
\caption{ResNet-18 on CIFAR-10-C}
\label{fig:cka_data_cifar10_methods_colors}
\end{subfigure}
\hfill
\begin{subfigure}[t]{0.33\textwidth}
\centering
\includegraphics[width=\textwidth]{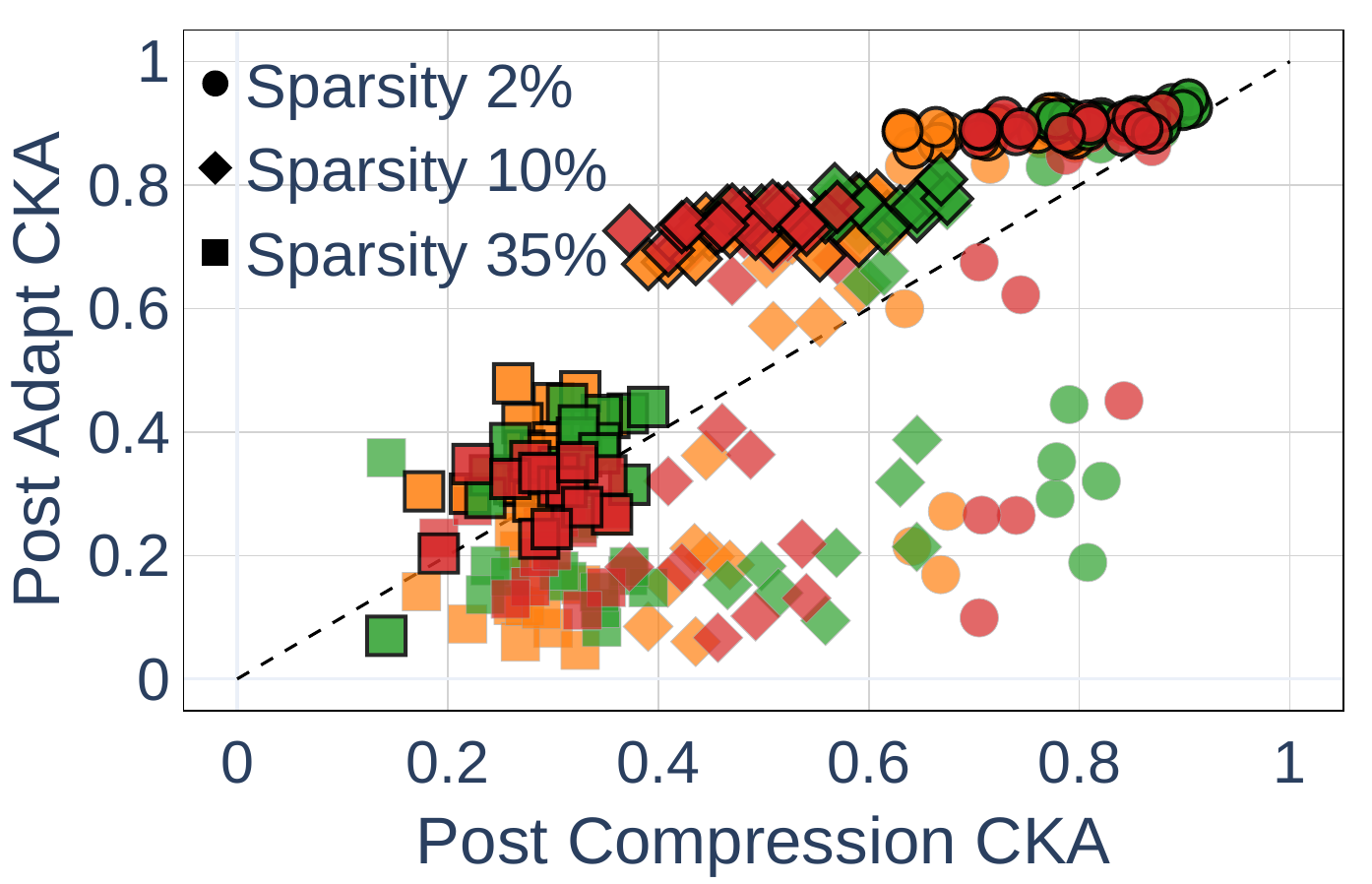}
\caption{ResNet-18 on ImageNet-C}
\label{fig:cka_data_imagenet_methods_colors}
\end{subfigure}
\hfill
\begin{subfigure}[t]{0.33\textwidth}
\centering
\includegraphics[width=\textwidth]{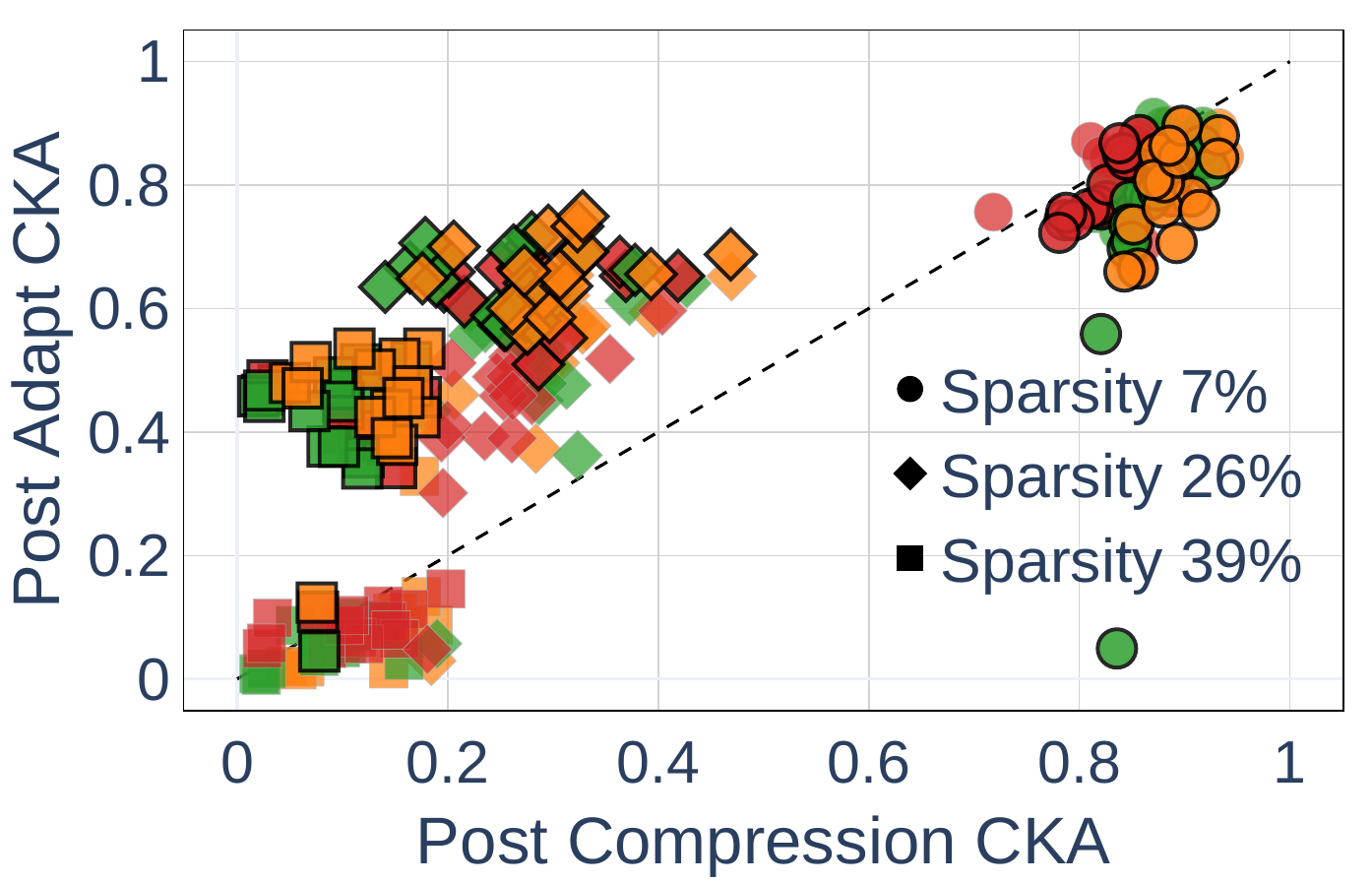}
\caption{ViT-Base on ImageNet-C}
\label{fig:cka_data_vit_methods_colors}
\end{subfigure}

\vspace{0.3em}

\begin{subfigure}[t]{0.33\textwidth}
\centering
\includegraphics[width=\textwidth]{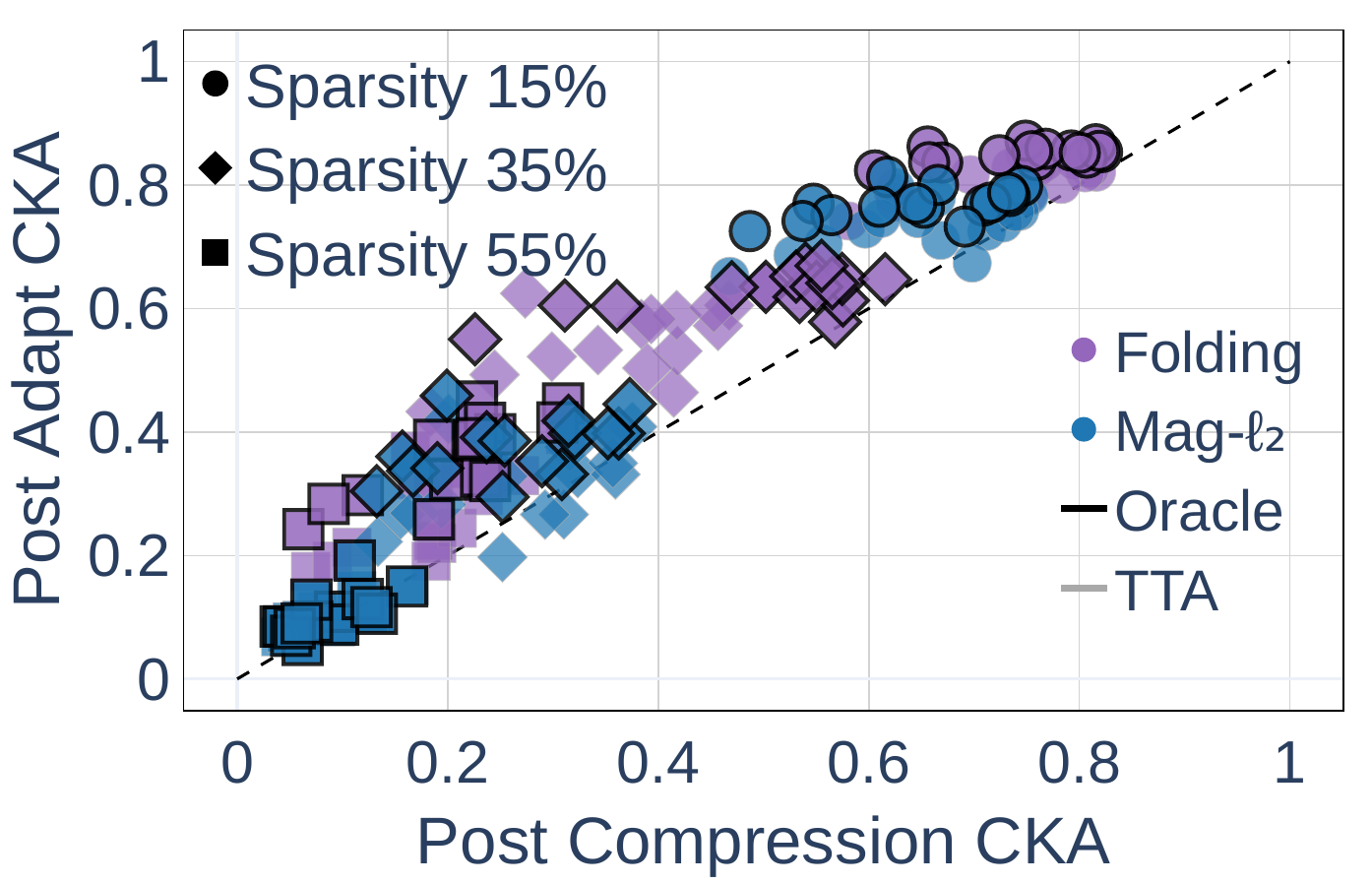}
\caption{ResNet-18 on CIFAR-10-C}
\label{fig:cka_free_cifar10_methods_colors}
\end{subfigure}
\hfill
\begin{subfigure}[t]{0.33\textwidth}
\centering
\includegraphics[width=\textwidth]{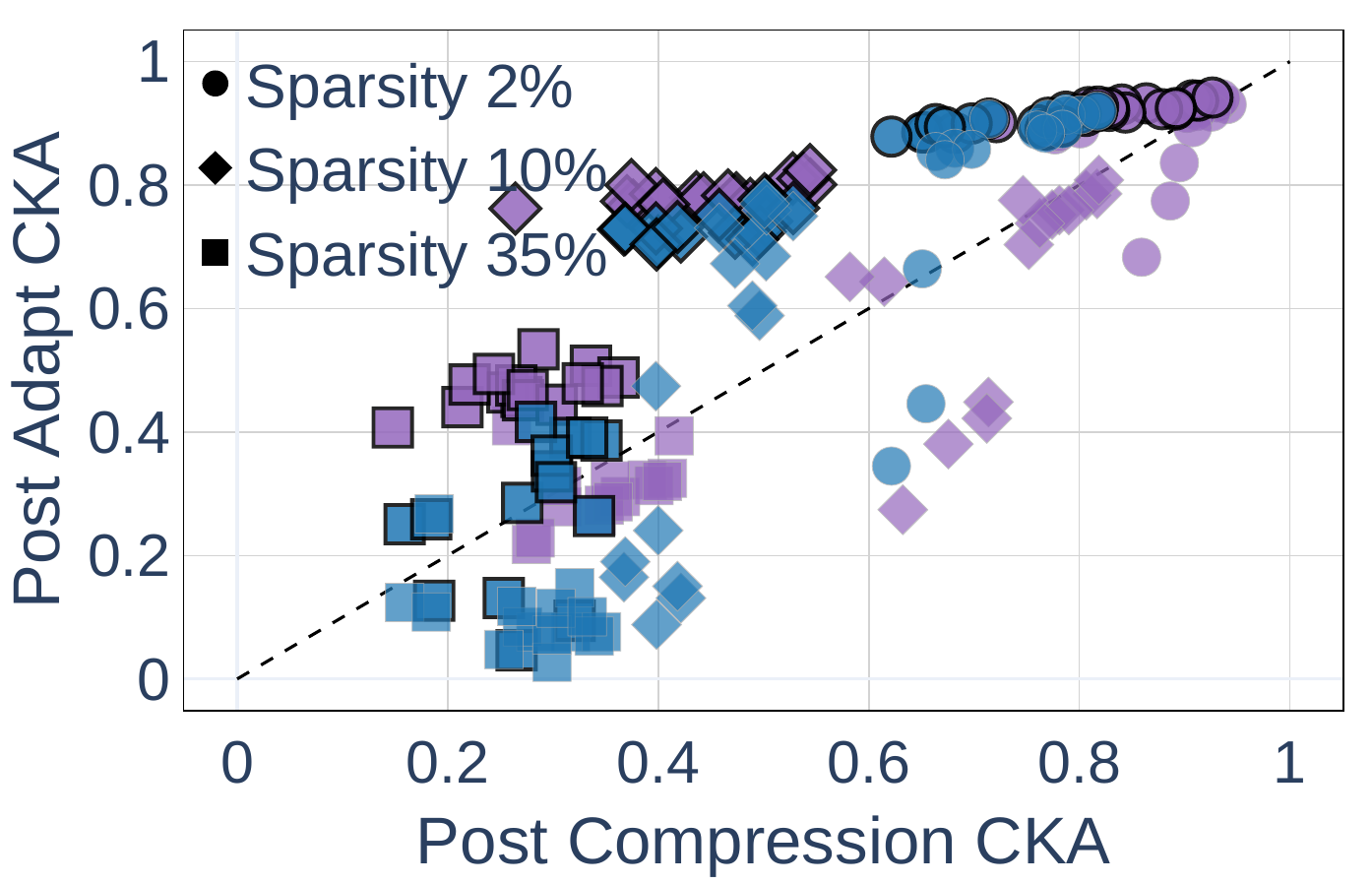}
\caption{ResNet-18 on ImageNet-C}
\label{fig:cka_free_imagenet_methods_colors}
\end{subfigure}
\hfill
\begin{subfigure}[t]{0.33\textwidth}
\centering
\includegraphics[width=\textwidth]{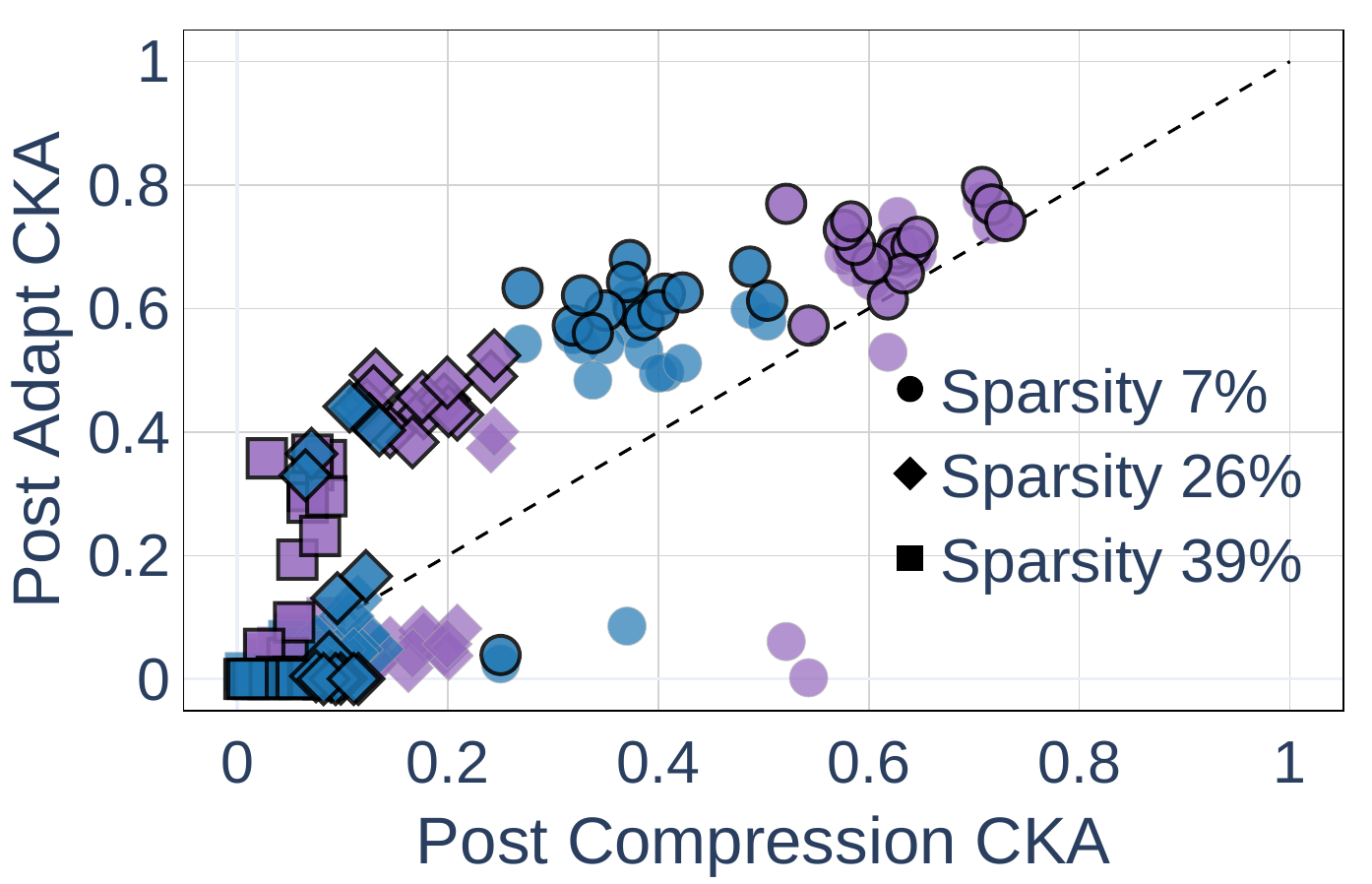}
\caption{ViT-Base on ImageNet-C}
\label{fig:cka_free_vit_methods_colors}
\end{subfigure}
\caption{\textbf{Test-time adaptation (\tta) degrades representations more than supervised adaptation (Oracle), with the gap depending on the compression method.}
We measure worst-layer CKA between dense and compressed models before (x-axis) and after adaptation (y-axis). Points above the identity line ($y{=}x$) indicate that adaptation brings representations closer to the dense model. Across all settings, both \tta and Oracle generally improve representations relative to the compressed model (most points lie above the identity line). However, \tta exhibits substantially weaker recovery than Oracle, with many points lying closer to or below the identity line, especially as compression increases. Notably, this degradation is already visible at low sparsity levels, indicating early loss of adaptable representations. The extent of this effect depends on the compression method: Wanda and Taylor preserve representations that remain partially recoverable, whereas OBD and Mag-$\ell_2$ exhibit markedly weaker recovery under \tta. Folding retains higher recoverability at larger sparsity. This gap is particularly strong on ImageNet-C, where \tta fails to recover representations even at moderate sparsity. \emph{Top row} (a--c): data-dependent pruning (Wanda, Taylor, OBD). \emph{Bottom row} (d--f): data-free compression (Folding, Mag-$\ell_2$).}
\label{fig:cka_main}
\end{figure*}

\begin{figure*}[t]

\centering
\begin{subfigure}[t]{0.33\textwidth}
\centering
\includegraphics[width=\textwidth]{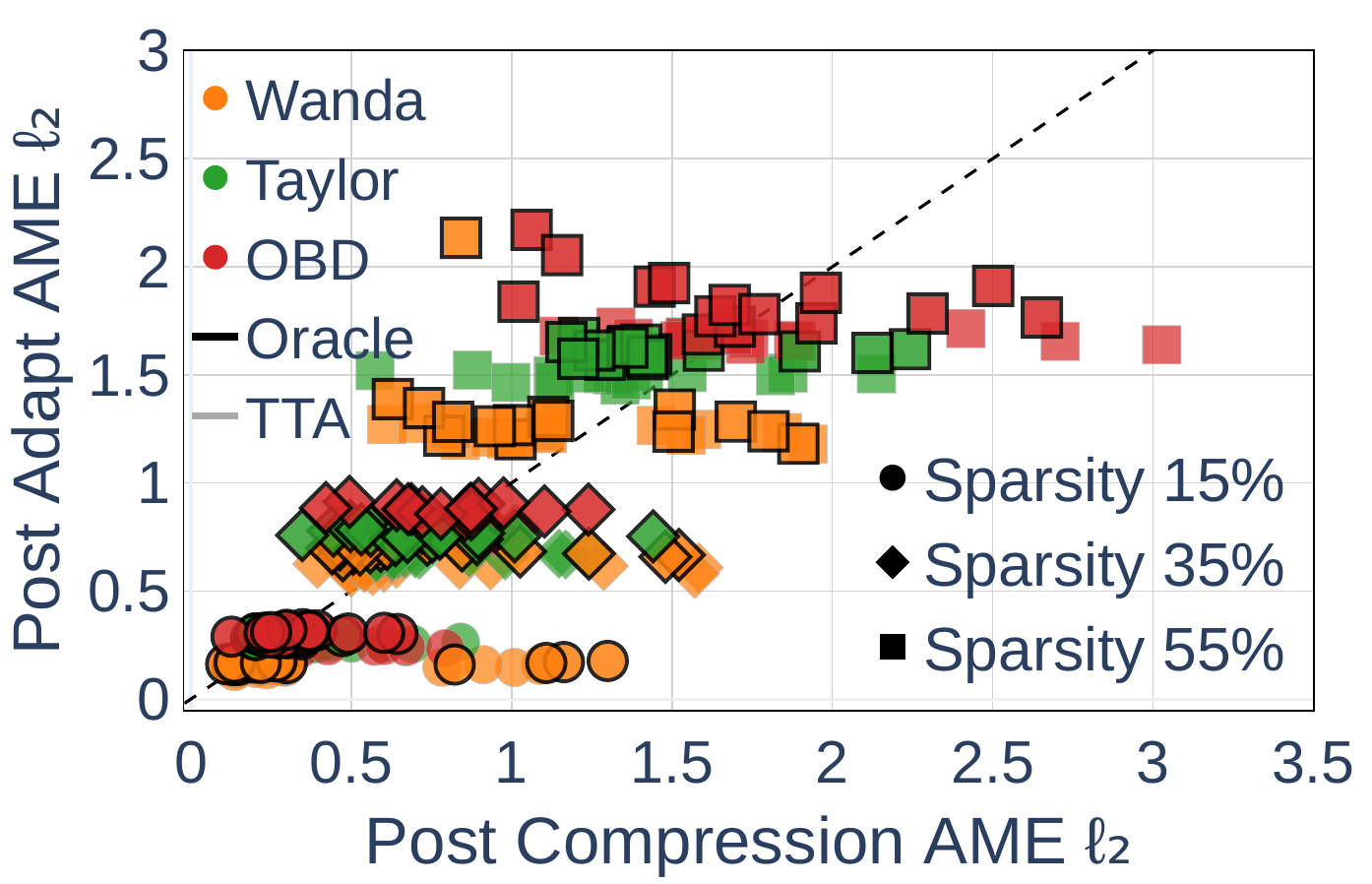}
\caption{ResNet-18 on CIFAR-10-C}
\label{fig:ame_data_cifar10_color_methods}
\end{subfigure}
\hfill
\begin{subfigure}[t]{0.33\textwidth}
\centering
\includegraphics[width=\textwidth]{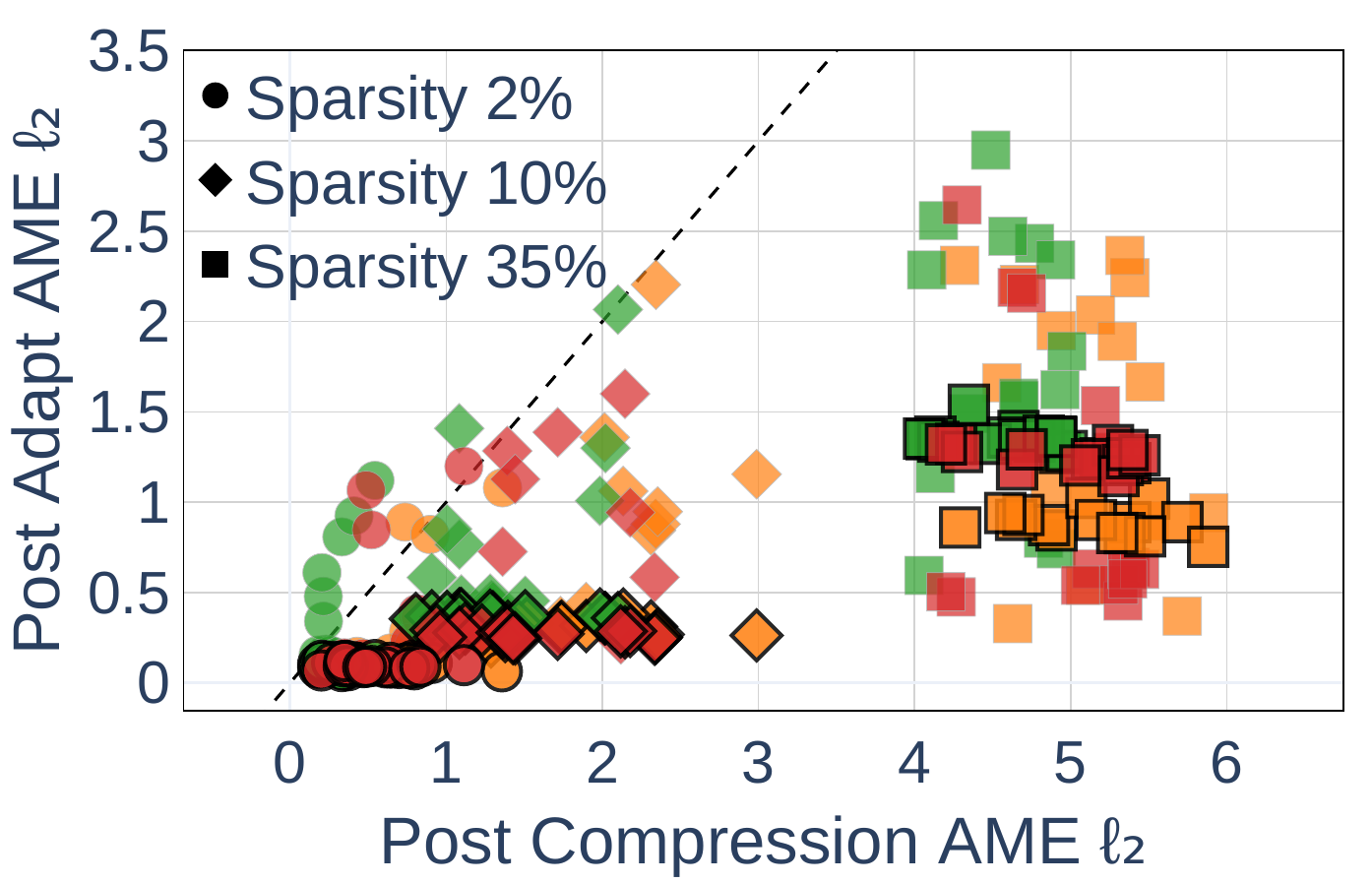}
\caption{ResNet-18 on ImageNet-C}
\label{fig:ame_data_imagenet_color_methods}
\end{subfigure}
\hfill
\begin{subfigure}[t]{0.33\textwidth}
\centering
\includegraphics[width=\textwidth]{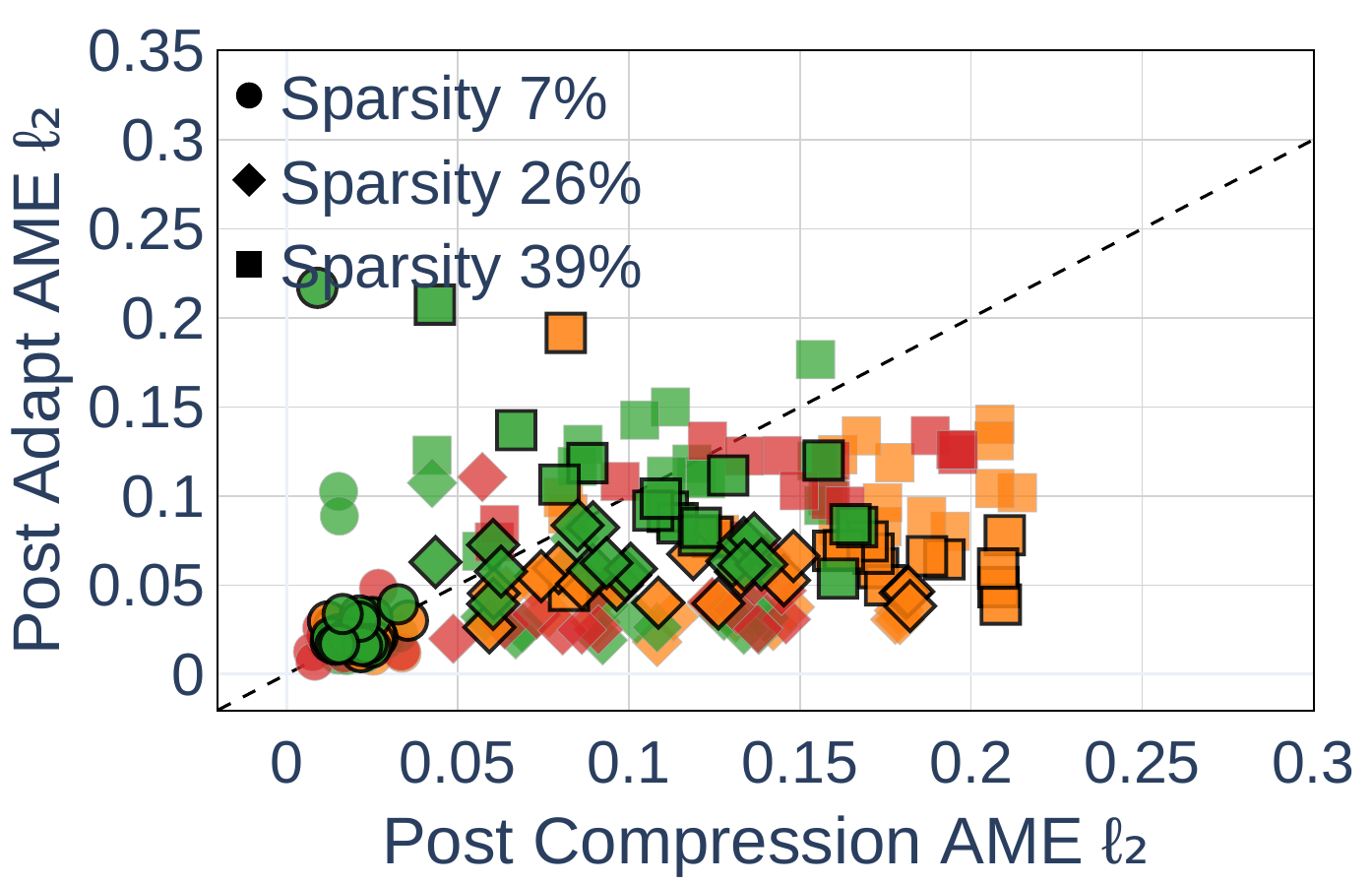}
\caption{ViT-Base on ImageNet-C}
\label{fig:ame_data_vit_color_methods}
\end{subfigure}

\vspace{0.3em}

\begin{subfigure}[t]{0.33\textwidth}
\centering
\includegraphics[width=\textwidth]{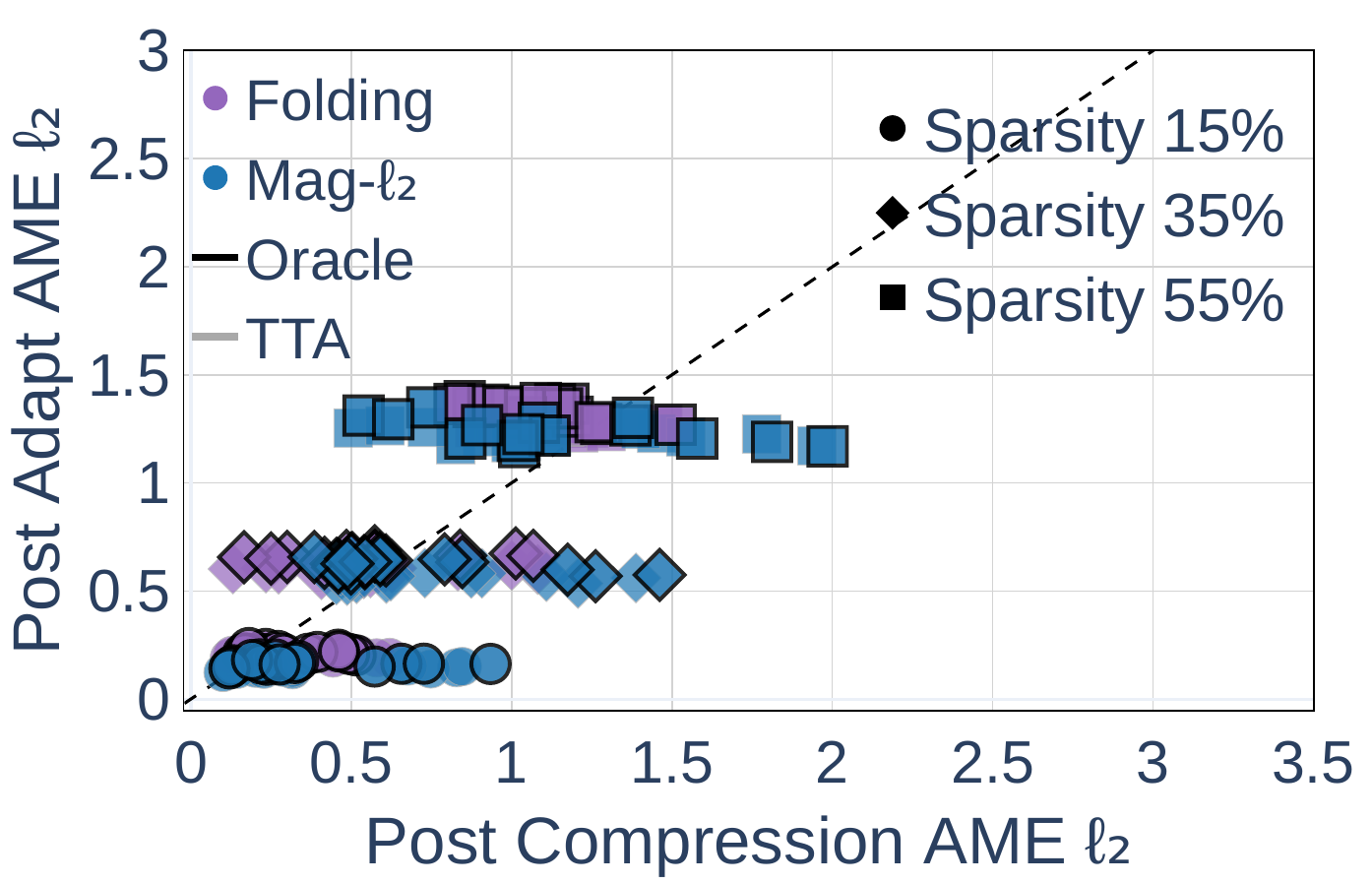}
\caption{ResNet-18 on CIFAR-10-C}
\label{fig:ame_free_cifar10_color_methods}
\end{subfigure}
\hfill
\begin{subfigure}[t]{0.33\textwidth}
\centering
\includegraphics[width=\textwidth]{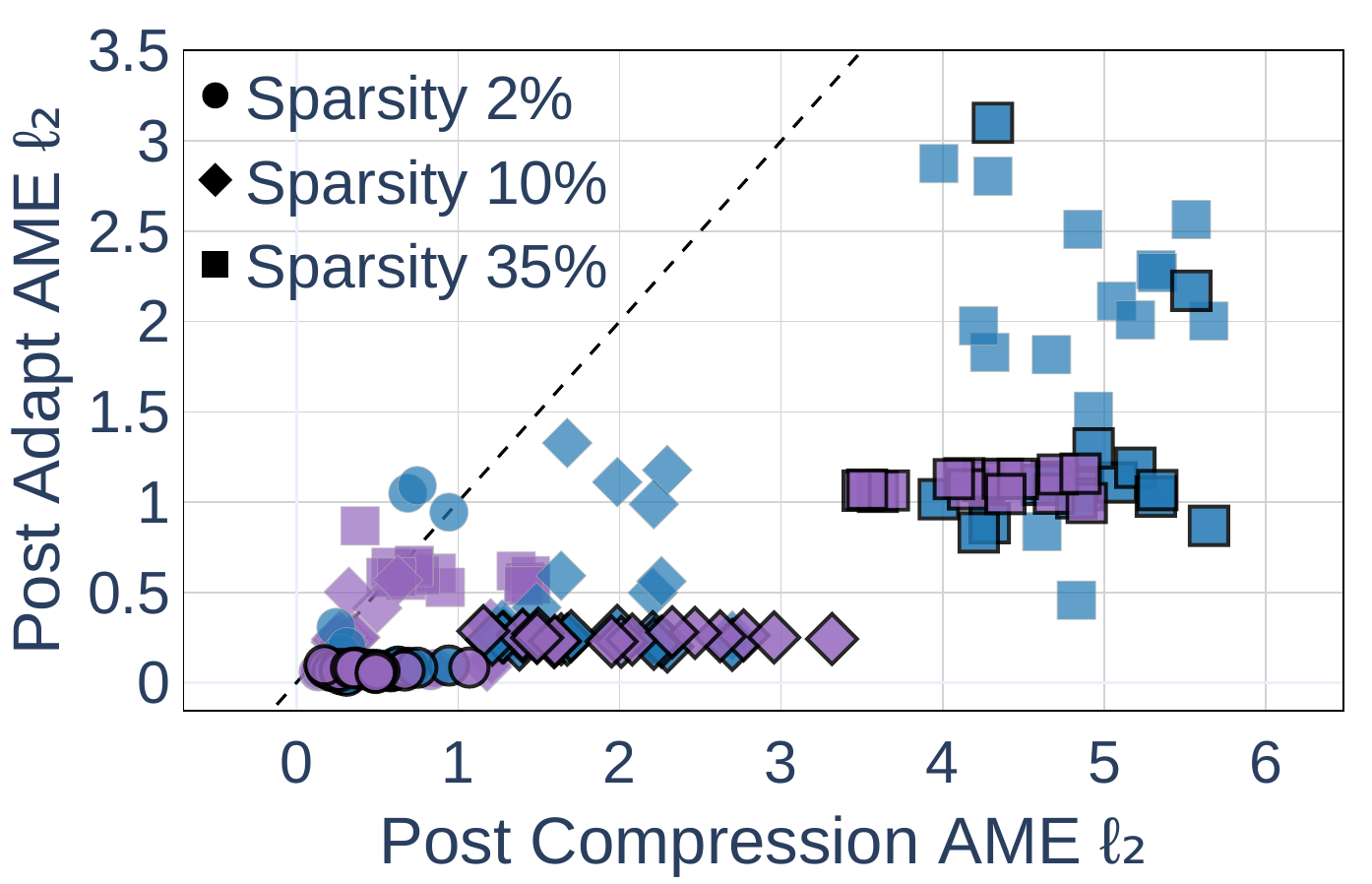}
\caption{ResNet-18 on ImageNet-C}
\label{fig:ame_free_imagenet_color_methods}
\end{subfigure}
\hfill
\begin{subfigure}[t]{0.33\textwidth}
\centering
\includegraphics[width=\textwidth]{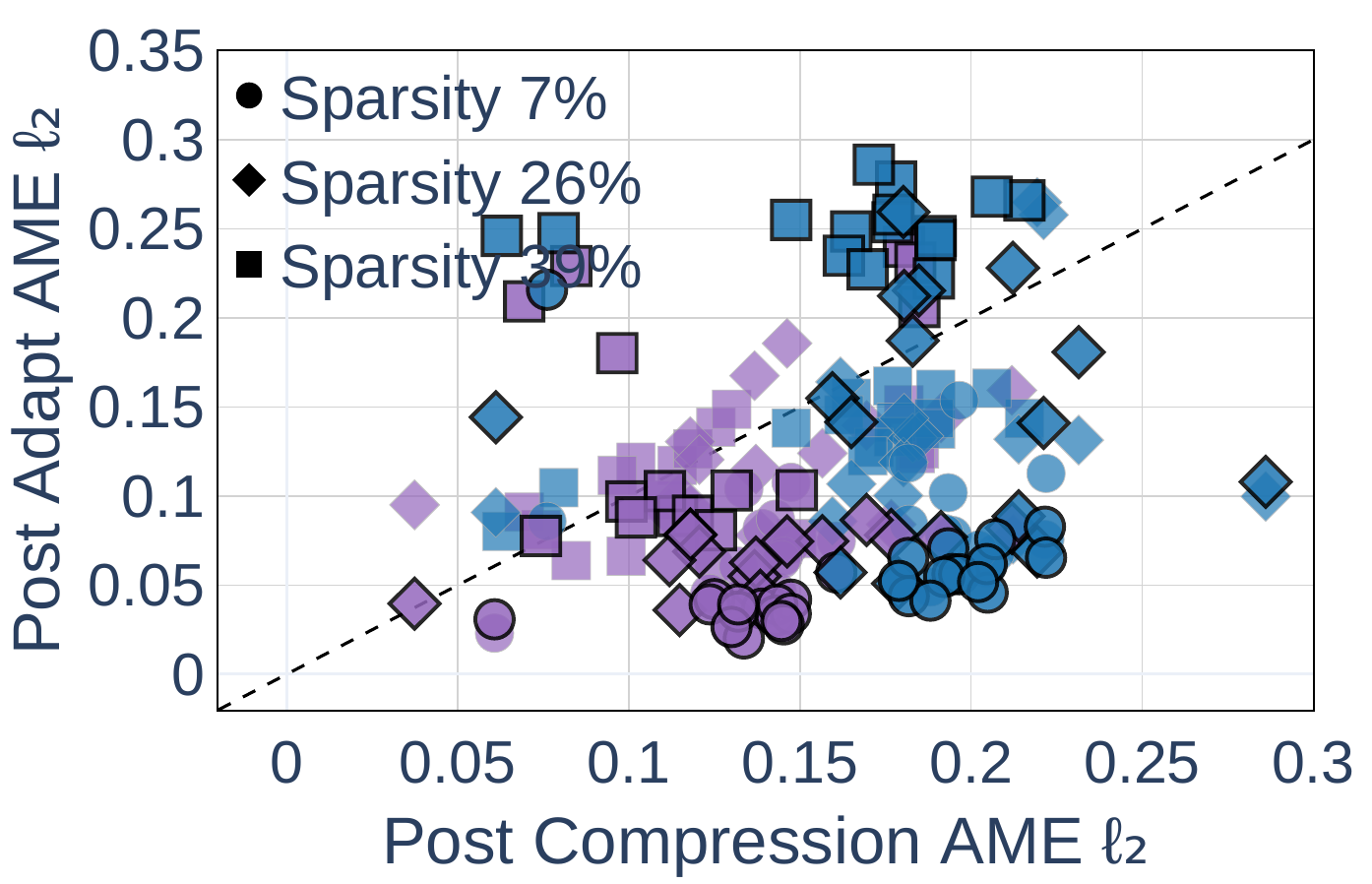}
\caption{ViT-Base on ImageNet-C}
\label{fig:ame_free_vit_color_methods}
\end{subfigure}

\caption{\textbf{Test-time adaptation (\tta) only partially restores output expressivity, with recovery strongly dependent on the compression method.}
We measure expressivity via Activation Map Entropy (AME) and report the $\ell_2$ distance between entropy vectors of compressed and dense models before (x-axis) and after adaptation (y-axis). Lower values indicate outputs closer to the dense model. Points below the identity line ($y{=}x$) correspond to successful recovery of output expressivity, the lower the better. Across all settings, both \tta and supervised adaptation (Oracle) reduce the AME distance, indicating partial restoration of output diversity. However, Oracle consistently achieves stronger and more stable recovery, while \tta exhibits weaker and less consistent improvements, especially at higher compression levels and on ImageNet-C. The recovery behavior varies across compression methods. Wanda and Taylor enable moderate recovery under \tta, whereas OBD and Mag-$\ell_2$ show unstable recovery, often failing to restore output expressivity. In contrast, Folding yields more consistent recovery patterns, but still remains below Oracle.
\emph{Top row} (a--c): data-dependent pruning (Wanda, Taylor, OBD). \emph{Bottom row} (d--f): data-free compression (Folding, Mag-$\ell_2$).}
\label{fig:ame_main}
\end{figure*}

\subsection{Expressivity and diversity}

Structured compression can reduce the diversity of internal representations and collapse their effective rank, thereby restricting the range of functions the model can realize during adaptation. We ask: \textit{To what extent can adaptation recover the representational expressivity and diversity lost due to structured compression?} To answer this question, we quantify representational recovery using Centered Kernel Alignment (CKA) to measure alignment with the dense model and Activation Map Entropy (AME) to capture the diversity of layer-wise activations. We additionally test whether the representational collapse is an artifact of the similarity index, and whether model size explains the loss of adaptability.

\textbf{Centered Kernel Alignment.}
We quantify representational similarity between dense and compressed models using Centered Kernel Alignment (\cka)~\citep{kornblith2019similarity}, computed on post-residual hidden representations and summarized by the worst-layer \cka across all layers. This worst-layer statistic identifies the most degraded layer under compression. As reported in~\citet{tishby2015deep}, information lost at any intermediate layer cannot be recovered by subsequent layers; the most degraded layer acts as a representational bottleneck that may constrain the model's capacity for adaptation. The information-bottleneck principle~\citep{tishby2015deep} governs the mutual information between the input and the layer-$l$ activation, not the kernel alignment that \cka measures; we use the information-bottleneck argument as an intuition for the monotone non-increase of recoverable information along the forward pass and motivate the worst-layer statistic empirically by reporting its correlation with adaptation accuracy (Fig.~\ref{fig:cka_spearman_appendix}).

On ResNet-18 (Fig.~\ref{fig:cka_data_imagenet_methods_colors},~\ref{fig:cka_free_imagenet_methods_colors}), we observe that \tta exhibits representational collapse already at very low sparsity (as early as 2\% on ImageNet-C), while Oracle consistently recovers representations closer to the dense model. This gap persists and widens with increasing compression, indicating that \tta fails to restore the most degraded layers. This behavior is consistent with a self-reinforcing degradation under entropy minimization~\citep{niu2023towards}: since gradients are derived from the model’s predictions damaged by compression, prediction errors may influence the \tta adaptation signal, reinforcing representational distortions rather than correcting them. In contrast, Oracle relies on ground-truth labels, which may provide a more robust gradient signal and explain its consistently stronger representation recovery. We formalize this hypothesis in the gradient degeneracy analysis (Sec.~\ref{sec:theory}).

On ViT-Base (Fig.~\ref{fig:cka_free_vit_methods_colors}), the impact of the compression method becomes more pronounced. At 39\% sparsity, Mag-$\ell_2$ leads to near-complete representational collapse (\cka $\approx 0$), whereas Folding recovers substantially higher post-adaptation alignment with the dense model. This difference may arise from how the two methods transform the activation Gram matrices analyzed by \cka~\citep{kornblith2019similarity}: Mag-$\ell_2$ removes entire units and their contributions, while Folding aggregates correlated units via clustering~\citep{wang2025forget}, better preserving the variance structure of the representations. This observation is consistent with~\citet{hu2024sp3}, who show that pruning in the original feature dimension disrupts principal components that span multiple dimensions, motivating projection-based approaches.

\textbf{Activation Map Entropy.}
While \cka captures representational \emph{expressivity} via alignment with the dense model~\citep{kornblith2019similarity}, it does not characterize the spectral diversity of layer-wise activations. To complement this, we define Activation Map Entropy (\ame) building on the matrix-based entropy framework of~\citet{skean2025layer} as a measure of representational \emph{diversity}. For each layer $l$, \ame is defined as the Shannon entropy of the normalized eigenvalues of the activation Gram matrix $G_l \in \mathbb{R}^{C_l \times C_l}$, quantifying the effective rank of the representation~\citep{roy2007effective}.

Structured compression reduces $G_l$ from $C_l \times C_l$ to $C'_l \times C'_l$, imposing an upper bound on entropy at $\log C'_l$. Since \tta methods~\citep{niu2023towards, wang2021tent} update only normalization parameters, they cannot increase the rank of $G_l$ beyond $C'_l$. Compression therefore imposes an upper bound on the recoverable diversity of representations. Consistent with prior observations that pruning reduces activation entropy~\citep{liao2024nepenthe}, we expect this reduction to propagate to the eigenspectrum of $G_l$.

Empirically, on CIFAR-10-C (Fig.~\ref{fig:ame_data_cifar10_color_methods},~\ref{fig:ame_free_cifar10_color_methods}), both Oracle and \tta recover \ame at low sparsity (15\%), but diverge at higher sparsity (35\%, 55\%), where points form horizontal saturation bands. These bands reflect the entropy ceiling imposed by reduced dimensionality, limiting further recovery regardless of the adaptation method. On ImageNet-C with ResNet-18 (Fig.~\ref{fig:ame_data_imagenet_color_methods},~\ref{fig:ame_free_imagenet_color_methods}), this limitation becomes more pronounced: at 35\% sparsity, most points lie below the identity line, indicating partial recovery, with Oracle consistently achieving tighter and lower-distance clusters than \tta. This suggests that while both methods operate under the same dimensionality constraint, Oracle more effectively redistributes the available eigenvalue mass toward the dense reference.

On ViT-Base (Fig.~\ref{fig:ame_data_vit_color_methods},~\ref{fig:ame_free_vit_color_methods}), \ame distances are substantially smaller and less sensitive to compression. This can be attributed to FFN-only compression, which preserves the post-residual dimensionality fixed at 768~\citep{dosovitskiy2020image}, keeping the size of $G_l$ invariant. As a result, the entropy ceiling remains largely unchanged, and residual connections further mitigate the impact of compression on the activation spectrum. Nevertheless, compression methods still differ: Folding achieves lower post-adaptation \ame distances than Mag-$\ell_2$ and consistently outperforms \tta, aligning with the trends observed in the \cka analysis (Fig.~\ref{fig:cka_free_vit_methods_colors}).

\textbf{Metric Similarity Analysis.} 
To further support our analysis, we adopt the spectral-ratio pseudo-distance $d_{\mathrm{SR}}$~\citep{caycogajic2026geometry}, which compares the intrinsic geometry of neural representations under the manifold hypothesis, whereas \cka compares their extrinsic geometry. Following~\citet{caycogajic2026geometry}, we apply $d_{\mathrm{SR}}$ to the Riemannian pullback metrics of the dense and the compressed model, which measure how strongly each model's output responds to small perturbations of the input, rather than to the activation Gram matrices $G_l$ analyzed by \cka and \ame. $d_{\mathrm{SR}}$ ranks post-compression representation consistently with worst-layer \cka (Spearman $\rho=-0.925$ on ResNet-18/ImageNet-C and $-0.665$ on ViT-Base/ImageNet-C; the negative sign is expected when a distance metric is compared with a similarity; see Fig.~\ref{fig:msa_original}), indicating that the representational collapse observed on ImageNet-C is not an artifact of the similarity index~\citep{davari2022reliability}.

\textbf{Disentangling compression from capacity.} We compare compressed against \emph{smaller-dense} models of equal architecture trained from scratch: at increasing sparsity, \tta and its Oracle adaptation show an earlier collapse compared to the smaller-dense models (Fig.~\ref{fig:scale_study}), indicating that the loss of adaptability is induced by the structured compression rather than by the reduced model size.

Structured compression imposes intrinsic limits on both representational alignment and diversity that adaptation cannot fully overcome. While Oracle approaches these limits, \tta remains constrained by degraded signals and reduced capacity, resulting in systematically weaker recovery.

\begin{figure*}[t]
\centering
\begin{subfigure}[t]{0.32\textwidth}
\centering
\includegraphics[width=\textwidth]{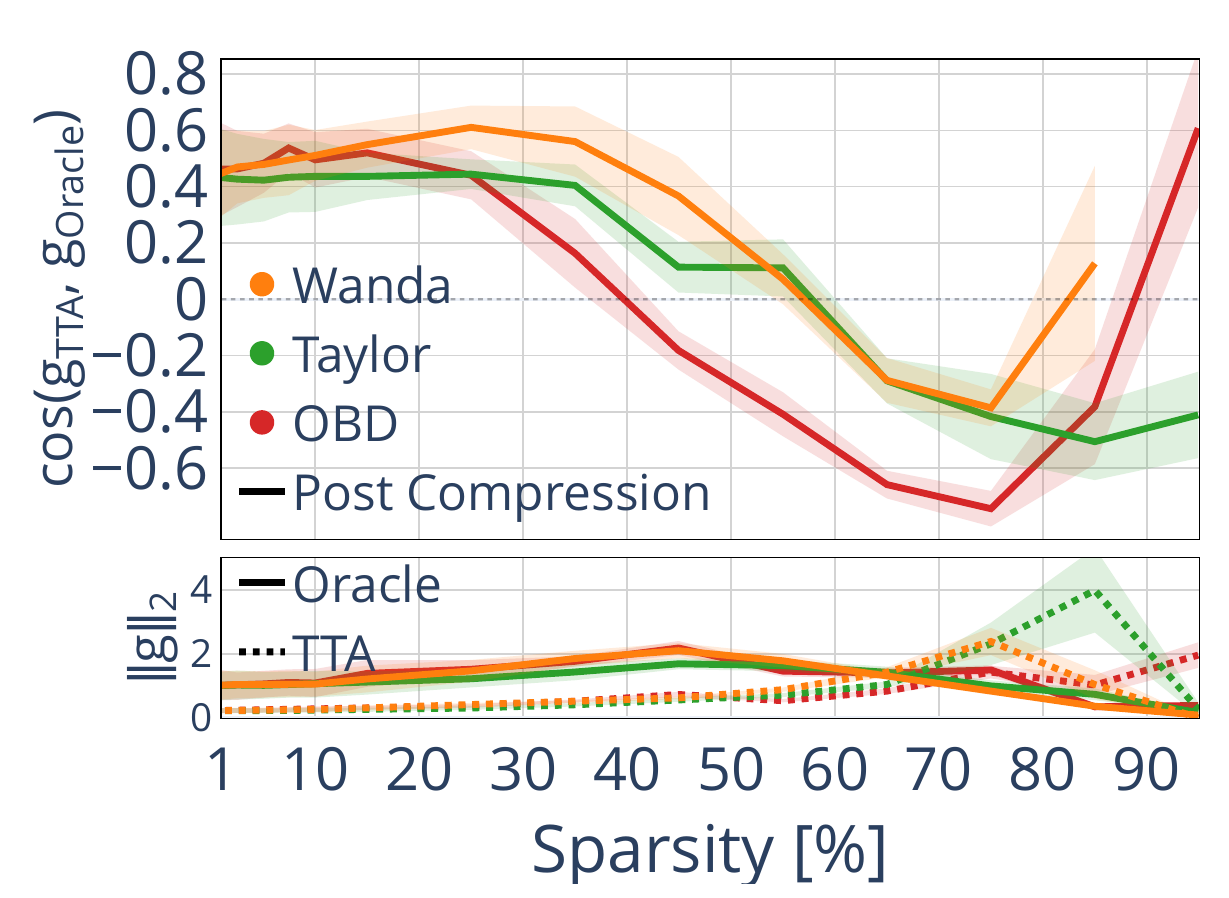}
\caption{ResNet-18 on CIFAR-10-C}
\label{fig:cs_data_cifar10}
\end{subfigure}
\hfill
\begin{subfigure}[t]{0.32\textwidth}
\centering
\includegraphics[width=\textwidth]{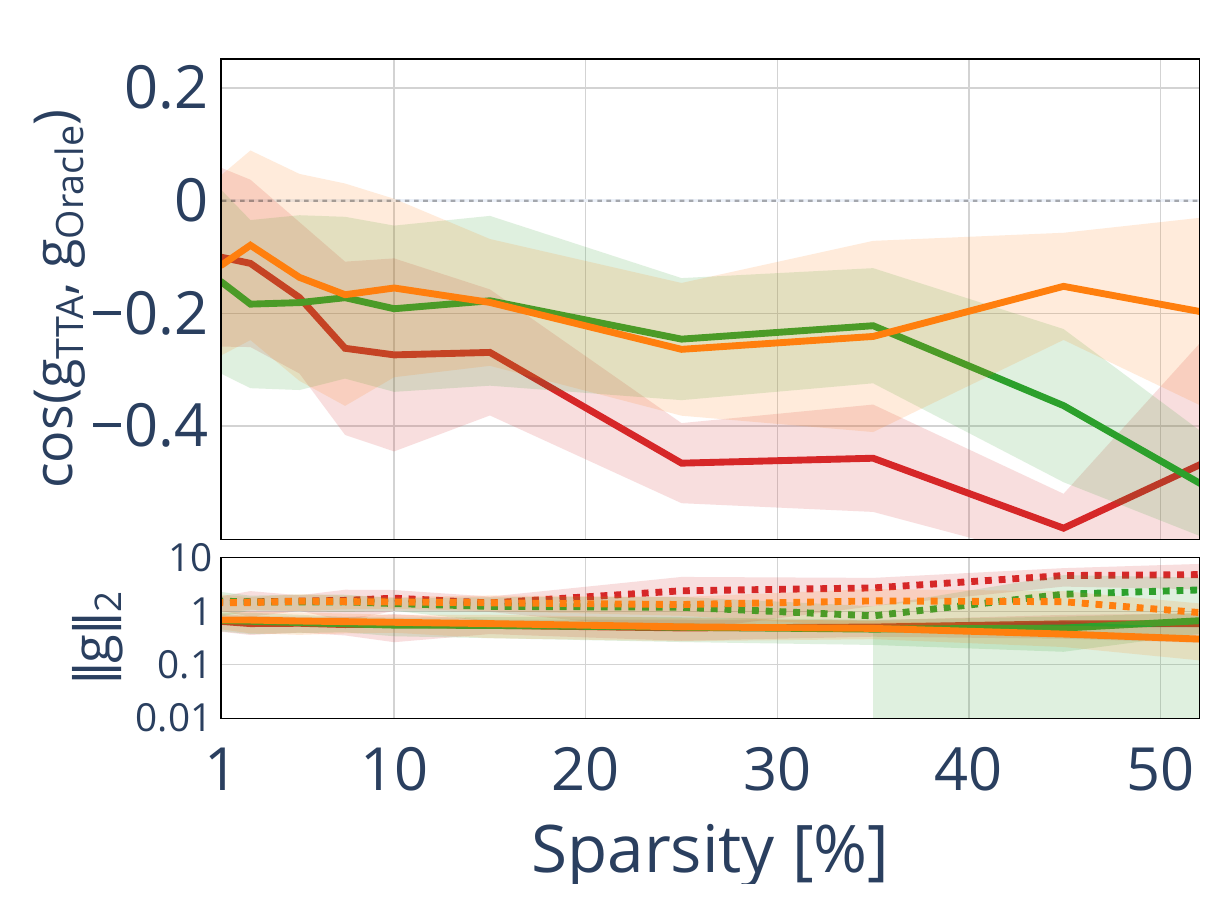}
\caption{ResNet-18 on ImageNet-C}
\label{fig:cs_data_imagenet}
\end{subfigure}
\hfill
\begin{subfigure}[t]{0.32\textwidth}
\centering
\includegraphics[width=\textwidth]{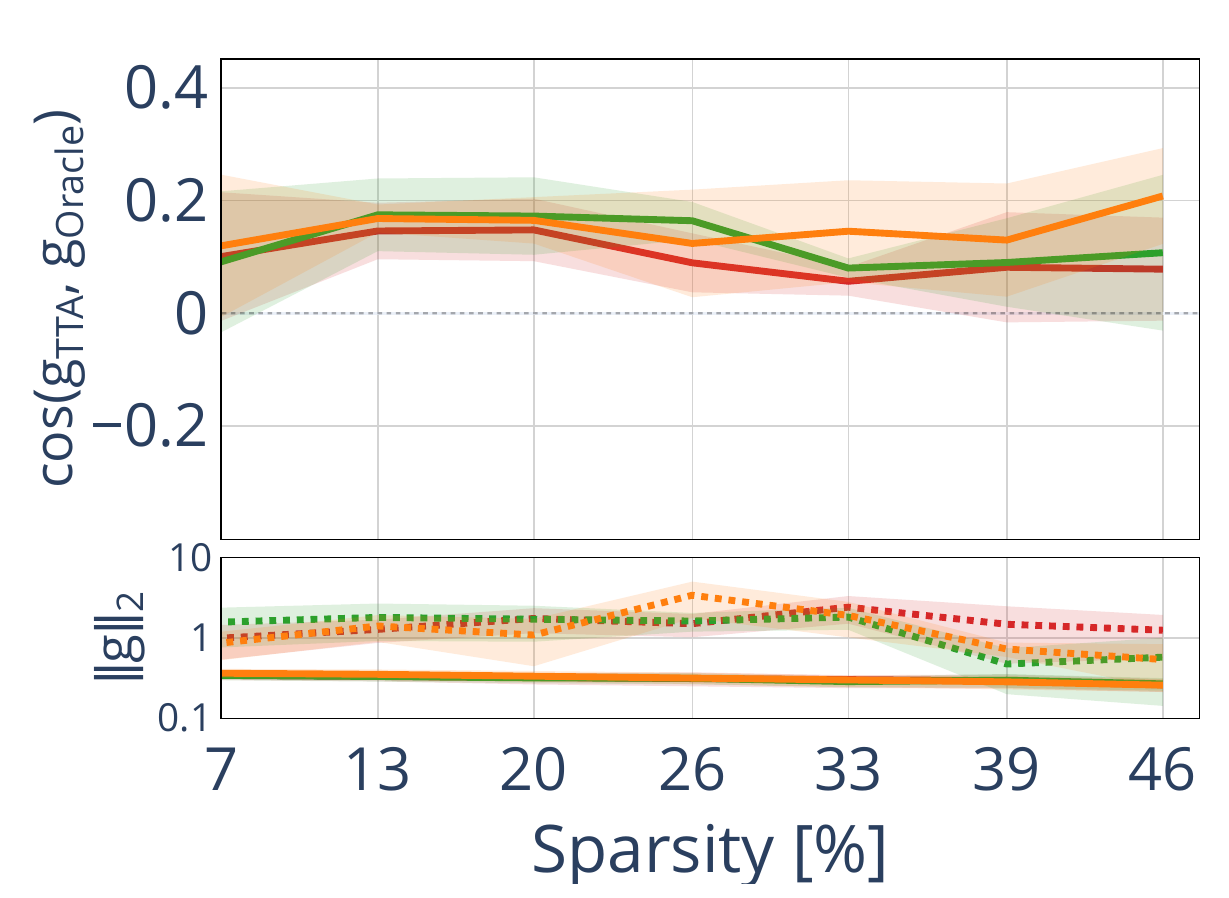}
\caption{ViT-Base on ImageNet-C}
\label{fig:cs_data_vit}
\end{subfigure}

\vspace{0.3em}

\begin{subfigure}[t]{0.32\textwidth}
\centering
\includegraphics[width=\textwidth]{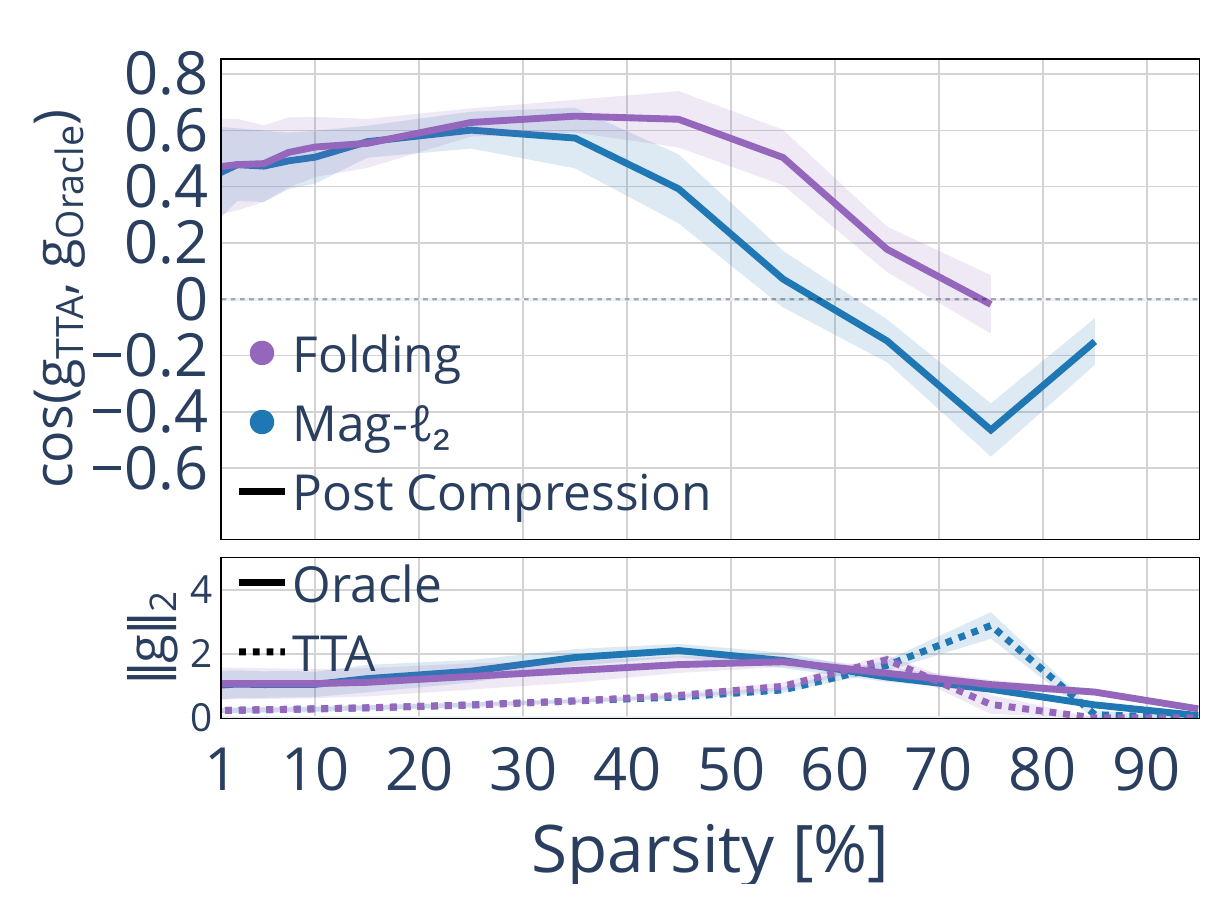}
\caption{ResNet-18 on CIFAR-10-C}
\label{fig:cs_free_cifar10}
\end{subfigure}
\hfill
\begin{subfigure}[t]{0.32\textwidth}
\centering
\includegraphics[width=\textwidth]{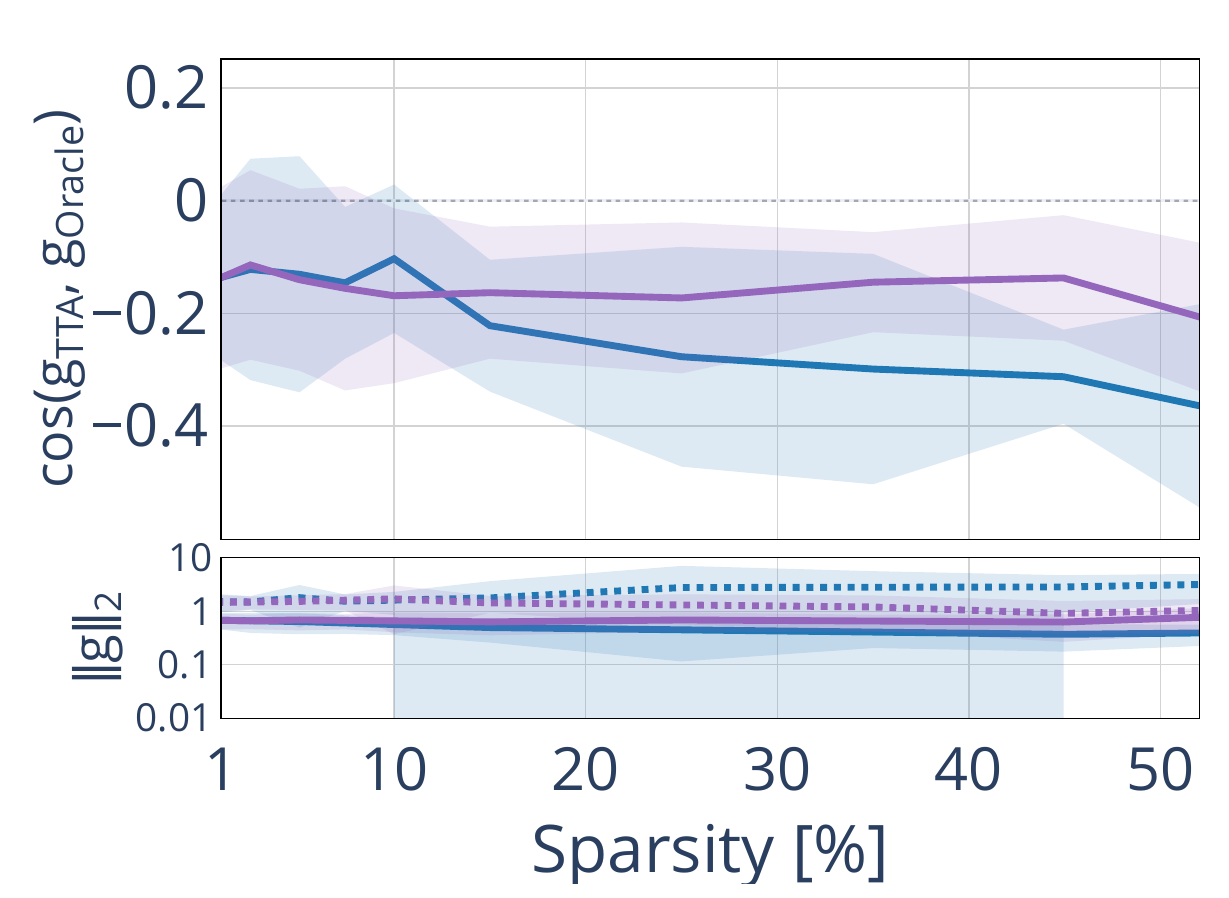}
\caption{ResNet-18 on ImageNet-C}
\label{fig:cs_free_imagenet}
\end{subfigure}
\hfill
\begin{subfigure}[t]{0.32\textwidth}
\centering
\includegraphics[width=\textwidth]{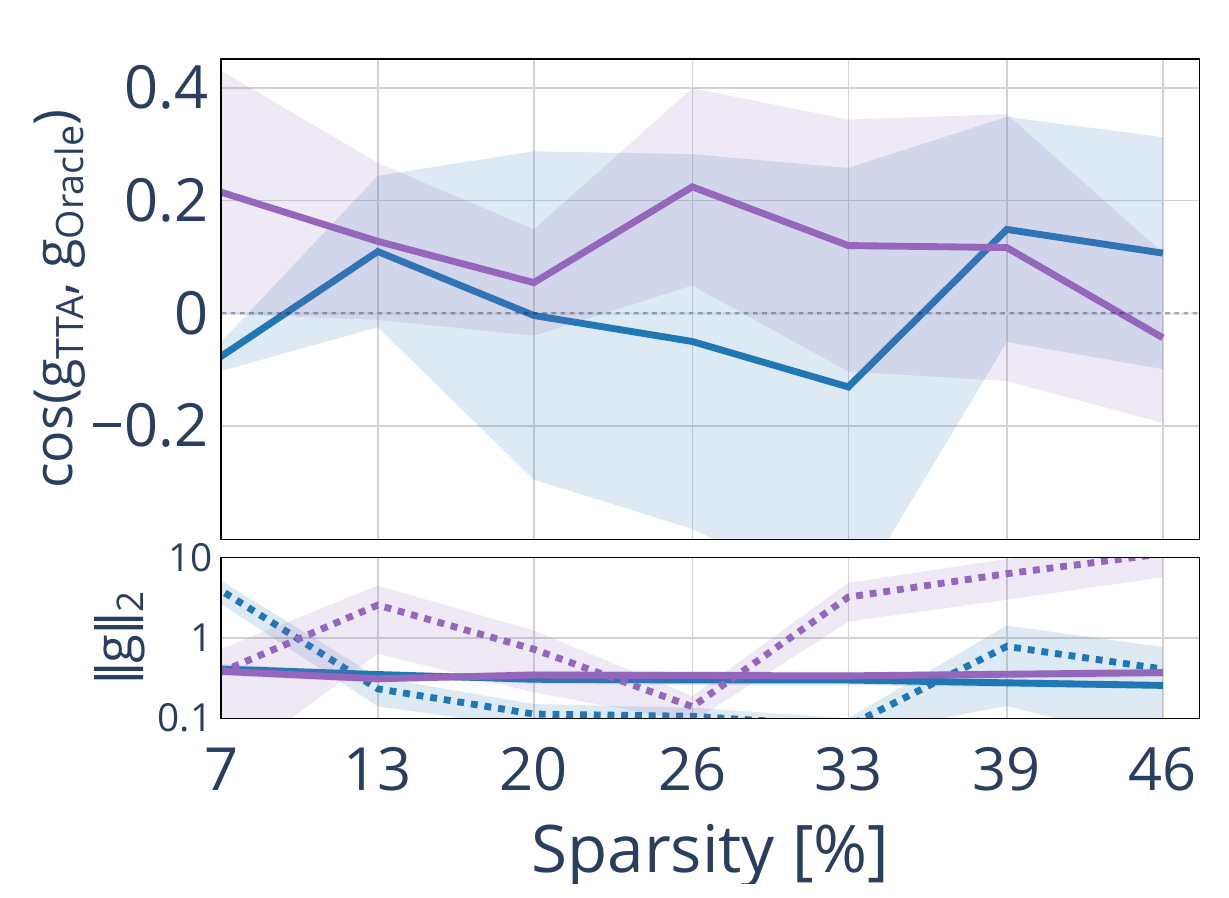}
\caption{ViT-Base on ImageNet-C}
\label{fig:cs_free_vit}
\end{subfigure}
\caption{\textbf{Structured compression increasingly misaligns the \tta gradient from the supervised gradient direction (Oracle).} We report cosine similarity $\cos(\mathbf{g}_{\text{\tta}},\mathbf{g}_{\text{Oracle}})$ (Eq.~\ref{eq:grad_cosine}, top panels) and gradient $\ell_2$ norms (bottom panels), at the post-compression, pre-adaptation parameters $\phi_0$ averaged across all corruptions. At low sparsity, the \tta and Oracle gradients are positively aligned, suggesting that the \tta objective initially provides a reliable surrogate for the supervised adaptation direction. As sparsity increases, this alignment deteriorates and can reverse sign, indicating that the \tta updates conflict with the supervised ones, a more severe failure mode than the loss of adaptation signal ($\cos \approx 0$). \tta gradient norms vanish (degeneracy) or remain large while misaligned (active divergence), both leading to adaptation collapse reflecting the \tta accuracy collapse observed in Fig.~\ref{fig:hero}; lines terminate when the metric becomes undefined at high sparsities. The SPA objective on ViT-Base preserves weakly positive alignment across all tested sparsities, consistent with the smaller Oracle-\tta gap in Fig.~\ref{fig:hero}. \emph{Top row}~(a--c): data-dependent pruning (Wanda, Taylor, OBD). \emph{Bottom row}~(d--f): data-free compression (Folding, Mag-$\ell_2$).}
\label{fig:cosine_similarity}
\end{figure*}

\subsection{Subspace compatibility}

Structured compression reduces the set of trainable parameters available during adaptation and may also alter how well the remaining update directions align with the target-domain loss.
We ask: \emph{Does compression reduce the fraction of useful adaptation signal accessible within the admissible update subspace, and can this help explain the growing gap between unsupervised test-time adaptation and supervised target-domain adaptation?}

\paragraph{Gradient cosine similarity.} 

Structured compression reduces the dimensionality of the admissible update $\Delta \in \mathcal{U}$ (Eq.~\ref{eq:tta}) by eliminating adaptable parameters corresponding to pruned~\citep{li2017_pruning_filters} or folded~\citep{wang2025forget} units. Since pruned deep networks suffer disproportionately under distribution shift~\citep{liebenwein2021lost}, adaptability may depend not only on the preserved $\dim(\mathcal{U})$ but also on whether the adaptation gradient directions preserve their alignment with the gradient of the supervised loss on the target domain. We quantify this alignment via the cosine similarity between the accumulated \tta and Oracle gradients, restricted to $\mathcal{U}$: 
\begin{equation}
\cos(\mathbf{g}_{\text{\tta}},\, \mathbf{g}_{\text{Oracle}}) = \frac{\mathbf{g}_{\text{\tta}} \cdot \mathbf{g}_{\text{Oracle}}}{\|\mathbf{g}_{\text{\tta}}\| \cdot \|\mathbf{g}_{\text{Oracle}}\|},
\label{eq:grad_cosine}
\end{equation}
where $\mathbf{g}_{\text{\tta}} = \frac{1}{N}\sum_{i=1}^{N} \nabla_{\phi}\mathcal{L}_{\text{\tta}}(x_i;\phi_0)$ and $\mathbf{g}_{\text{Oracle}} = \frac{1}{N}\sum_{i=1}^{N} \nabla_{\phi}\mathcal{L}_{\text{CE}}(x_i, y_i;\phi_0)$ are the adaptation set gradients averaged over all $N$ corruption samples. We compute a single cosine similarity metric on these adaptation set gradients, rather than averaging per-sample~\citep{chen2025grata} or per-task~\citep{yu2020gradient} similarities, to assess whether the \tta loss provides a meaningful proxy of the supervised optimization direction at the pre-adaptation parameters $\phi_0$. The metric is undefined when either gradient norm vanishes, which implies an adaptation collapse regime, a condition we analyze theoretically below. 

Empirically, Fig.~\ref{fig:cosine_similarity} reveals three distinct regimes of gradient alignment under structured compression. On CIFAR-10-C (Fig.~\ref{fig:cs_data_cifar10}, \ref{fig:cs_free_cifar10}), the \tta and Oracle gradients are positively aligned at low sparsity, indicating that the \tta updates initially point in a direction consistent with the supervised optimization objective. As sparsity increases, however, the cosine similarity crosses zero and becomes strongly negative, reaching its minimum value between $70\%$ and $80\%$ sparsity across all compression criteria. This sign reversal suggests that the \tta update actively drives the model \emph{away} from the supervised optimum, constituting a more severe failure mode than the simple loss of an adaptation signal, which would correspond to $\cos \approx 0$. This phenomenon is compounded by the progressively widening discrepancy in gradient norms. The emergence of this misalignment coincides with the sparsity level at which \tta accuracy begins to deteriorate sharply (Fig.~\ref{fig:hero}),  offering a gradient lens explanation for the observed adaptation degradation. At extreme sparsities ($\geq$ 75\%), the metric becomes undefined when the \tta gradient norm vanishes for all corruptions, \eg Mag-$\ell_2$ beyond 85\%. These results support our entropy gradient analysis (Eq.~\ref{eq:ent_grad}, \ref{eq:ent_grad_bound}): at equal sparsity, compressed models driven to near-uniform predictions produce weak adaptation signals, in contrast to the supervised signals which do not vanish. 

On ImageNet-C with ResNet-18 (Fig.~\ref{fig:cs_data_imagenet}, \ref{fig:cs_free_imagenet}), gradient alignment is negative already at 1\%, suggesting that SAR's loss objective may be a poor surrogate of the supervised loss on this harder distribution shift, even before significant model capacity is removed by compression. The gradient norms in the bottom panel reveal a failure mode distinct from CIFAR-10-C: $\|\mathbf{g}_{\text{\tta}}\|$ systematically exceeds $\|\mathbf{g}_{\text{Oracle}}\|$ across all sparsities. This result, combined with the negative cosine similarity, indicates an \emph{active divergence} regime: the SAR-based objective produces large updates in a direction that conflicts with the supervised one. This adaptation regime, associated with structured compression, may actively damage the model consistent with the severe Oracle-\tta gap previously observed (Fig.~\ref{fig:hero}). A causal ablation of this mechanism (Fig.~\ref{fig:ablation_recovery}) indicates that the confident-but-wrong samples admitted by SAR's reliability filter drive the misalignment: excluding them with a correctness filter computed from the ground-truth labels restores the alignment to $\approx{+}0.9$ across all compression criteria and architectures. 

On ViT-Base with SPA (Fig.~\ref{fig:cs_data_vit}, \ref{fig:cs_free_vit}), for data-dependent criteria, cosine similarity remains weakly positive across all sparsities, suggesting that the consistency-based SPA objective maintains partial alignment with the supervised direction even at 46\% sparsity. No sign reversal occurs, consistent with the smaller Oracle-\tta accuracy gap observed in Fig.~\ref{fig:hero}. Data-free compression exhibits more unstable adaptation behavior: Mag-$\ell_2$ produces negative mean alignment at several sparsity levels, and both Folding and Mag-$\ell_2$ exhibit high variance across corruptions. Gradient alignment remains fragile and corruption-dependent under data-free compression, where no calibration data guides unit selection. Decomposing these curves by corruption family (Figs.~\ref{fig:cs_noise} --\ref{fig:cs_digital}) shows similar gradient alignment regimes within each family.

\begin{figure*}[h]
\centering
\begin{subfigure}[t]{0.32\textwidth}
\centering
\includegraphics[width=\textwidth]{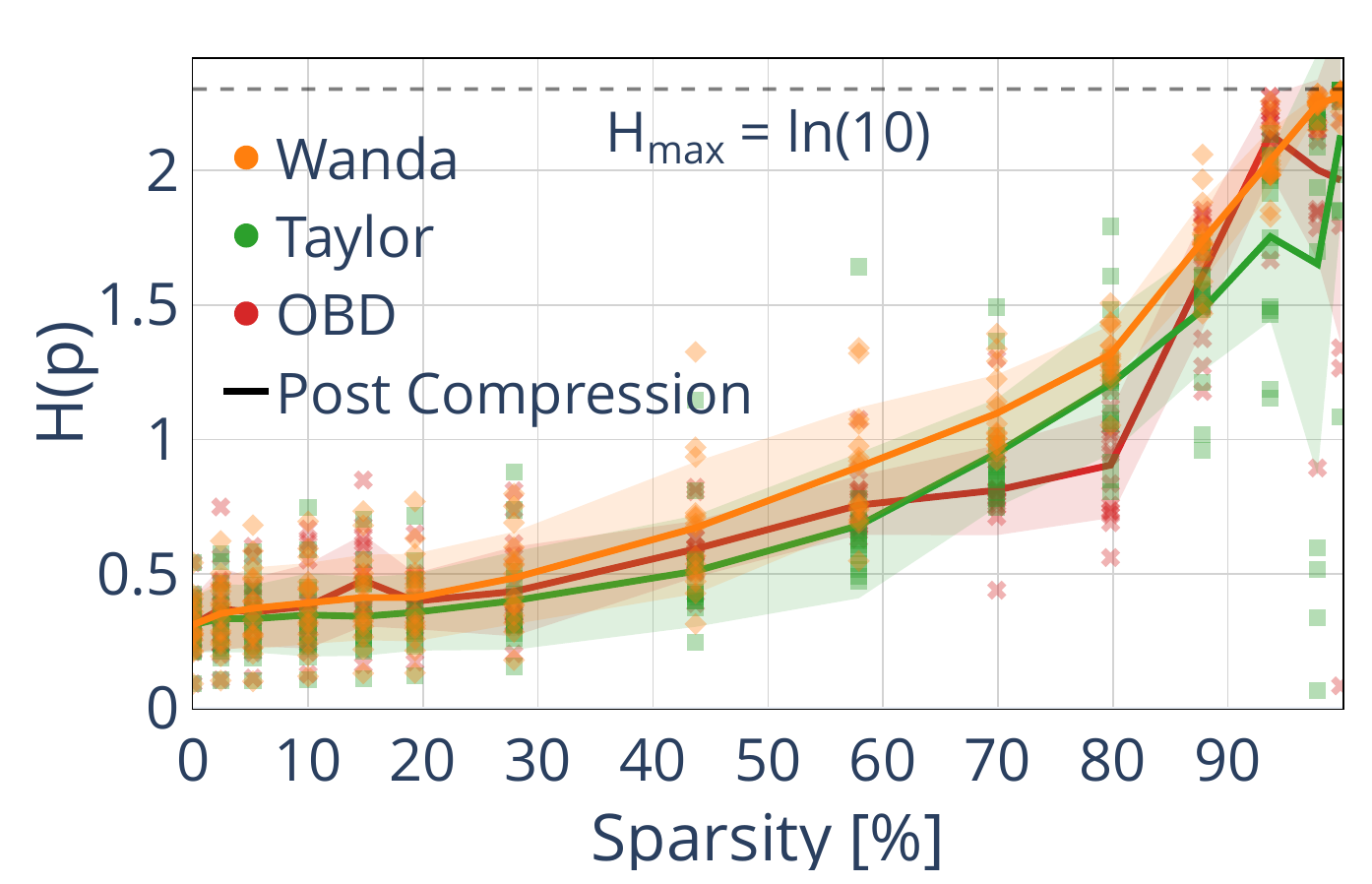}
\caption{ResNet-18 on CIFAR-10-C}
\label{fig:ent_data_cifar10}
\end{subfigure}
\hfill
\begin{subfigure}[t]{0.32\textwidth}
\centering
\includegraphics[width=\textwidth]{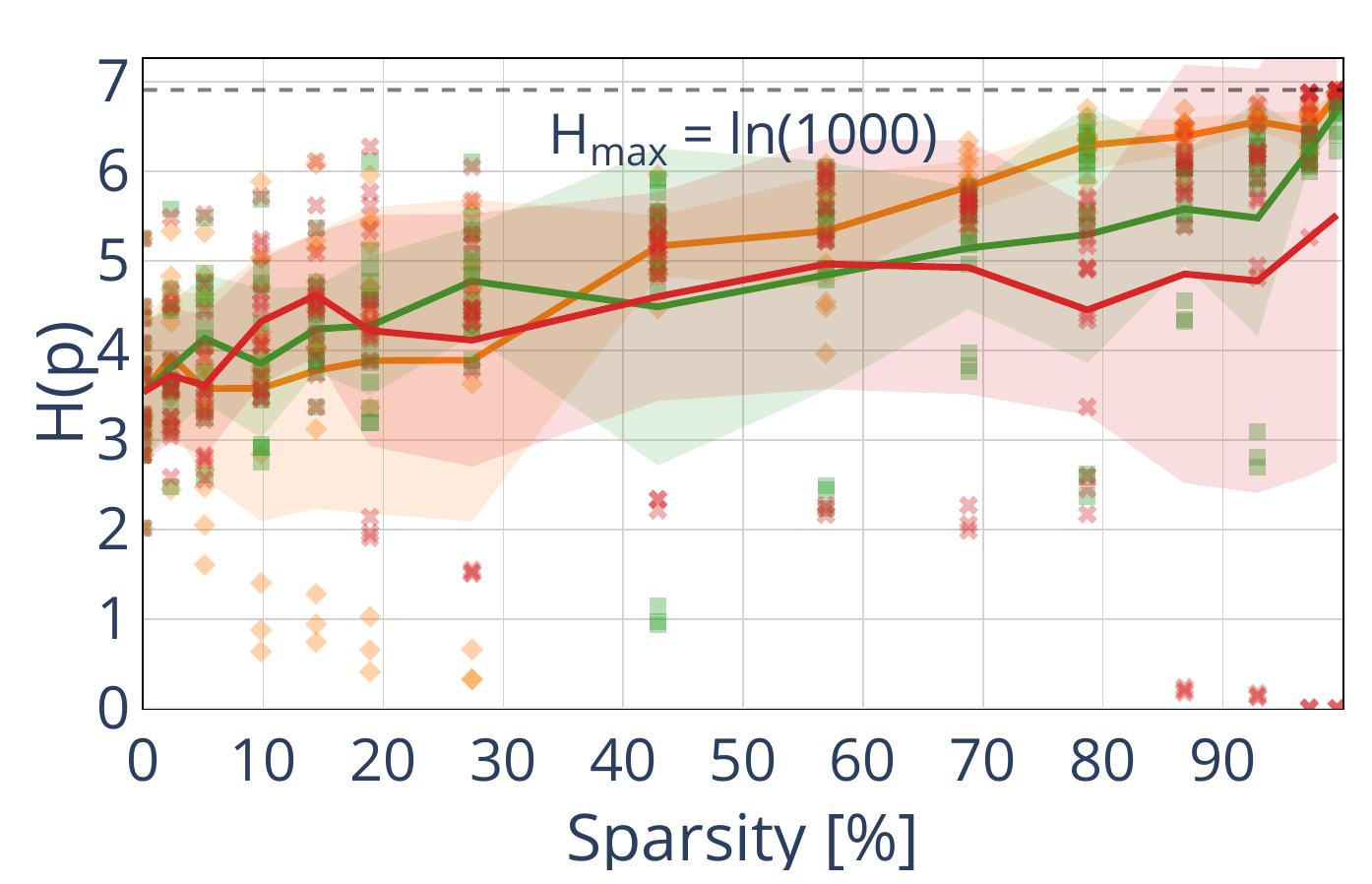}
\caption{ResNet-18 on ImageNet-C}
\label{fig:ent_data_imagenet}
\end{subfigure}
\hfill
\begin{subfigure}[t]{0.32\textwidth}
\centering
\includegraphics[width=\textwidth]{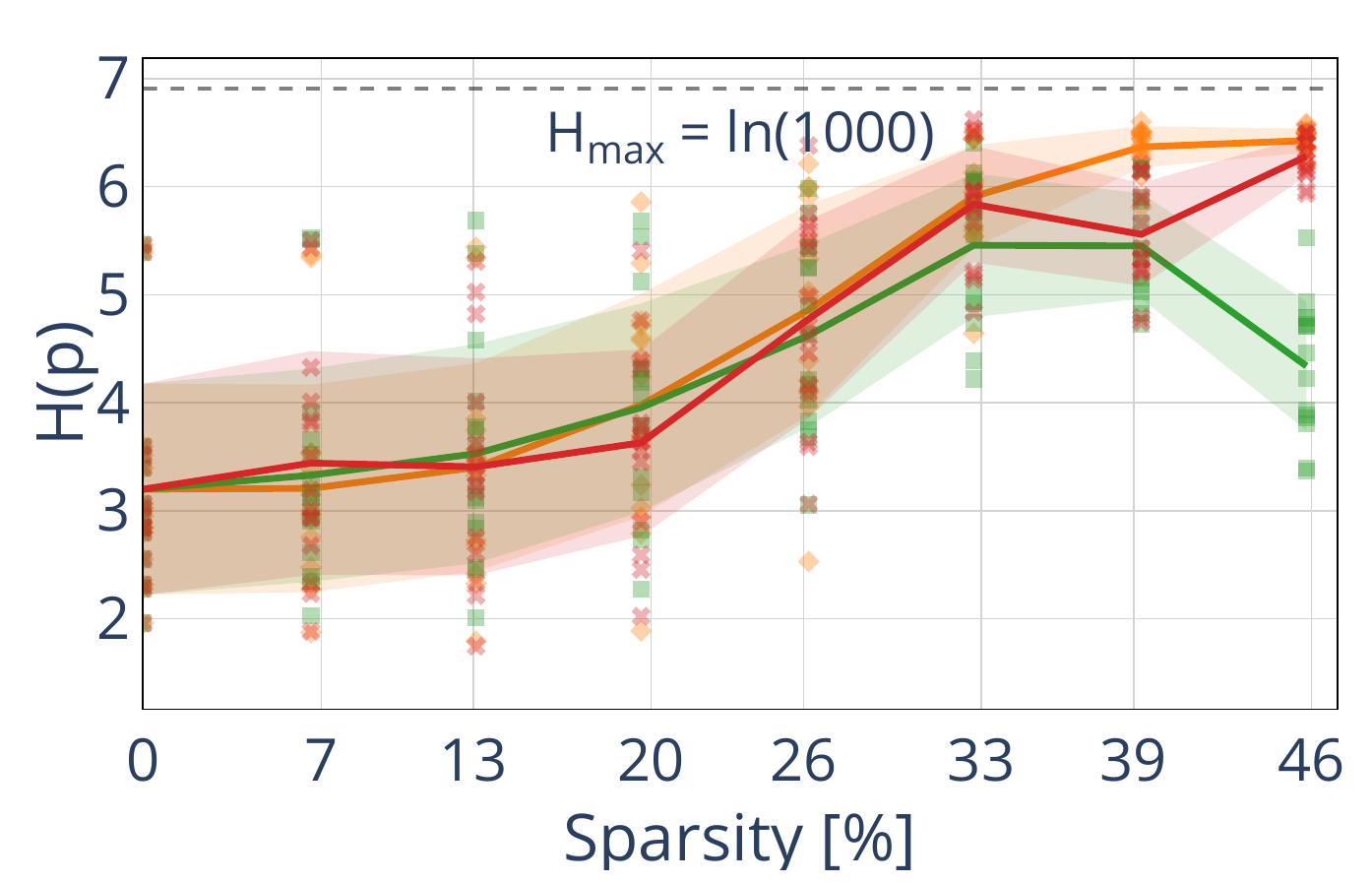}
\caption{ViT-Base on ImageNet-C}
\label{fig:ent_data_vit}
\end{subfigure}

\vspace{0.3em}

\begin{subfigure}[t]{0.32\textwidth}
\centering
\includegraphics[width=\textwidth]{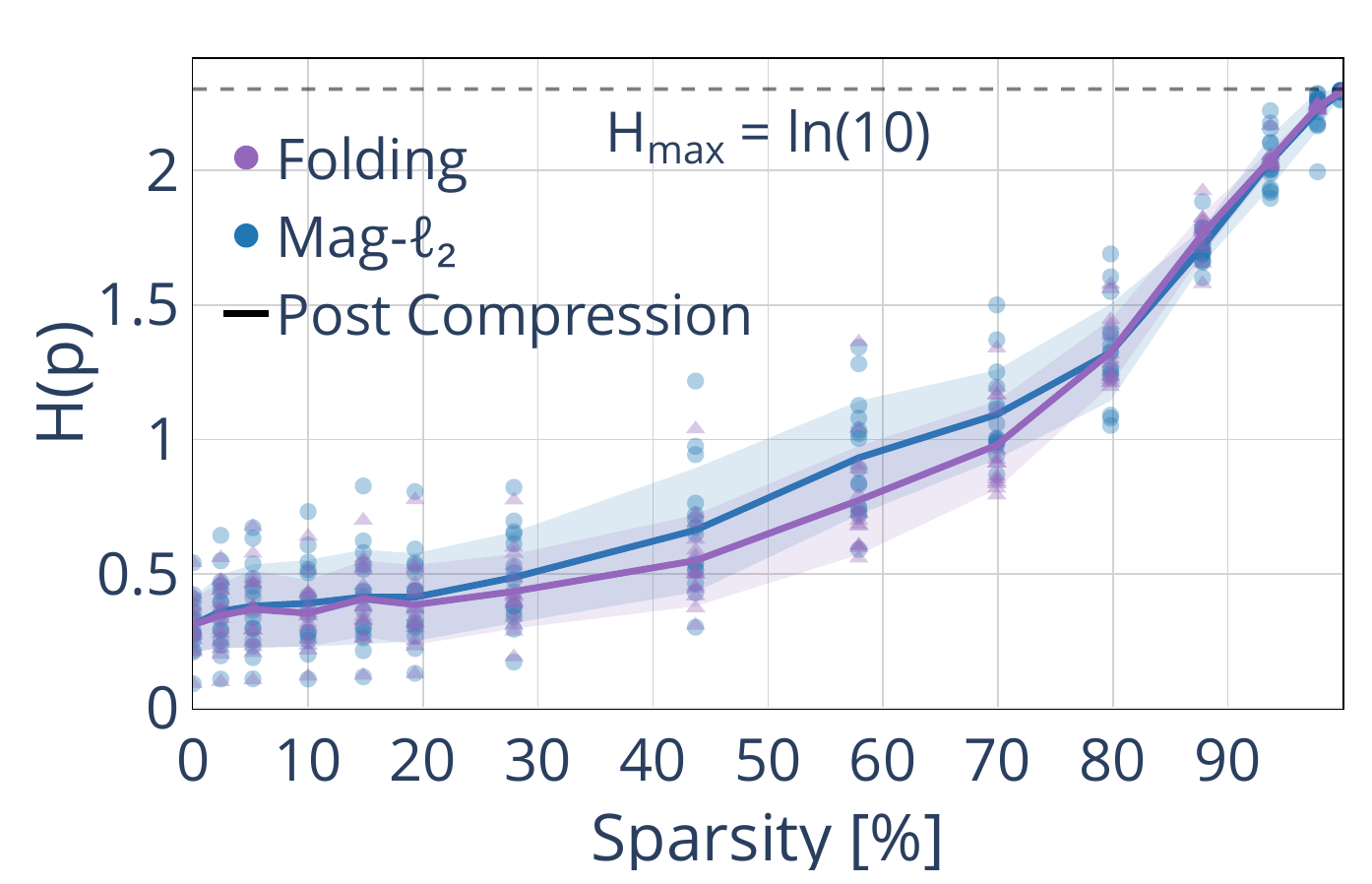}
\caption{ResNet-18 on CIFAR-10-C}
\label{fig:ent_free_cifar10}
\end{subfigure}
\hfill
\begin{subfigure}[t]{0.32\textwidth}
\centering
\includegraphics[width=\textwidth]{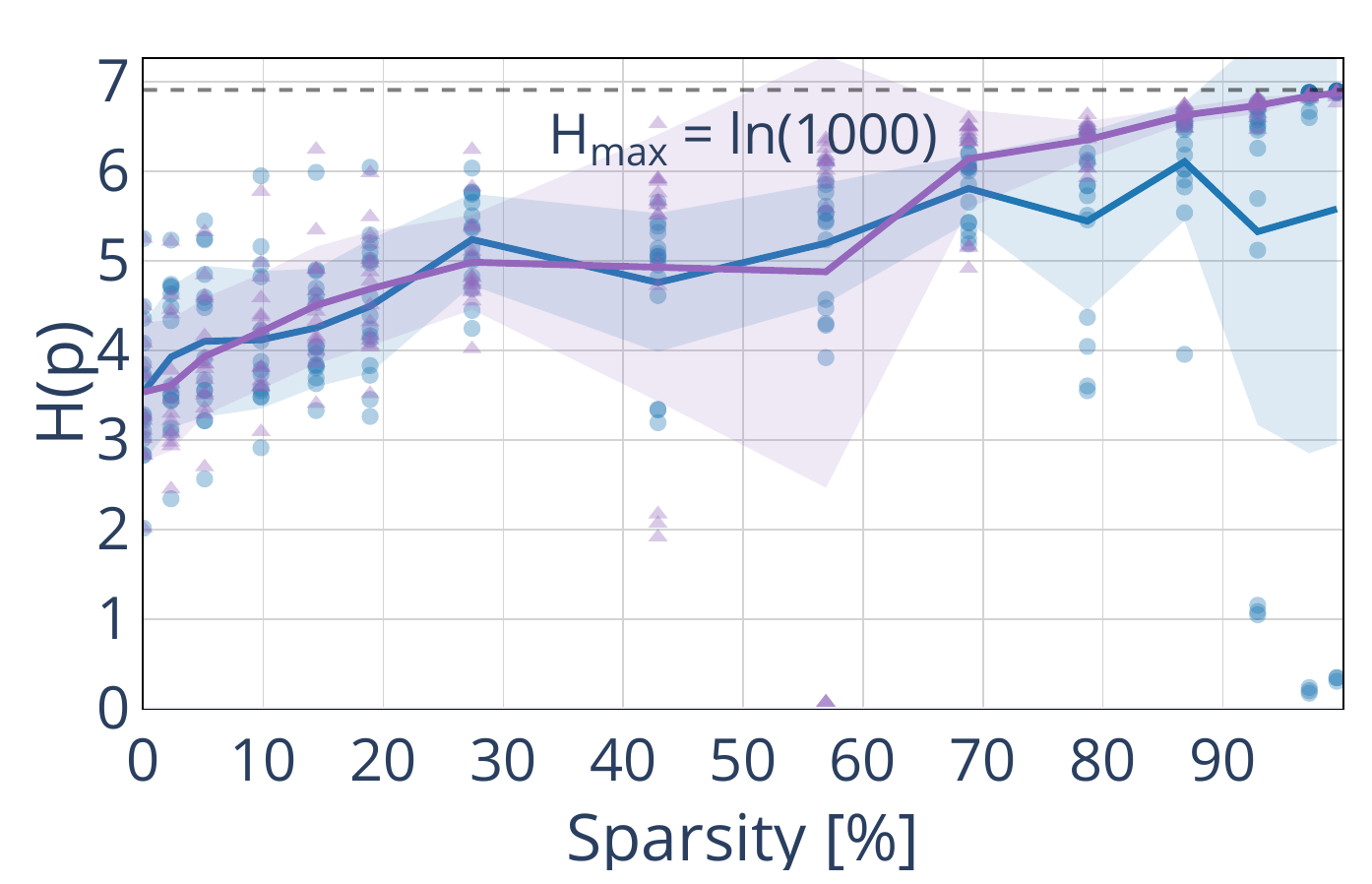}
\caption{ResNet-18 on ImageNet-C}
\label{fig:ent_free_imagenet}
\end{subfigure}
\hfill
\begin{subfigure}[t]{0.32\textwidth}
\centering
\includegraphics[width=\textwidth]{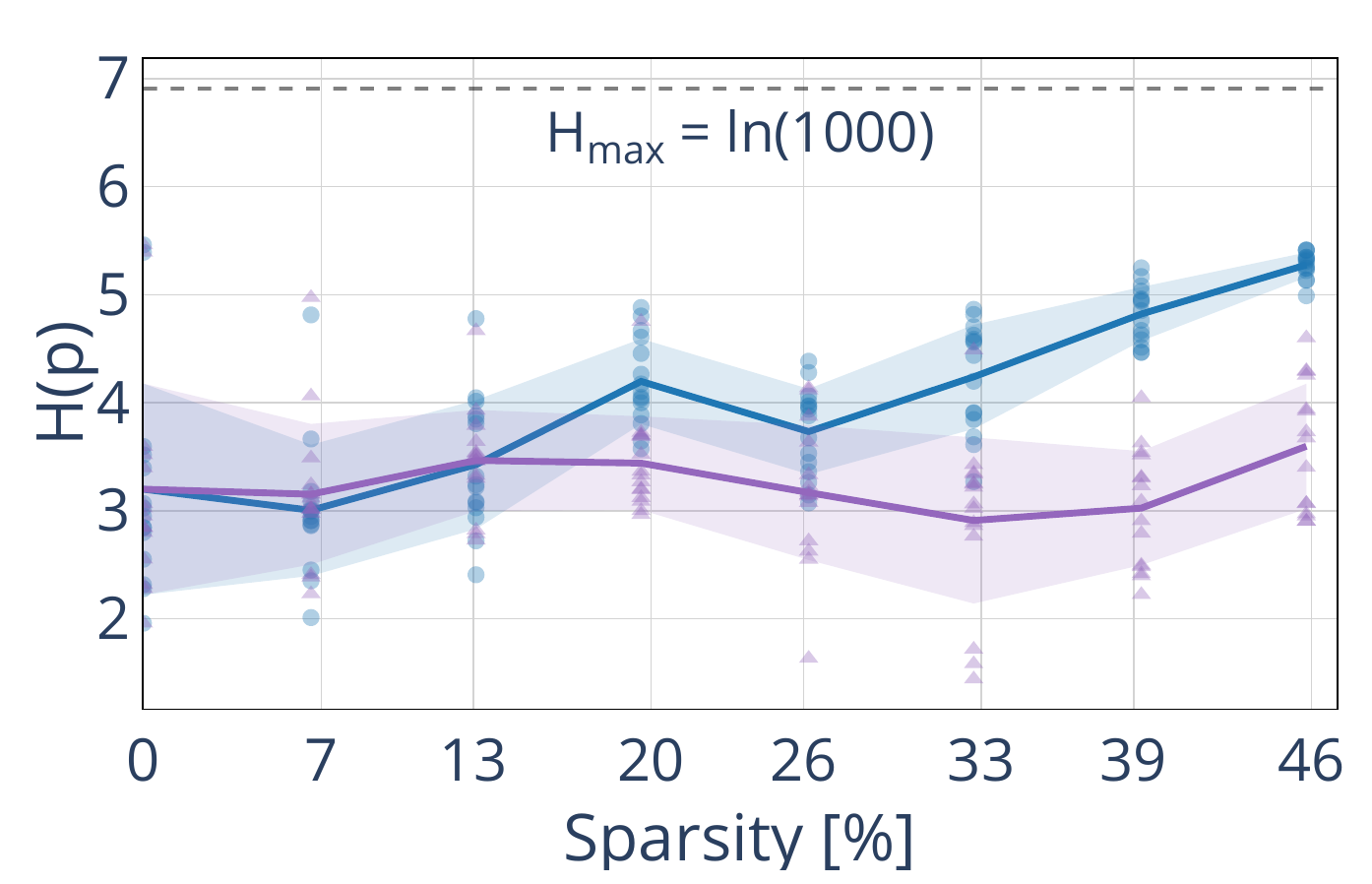}
\caption{ViT-Base on ImageNet-C}
\label{fig:ent_free_vit}
\end{subfigure}
\caption{\textbf{Prediction entropy $H(\mathbf{p})$ increases with sparsity in most settings, empirically supporting the gradient degeneracy mechanism of Eq.~\ref{eq:ent_loss} and Eq.~\ref{eq:ent_grad_bound}.} Each dot represents the mean prediction entropy $H(\mathbf{p}) = -\sum_{c} p_c \ln p_c$ of a single corruption type at severity 5; solid lines and shaded bands show the mean and $\pm \sigma$ across all corruptions. The dashed line marks $H_{max} = \ln C$ (uniform distribution). \textbf{Top row}~(a--c): data-dependent pruning (Wanda, Taylor, OBD). \textbf{Bottom row}~(d--f): data-free compression (Mag-$\ell_2$, Folding).}
\label{fig:prediction_entropy}
\end{figure*}

\paragraph{Objective-Induced Gradient Degeneracy.}
\label{sec:theory}
Even if the adaptation subspace is aligned with the target-domain loss, the \tta objective may fail to produce informative gradients under structured compression. We show that this degeneracy arises in both entropy- and consistency-based \tta objectives.

\textit{Entropy minimization.} \tta methods such as TENT~\citep{wang2021tent} and SAR~\citep{niu2023towards} optimize:
\begin{equation}
\mathcal L_{\mathrm{ent}}(p)
=
-\sum_{c=1}^{C} p_c \log p_c,
\label{eq:ent_loss}
\end{equation}
where $p=(p_1,\dots,p_C)$ is the predictive distribution.
Using $\sum_{c} p_c = 1$ and thus $\sum_{c}\nabla_{\phi} p_c = 0$, differentiating gives:
\begin{equation}
\nabla_{\phi}\mathcal L_{\mathrm{ent}} = -\sum_{c=1}^{C} (\log p_c + 1)\nabla_{\phi} p_c = -\sum_{c=1}^{C} \log(Cp_c)\,\nabla_{\phi} p_c .
\label{eq:ent_grad}
\end{equation}
When $p=(1/C,\dots,1/C)$, all coefficients $\log(Cp_c)$ vanish and $\nabla_{\phi}\mathcal L_{\mathrm{ent}} = \mathbf{0}$. More generally:
\begin{equation}
\|\nabla_{\phi}\mathcal L_{\mathrm{ent}}\|_2
\le
\left(\sum_{c=1}^{C} |\log(Cp_c)|\right)
\max_{c}\|\nabla_{\phi} p_c\|_2 .
\label{eq:ent_grad_bound}
\end{equation}
If compression pushes predictions toward a near-uniform regime, the entropy signal weakens (\emph{gradient degeneracy}), as observed on CIFAR-10-C (Fig.~\ref{fig:cs_data_cifar10},~\ref{fig:cs_free_cifar10}). Conversely, when compression preserves confident but incorrect predictions, $|\log(Cp_c)|$ remains large, producing misaligned high-norm gradients (\emph{active divergence}), as observed on ImageNet-C (Fig.~\ref{fig:cs_data_imagenet},~\ref{fig:cs_free_imagenet}). Prediction entropy measurements indicate that $H(\mathbf{p})$ increasingly approaches $H_{max}$ with higher sparsity in most settings (Fig.~\ref{fig:prediction_entropy}), further supporting the gradient degeneracy mechanism under compression.

\textit{KL-consistency.} 
SPA~\citep{niu2025self} minimizes the KL divergence between a detached clean prediction $q = \mathrm{softmax}(f(x;\phi))$ and an augmented-view prediction $p_{\mathrm{aug}} = \mathrm{softmax}(g(x_{\mathrm{aug}};\phi))$:
\begin{equation}
\mathcal L_{\mathrm{KL}} = \sum_{c=1}^{C} q_c \log \frac{q_c}{p_{\mathrm{aug},c}}.
\label{eq:kl_loss}
\end{equation}
Since $q$ is detached, $\nabla_{\phi}\mathcal L_{\mathrm{KL}} = -\sum_{c} (q_c / p_{\mathrm{aug},c})\,\nabla_{\phi} p_{\mathrm{aug},c}$. When $q \approx p_{\mathrm{aug}}$, each ratio $q_c/p_{\mathrm{aug},c} \approx 1$, and together with $\sum_c \nabla_\phi p_{\mathrm{aug},c} = \mathbf{0}$ this implies that the gradient vanishes. Compression triggers this by driving both views toward uniform predictions.

\textit{Supervised cross-entropy (Oracle).} 
By contrast, the supervised gradient for a labeled example $(x,y)$:
\begin{equation}
\nabla_{\phi}\mathcal L_{\mathrm{CE}} = -\frac{1}{p_y}\nabla_{\phi} p_y,
\label{eq:ce_grad}
\end{equation}
concentrates on a single direction scaled by $1/p_y$, which grows as confidence drops. Even when compression degrades predictions, supervised updates thus remain informative where entropy or consistency signals vanish. Together, Eqs.~\ref{eq:ent_grad}--\ref{eq:ce_grad} reveal two complementary mechanisms behind the post-compression Oracle-\tta gap: (i)~compression removes or misaligns parameter directions useful for target-domain adaptation, and (ii)~unsupervised \tta objectives simultaneously lose their gradient signal as predictions approach an uninformative regime.

\section{Discussion, Outlook and Limitations}

This paper analyzes the interaction between structured compression and test-time adaptation, a combination that is common in edge deployment but has received limited attention. Our study reveals that the standard compress-and-deploy pipeline can induce \emph{silent plasticity loss}: compressed models that preserve source-domain accuracy yet progressively lose the ability to adapt at test time via unsupervised objectives. This finding extends prior observations that compression harms robustness under distribution shift~\citep{liebenwein2021lost} and disproportionately affects underrepresented subgroups~\citep{hooker2020characterising}, by showing that degradation extends to adaptability. Our theoretical analysis establishes that the gradients of both entropy-based and consistency-based \tta objectives can degenerate under compression, while the supervised cross-entropy gradient remains more informative at the same sparsity (Eq.~\ref{eq:ent_grad}, Eq.~\ref{eq:ce_grad}). 
 
\textbf{Practical guidelines.} Our results suggest two actionable considerations. First, the compression criterion should be selected for adaptability, not source accuracy alone; data-dependent criteria (Wanda, Taylor) preserve higher representational recoverability (Fig.~\ref{fig:cka_main}) and gradient alignment (Fig.~\ref{fig:cosine_similarity}) than Mag-$\ell_2$ and OBD, while Folding retains more adaptability among data-free methods. Second, the compression budget and \tta objective should be co-designed: on CIFAR-10-C, gradient alignment reverses sign between 45\% and 65\% sparsity for the pruning criteria (near 75\% for Folding); on ImageNet-C, entropy-based objectives report active divergence already at minimal sparsity, whereas consistency-based objectives maintain higher alignment across all analyzed sparsities. 

\textbf{Outlook.} There are several directions to extend this work. First, although our empirical evaluation covers both backpropagation-based and backpropagation-free \tta approaches (FOA~\citep{niu2024test}, PEA~\citep{xiao2026architectureagnostic}), our diagnostics are mainly gradient-based and do not yet cover backpropagation-free adaptation methods; we leave this as an open direction. The corresponding accuracy curves are reported in Fig.~\ref{fig:bp_free_hero}: the benefit of backpropagation-free adaptation likewise collapses toward the no-adaptation baseline as compression increases. Second, designing compression-aware \tta objectives that preserve adaptability under compression is left for future work. Third, extending this analysis to unstructured pruning, quantization, and their combinations would provide a more complete picture of the compression-adaptation interaction.

\textbf{Limitations.} Our study considers two architectures (ResNet-18 and ViT-Base) and two (one backpropagation-based, one backpropagation-free) \tta methods per architecture, covering both entropy- and consistency-based objectives; the generalization to larger models (\eg large language models) remains for future work. While the experiments are primarily conducted at maximum corruption severity (\ie 5), we additionally report results at intermediate severity 3 (Fig.~\ref{fig:cka_dense_prune_tta_sev3}) and across multiple seeds (Fig.~\ref{fig:hero_multiseed}). We hope this work sparks interest in designing structured compression strategies that explicitly preserve adaptability, narrowing the gap between efficient deployment and test-time adaptation.

\section*{Acknowledgments}
This research was funded in part by the Austrian Science Fund (FWF) within the DENISE doctoral school (grant DOI: 10.55776/DFH5) and the FFG COMET K1 Center ``Pro\textsuperscript{2}Future II'' (Cognitive and Sustainable Products and Production Systems of the Future), Contract No.~911655. The results presented in this paper were computed using the computational resources of the HLR Zentralen Informatikdienstes of Graz University of Technology and the Austrian Scientific Computing (ASC) infrastructure.

\bibliography{bibliography}
\bibliographystyle{collas2026_conference}

\newpage

\appendix
\section{Appendix}
\label{sec:appendix}

\begin{figure*}[h]
\centering
\begin{subfigure}[t]{0.32\textwidth}
\centering
\includegraphics[width=\textwidth]{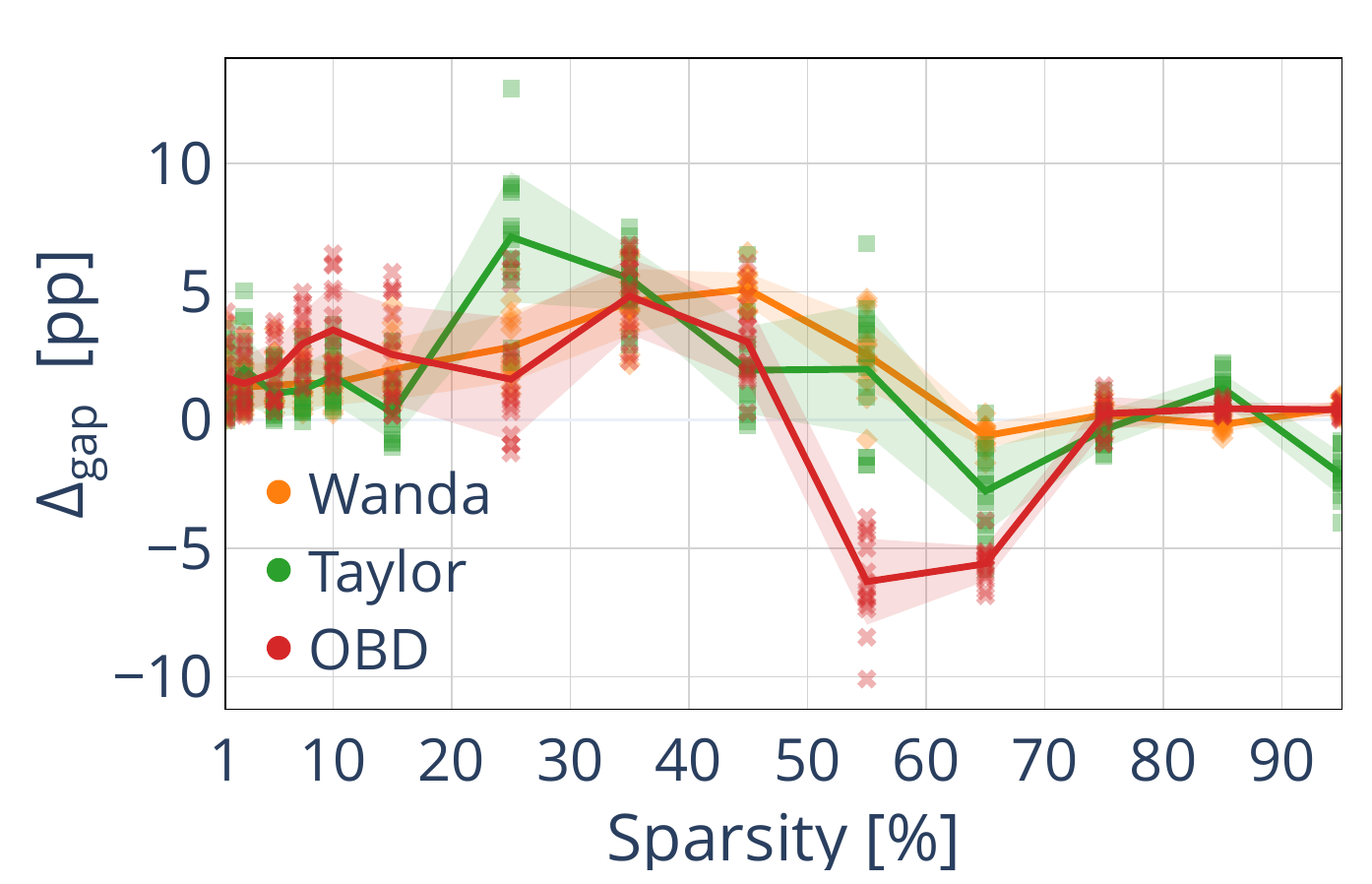}
\caption{ResNet-18 on CIFAR-10-C}
\label{fig:gap_data_cifar10}
\end{subfigure}
\hfill
\begin{subfigure}[t]{0.32\textwidth}
\centering
\includegraphics[width=\textwidth]{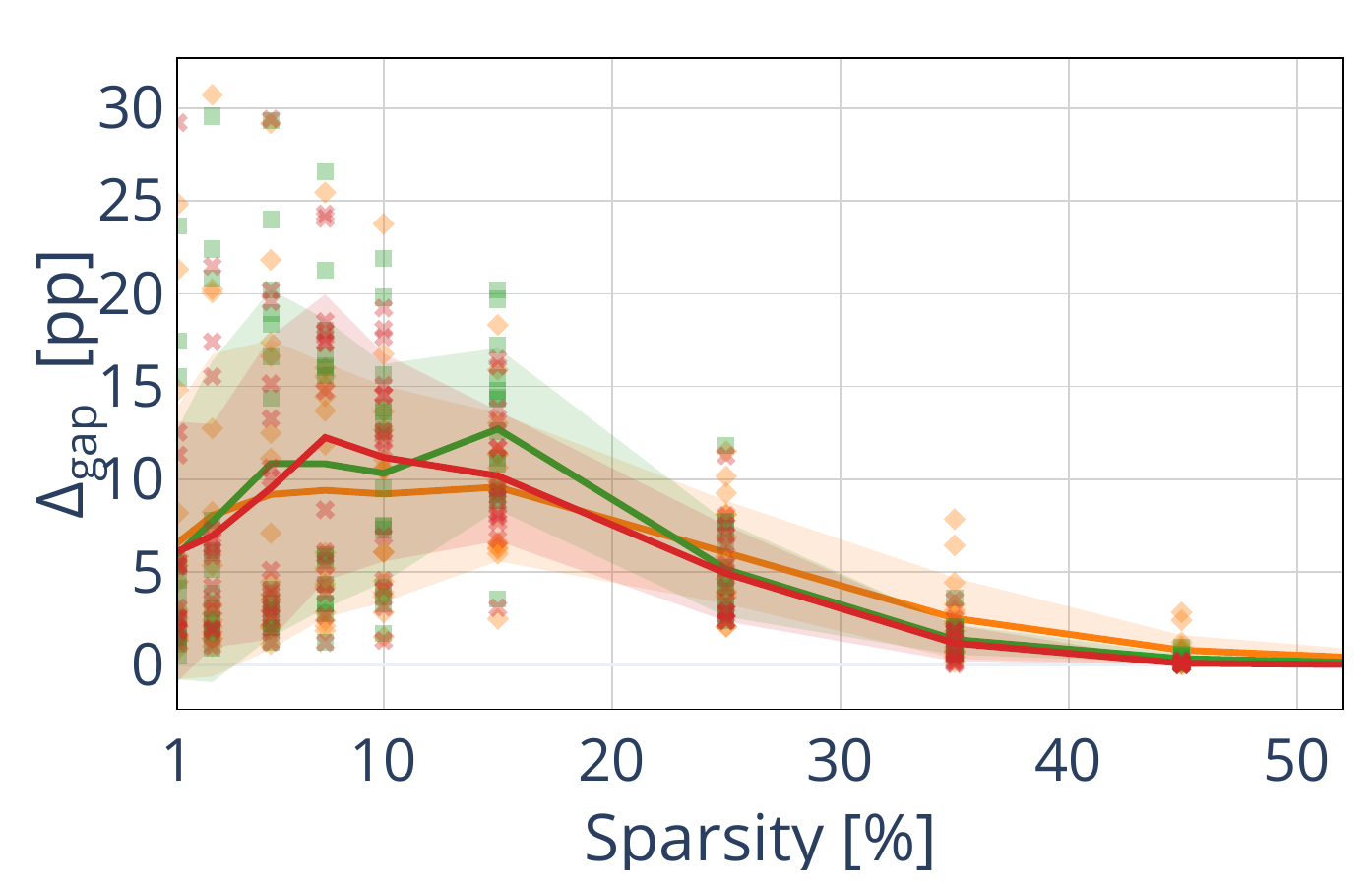}
\caption{ResNet-18 on ImageNet-C}
\label{fig:gap_data_imagenet}
\end{subfigure}
\hfill
\begin{subfigure}[t]{0.32\textwidth}
\centering
\includegraphics[width=\textwidth]{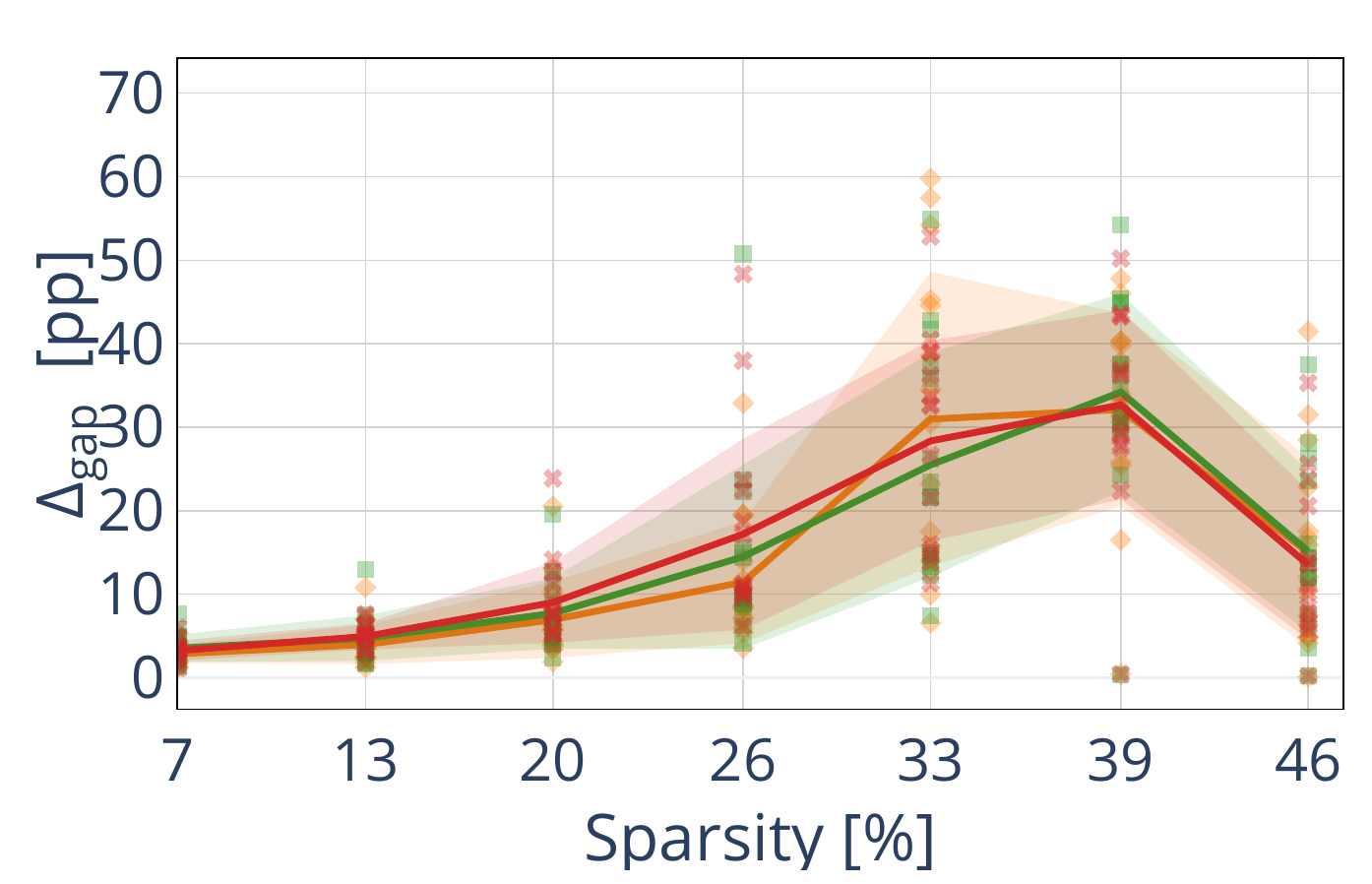}
\caption{ViT-Base on ImageNet-C}
\label{fig:gap_data_vit}
\end{subfigure}

\vspace{0.3em}

\begin{subfigure}[t]{0.32\textwidth}
\centering
\includegraphics[width=\textwidth]{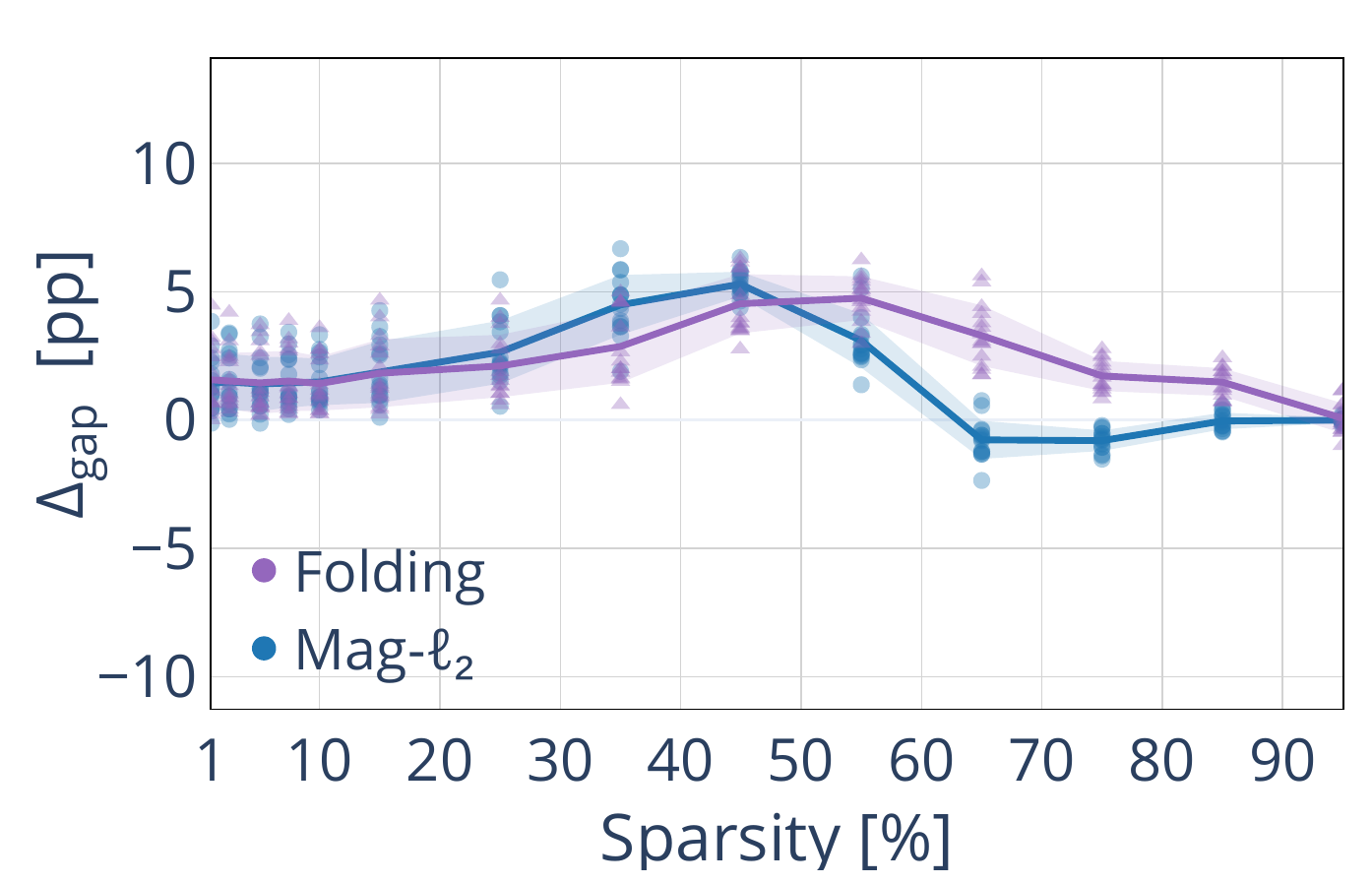}
\caption{ResNet-18 on CIFAR-10-C}
\label{fig:gap_free_cifar10}
\end{subfigure}
\hfill
\begin{subfigure}[t]{0.32\textwidth}
\centering
\includegraphics[width=\textwidth]{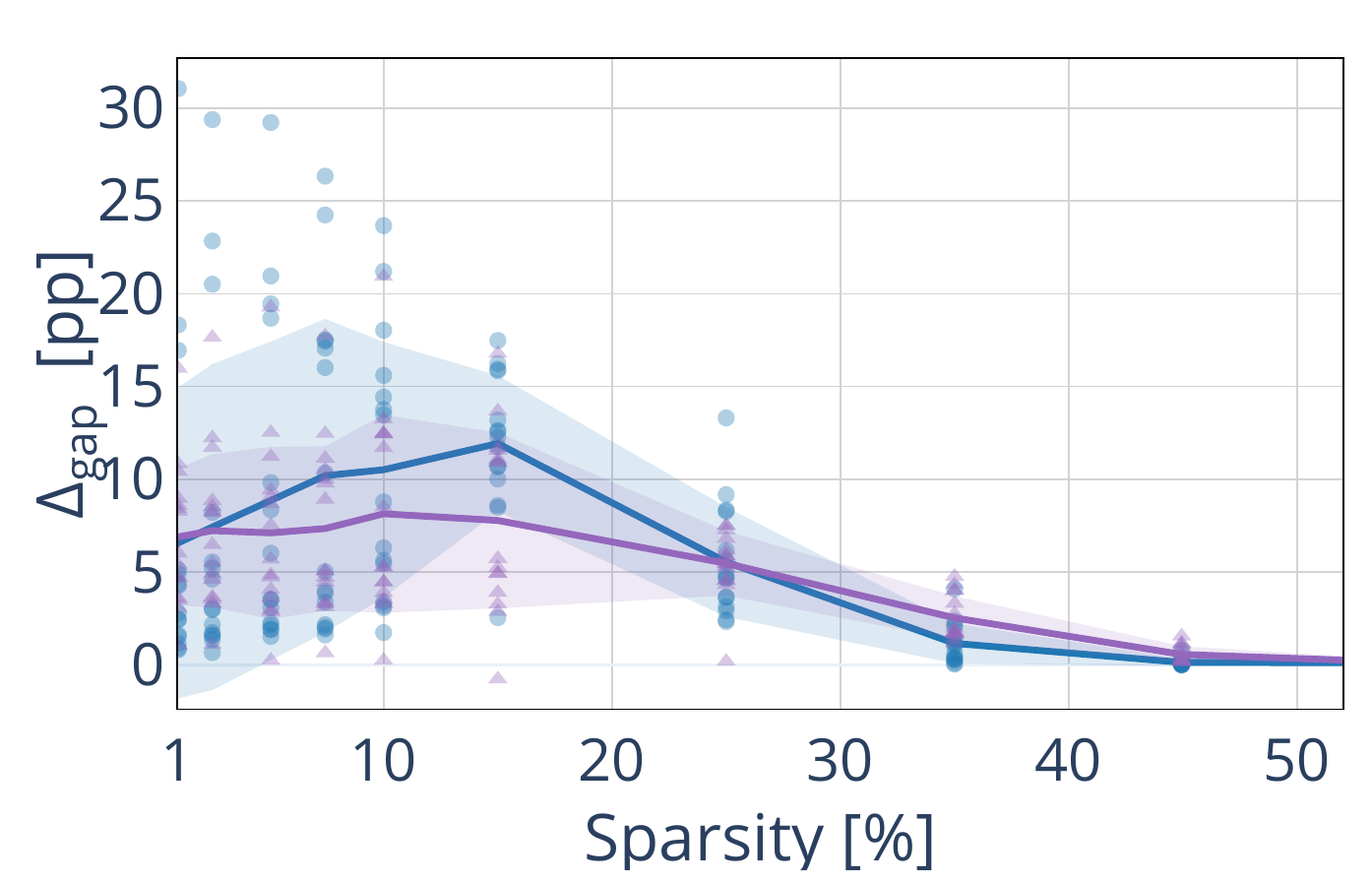}
\caption{ResNet-18 on ImageNet-C}
\label{fig:gap_free_imagenet}
\end{subfigure}
\hfill
\begin{subfigure}[t]{0.32\textwidth}
\centering
\includegraphics[width=\textwidth]{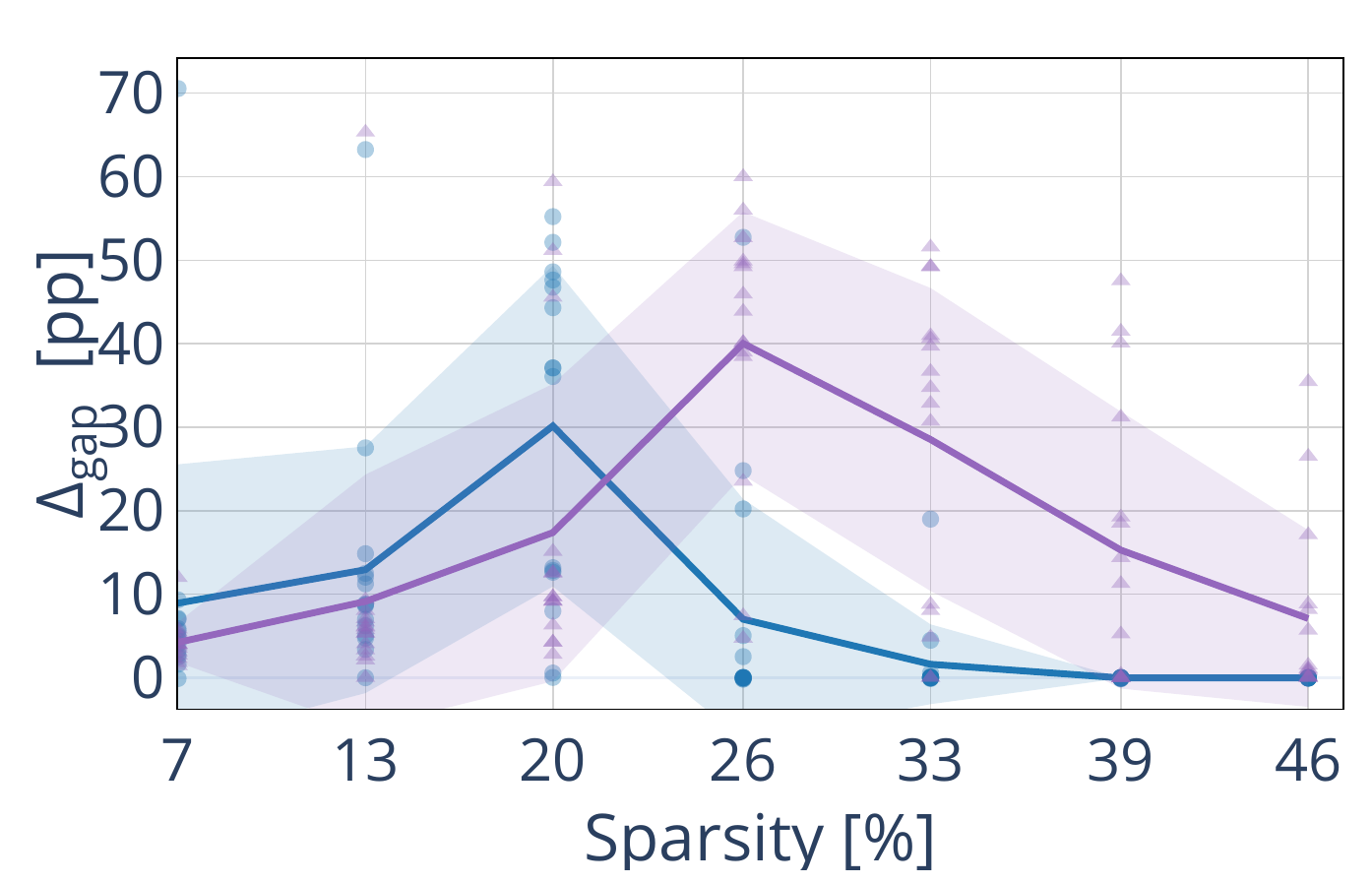}
\caption{ViT-Base on ImageNet-C}
\label{fig:gap_free_vit}
\end{subfigure}
\caption{\textbf{The per-corruption adaptation gap $\Delta_{\mathrm{gap}} = (\text{Oracle} - \text{TTA})_{\text{compressed}}$ generally widens with increasing compression, providing empirical support for \emph{silent plasticity loss}.} Each dot represents the adaptation gap for a single corruption type at a given sparsity level; solid lines and shaded bands show the mean and $\pm \sigma$ across all corruptions. \emph{Top row} (a--c): data-dependent pruning criteria (Wanda, Taylor, OBD). \emph{Bottom row} (d--f): data-free compression criteria (Folding, Mag-$\ell_2$).}
\label{fig:gap_scatter}
\end{figure*}

\begin{figure*}[h]
\centering

\begin{subfigure}[t]{0.48\textwidth}
\centering
\includegraphics[width=\textwidth]{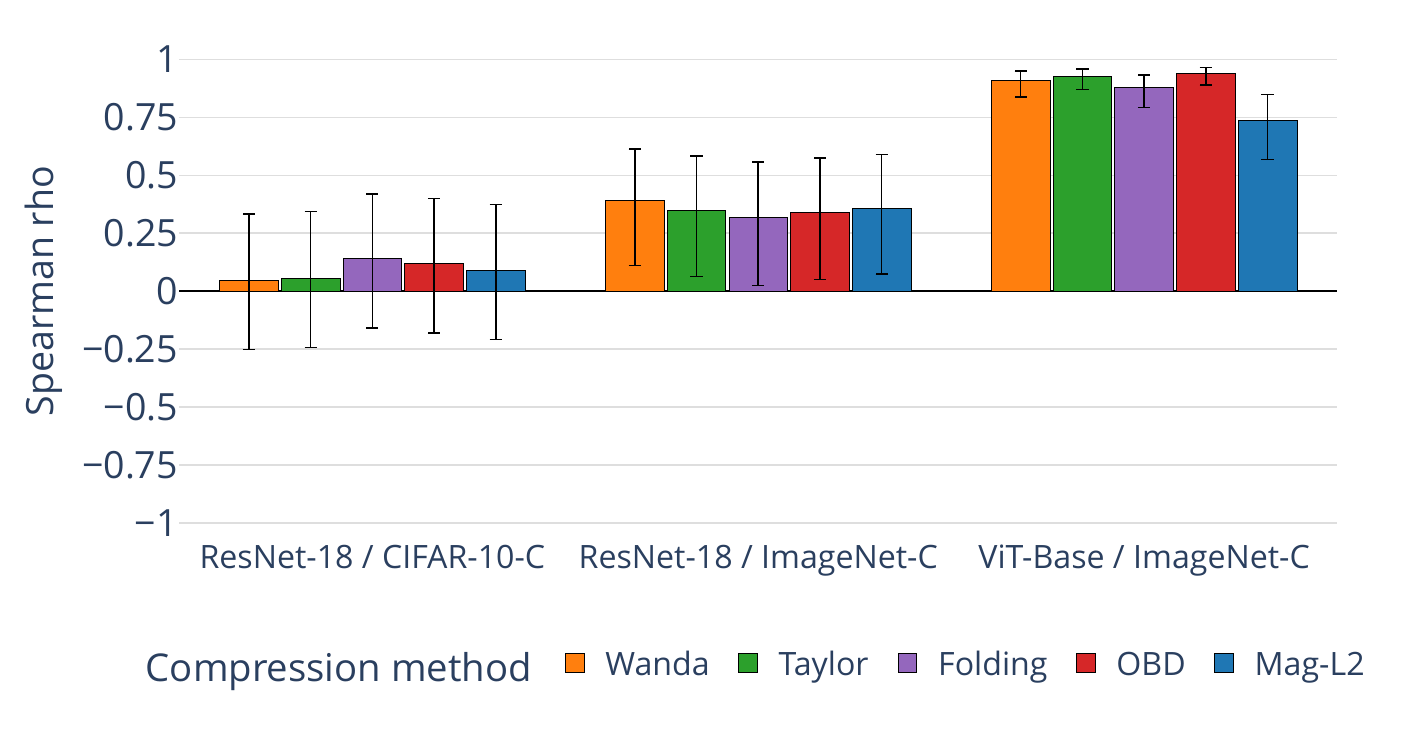}
\caption{Worst-layer \cka (severity~3).}
\label{fig:pearson_cka_sev3}
\end{subfigure}
\hfill
\begin{subfigure}[t]{0.48\textwidth}
\centering
\includegraphics[width=\textwidth]{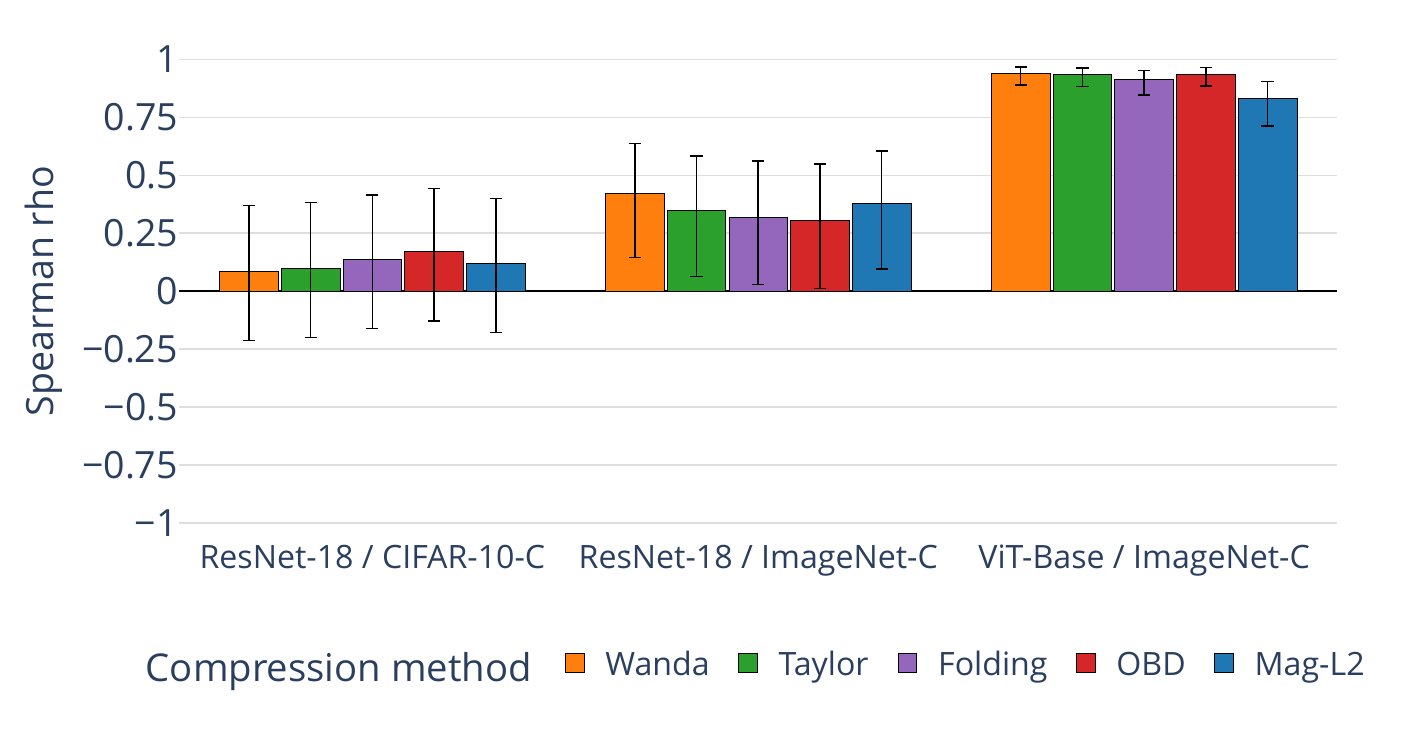}
\caption{Worst-layer \cka (severity~5).}
\label{fig:pearson_cka_sev5}
\end{subfigure}

\caption{Before adaptation, the worst-layer \cka of the compressed model is positively correlated with its post-adaptation accuracy, strongly on ViT-Base/ImageNet-C and only weakly on ResNet-18. Each bar reports the Spearman correlation $\rho$ between the pre-\tta worst-layer \cka at $\phi_0$ and the post-adaptation accuracy of SAR ResNet-18 or SPA ViT-Base over all corruptions. Black error bars are $95\%$ confidence intervals. \emph{Left to right}: severity~$3$, severity~$5$ corruptions level.}
\label{fig:cka_spearman_appendix}
\end{figure*} 

\begin{figure*}[t]
\centering
\begin{subfigure}[t]{0.32\textwidth}
\centering
\includegraphics[width=\textwidth]{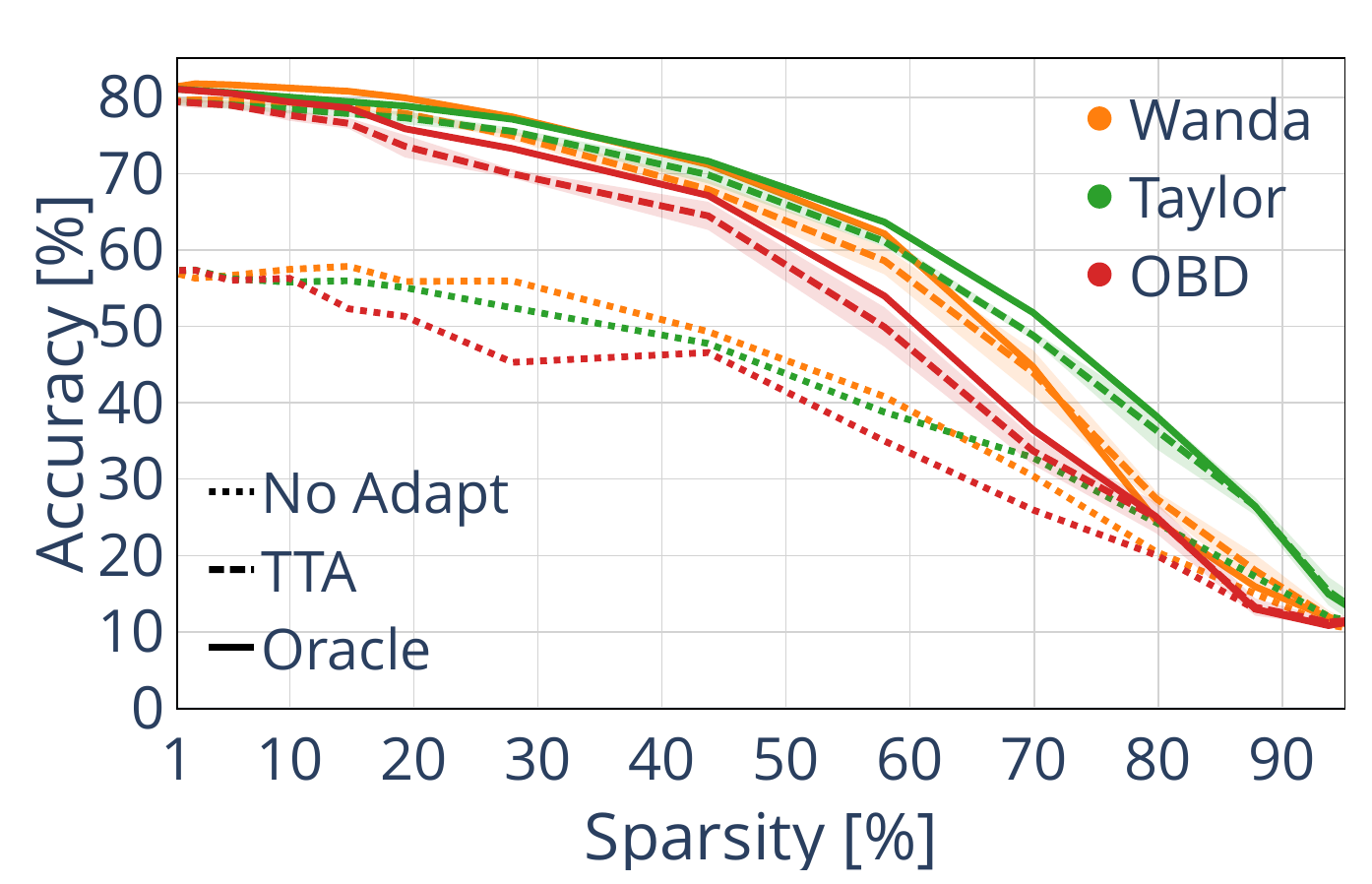}
\caption{ResNet-18 on CIFAR-10-C}
\label{fig:hero_data_cifar10}
\end{subfigure}
\hfill
\begin{subfigure}[t]{0.32\textwidth}
\centering
\includegraphics[width=\textwidth]{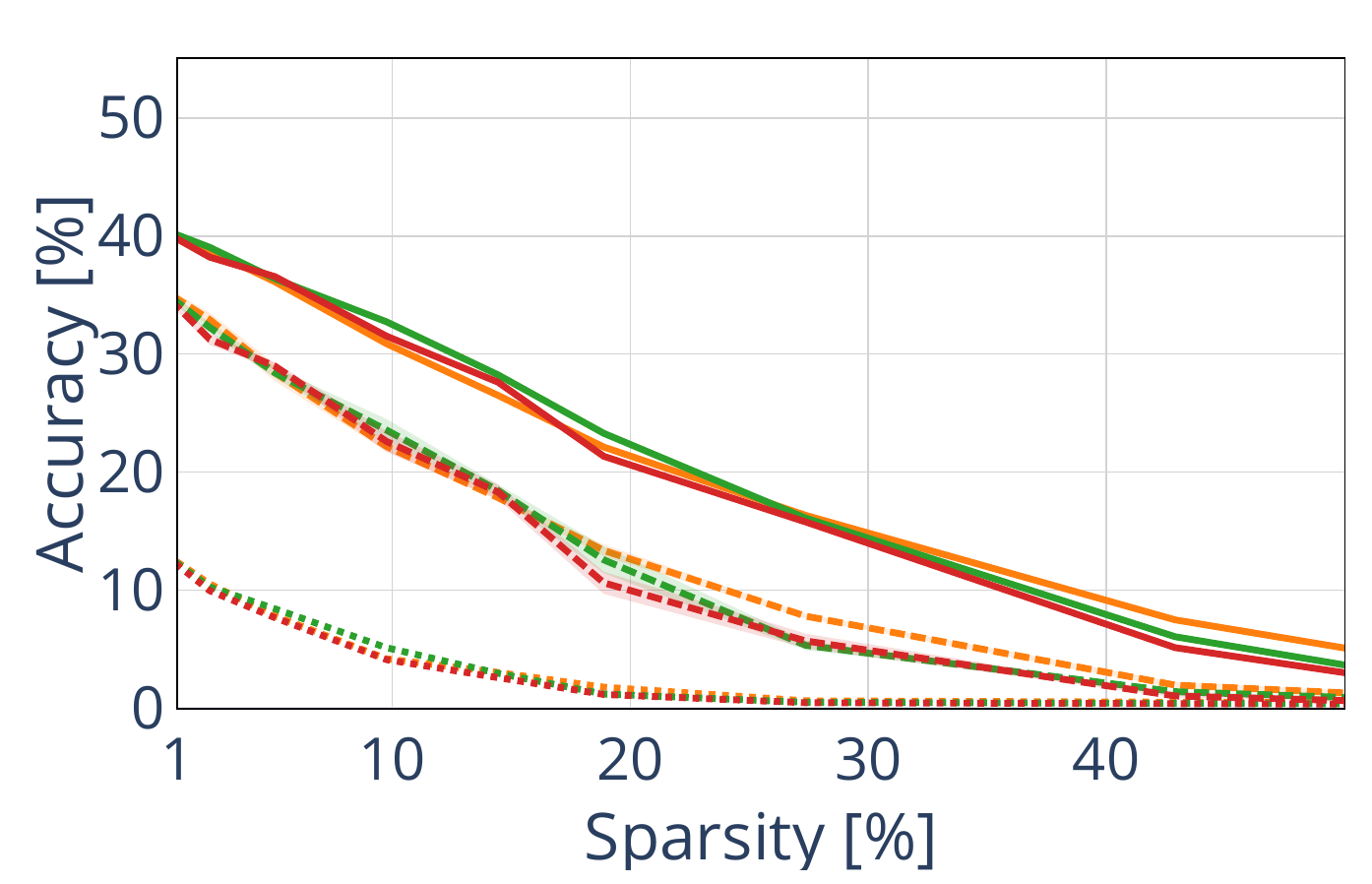}
\caption{ResNet-18 on ImageNet-C}
\label{fig:hero_data_imagenet_resnet}
\end{subfigure}
\hfill
\begin{subfigure}[t]{0.32\textwidth}
\centering
\includegraphics[width=\textwidth]{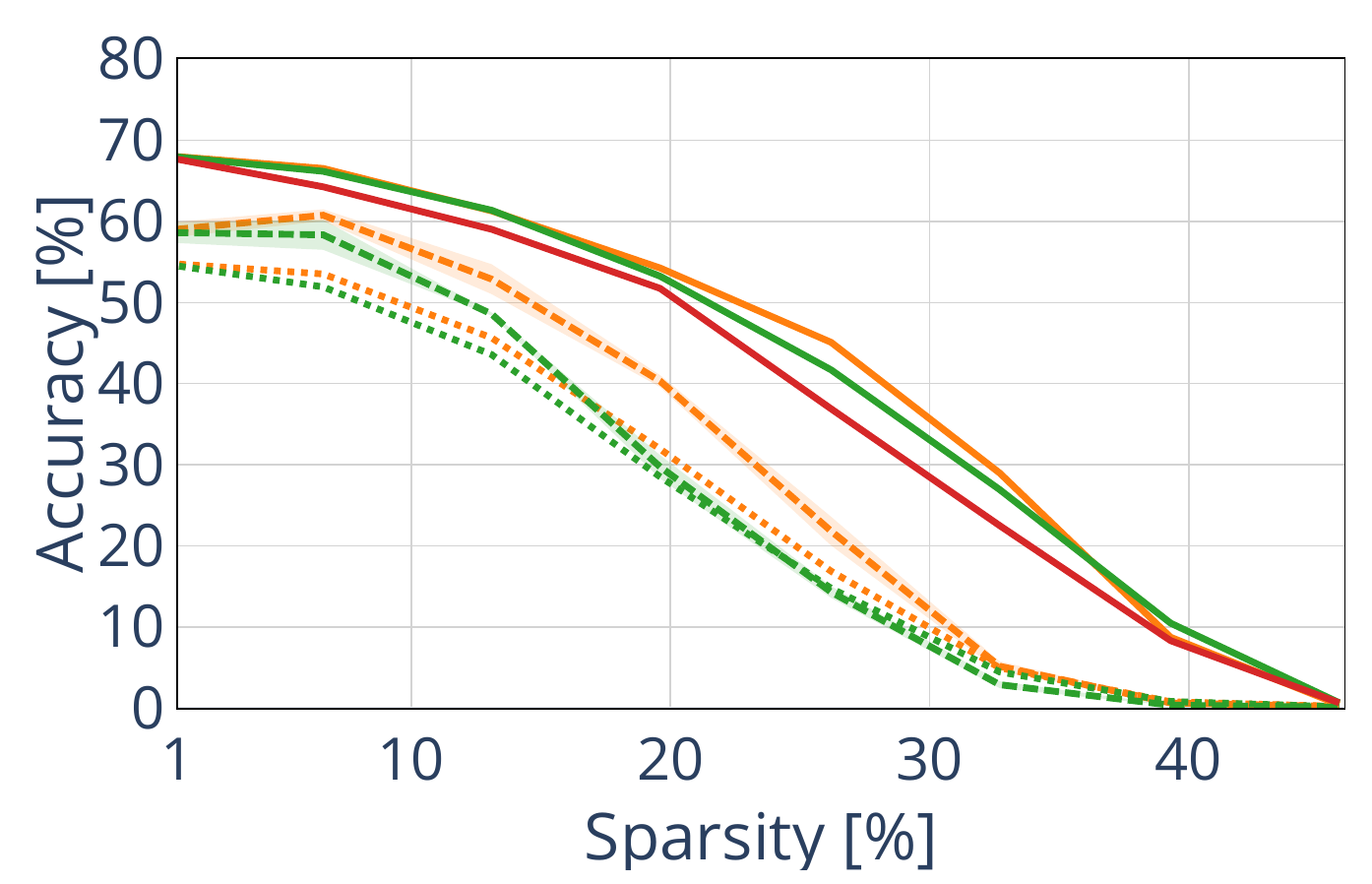}
\caption{ViT-Base on ImageNet-C}
\label{fig:hero_data_imagenet_vit}
\end{subfigure}

\vspace{0.3em}

\begin{subfigure}[t]{0.32\textwidth}
\centering
\includegraphics[width=\textwidth]{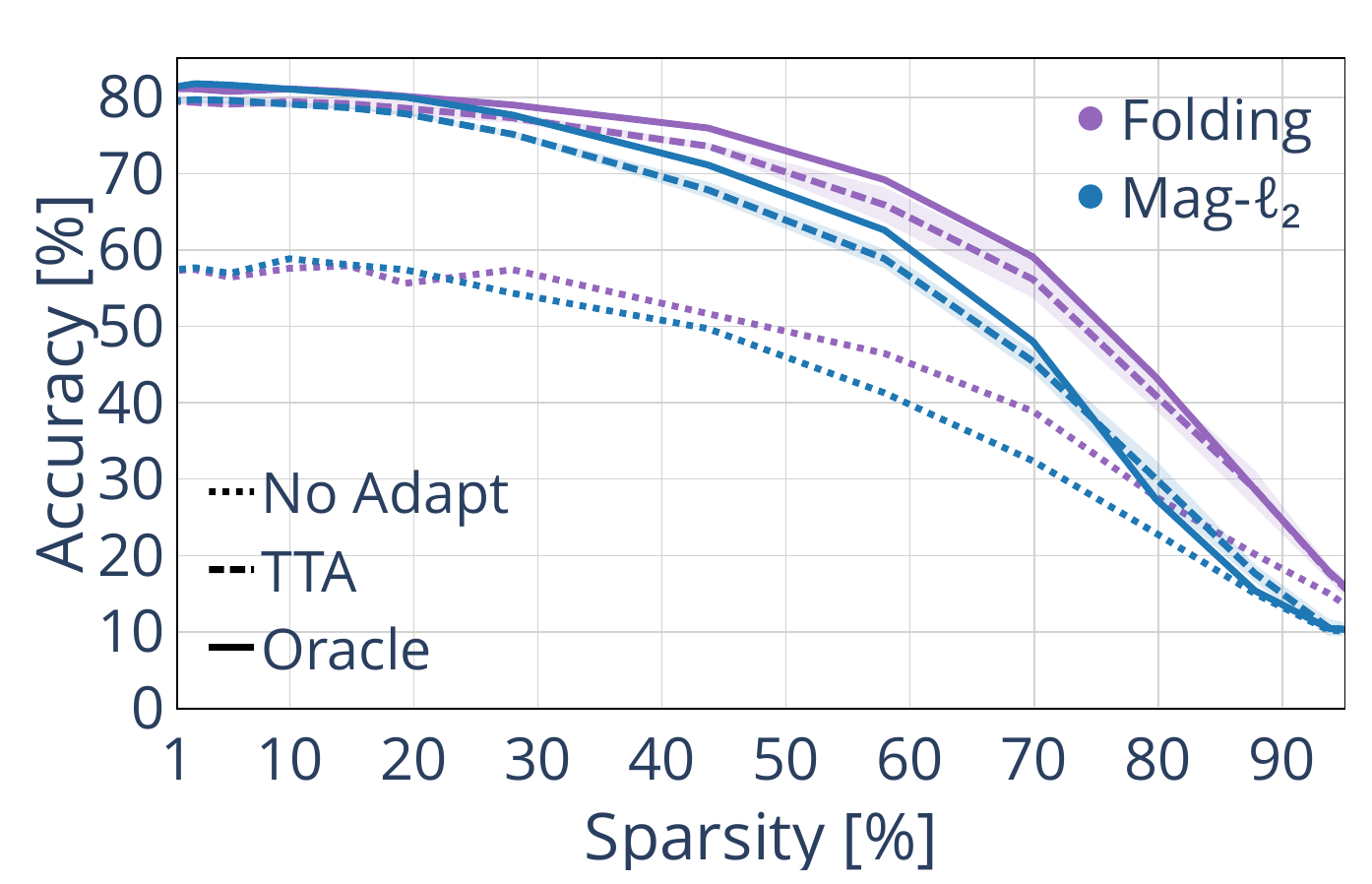}
\caption{ResNet-18 on CIFAR-10-C}
\label{fig:hero_free_cifar10}
\end{subfigure}
\hfill
\begin{subfigure}[t]{0.32\textwidth}
\centering
\includegraphics[width=\textwidth]{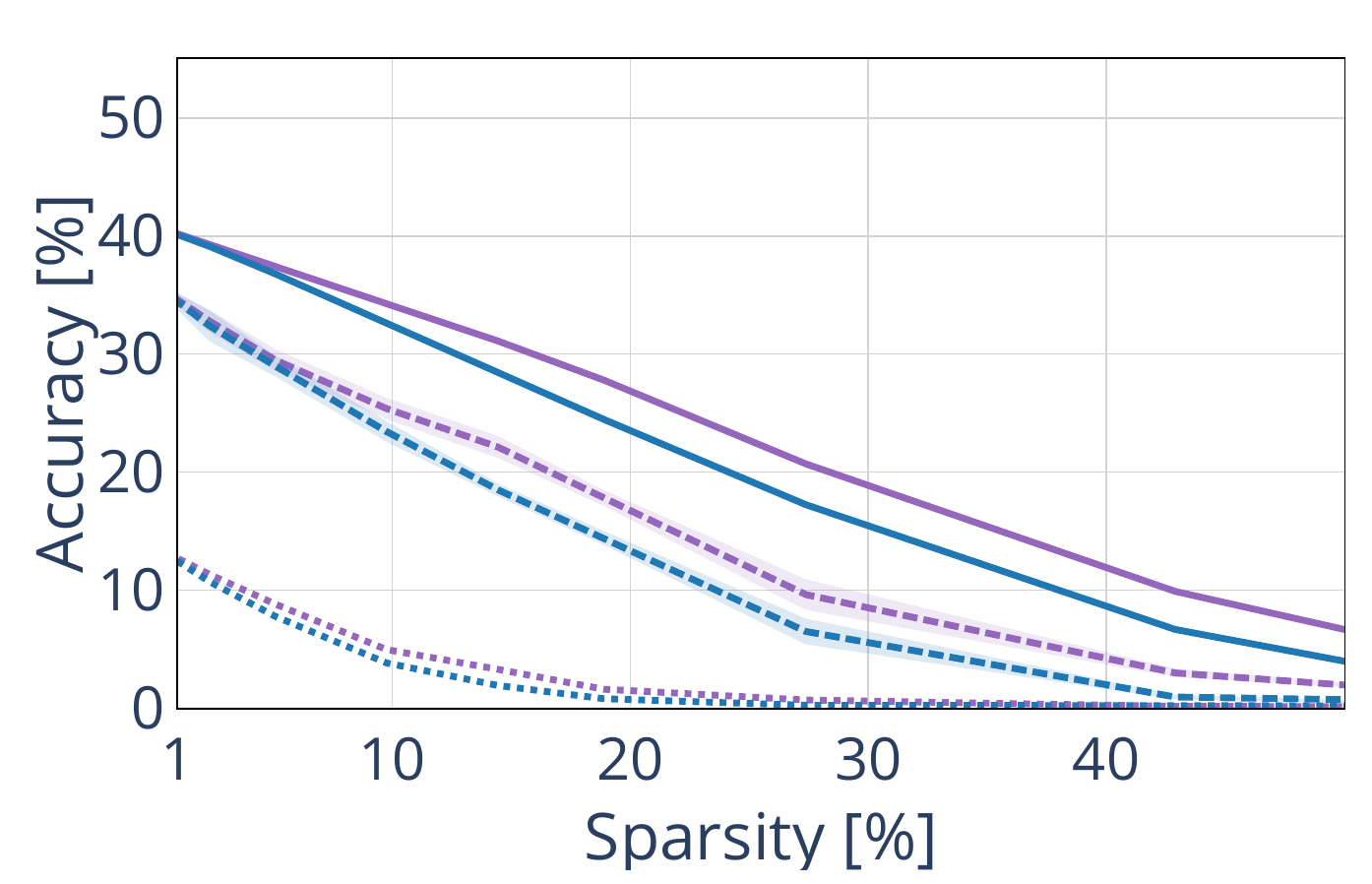}
\caption{ResNet-18 on ImageNet-C}
\label{fig:hero_free_imagenet_resnet}
\end{subfigure}
\hfill
\begin{subfigure}[t]{0.32\textwidth}
\centering
\includegraphics[width=\textwidth]{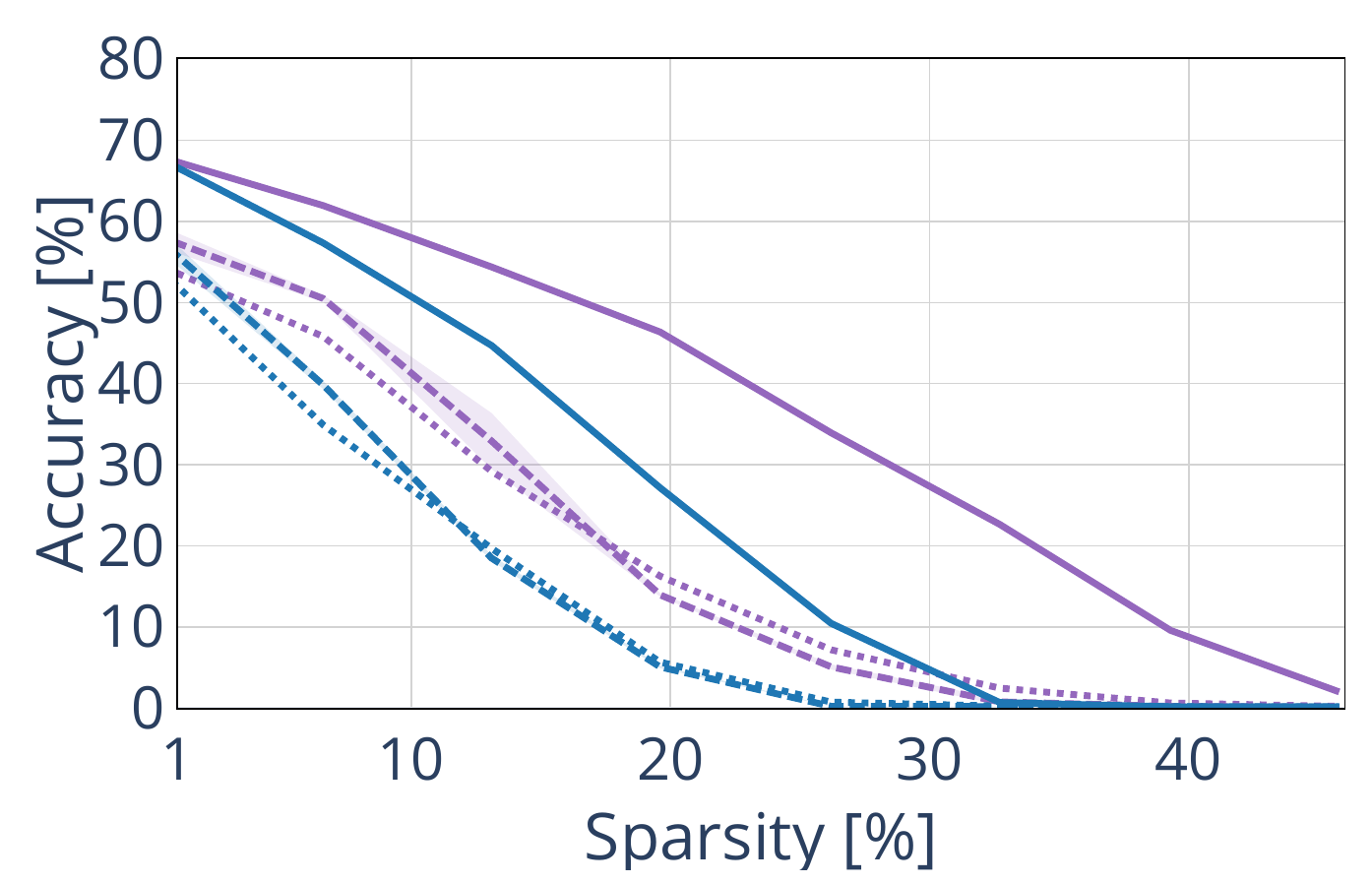}
\caption{ViT-Base on ImageNet-C}
\label{fig:hero_free_imagenet_vit}
\end{subfigure}
\caption{\textbf{Multi-seed gap between test-time adaptation (SAR) and the matched supervised baseline (Oracle-SAR), with SAR applied to both architectures.} SAR and Oracle-SAR share the same optimization settings and differ only in the loss definitions. Shaded regions indicate $\pm\sigma$ across three random seeds of the per-seed corruption-averaged accuracy. Same layout as Fig.~\ref{fig:hero}.
}
\label{fig:hero_multiseed}
\end{figure*}

\begin{figure*}[t]
\centering
\begin{subfigure}[t]{0.32\textwidth}
\centering
\includegraphics[width=\textwidth]{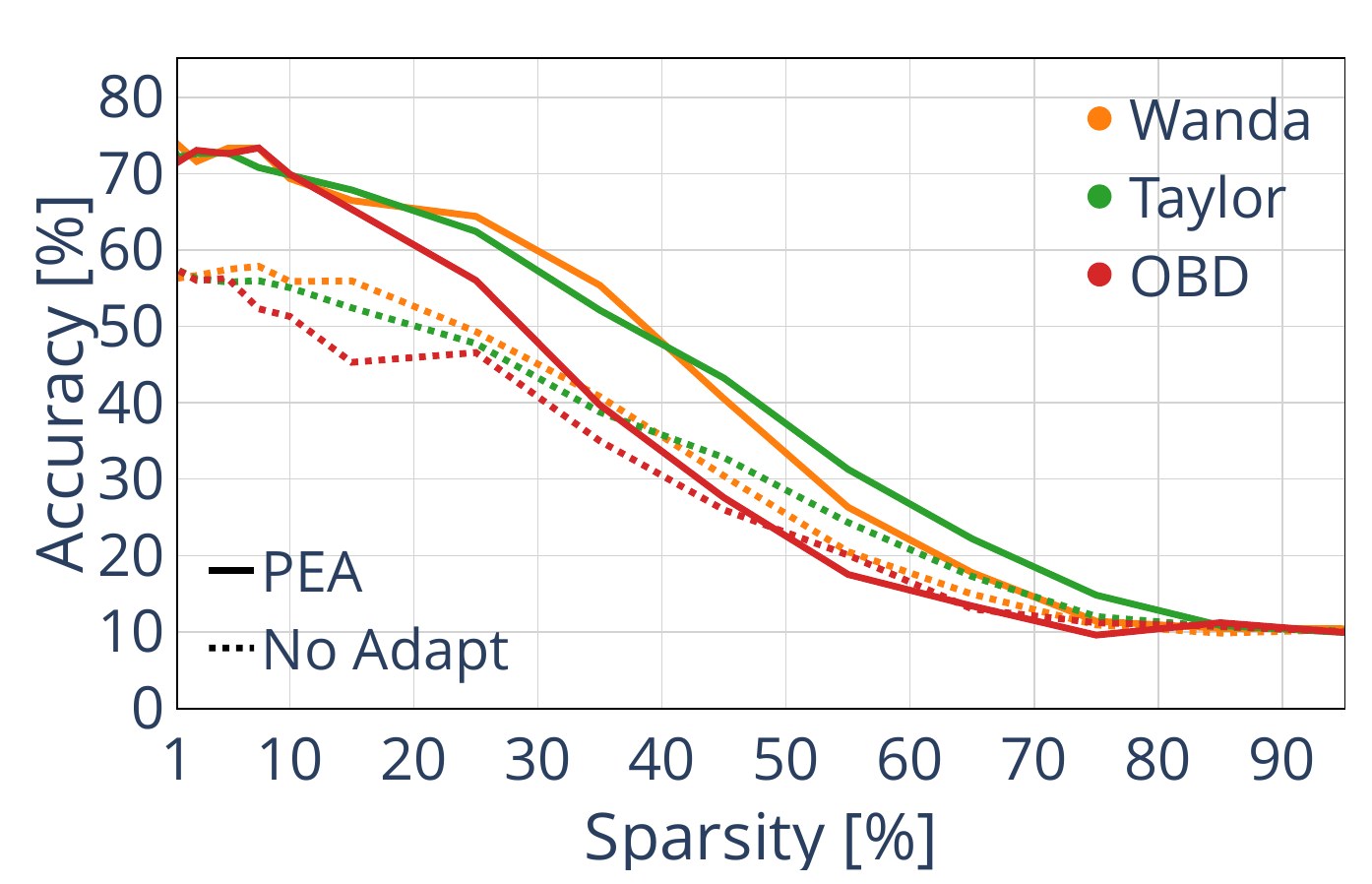}
\caption{ResNet-18 on CIFAR-10-C}
\label{fig:bp_free_data_cifar10}
\end{subfigure}
\hfill
\begin{subfigure}[t]{0.32\textwidth}
\centering
\includegraphics[width=\textwidth]{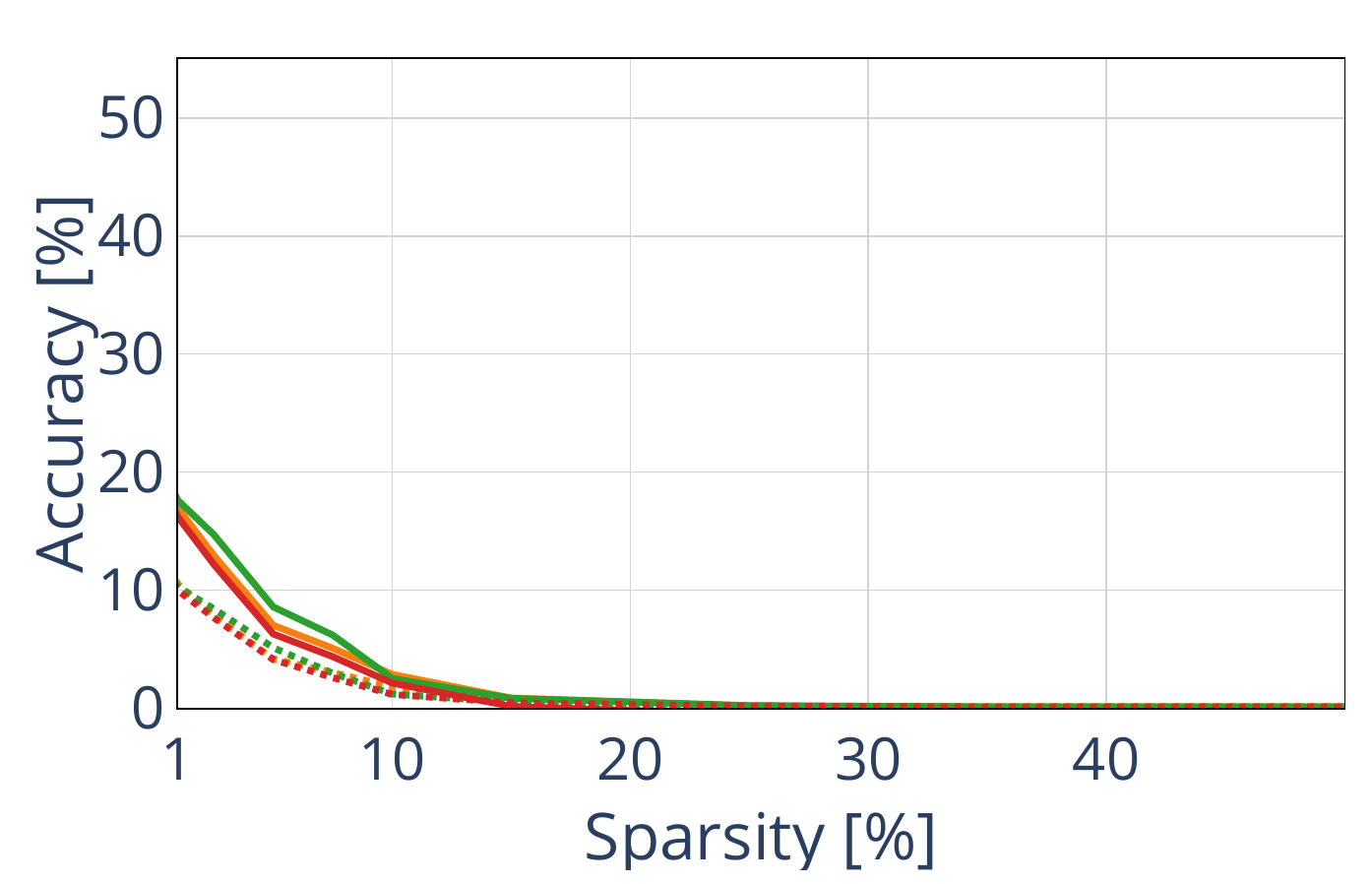}
\caption{ResNet-18 on ImageNet-C}
\label{fig:bp_free_data_imagenet_resnet}
\end{subfigure}
\hfill
\begin{subfigure}[t]{0.32\textwidth}
\centering
\includegraphics[width=\textwidth]{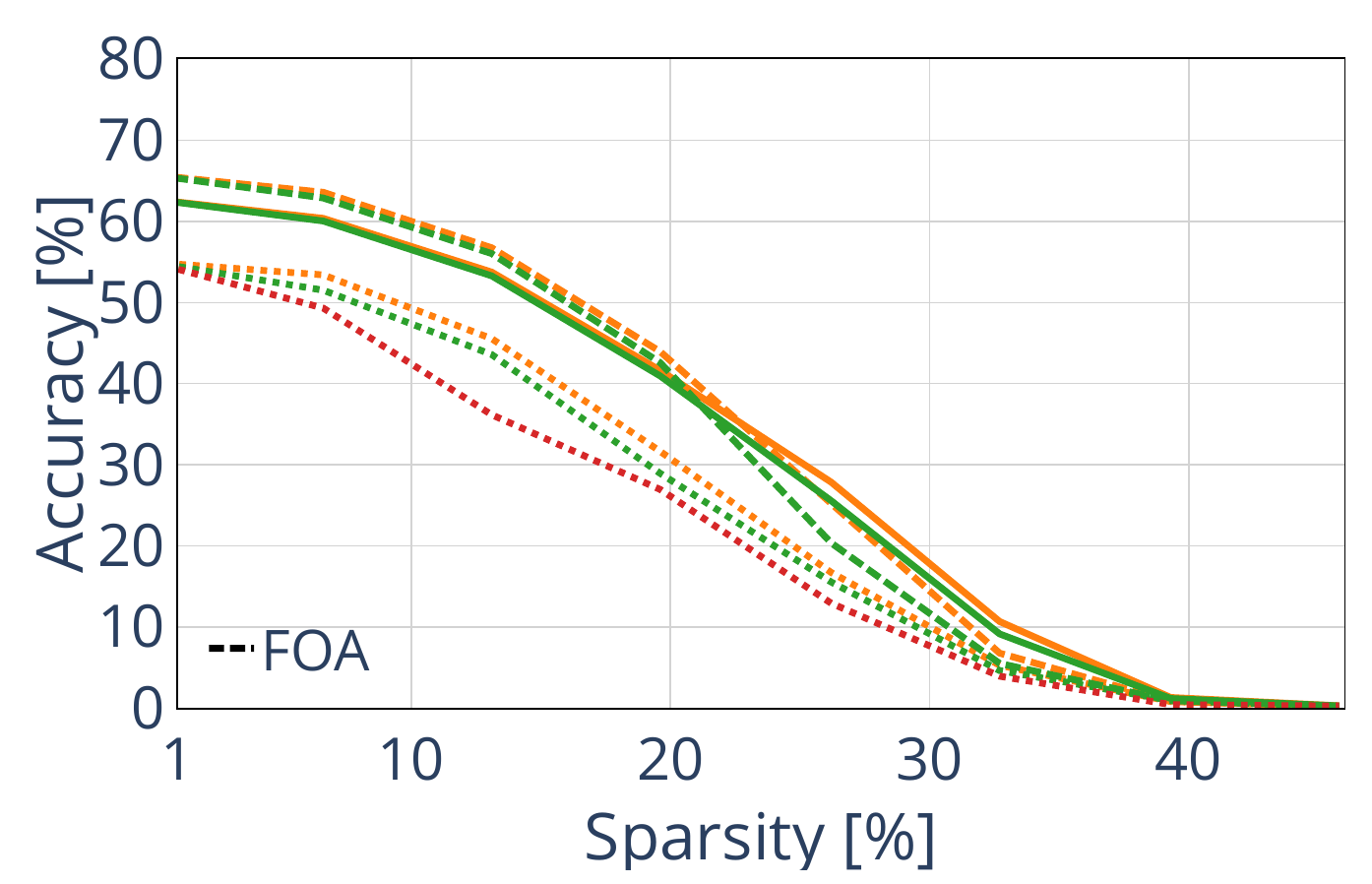}
\caption{ViT-Base on ImageNet-C}
\label{fig:bp_free_data_imagenet_vit}
\end{subfigure}

\vspace{0.3em}

\begin{subfigure}[t]{0.32\textwidth}
\centering
\includegraphics[width=\textwidth]{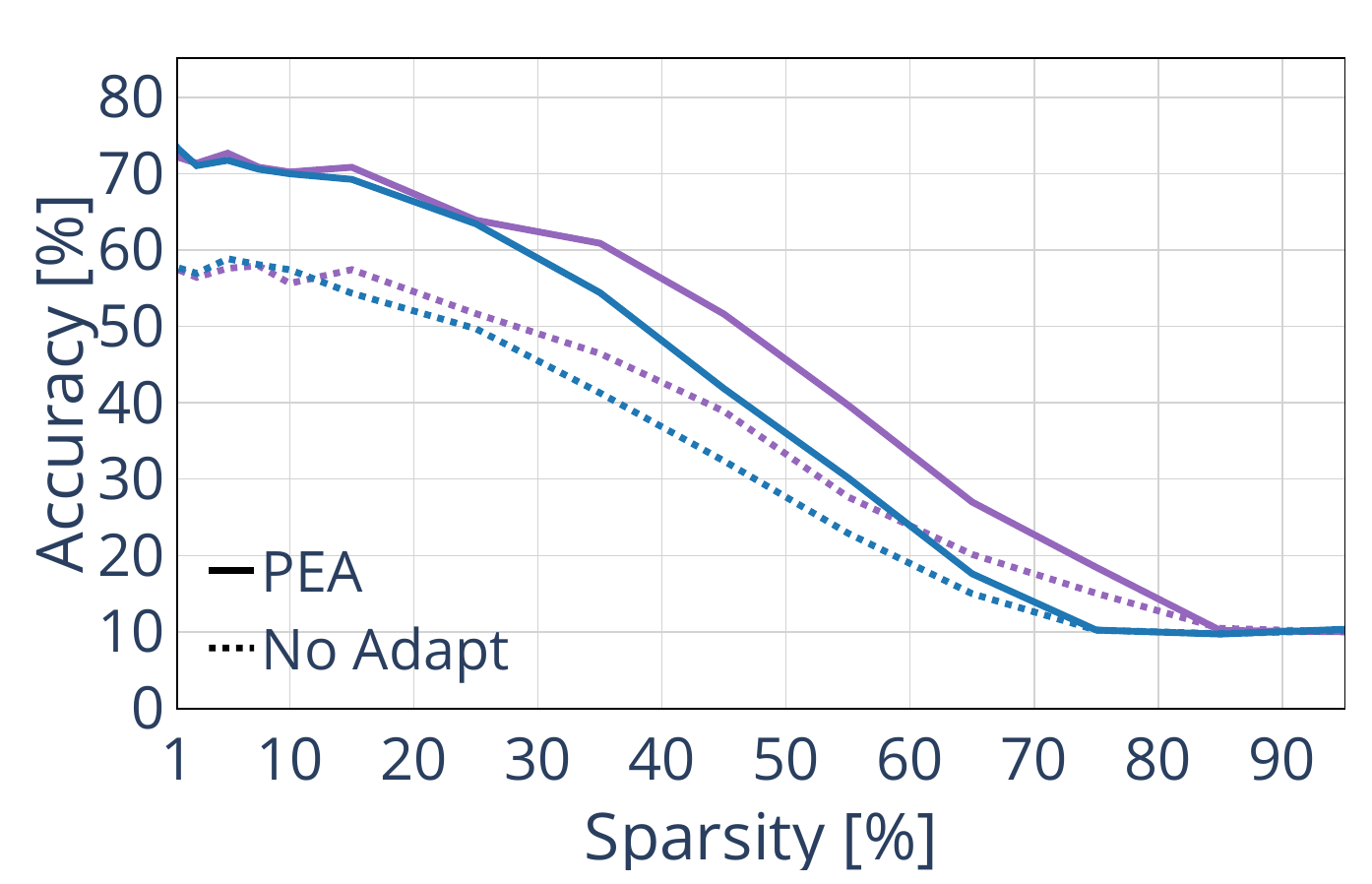}
\caption{ResNet-18 on CIFAR-10-C}
\label{fig:bp_free_free_cifar10}
\end{subfigure}
\hfill
\begin{subfigure}[t]{0.32\textwidth}
\centering
\includegraphics[width=\textwidth]{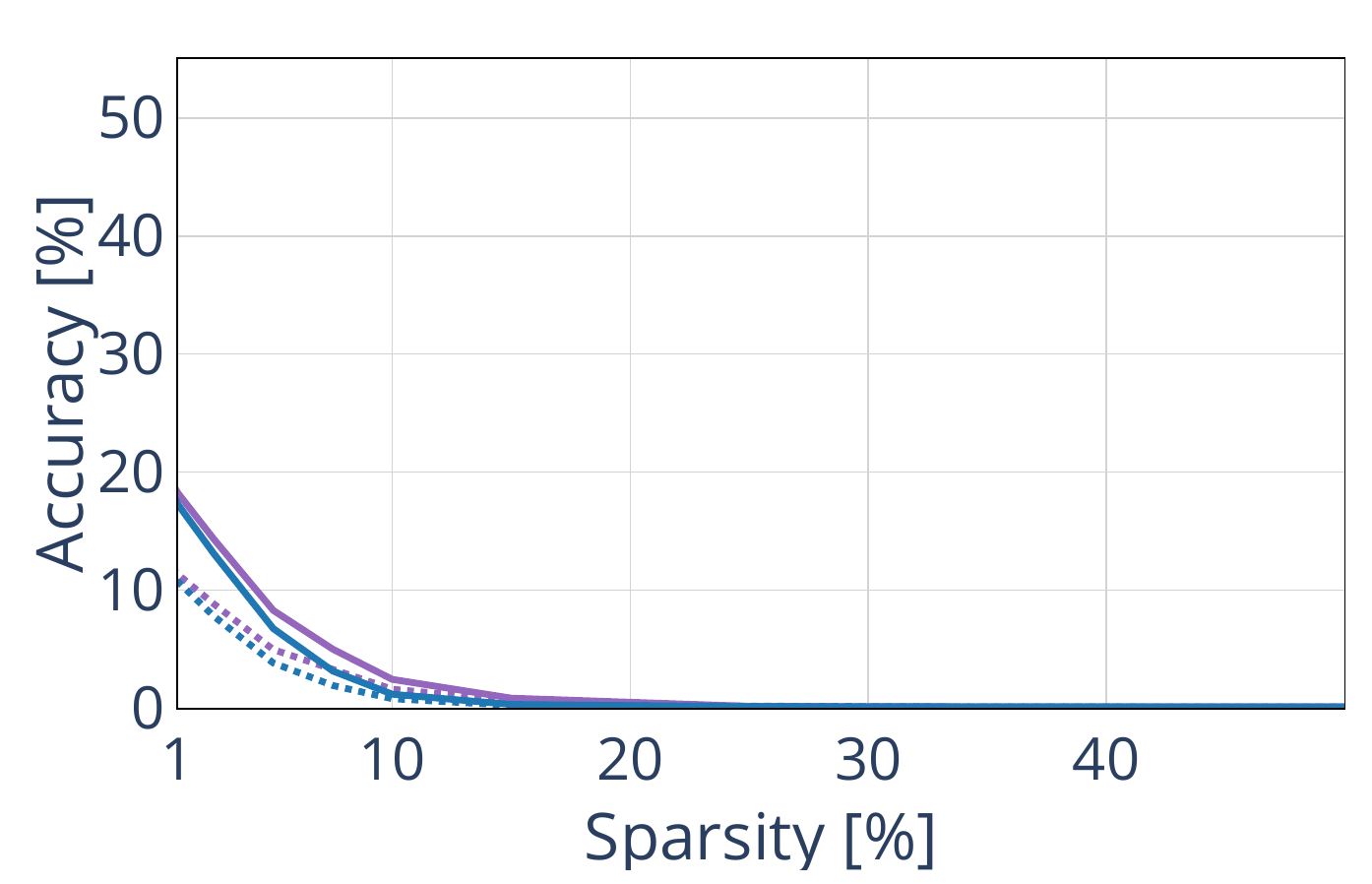}
\caption{ResNet-18 on ImageNet-C}
\label{fig:bp_free_free_imagenet_resnet}
\end{subfigure}
\hfill
\begin{subfigure}[t]{0.32\textwidth}
\centering
\includegraphics[width=\textwidth]{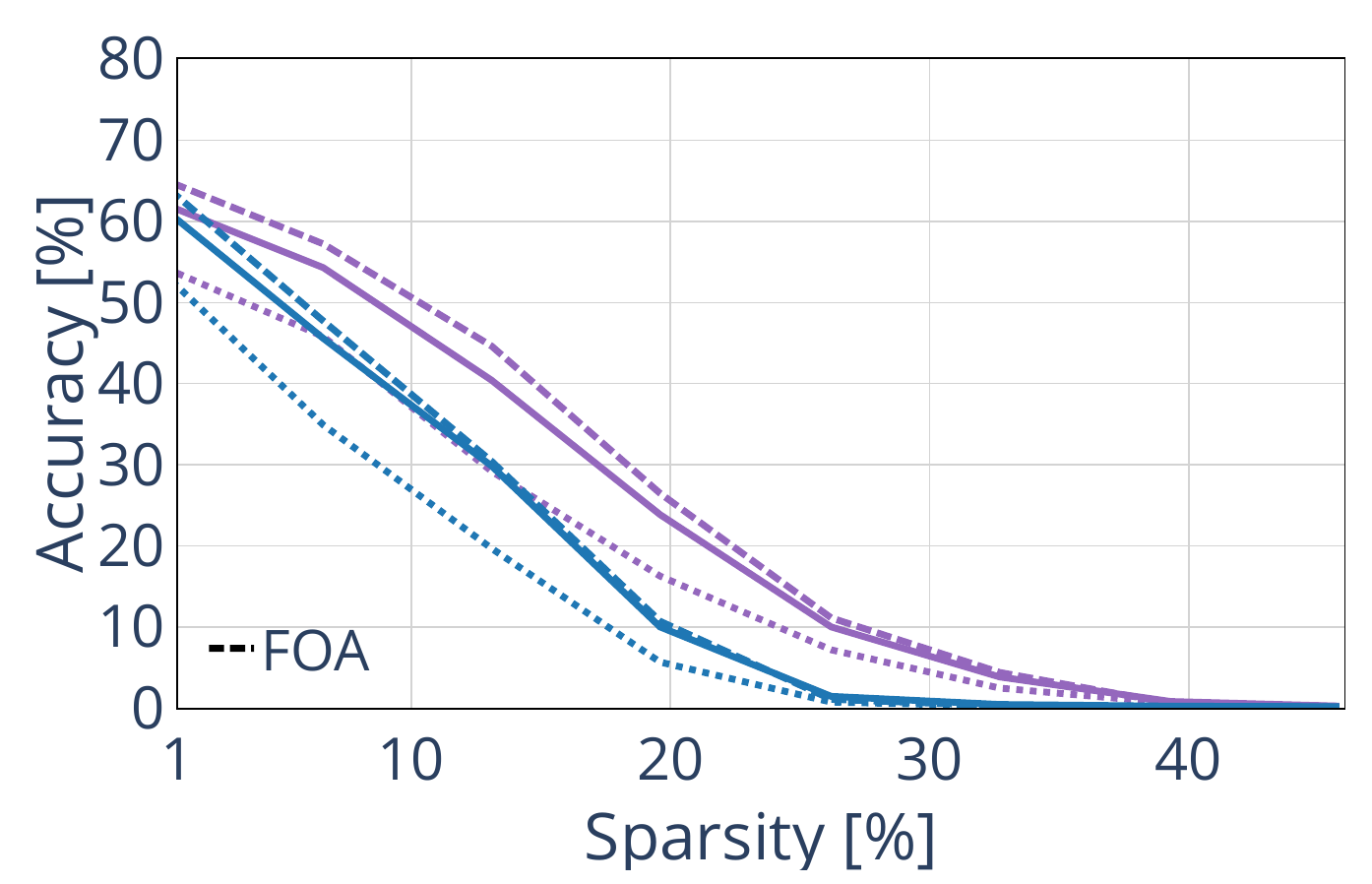}
\caption{ViT-Base on ImageNet-C}
\label{fig:bp_free_free_imagenet_vit}
\end{subfigure}
\caption{\textbf{Adaptation accuracy curves under backpropagation-free \tta (PEA and FOA) methods.} PEA~\citep{xiao2026architectureagnostic} and FOA~\citep{niu2024test} post-adaptation accuracy averaged across all corruptions with severity 5. PEA is applied to both ResNet-18 and ViT-Base; FOA is applied to ViT-Base only since it relies on learnable prompt embeddings, not available for the ResNet-18 models analyzed in this work. Same layout as Fig.~\ref{fig:hero}: {Top row} (a--c): data-dependent pruning (Wanda, Taylor, OBD). \emph{Bottom row} (d--f): data-free compression (Folding, Mag-$\ell_2$).}
\label{fig:bp_free_hero}
\end{figure*}

\begin{figure*}[h]
\centering
\begin{subfigure}[t]{0.32\textwidth}
\centering
\includegraphics[width=\textwidth]{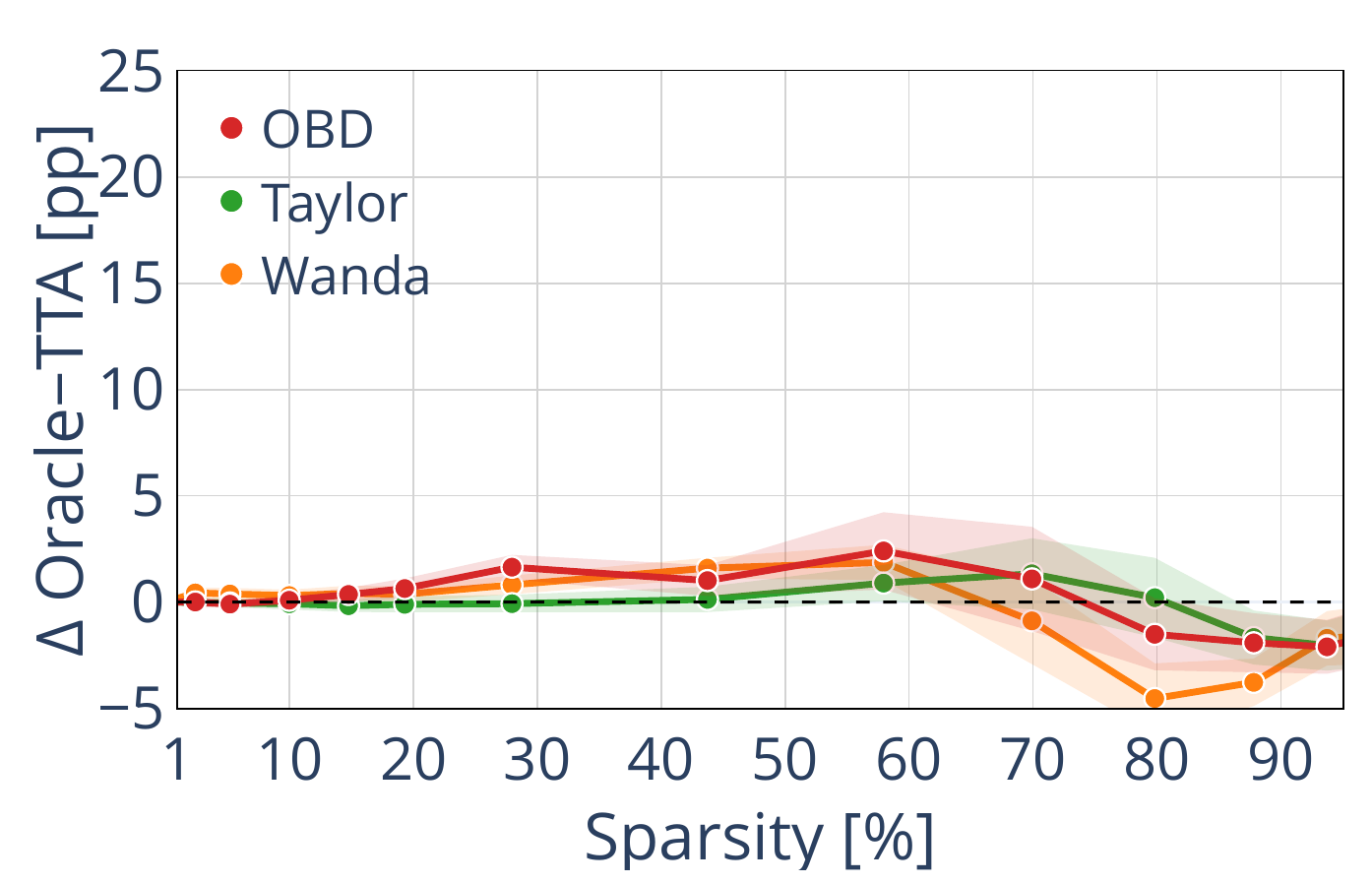}
\caption{ResNet-18 on CIFAR-10-C}
\label{fig:cig_data_cifar10}
\end{subfigure}
\hfill
\begin{subfigure}[t]{0.32\textwidth}
\centering
\includegraphics[width=\textwidth]{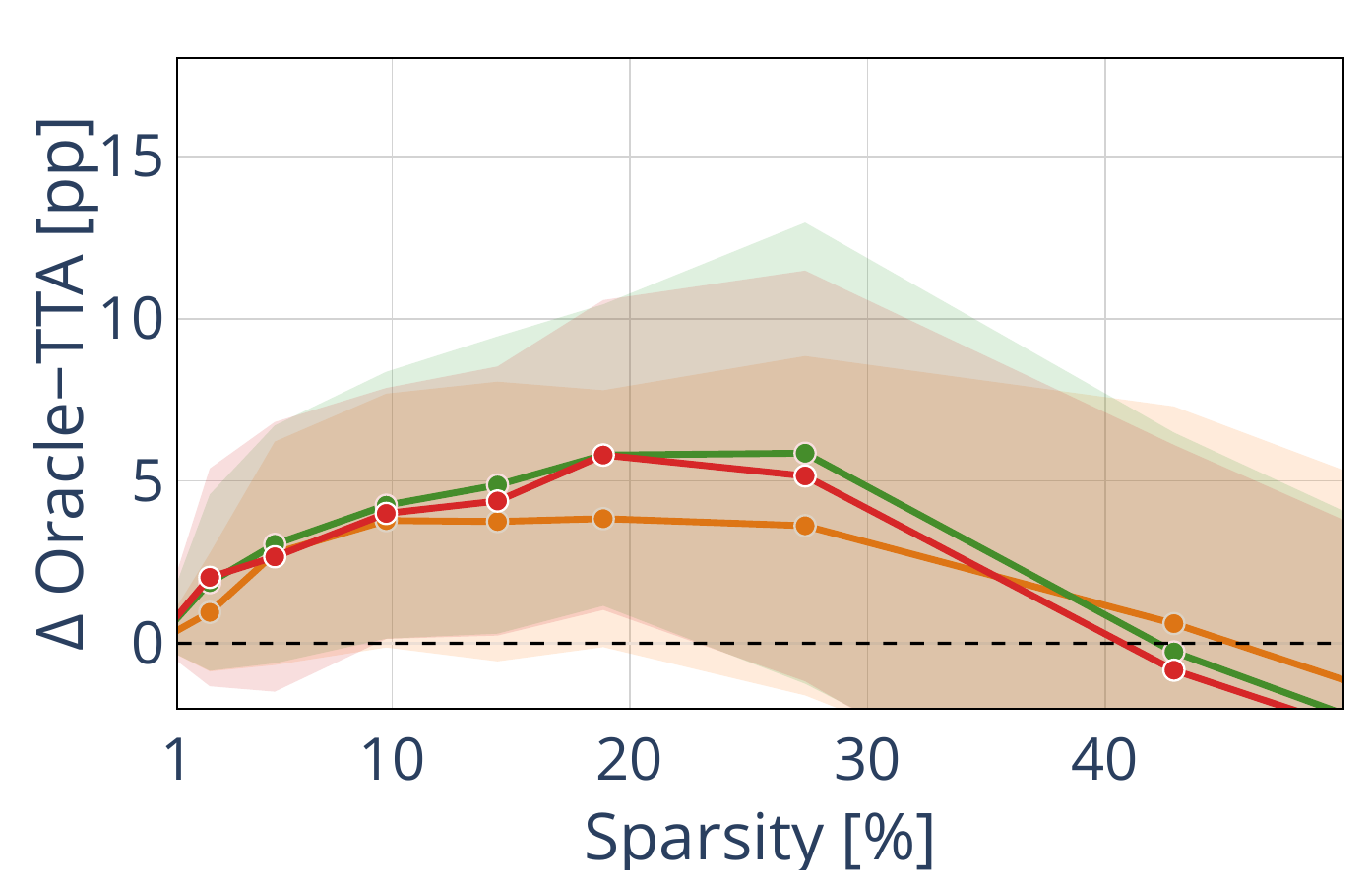}
\caption{ResNet-18 on ImageNet-C}
\label{fig:cig_data_imagenet}
\end{subfigure}
\hfill
\begin{subfigure}[t]{0.32\textwidth}
\centering
\includegraphics[width=\textwidth]{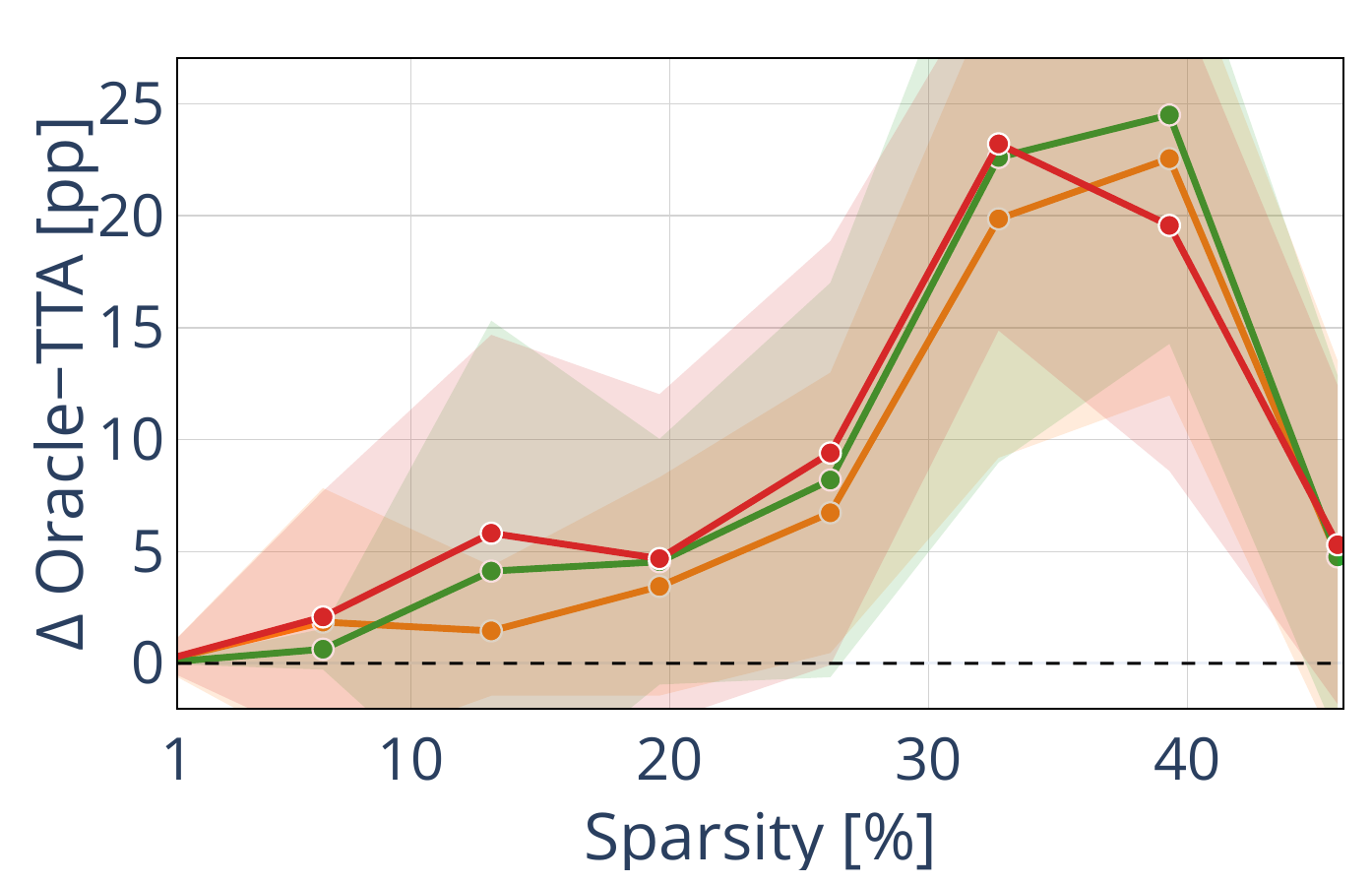}
\caption{ViT-Base on ImageNet-C}
\label{fig:cig_data_vit}
\end{subfigure}

\vspace{0.3em}

\begin{subfigure}[t]{0.32\textwidth}
\centering
\includegraphics[width=\textwidth]{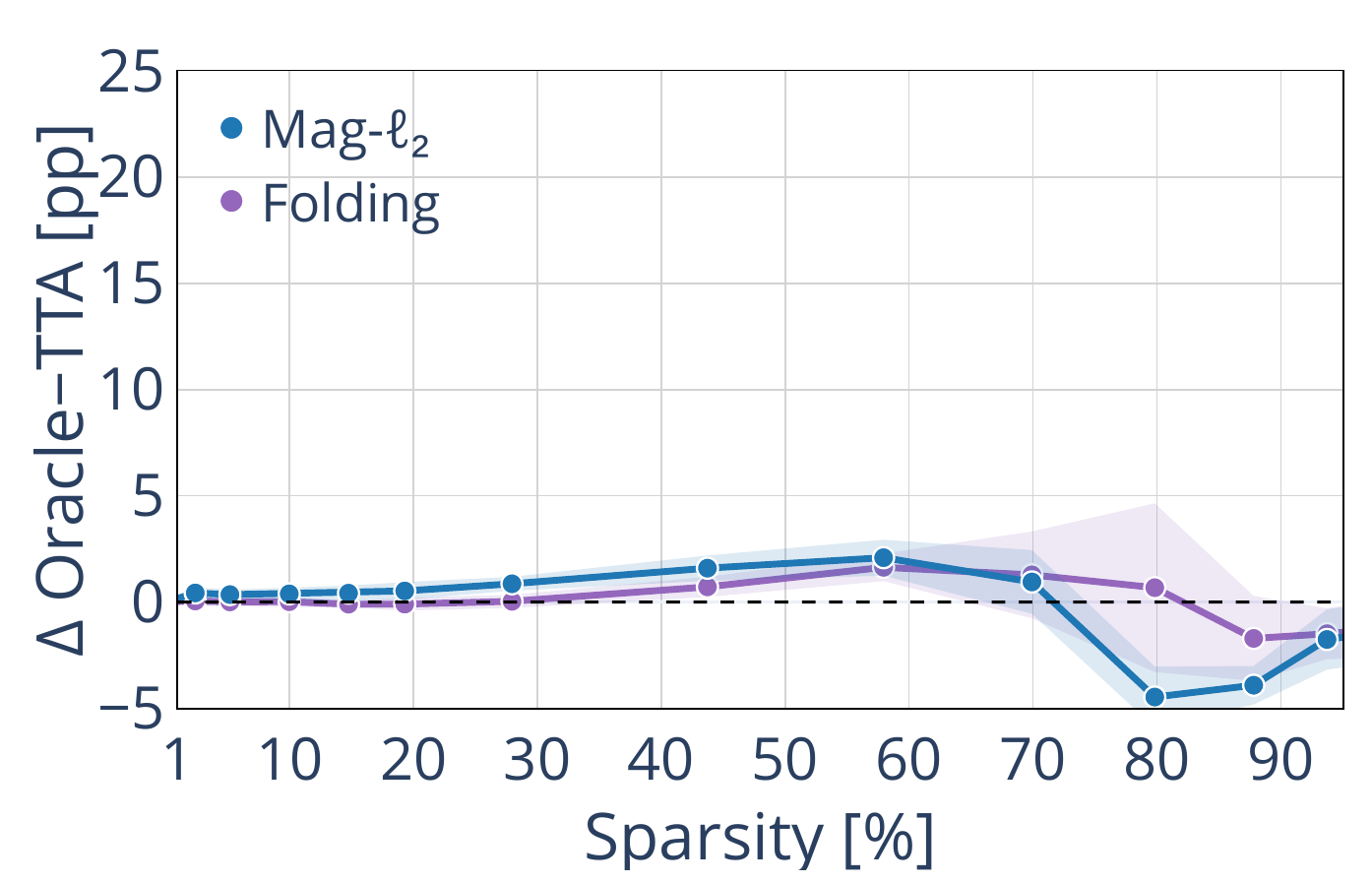}
\caption{ResNet-18 on CIFAR-10-C}
\label{fig:cig_free_cifar10}
\end{subfigure}
\hfill
\begin{subfigure}[t]{0.32\textwidth}
\centering
\includegraphics[width=\textwidth]{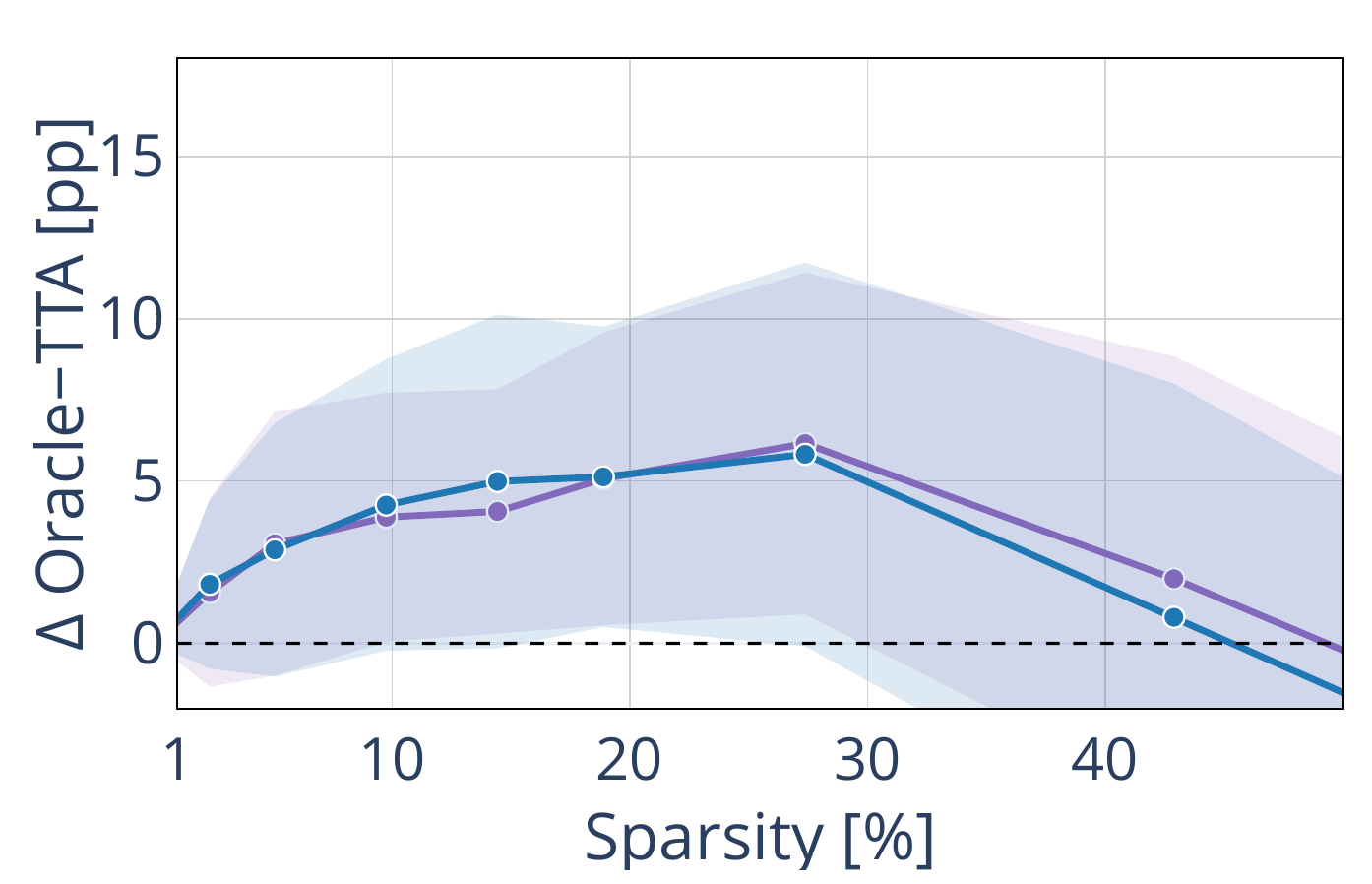}
\caption{ResNet-18 on ImageNet-C}
\label{fig:cig_free_imagenet}
\end{subfigure}
\hfill
\begin{subfigure}[t]{0.32\textwidth}
\centering
\includegraphics[width=\textwidth]{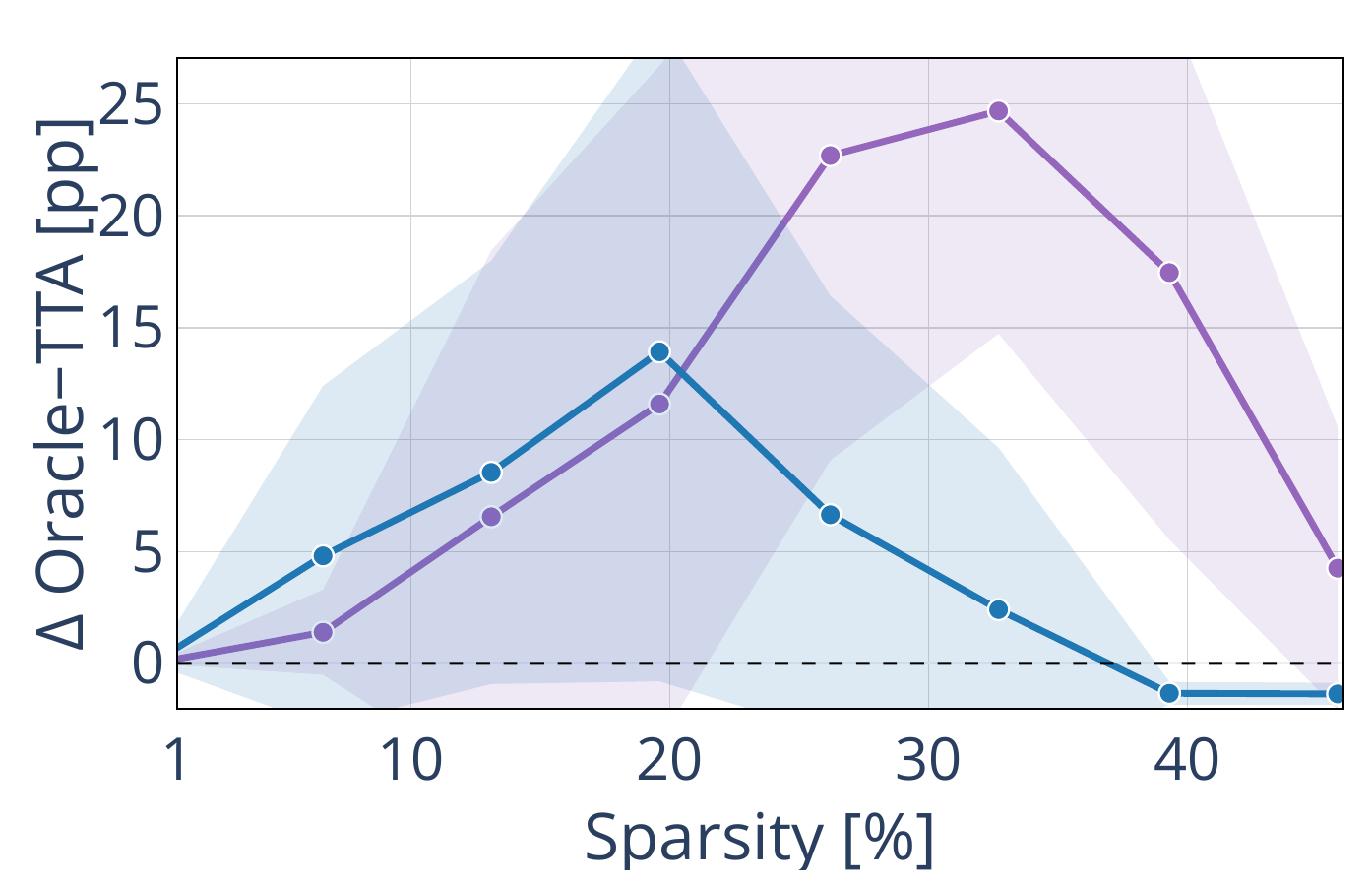}
\caption{ViT-Base on ImageNet-C}
\label{fig:cig_free_vit}
\end{subfigure}
\caption{\textbf{Compression-induced Oracle--\tta gap} $\Delta G(r) = [\text{Oracle}-\text{\tta}](r) - [\text{Oracle}-\text{\tta}](r=0.0)$. Solid lines are corruption-averaged means with $\pm \sigma$ bands across all corruptions; a positive value indicates that compression amplifies the Oracle--\tta gap beyond the dense baseline. Same layout as Fig.~\ref{fig:hero}.}
\label{fig:compression_induced_gap}
\end{figure*}

\begin{figure*}[h]
\centering
\begin{subfigure}[t]{0.32\textwidth}\centering\includegraphics[width=\textwidth]{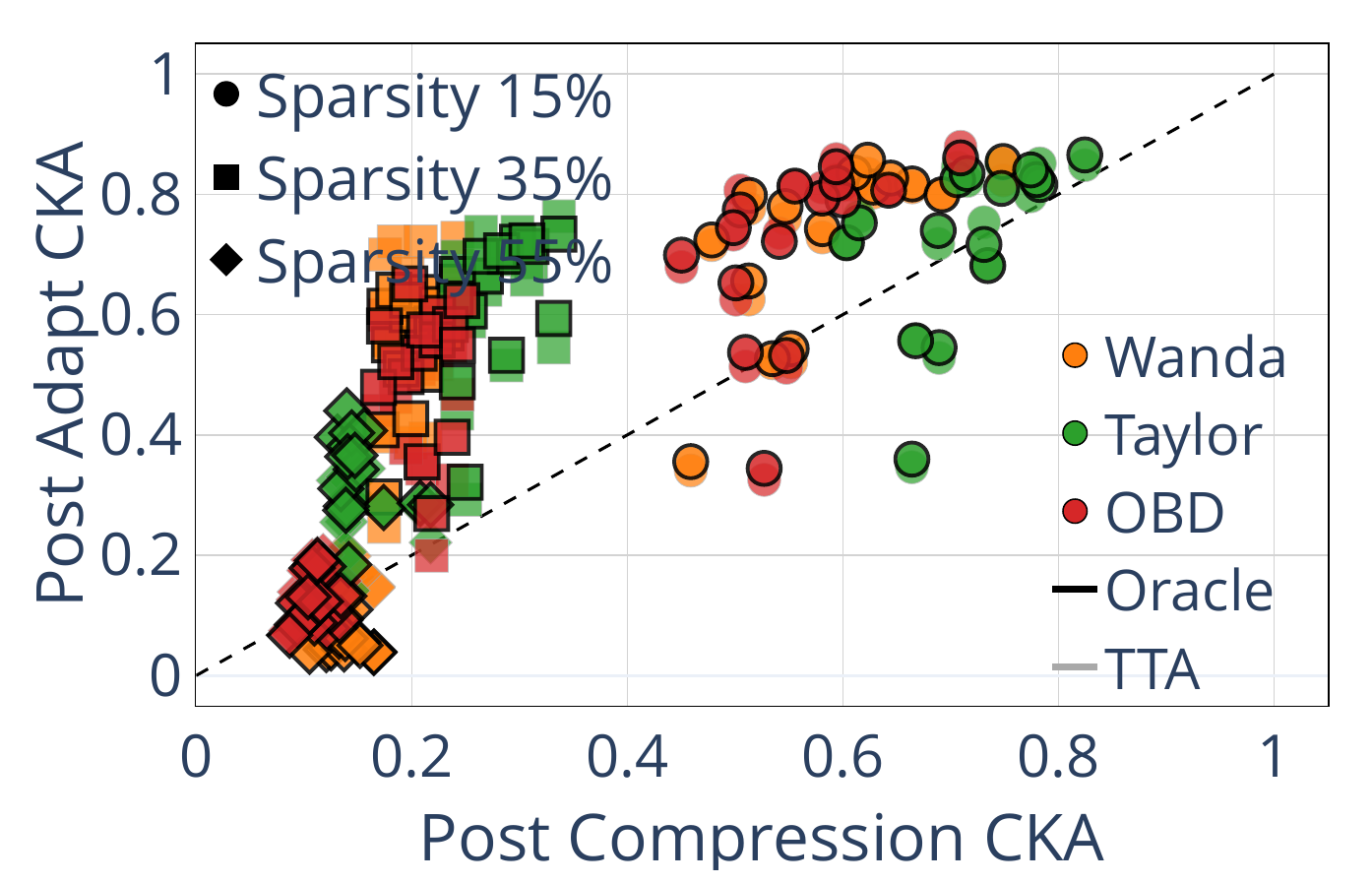}\caption{ResNet-18 on CIFAR-10-C}\label{fig:cka_dpt_s3_d_c}\end{subfigure}
\hfill
\begin{subfigure}[t]{0.32\textwidth}\centering\includegraphics[width=\textwidth]{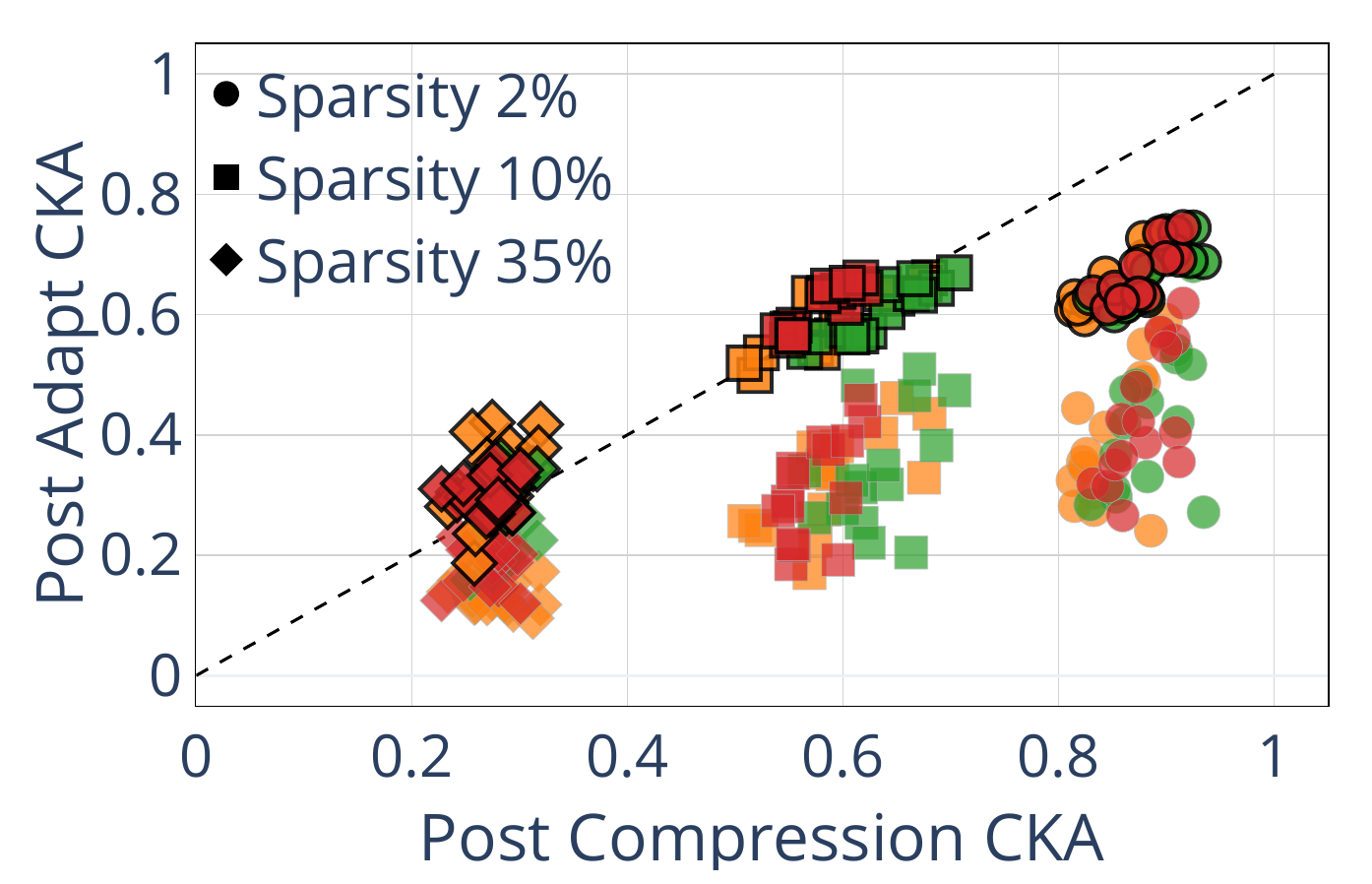}\caption{ResNet-18 on ImageNet-C}\label{fig:cka_dpt_s3_d_ir}\end{subfigure}
\hfill
\begin{subfigure}[t]{0.32\textwidth}\centering\includegraphics[width=\textwidth]{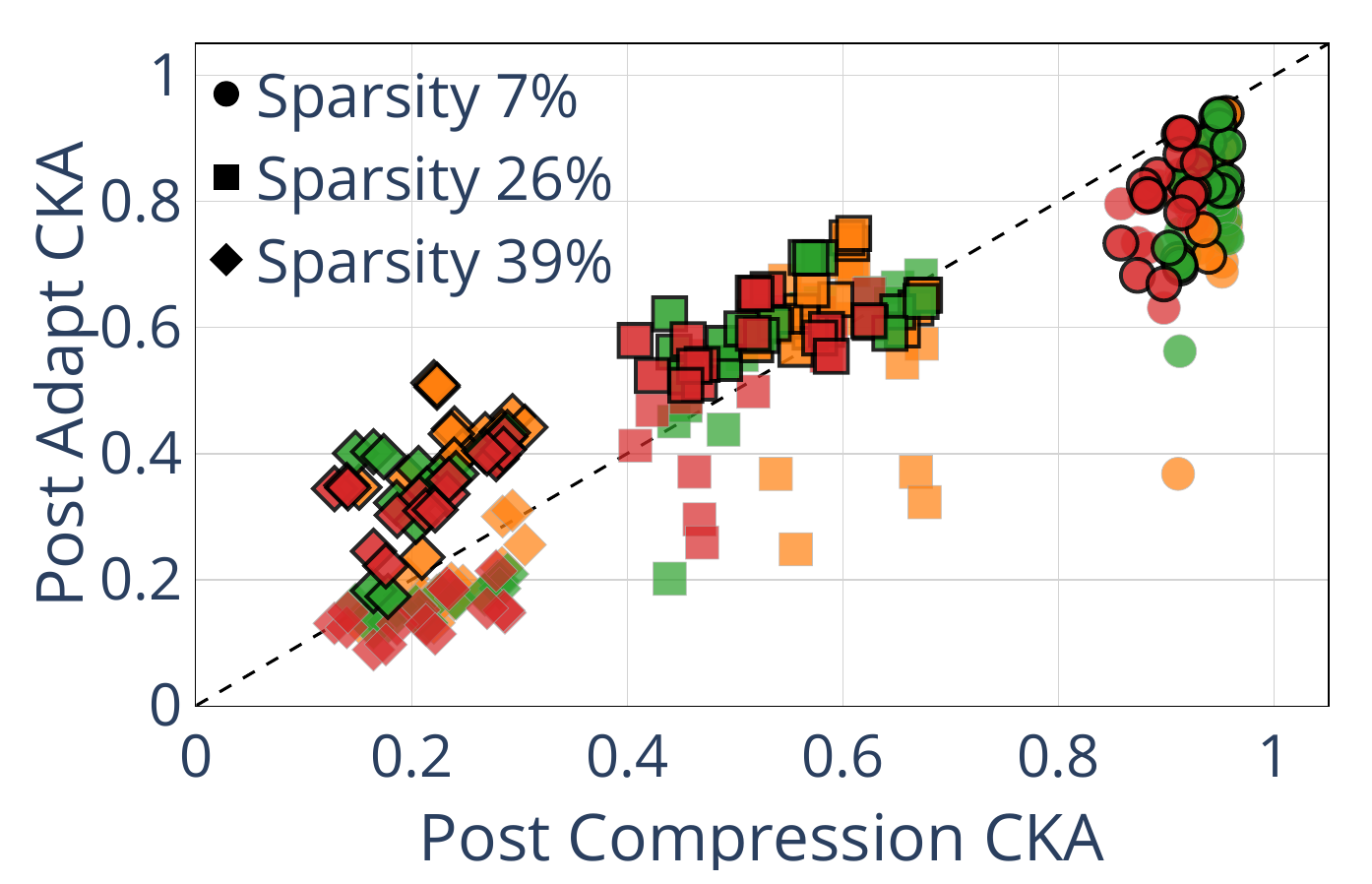}\caption{ViT-Base on ImageNet-C}\label{fig:cka_dpt_s3_d_iv}\end{subfigure}

\vspace{0.3em}

\begin{subfigure}[t]{0.32\textwidth}\centering\includegraphics[width=\textwidth]{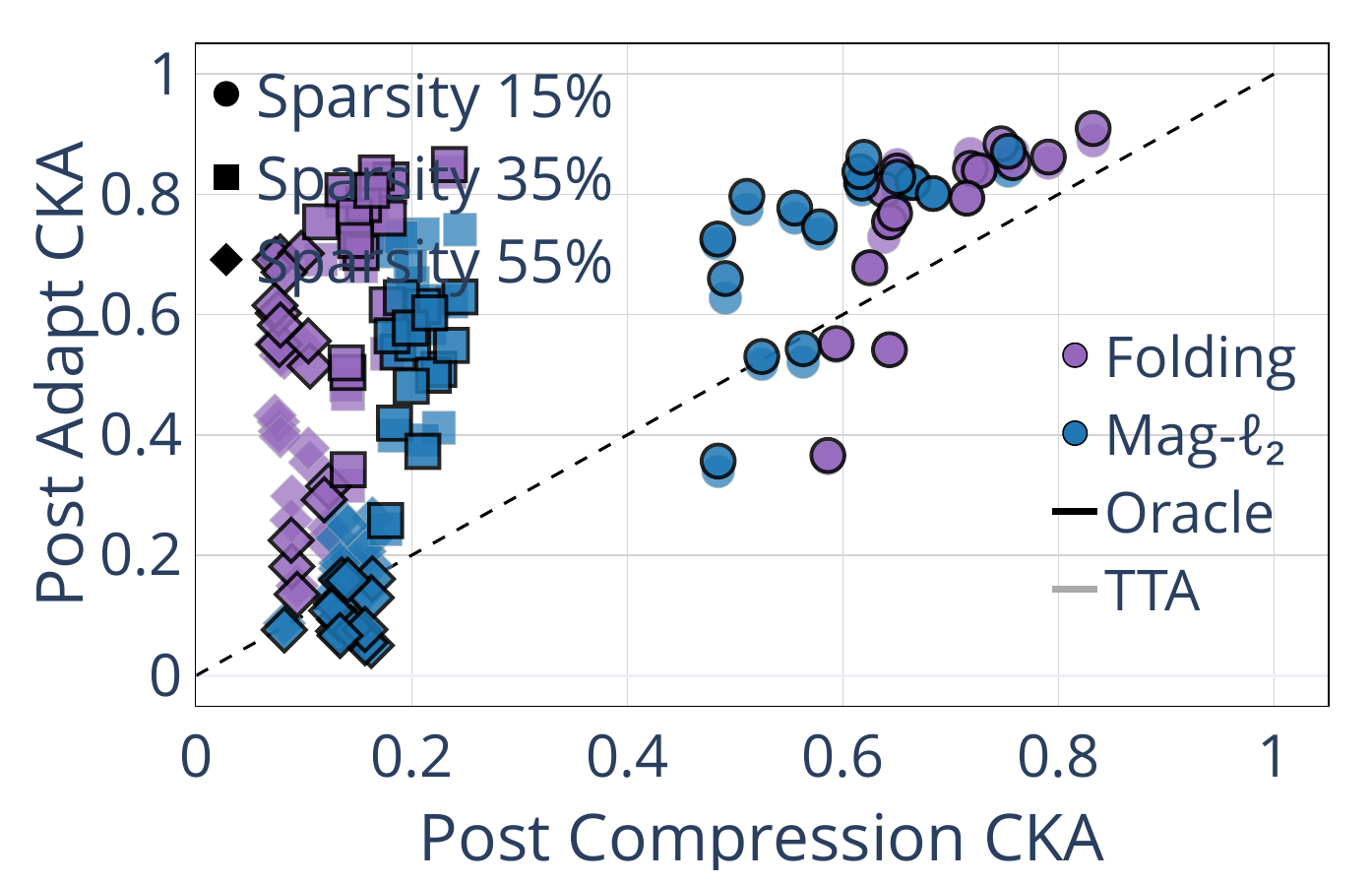}\caption{ResNet-18 on CIFAR-10-C}\label{fig:cka_dpt_s3_f_c}\end{subfigure}
\hfill
\begin{subfigure}[t]{0.32\textwidth}\centering\includegraphics[width=\textwidth]{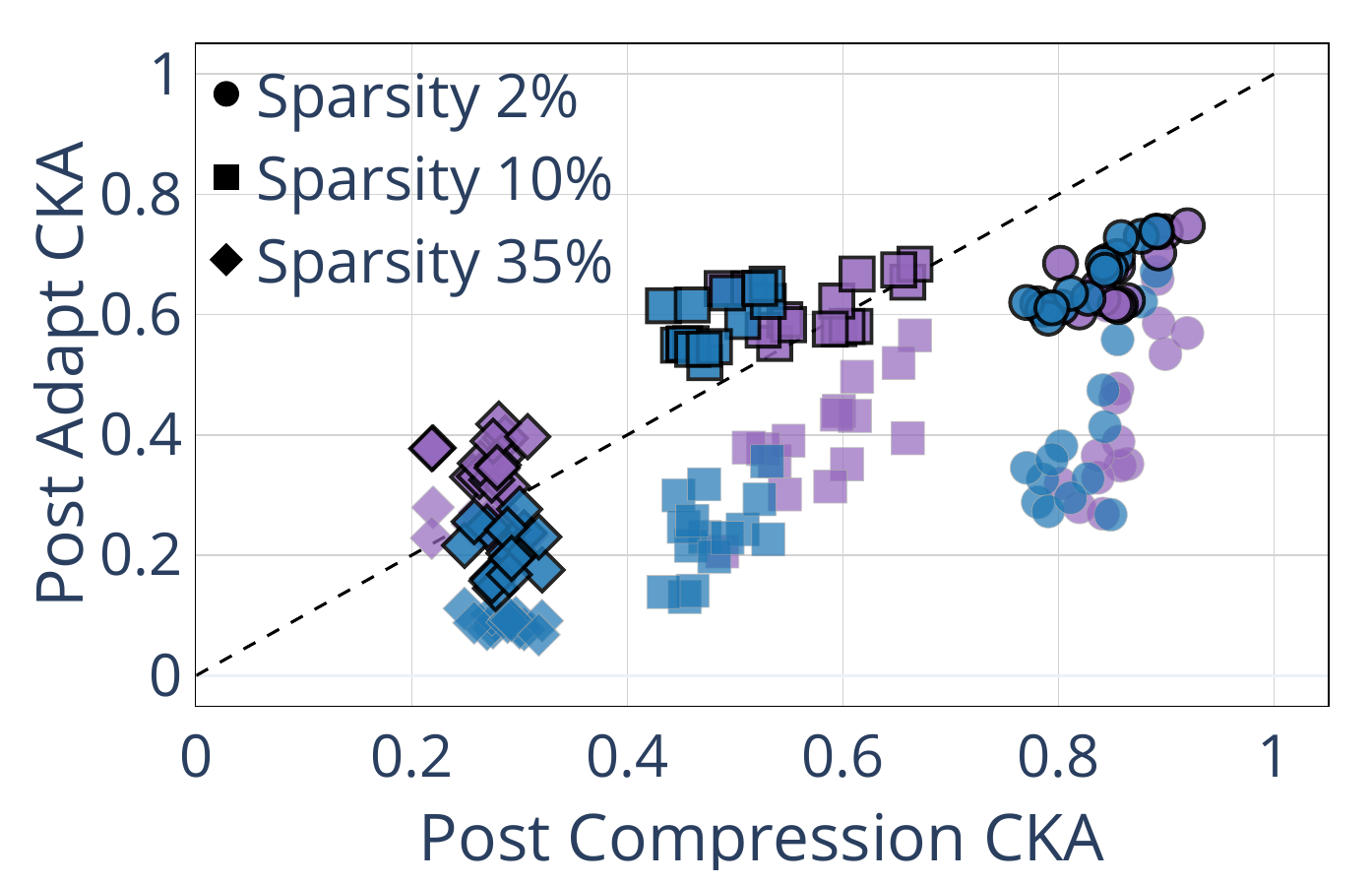}\caption{ResNet-18 on ImageNet-C}\label{fig:cka_dpt_s3_f_ir}\end{subfigure}
\hfill
\begin{subfigure}[t]{0.32\textwidth}\centering\includegraphics[width=\textwidth]{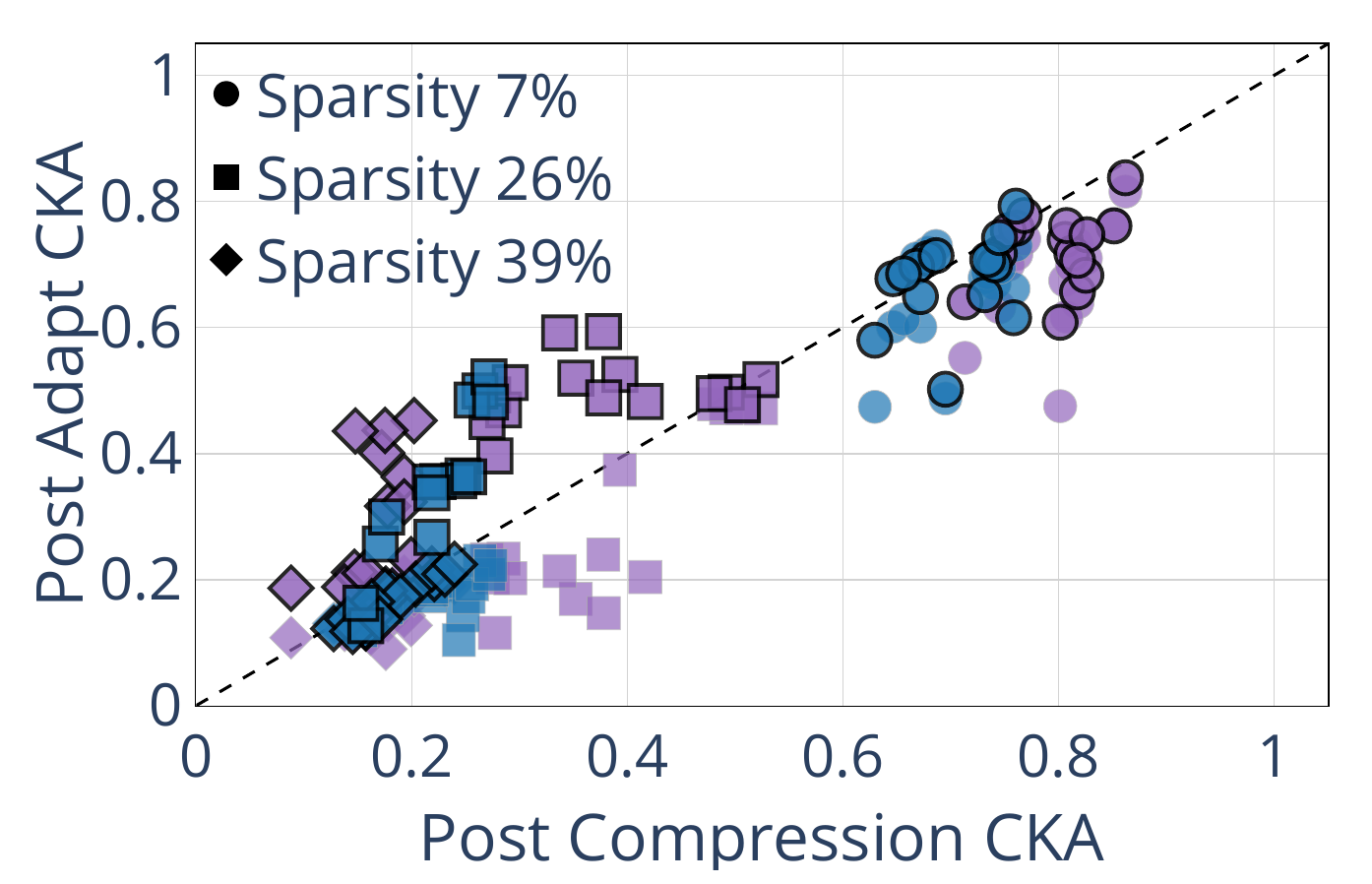}\caption{ViT-Base on ImageNet-C}\label{fig:cka_dpt_s3_f_iv}\end{subfigure}
\caption{\textbf{Multi-seed worst-layer \cka, \emph{dense-prune-TTA}, severity~3.} Same layout as Fig.~\ref{fig:cka_main}.}
\label{fig:cka_dense_prune_tta_sev3}
\end{figure*}

\begin{figure*}[h]
\centering

\begin{subfigure}[t]{0.24\textwidth}\centering\includegraphics[width=\textwidth]{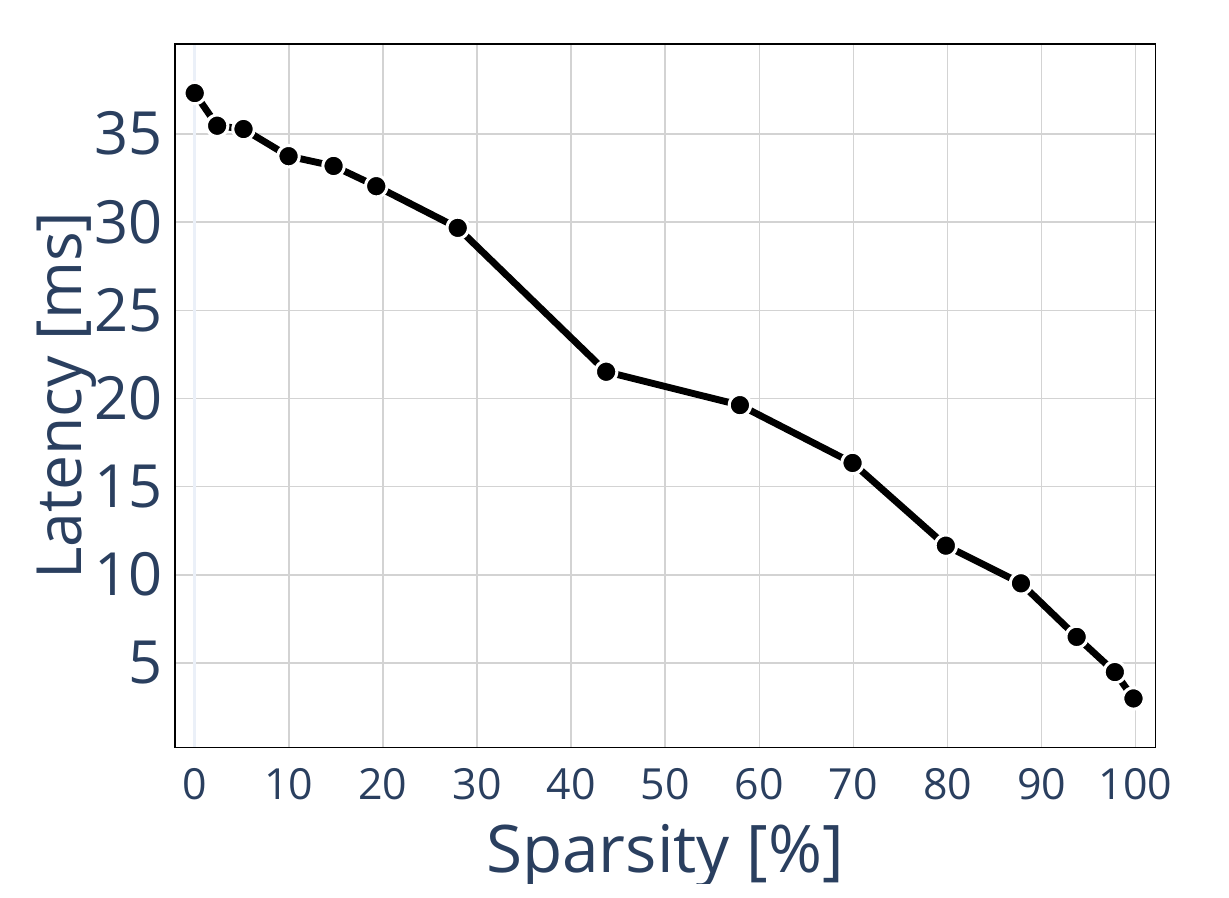}\caption{RN-18 CIFAR: Latency}\label{fig:overhead_lat_rn18c10}\end{subfigure}
\hfill
\begin{subfigure}[t]{0.24\textwidth}\centering\includegraphics[width=\textwidth]{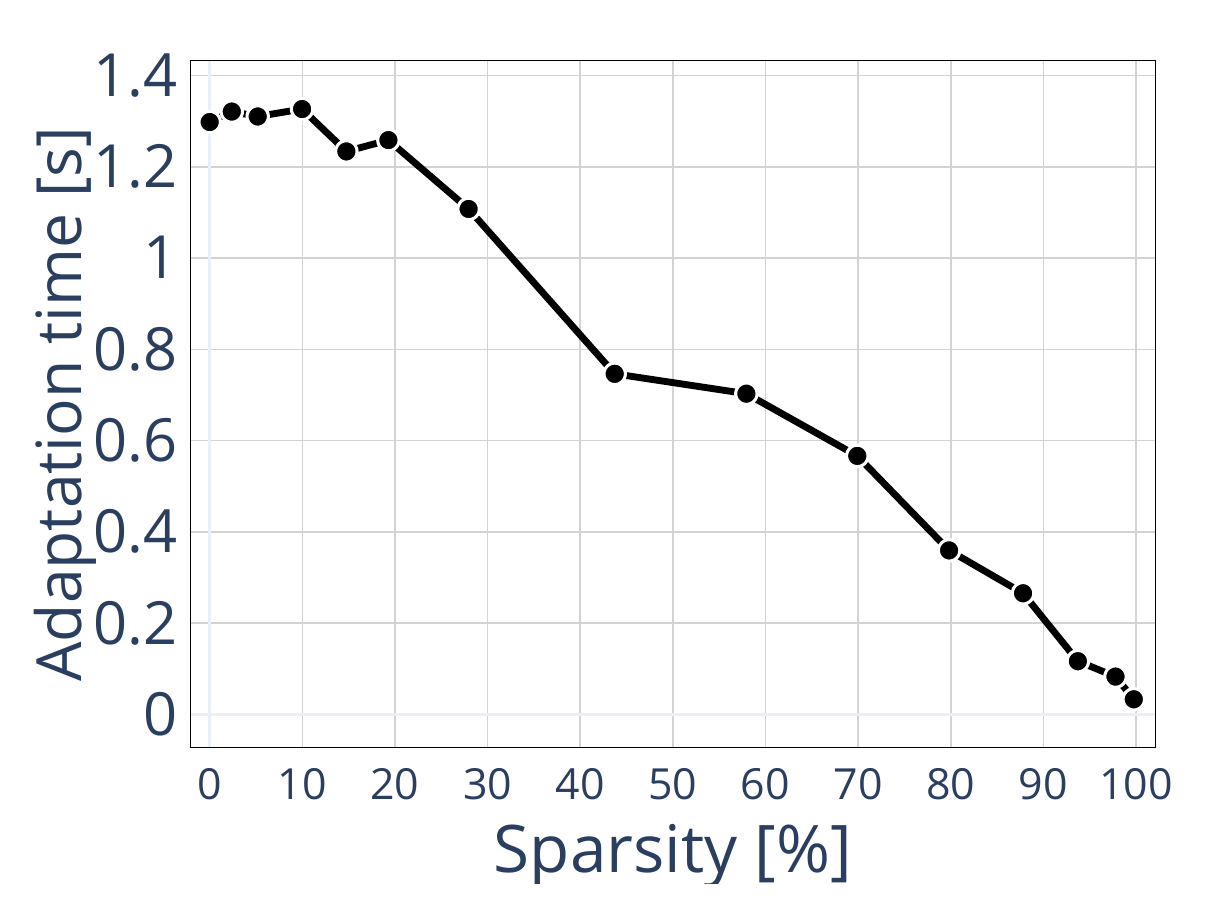}\caption{RN-18 CIFAR: Adapt. time}\label{fig:overhead_adapt_rn18c10}\end{subfigure}
\hfill
\begin{subfigure}[t]{0.24\textwidth}\centering\includegraphics[width=\textwidth]{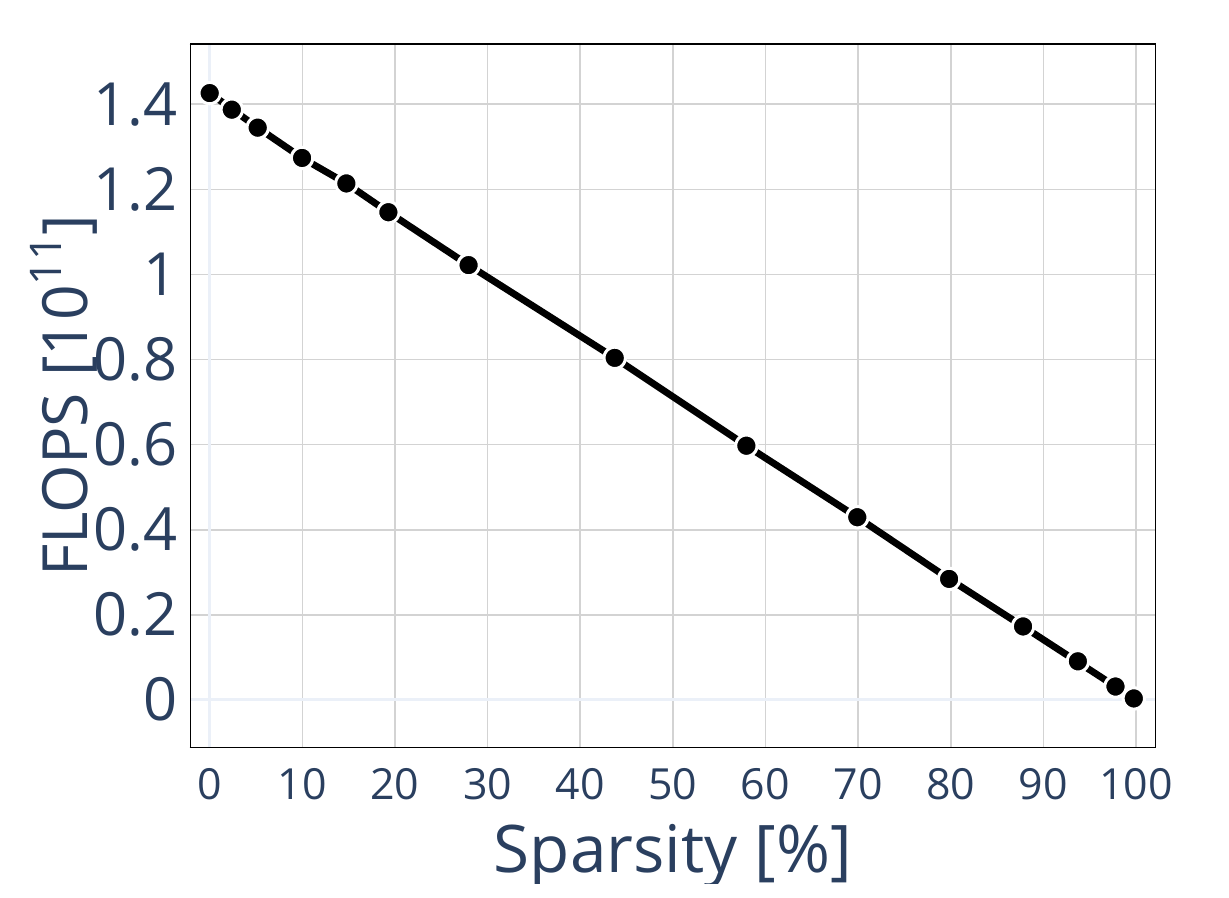}\caption{RN-18 CIFAR: FLOPS}\label{fig:overhead_flops_rn18c10}\end{subfigure}
\hfill
\begin{subfigure}[t]{0.24\textwidth}\centering\includegraphics[width=\textwidth]{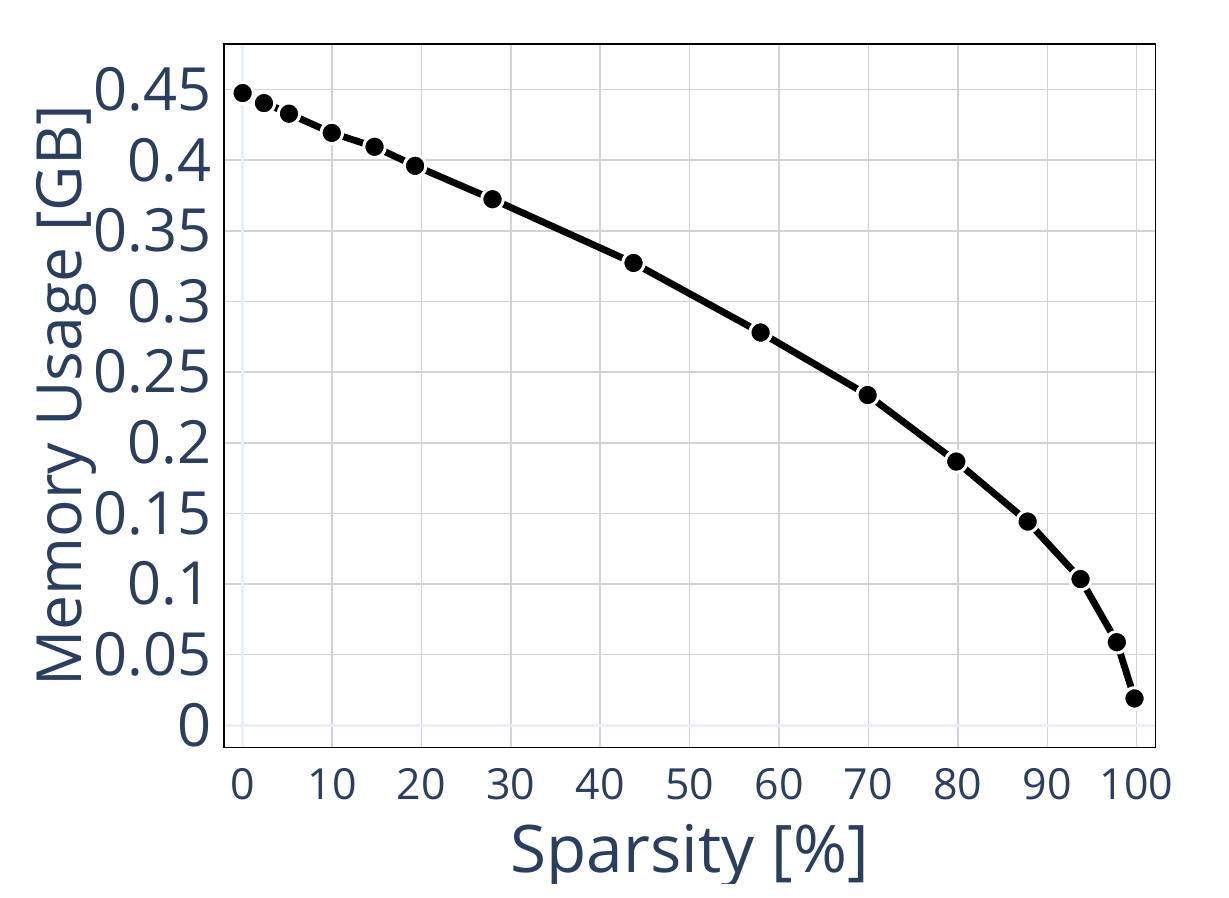}\caption{RN-18 CIFAR: Memory}\label{fig:overhead_mem_rn18c10}\end{subfigure}

\vspace{0.3em}

\begin{subfigure}[t]{0.24\textwidth}\centering\includegraphics[width=\textwidth]{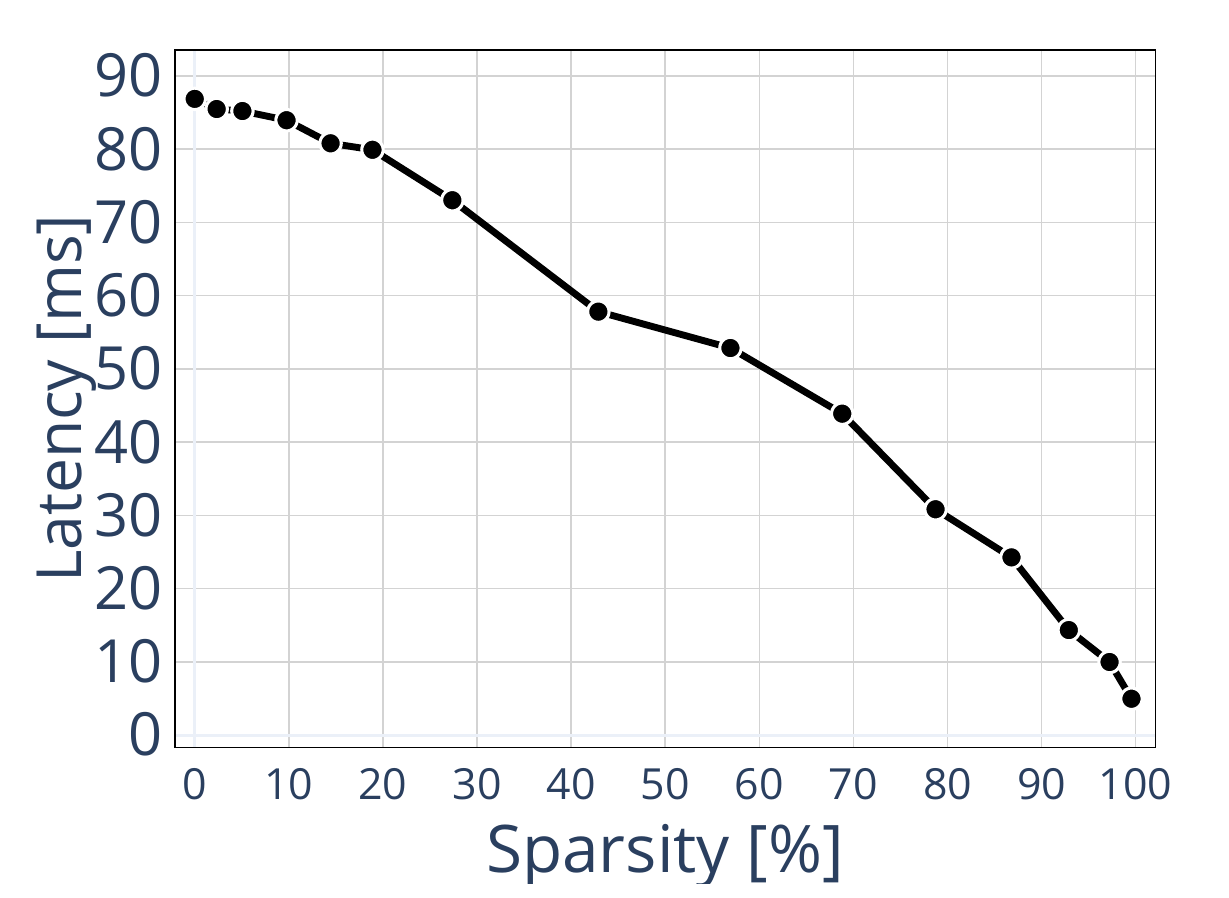}\caption{RN-18 ImageNet: Latency}\label{fig:overhead_lat_rn18im}\end{subfigure}
\hfill
\begin{subfigure}[t]{0.24\textwidth}\centering\includegraphics[width=\textwidth]{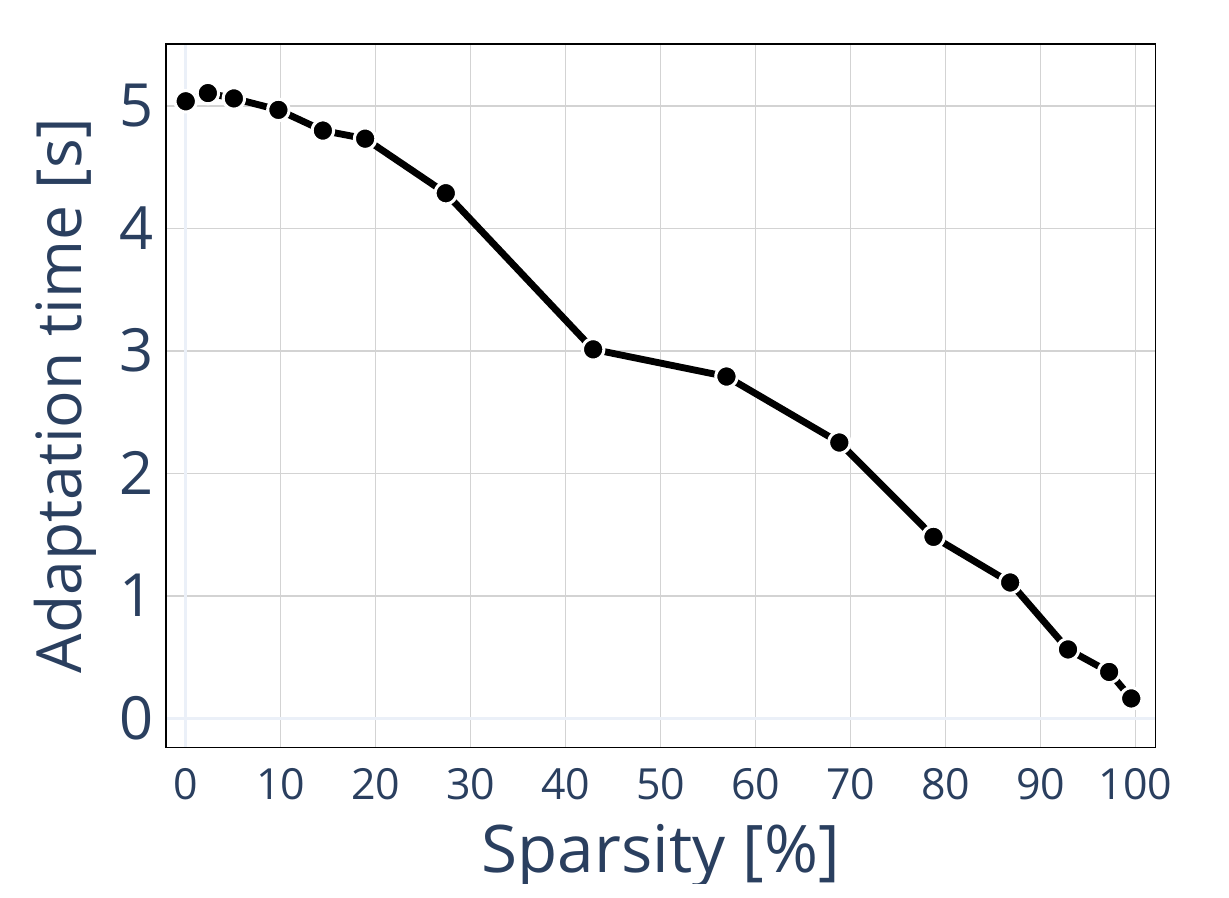}\caption{RN-18 ImageNet: Adapt. time}\label{fig:overhead_adapt_rn18im}\end{subfigure}
\hfill
\begin{subfigure}[t]{0.24\textwidth}\centering\includegraphics[width=\textwidth]{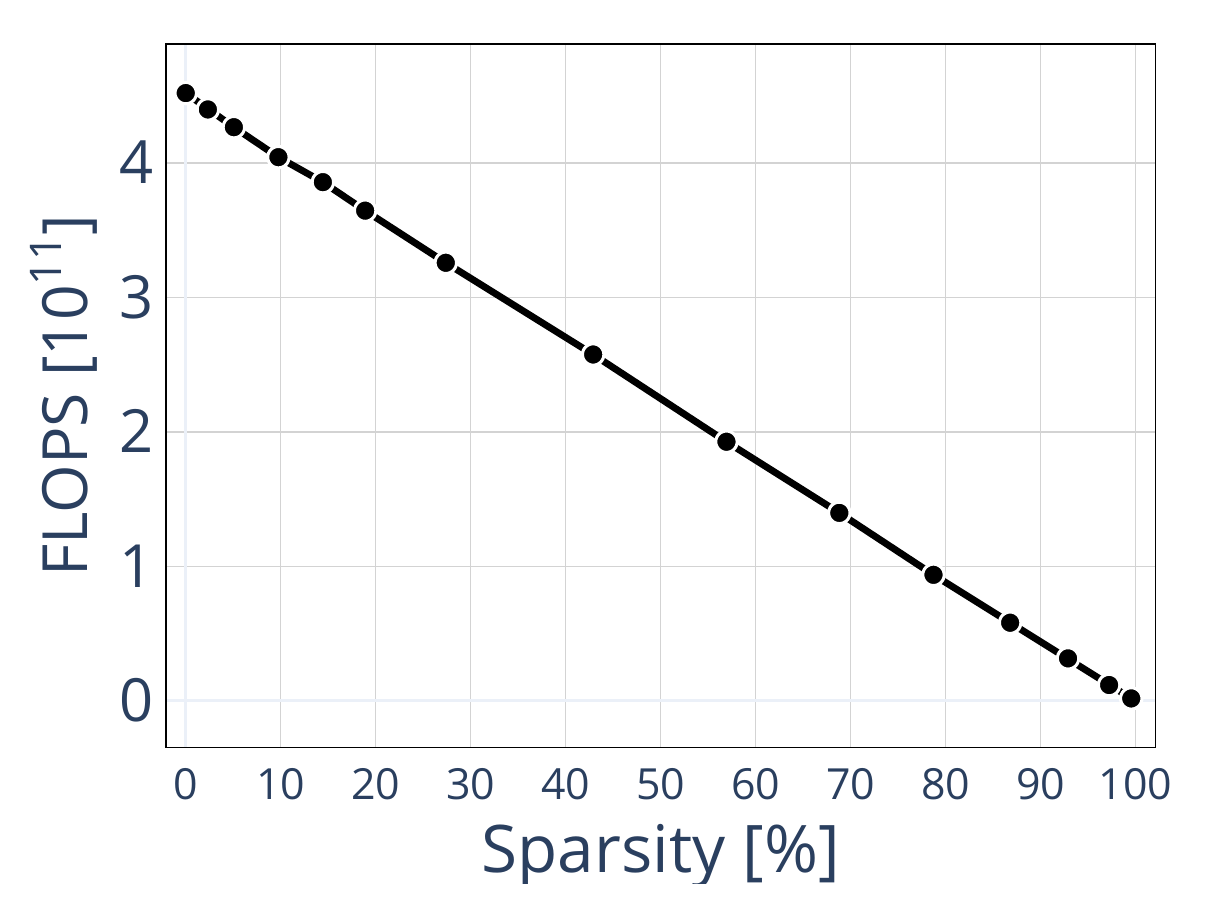}\caption{RN-18 ImageNet: FLOPS}\label{fig:overhead_flops_rn18im}\end{subfigure}
\hfill
\begin{subfigure}[t]{0.24\textwidth}\centering\includegraphics[width=\textwidth]{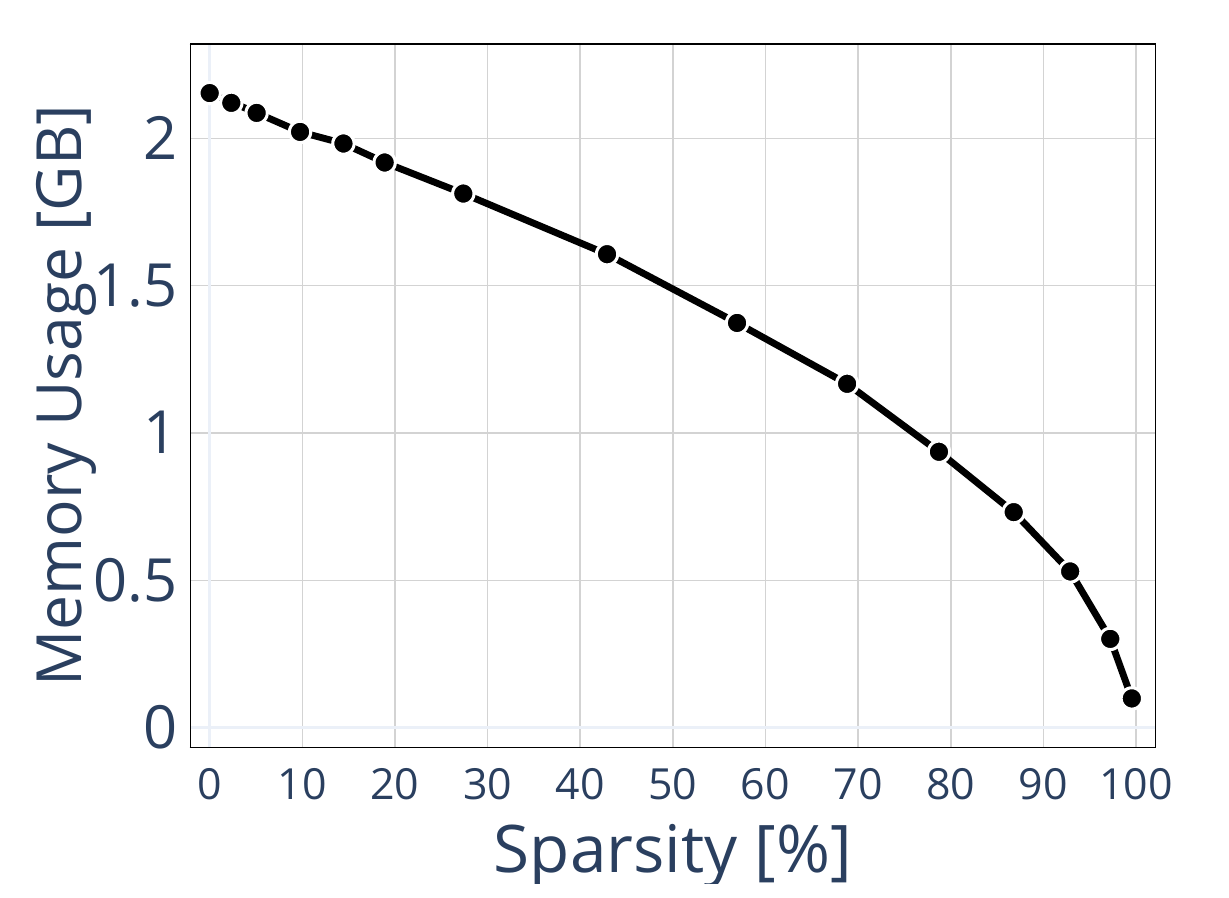}\caption{RN-18 ImageNet: Memory}\label{fig:overhead_mem_rn18im}\end{subfigure}

\vspace{0.3em}

\begin{subfigure}[t]{0.24\textwidth}\centering\includegraphics[width=\textwidth]{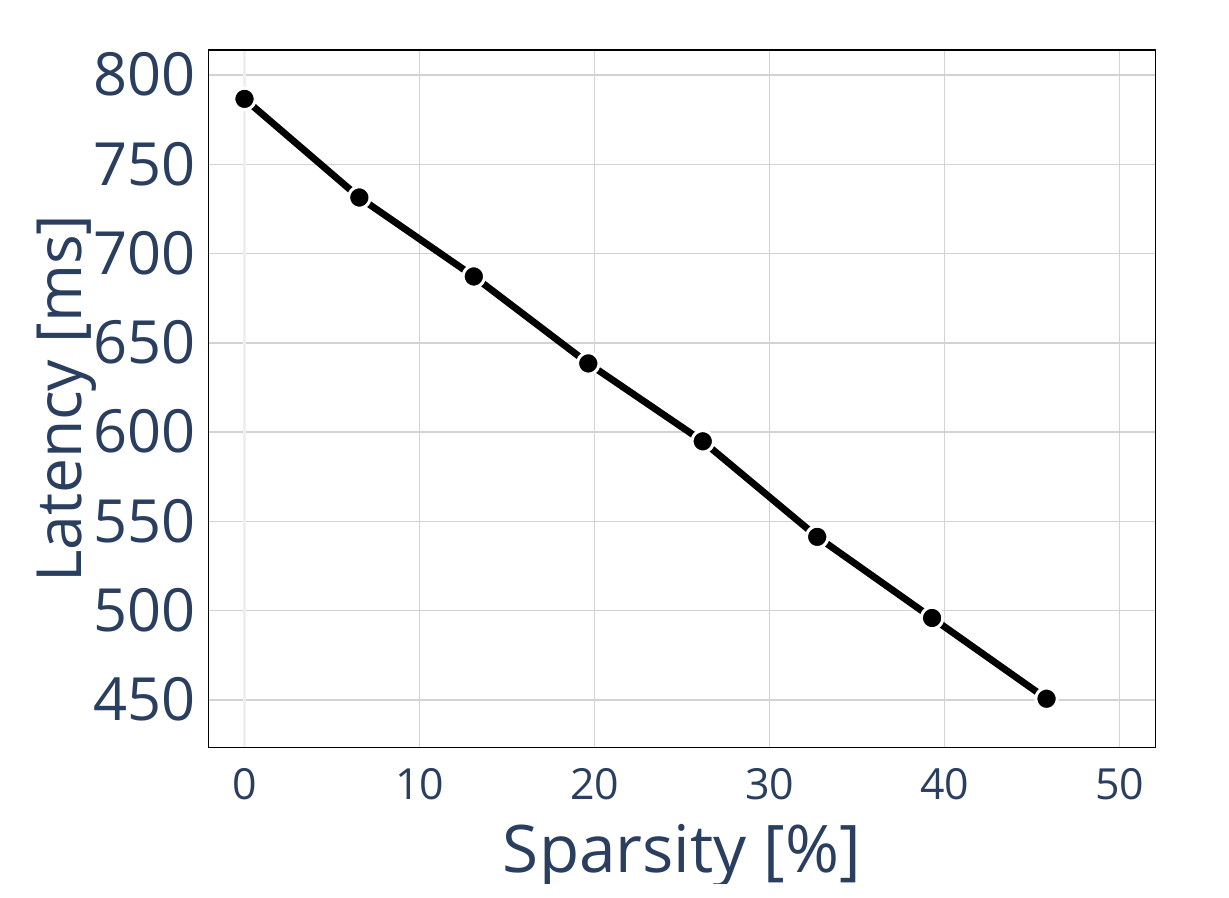}\caption{ViT-Base: Latency}\label{fig:overhead_lat_vitim}\end{subfigure}
\hfill
\begin{subfigure}[t]{0.24\textwidth}\centering\includegraphics[width=\textwidth]{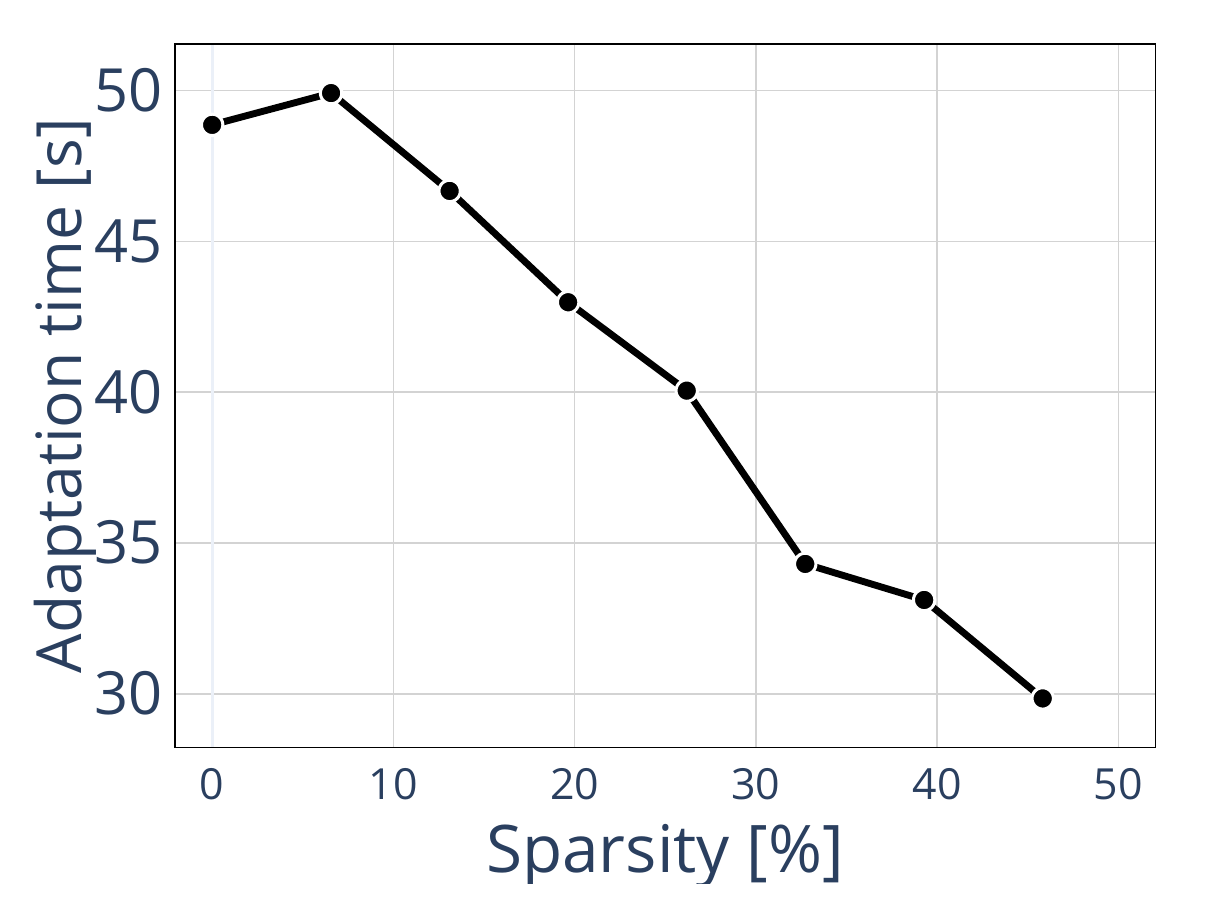}\caption{ViT-Base: Adapt. time}\label{fig:overhead_adapt_vitim}\end{subfigure}
\hfill
\begin{subfigure}[t]{0.24\textwidth}\centering\includegraphics[width=\textwidth]{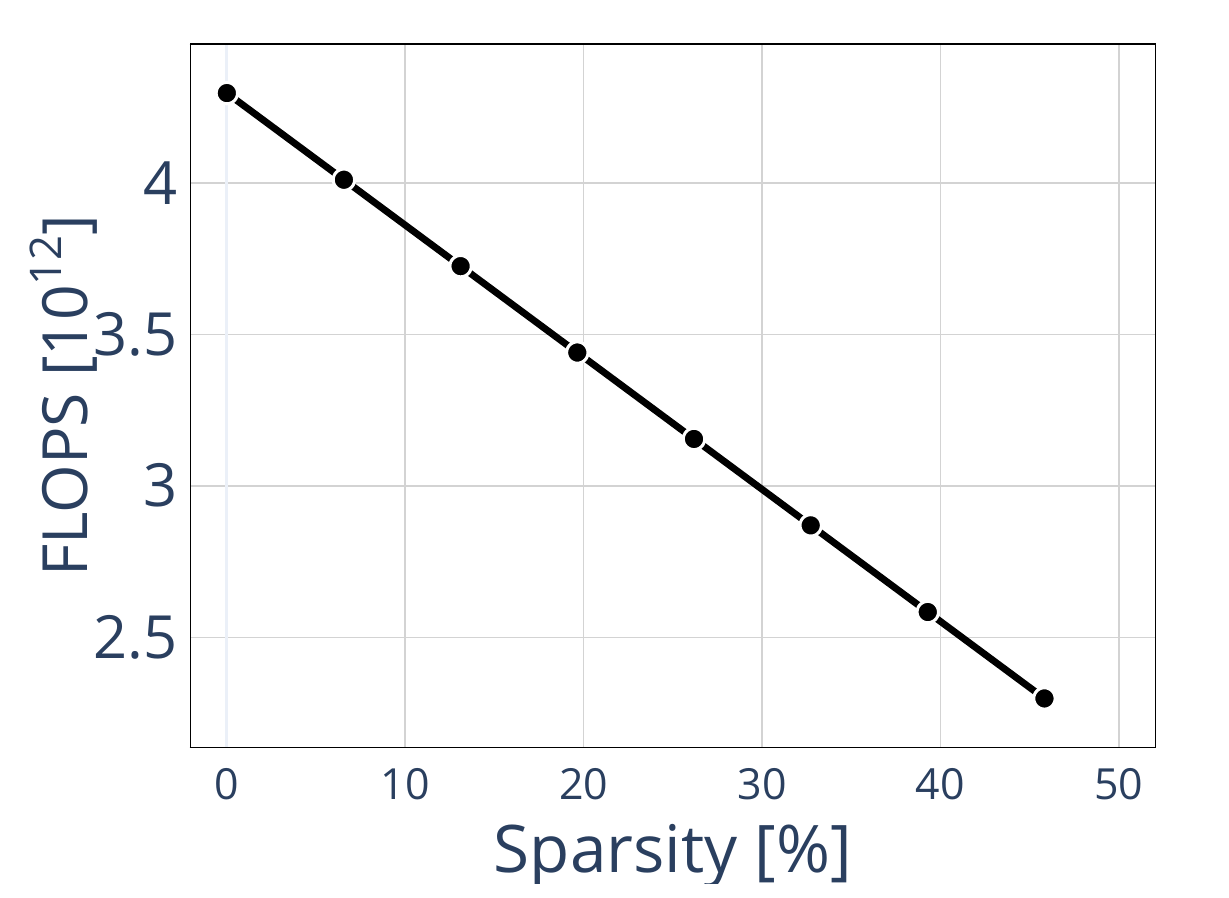}\caption{ViT-Base: FLOPS}\label{fig:overhead_flops_vitim}\end{subfigure}
\hfill
\begin{subfigure}[t]{0.24\textwidth}\centering\includegraphics[width=\textwidth]{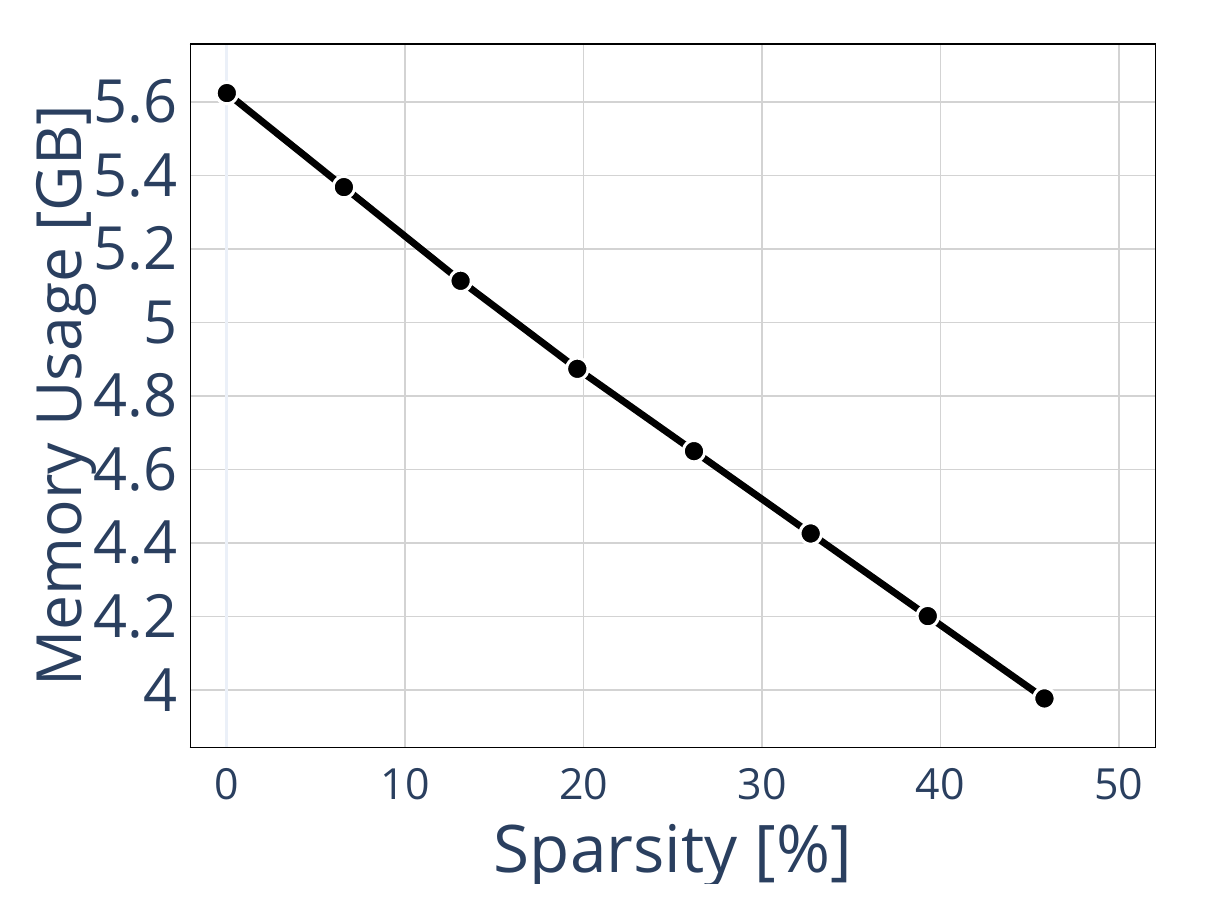}\caption{ViT-Base: Memory}\label{fig:overhead_mem_vitim}\end{subfigure}

\caption{\textbf{Compute, memory and latency profile across compression ratios.} For each architecture we report, as a function of sparsity, the CPU inference latency (ms), the full adaptation-step time (s). The analytical FLOPs and memory usage (GB) per adaptation step are also reported. All measurements use batch size $B{=}64$ on three channel inputs at the pretraining data resolution. The measurements are obtained on a single AMD EPYC 9654 96-core processor single-threaded. We use the evaluation protocol described in~\cite{slamanig2025llmsedgeparameterefficientfinetuning}. Under the uniform per-layer compression, for each compression ratio all the five structured compression methods studied in the paper yield architecturally equal smaller-dense ResNet-18 and ViT-Base networks, thus a single curve per architecture suffices; we report Magnitude-$\ell_2$ as a representative structured compression method.}
\label{fig:compute_overheads_appendix}
\end{figure*}

\begin{figure*}[h]
\centering
\begin{subfigure}[t]{0.32\textwidth}\centering\includegraphics[width=\textwidth]{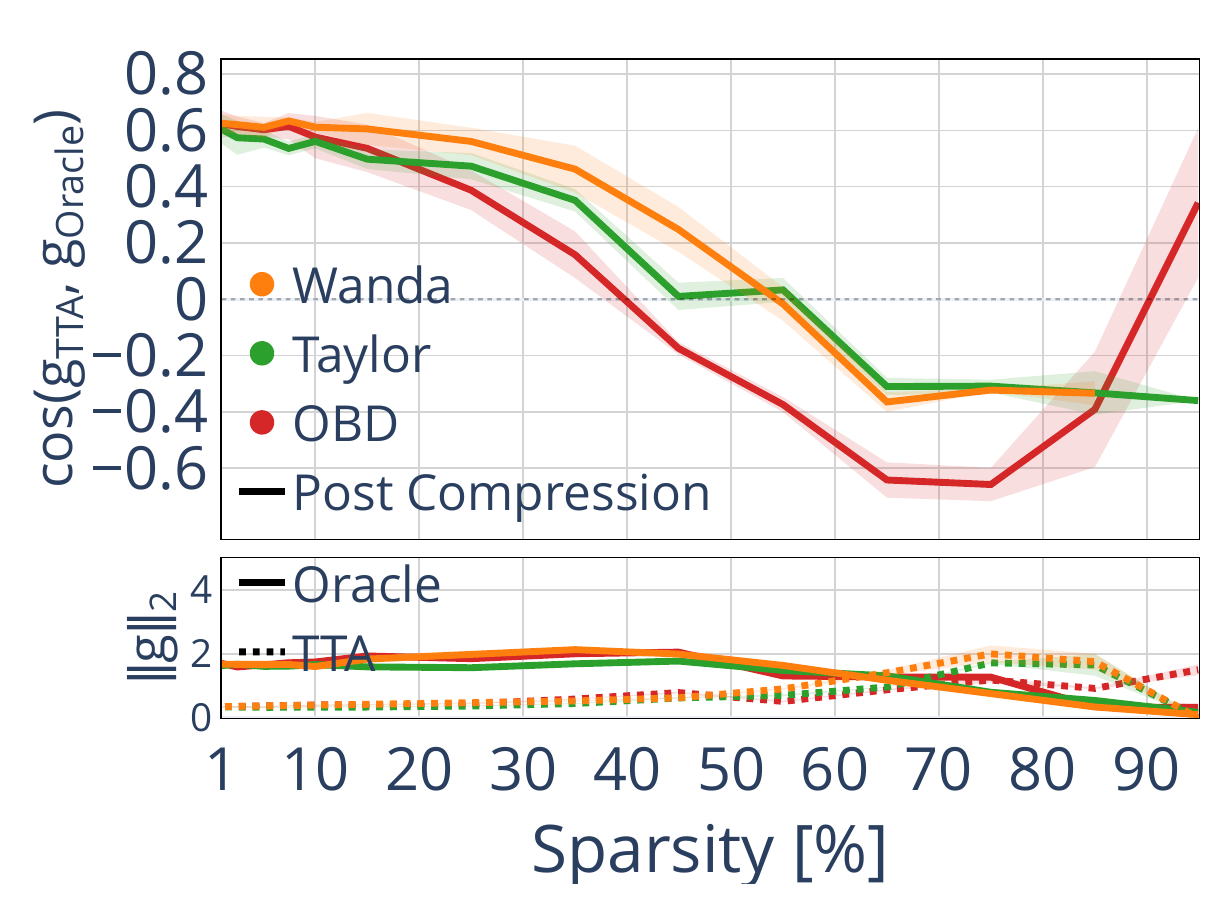}\caption{ResNet-18 / CIFAR-10-C.}\label{fig:cs_noise_data_cifar10}\end{subfigure}
\hfill
\begin{subfigure}[t]{0.32\textwidth}\centering\includegraphics[width=\textwidth]{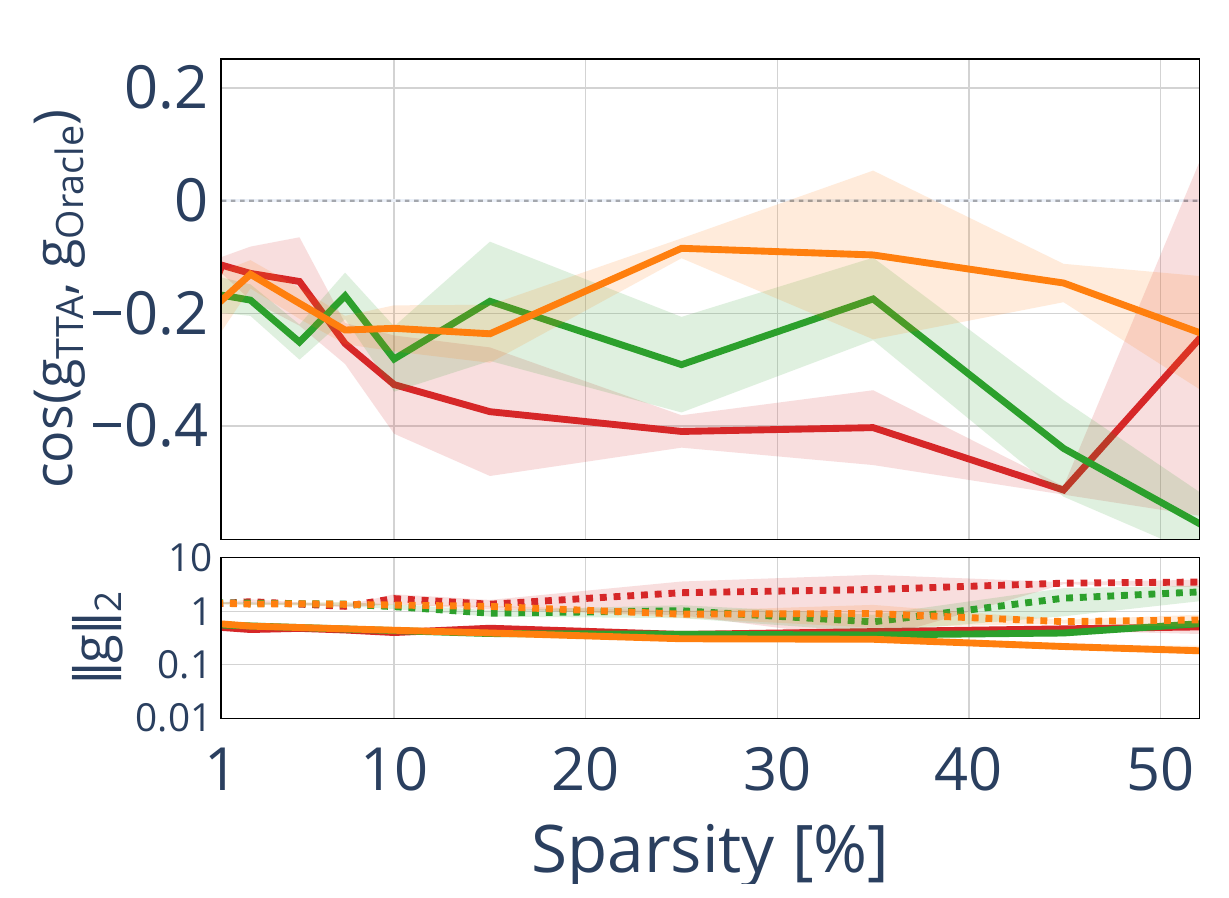}\caption{ResNet-18 / ImageNet-C.}\label{fig:cs_noise_data_imagenet}\end{subfigure}
\hfill
\begin{subfigure}[t]{0.32\textwidth}\centering\includegraphics[width=\textwidth]{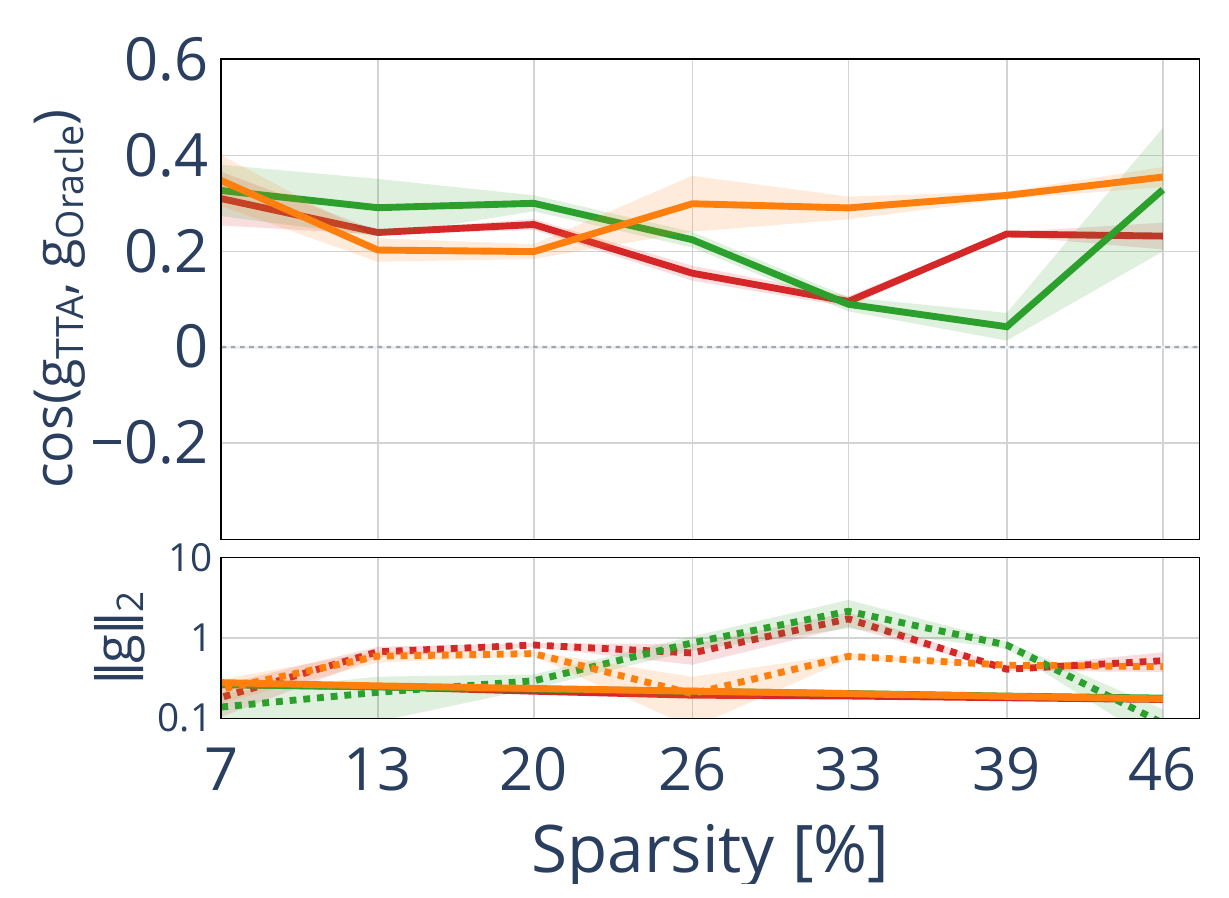}\caption{ViT-Base / ImageNet-C.}\label{fig:cs_noise_data_vit}\end{subfigure}

\vspace{0.3em}

\begin{subfigure}[t]{0.32\textwidth}\centering\includegraphics[width=\textwidth]{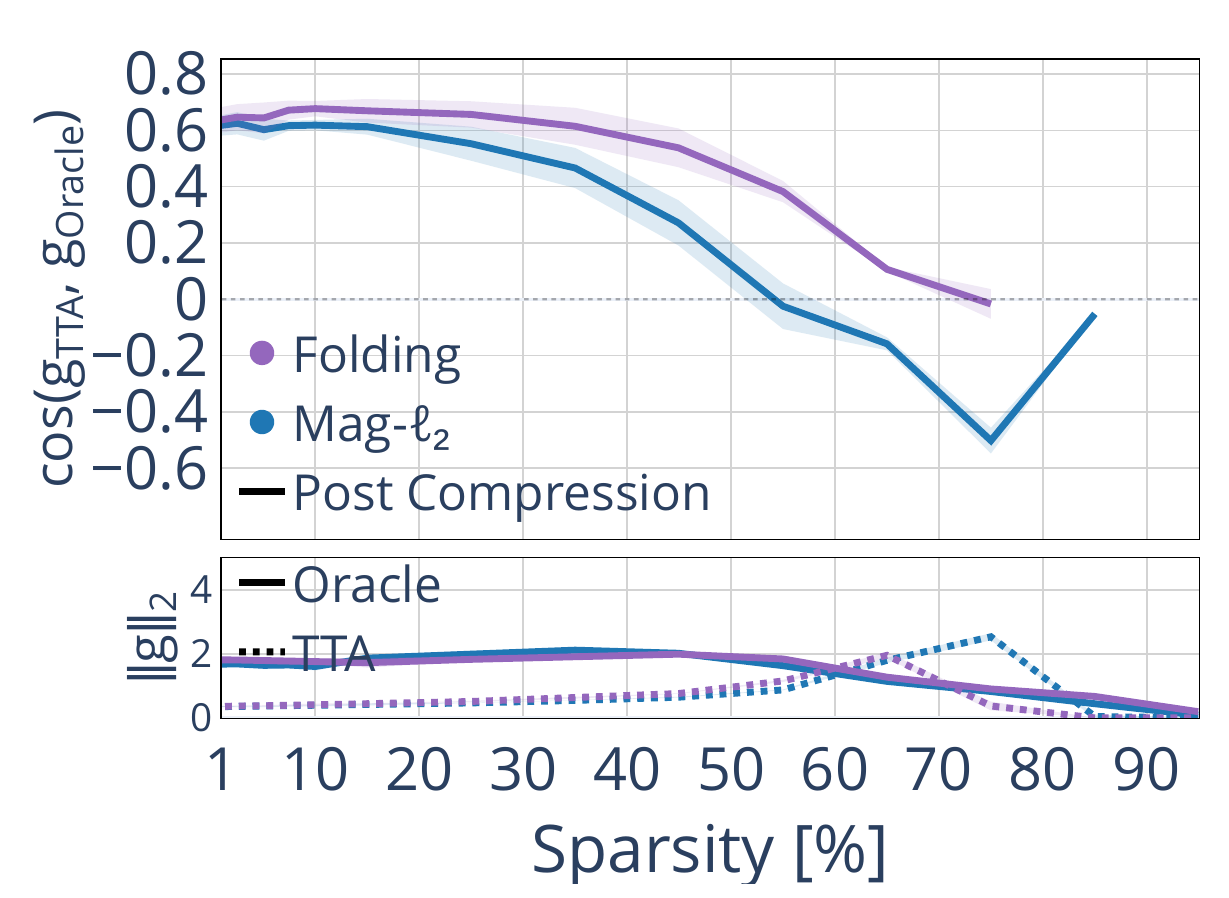}\caption{ResNet-18 / CIFAR-10-C.}\label{fig:cs_noise_free_cifar10}\end{subfigure}
\hfill
\begin{subfigure}[t]{0.32\textwidth}\centering\includegraphics[width=\textwidth]{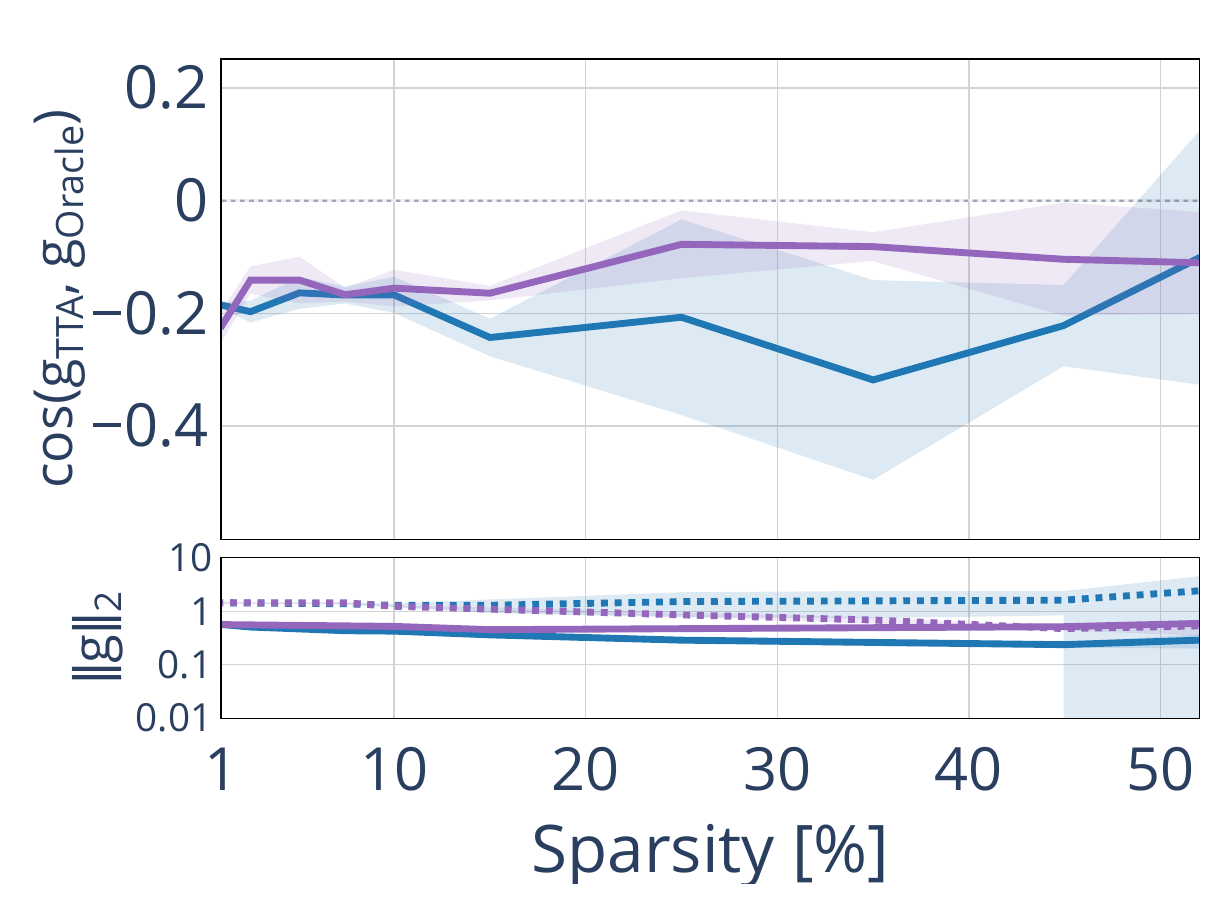}\caption{ResNet-18 / ImageNet-C.}\label{fig:cs_noise_free_imagenet}\end{subfigure}
\hfill
\begin{subfigure}[t]{0.32\textwidth}\centering\includegraphics[width=\textwidth]{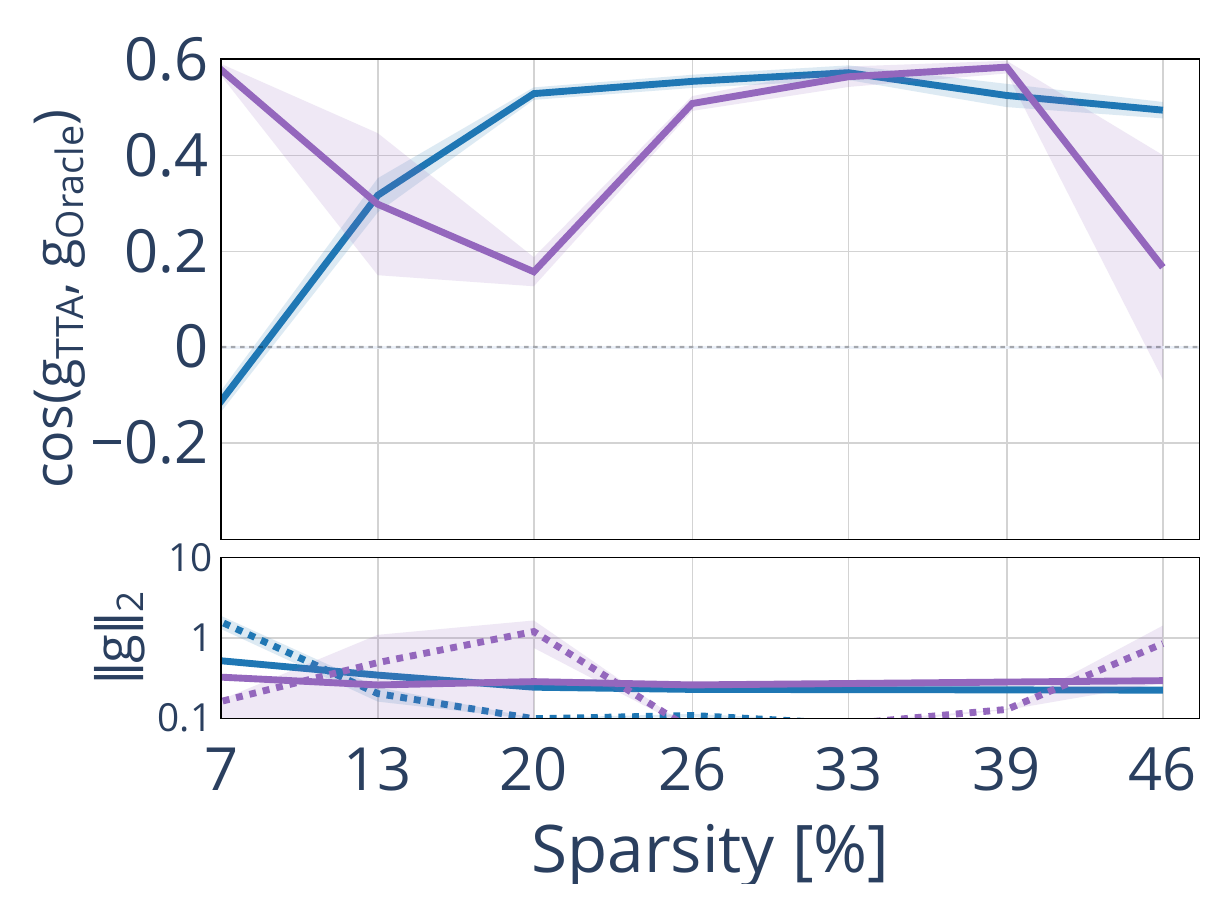}\caption{ViT-Base / ImageNet-C.}\label{fig:cs_noise_free_vit}\end{subfigure}

\caption{\textbf{Gradient alignment restricted to the \emph{Noise} corruption family} (Gaussian, Shot, Impulse). Same visualization scheme as Fig.~\ref{fig:cosine_similarity}; $\pm \sigma$ band across the three Noise corruptions.}
\label{fig:cs_noise}
\end{figure*}

\begin{figure*}[h]
\centering
\begin{subfigure}[t]{0.32\textwidth}\centering\includegraphics[width=\textwidth]{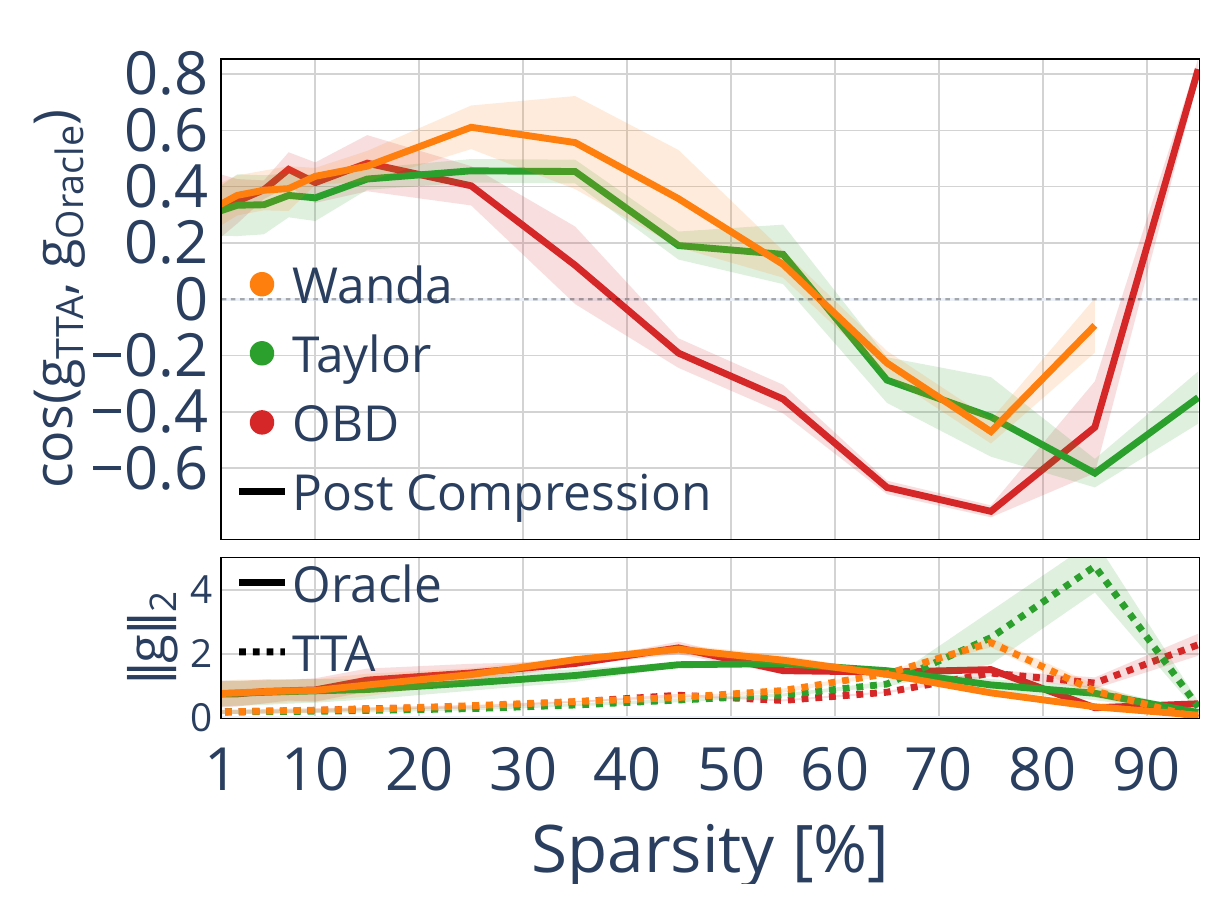}\caption{ResNet-18 / CIFAR-10-C.}\label{fig:cs_blur_data_cifar10}\end{subfigure}
\hfill
\begin{subfigure}[t]{0.32\textwidth}\centering\includegraphics[width=\textwidth]{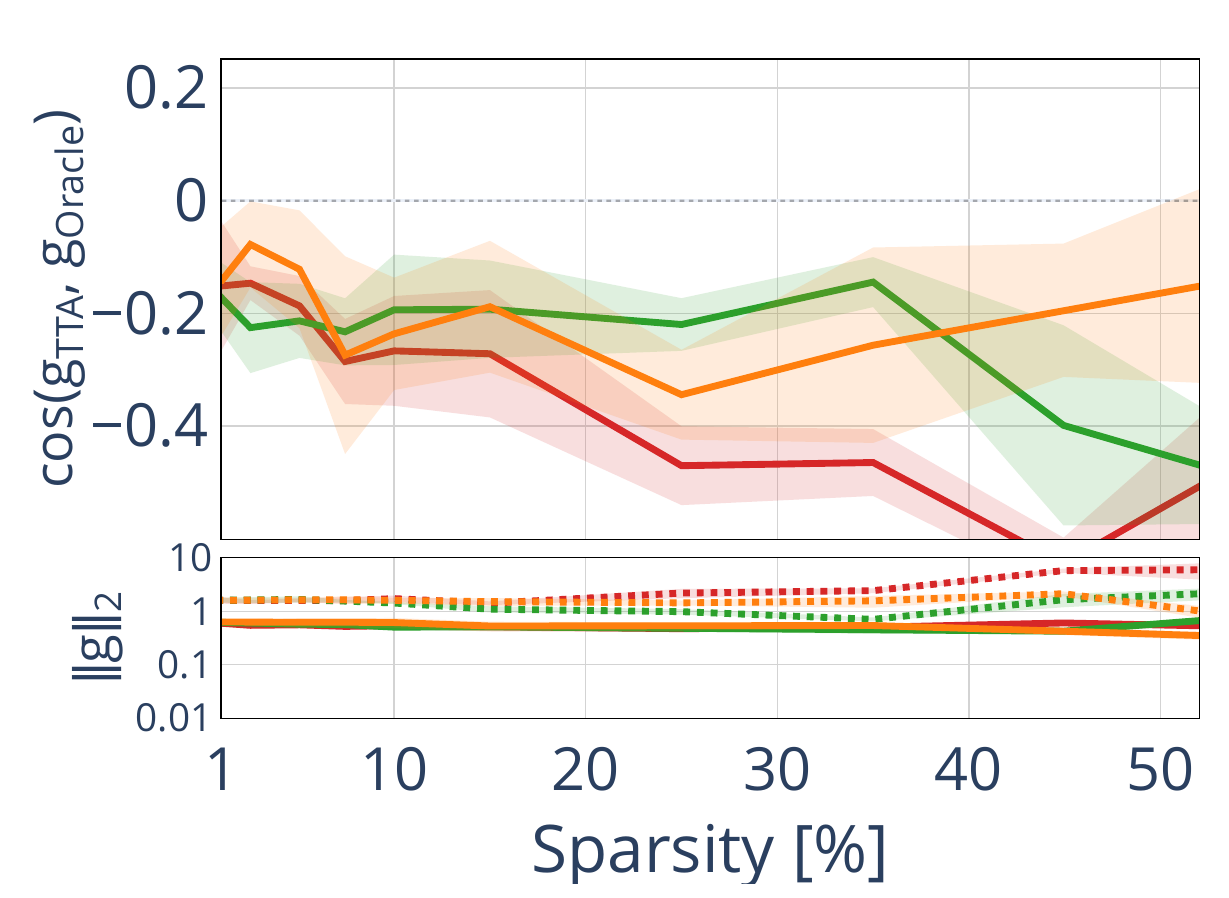}\caption{ResNet-18 / ImageNet-C.}\label{fig:cs_blur_data_imagenet}\end{subfigure}
\hfill
\begin{subfigure}[t]{0.32\textwidth}\centering\includegraphics[width=\textwidth]{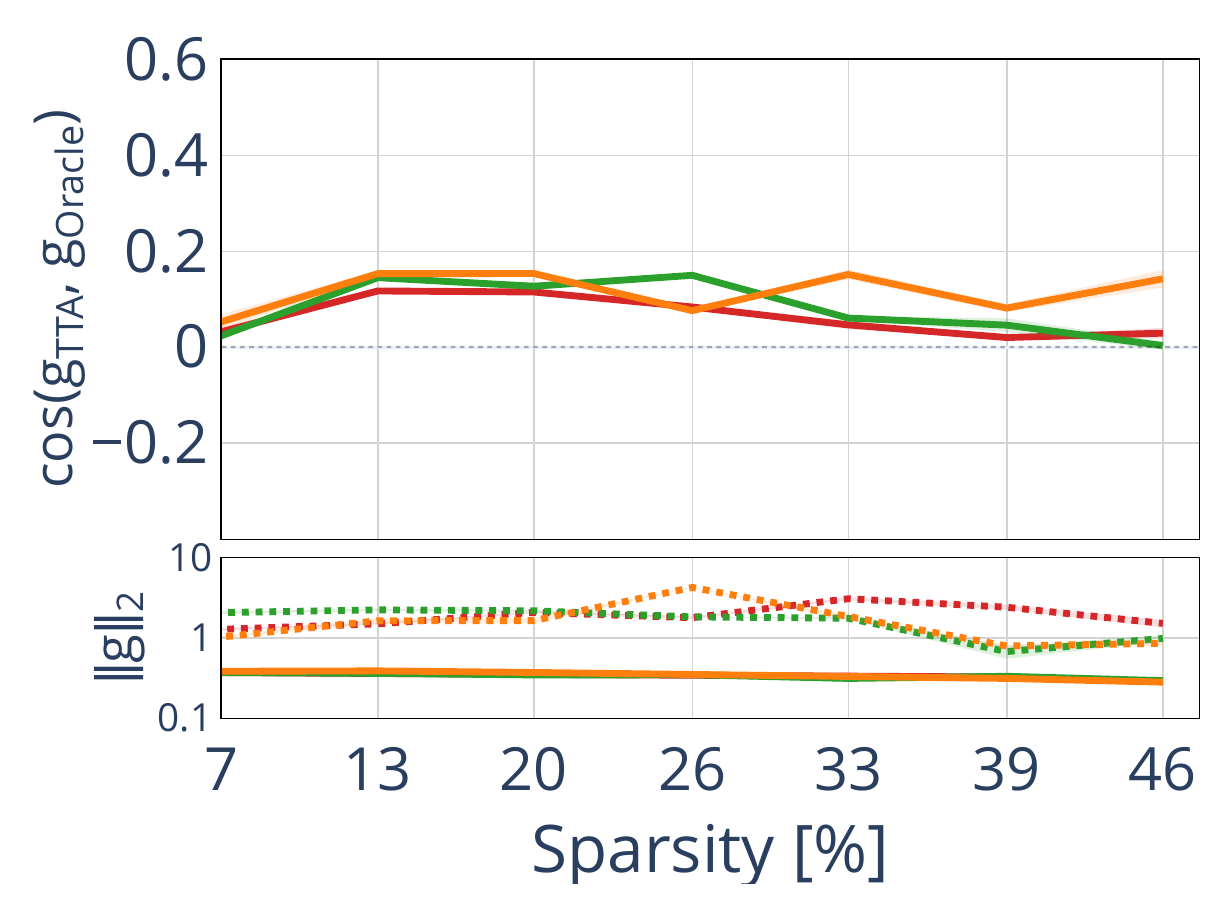}\caption{ViT-Base / ImageNet-C.}\label{fig:cs_blur_data_vit}\end{subfigure}

\vspace{0.3em}

\begin{subfigure}[t]{0.32\textwidth}\centering\includegraphics[width=\textwidth]{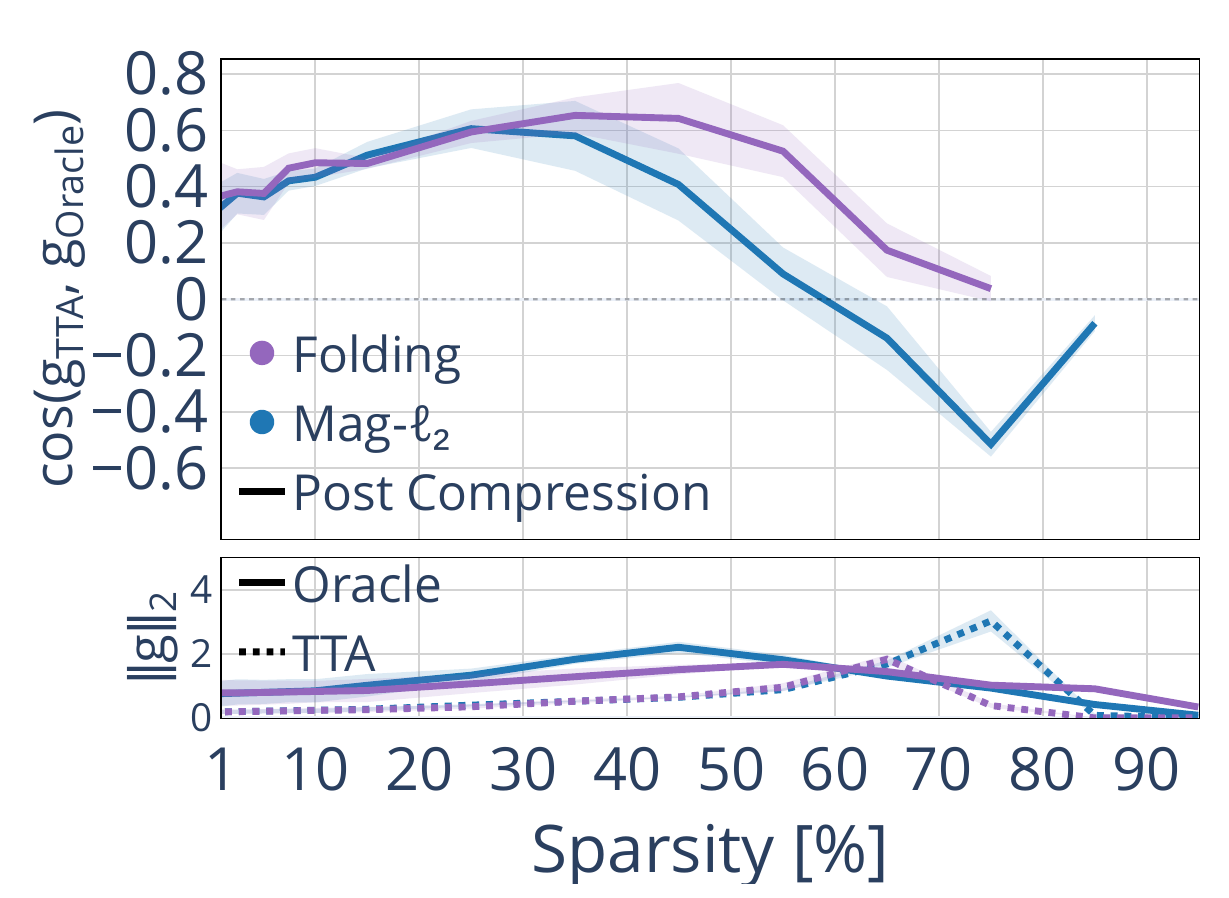}\caption{ResNet-18 / CIFAR-10-C.}\label{fig:cs_blur_free_cifar10}\end{subfigure}
\hfill
\begin{subfigure}[t]{0.32\textwidth}\centering\includegraphics[width=\textwidth]{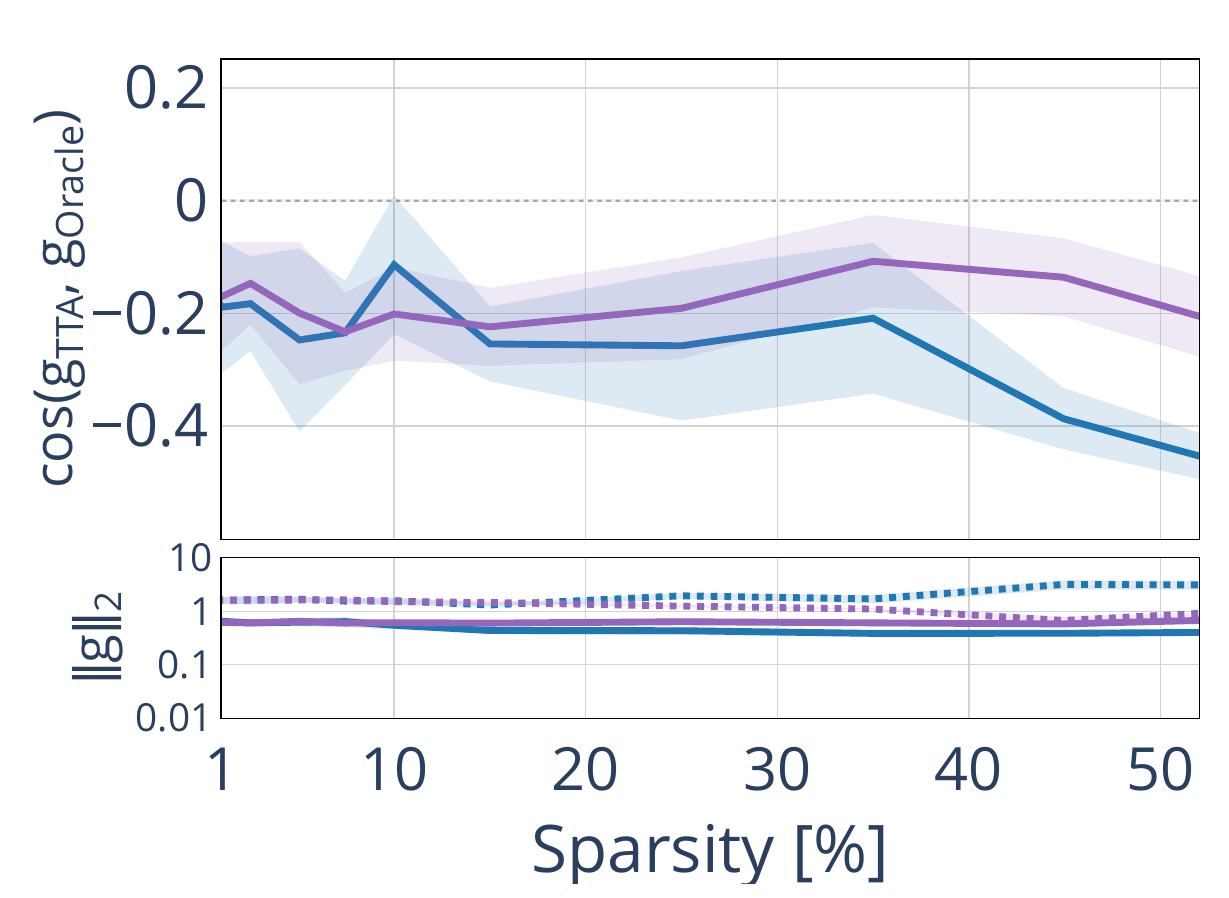}\caption{ResNet-18 / ImageNet-C.}\label{fig:cs_blur_free_imagenet}\end{subfigure}
\hfill
\begin{subfigure}[t]{0.32\textwidth}\centering\includegraphics[width=\textwidth]{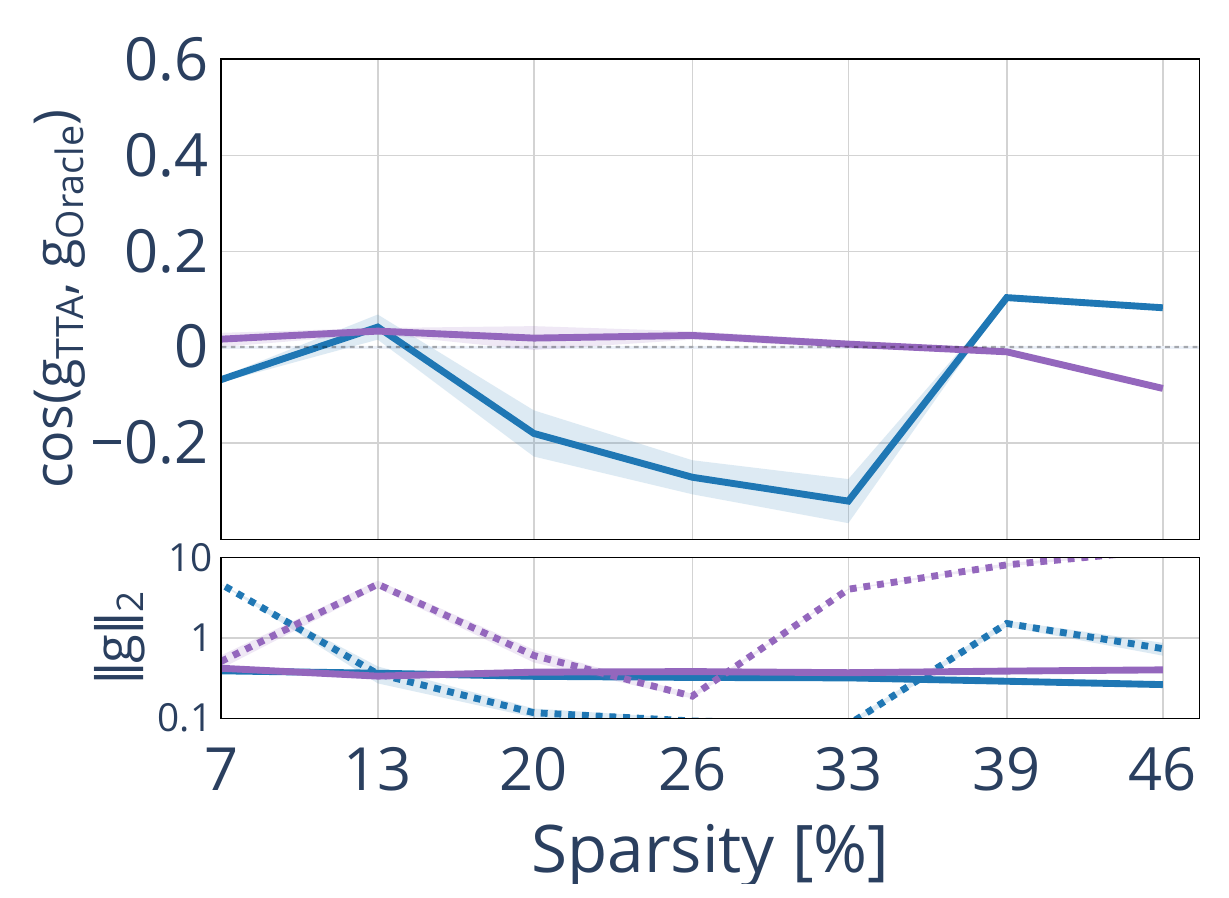}\caption{ViT-Base / ImageNet-C.}\label{fig:cs_blur_free_vit}\end{subfigure}

\caption{\textbf{Gradient alignment restricted to the \emph{Blur} corruption family} (Defocus, Glass, Motion, Zoom). Same visualization scheme as Fig.~\ref{fig:cosine_similarity}; $\pm \sigma$ band across the four Blur corruptions.}
\label{fig:cs_blur}
\end{figure*}

\begin{figure*}[h]
\centering
\begin{subfigure}[t]{0.32\textwidth}\centering\includegraphics[width=\textwidth]{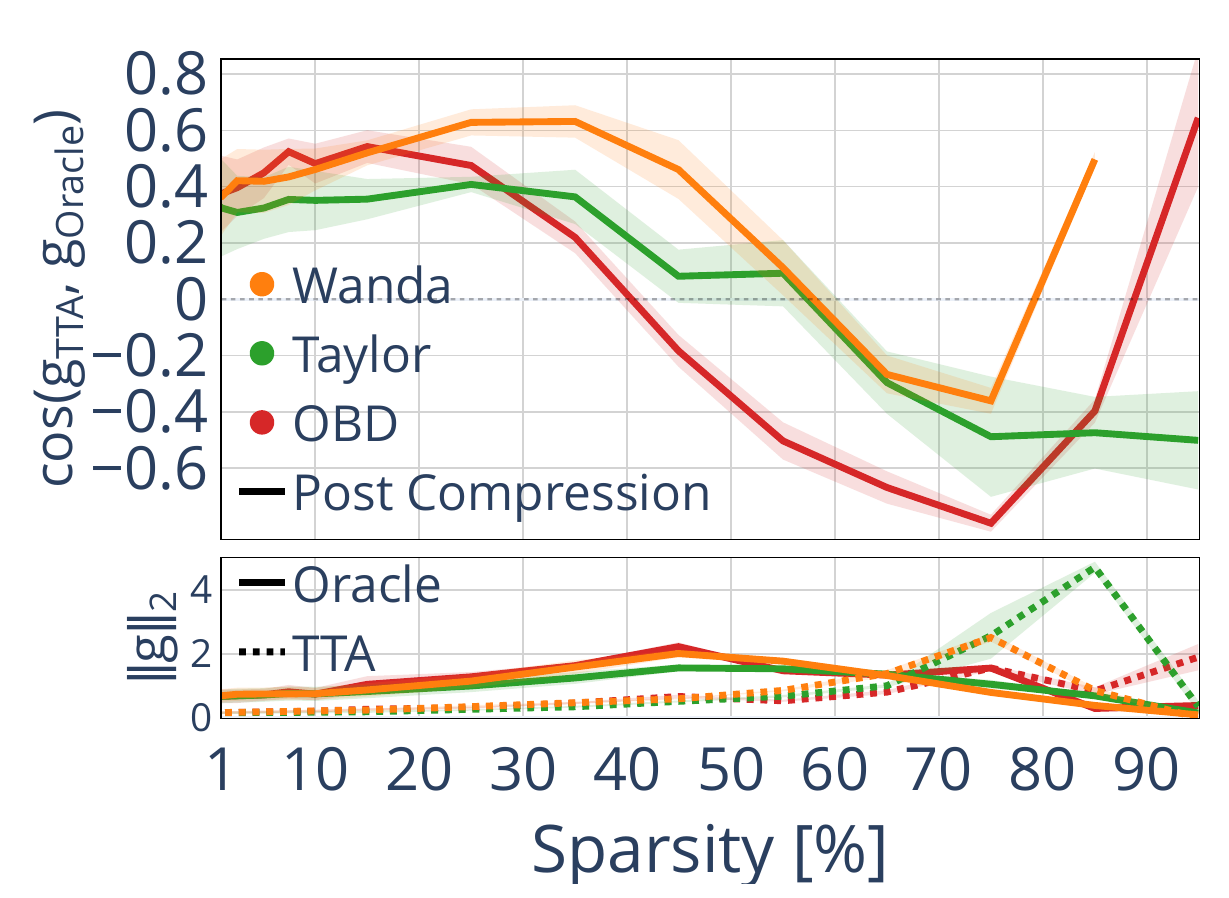}\caption{ResNet-18 / CIFAR-10-C.}\label{fig:cs_weather_data_cifar10}\end{subfigure}
\hfill
\begin{subfigure}[t]{0.32\textwidth}\centering\includegraphics[width=\textwidth]{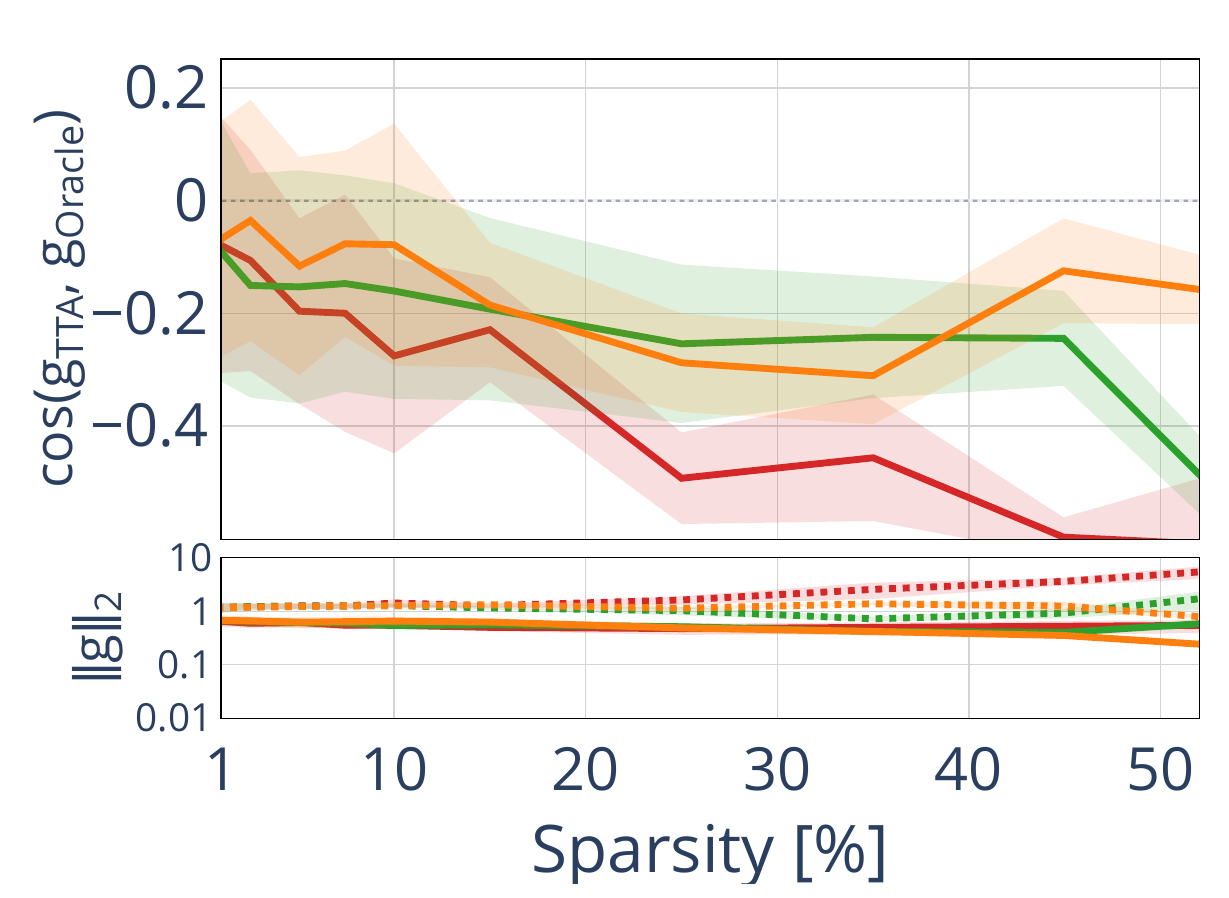}\caption{ResNet-18 / ImageNet-C.}\label{fig:cs_weather_data_imagenet}\end{subfigure}
\hfill
\begin{subfigure}[t]{0.32\textwidth}\centering\includegraphics[width=\textwidth]{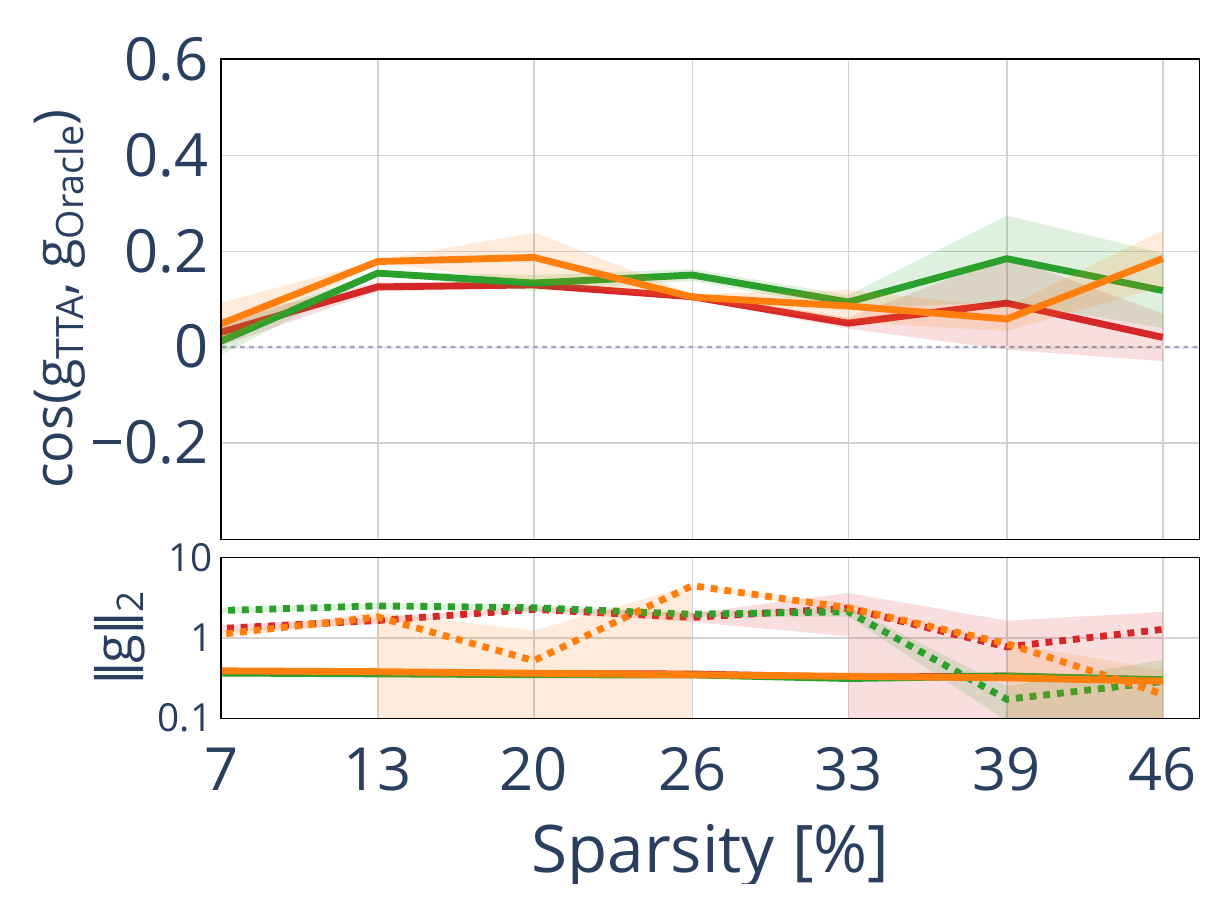}\caption{ViT-Base / ImageNet-C.}\label{fig:cs_weather_data_vit}\end{subfigure}

\vspace{0.3em}

\begin{subfigure}[t]{0.32\textwidth}\centering\includegraphics[width=\textwidth]{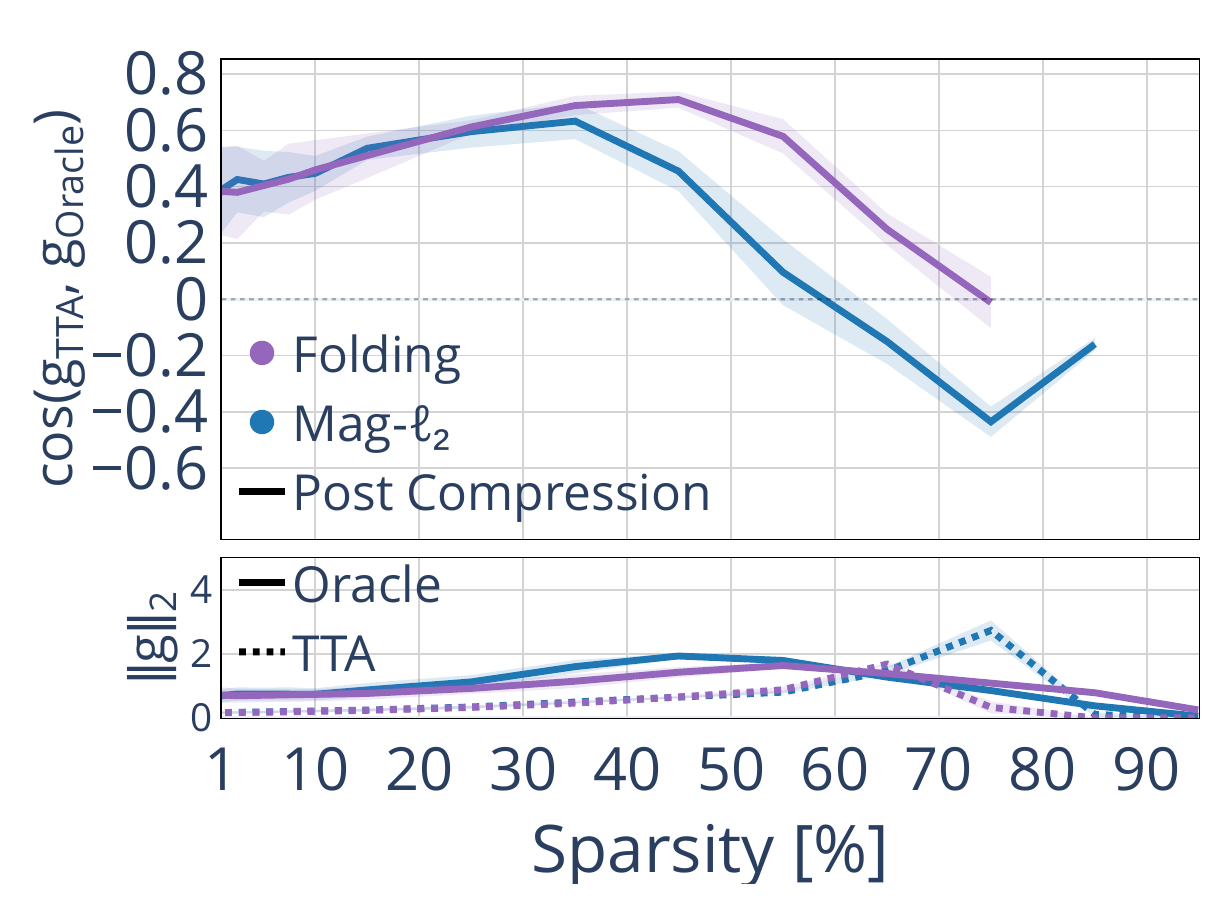}\caption{ResNet-18 / CIFAR-10-C.}\label{fig:cs_weather_free_cifar10}\end{subfigure}
\hfill
\begin{subfigure}[t]{0.32\textwidth}\centering\includegraphics[width=\textwidth]{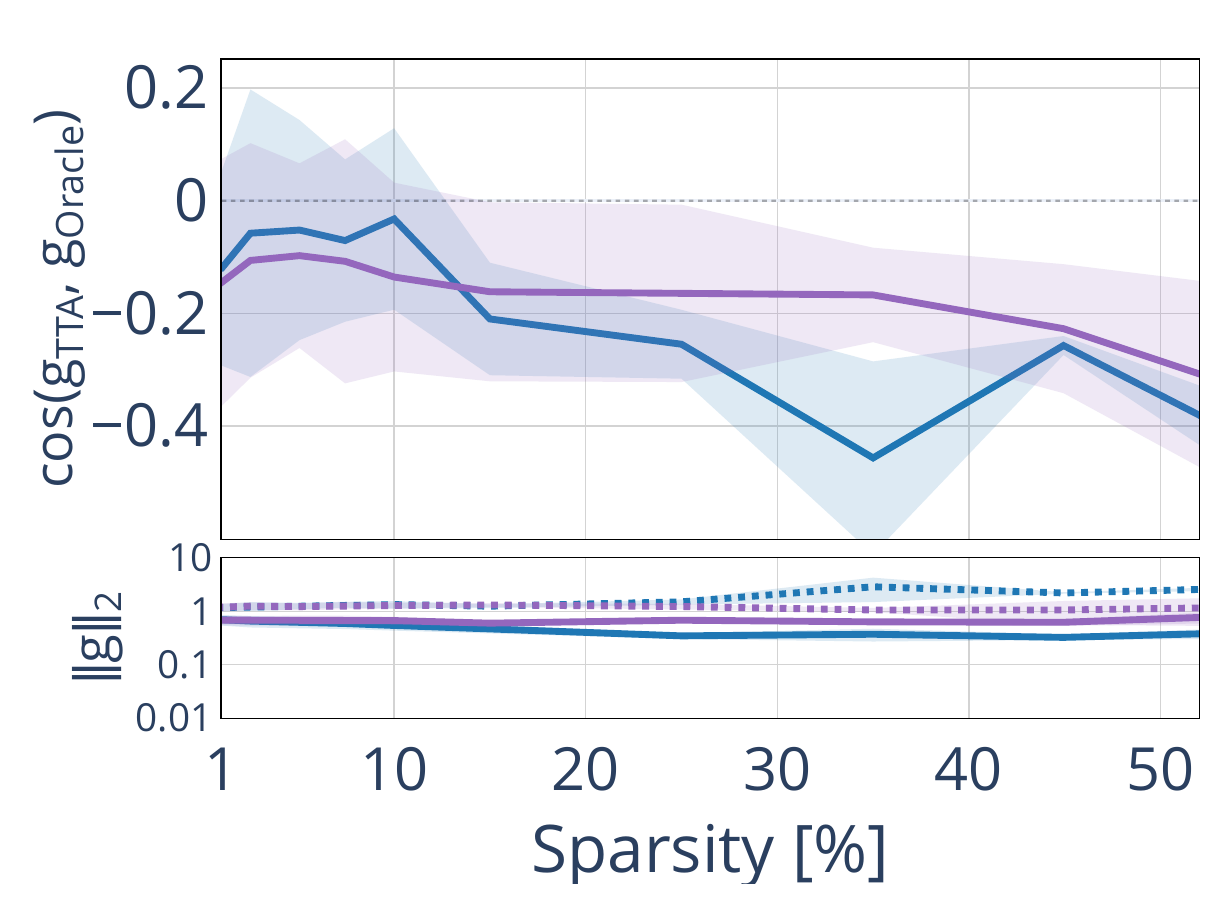}\caption{ResNet-18 / ImageNet-C.}\label{fig:cs_weather_free_imagenet}\end{subfigure}
\hfill
\begin{subfigure}[t]{0.32\textwidth}\centering\includegraphics[width=\textwidth]{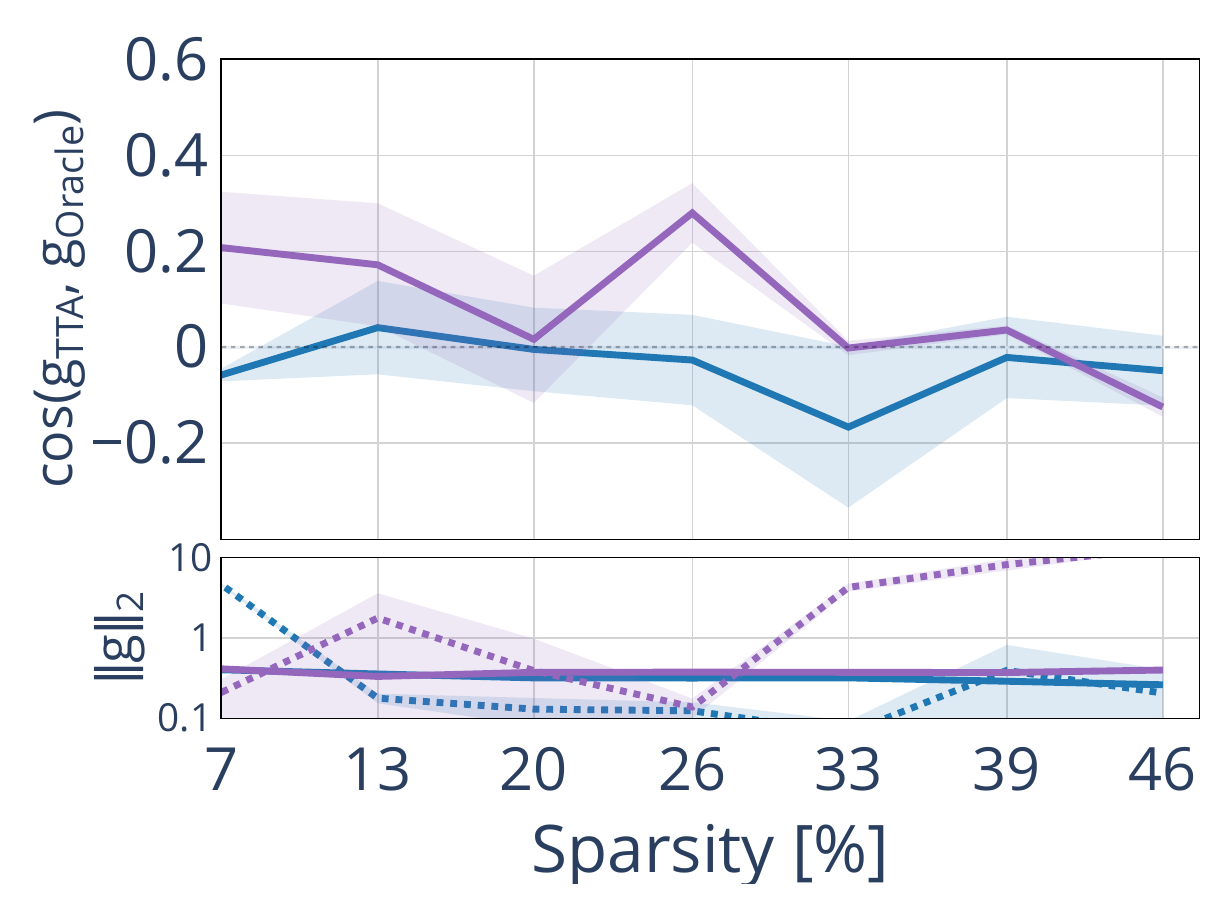}\caption{ViT-Base / ImageNet-C.}\label{fig:cs_weather_free_vit}\end{subfigure}

\caption{\textbf{Gradient alignment restricted to the \emph{Weather} corruption family} (Snow, Frost, Fog, Brightness). Same visualization scheme as Fig.~\ref{fig:cosine_similarity}; $\pm \sigma$ band across the four Weather corruptions.}
\label{fig:cs_weather}
\end{figure*}

\begin{figure*}[h]
\centering
\begin{subfigure}[t]{0.32\textwidth}\centering\includegraphics[width=\textwidth]{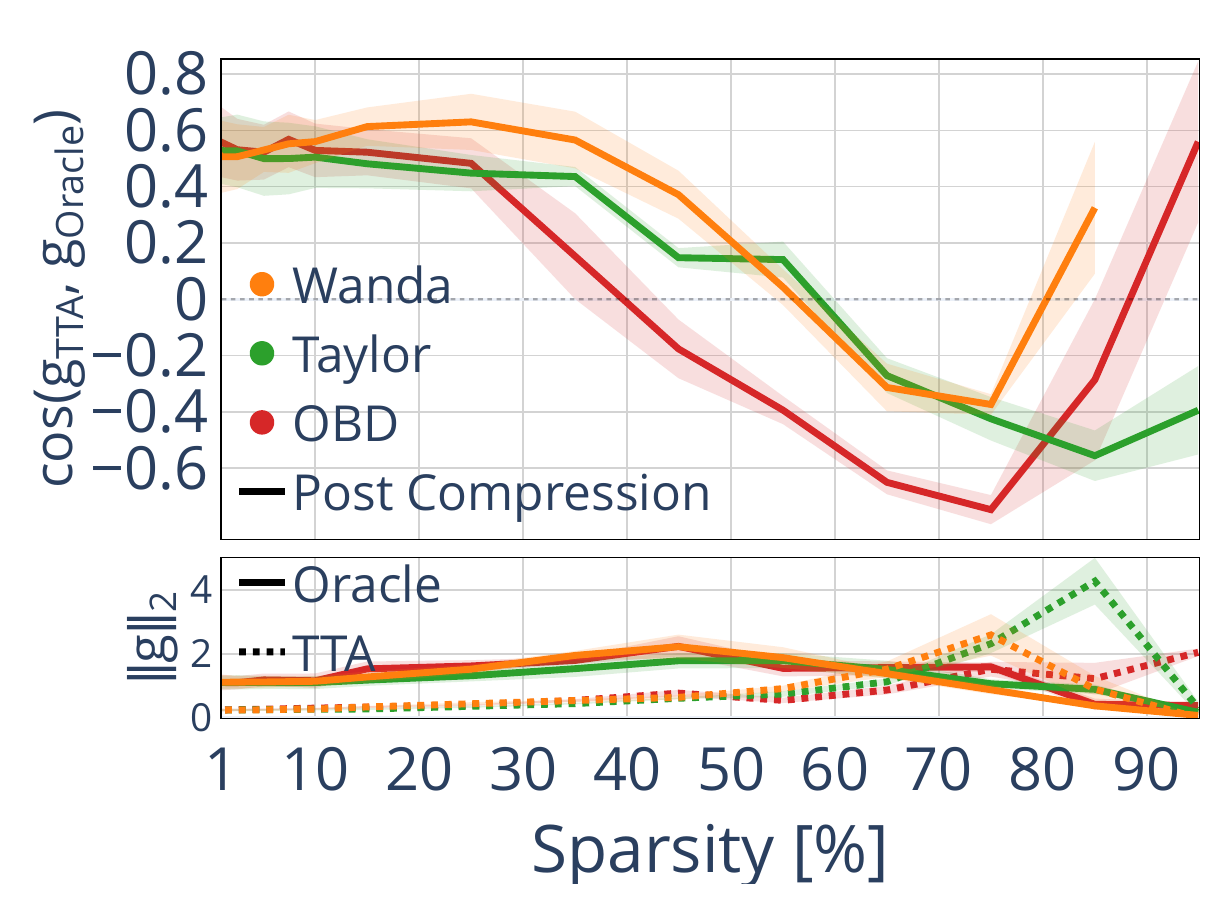}\caption{ResNet-18 / CIFAR-10-C.}\label{fig:cs_digital_data_cifar10}\end{subfigure}
\hfill
\begin{subfigure}[t]{0.32\textwidth}\centering\includegraphics[width=\textwidth]{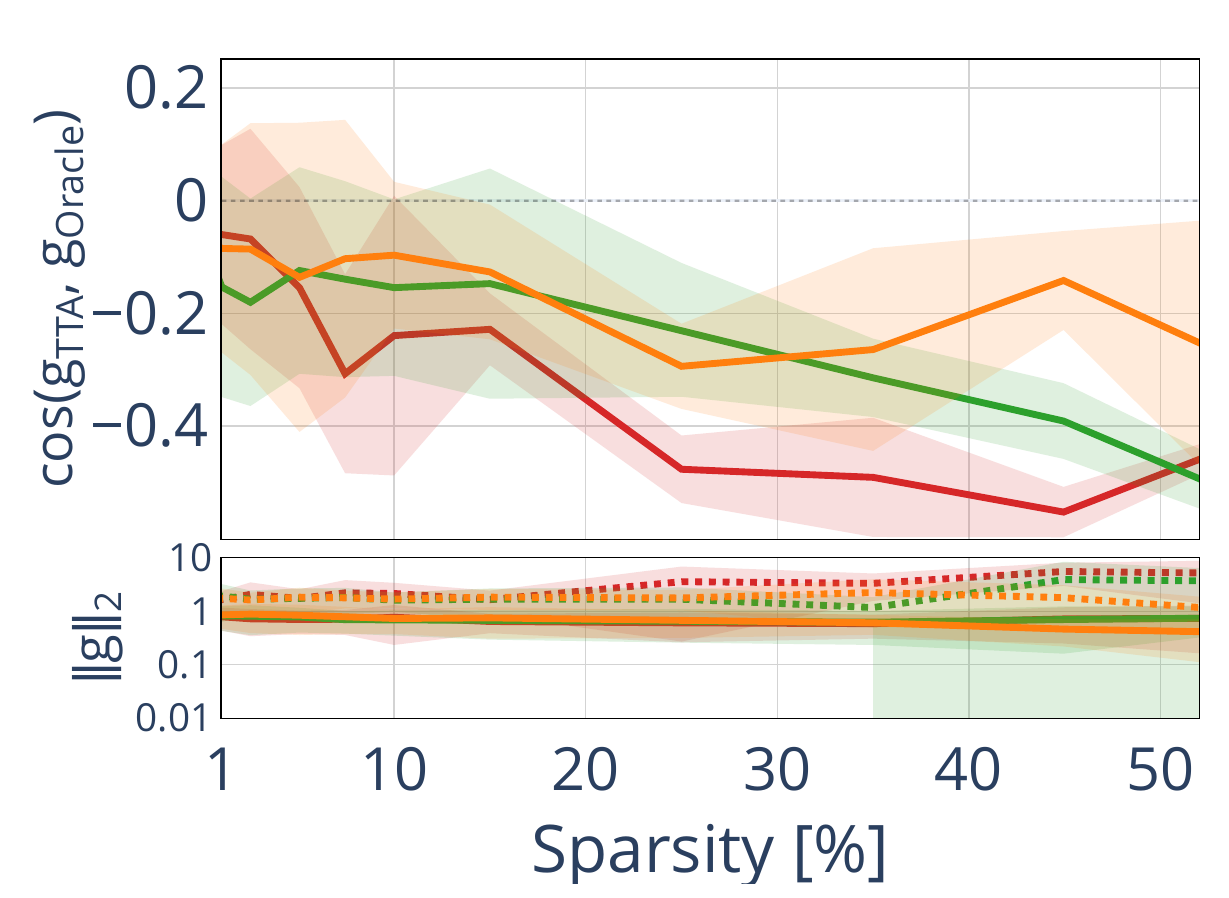}\caption{ResNet-18 / ImageNet-C.}\label{fig:cs_digital_data_imagenet}\end{subfigure}
\hfill
\begin{subfigure}[t]{0.32\textwidth}\centering\includegraphics[width=\textwidth]{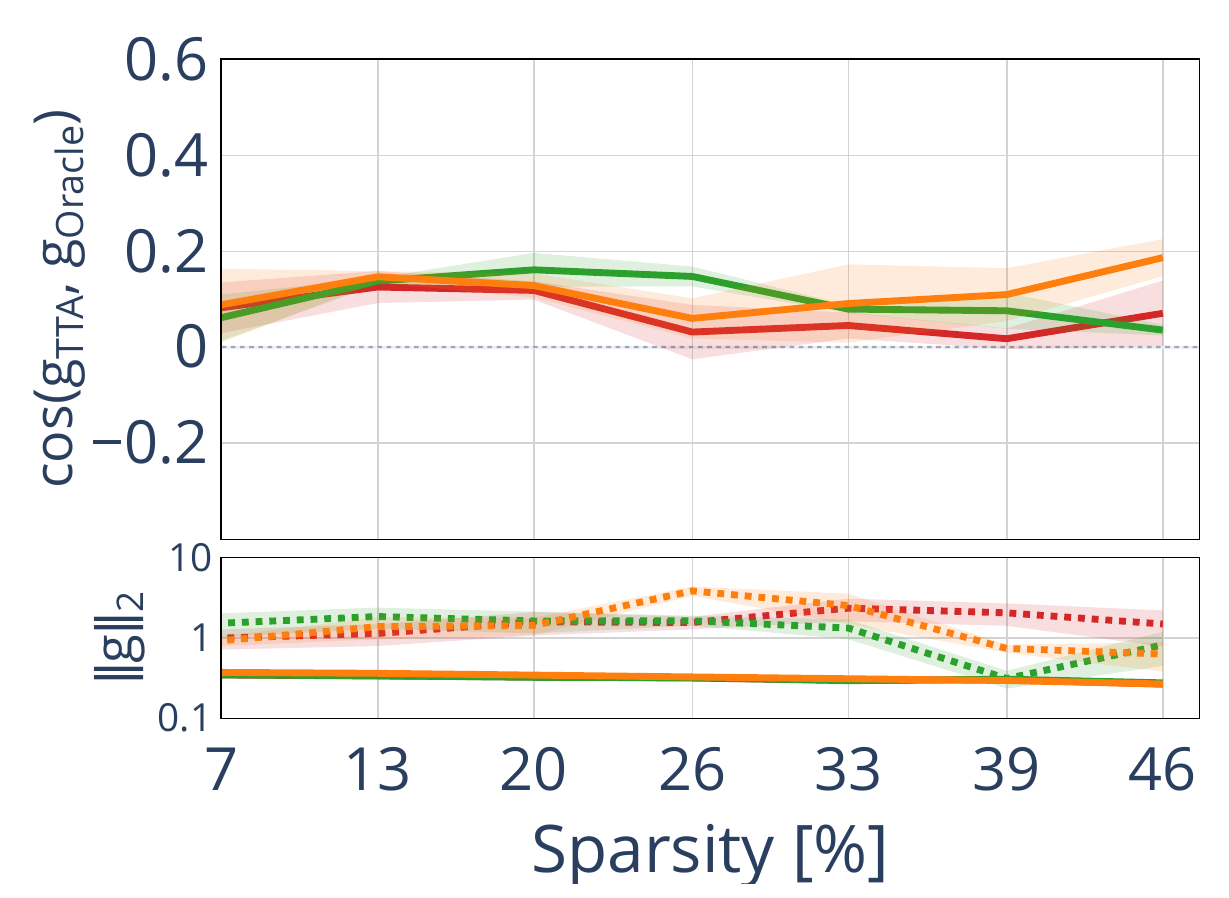}\caption{ViT-Base / ImageNet-C.}\label{fig:cs_digital_data_vit}\end{subfigure}

\vspace{0.3em}

\begin{subfigure}[t]{0.32\textwidth}\centering\includegraphics[width=\textwidth]{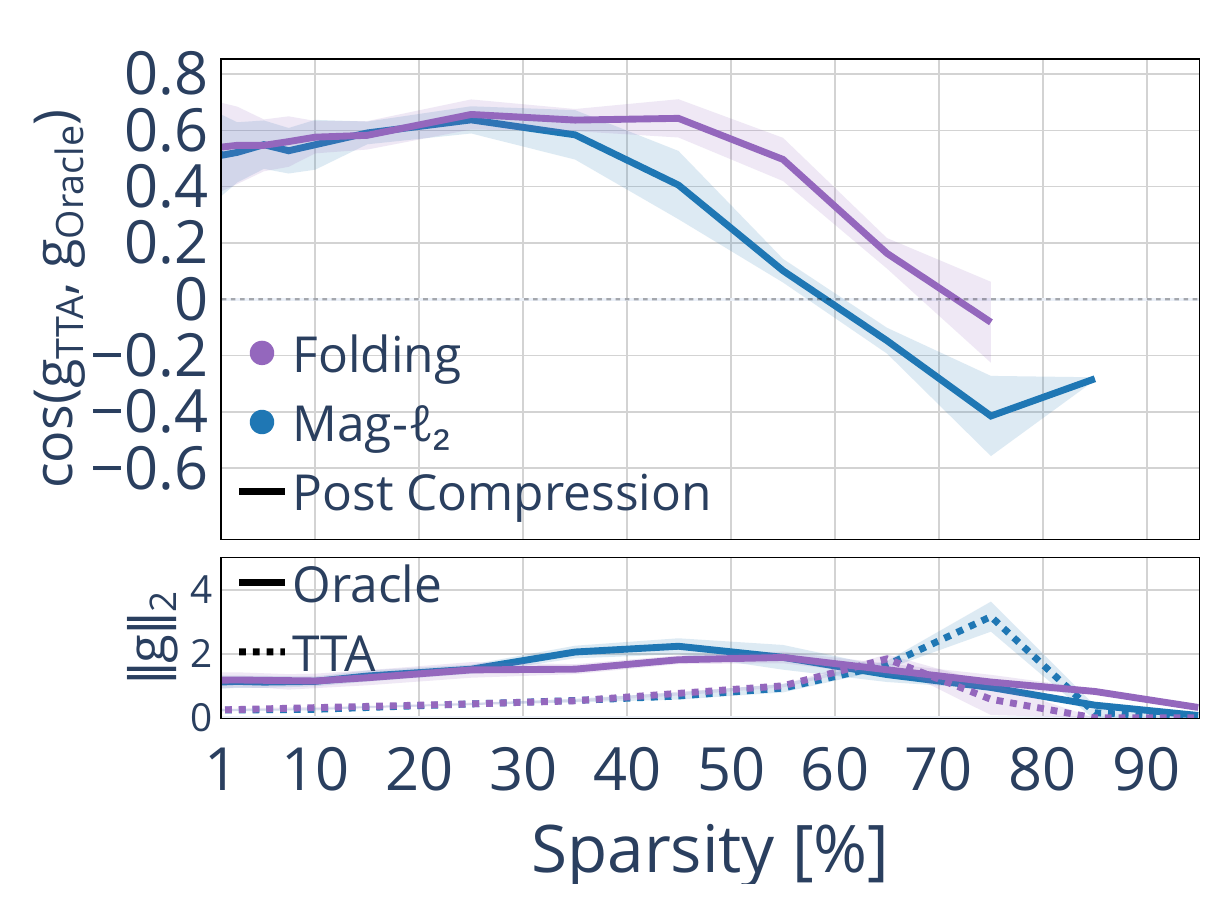}\caption{ResNet-18 / CIFAR-10-C.}\label{fig:cs_digital_free_cifar10}\end{subfigure}
\hfill
\begin{subfigure}[t]{0.32\textwidth}\centering\includegraphics[width=\textwidth]{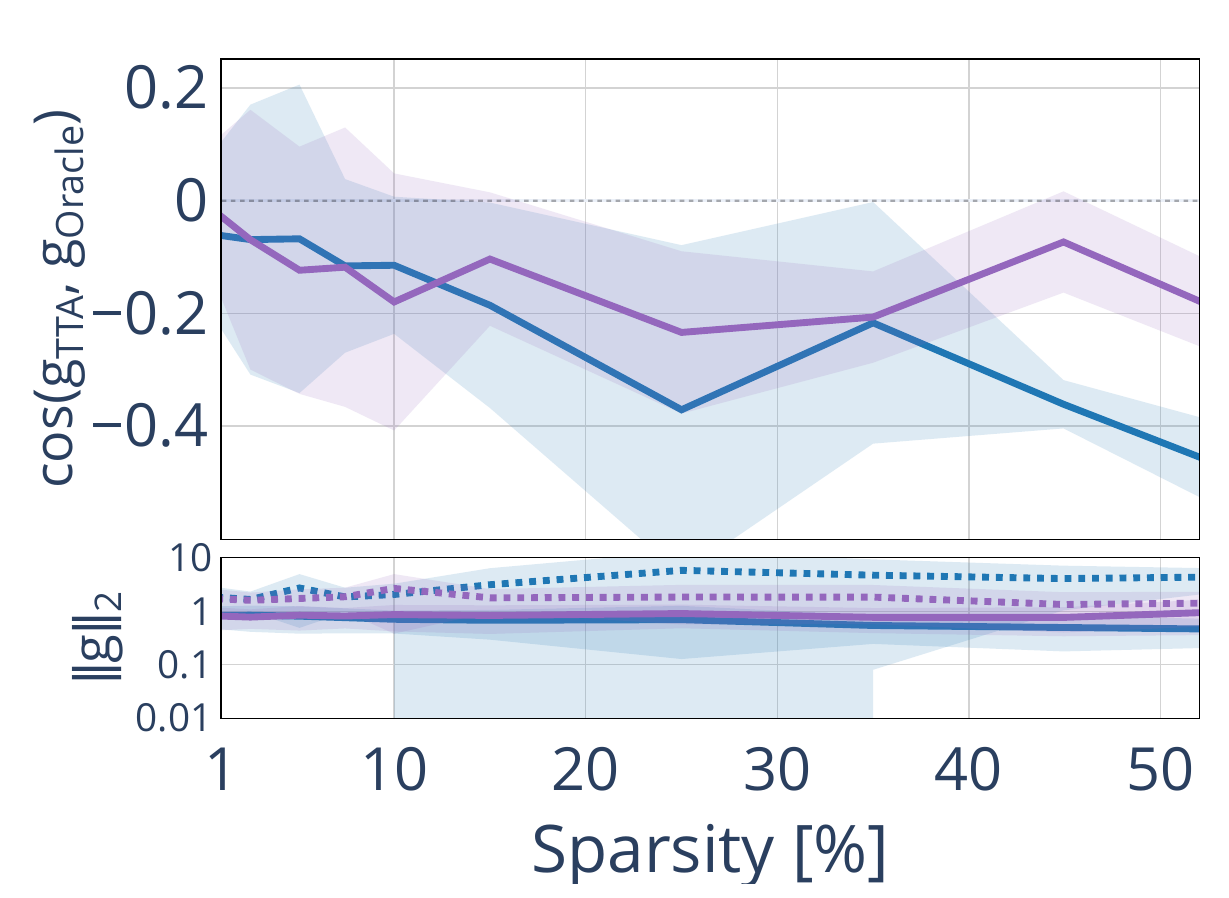}\caption{ResNet-18 / ImageNet-C.}\label{fig:cs_digital_free_imagenet}\end{subfigure}
\hfill
\begin{subfigure}[t]{0.32\textwidth}\centering\includegraphics[width=\textwidth]{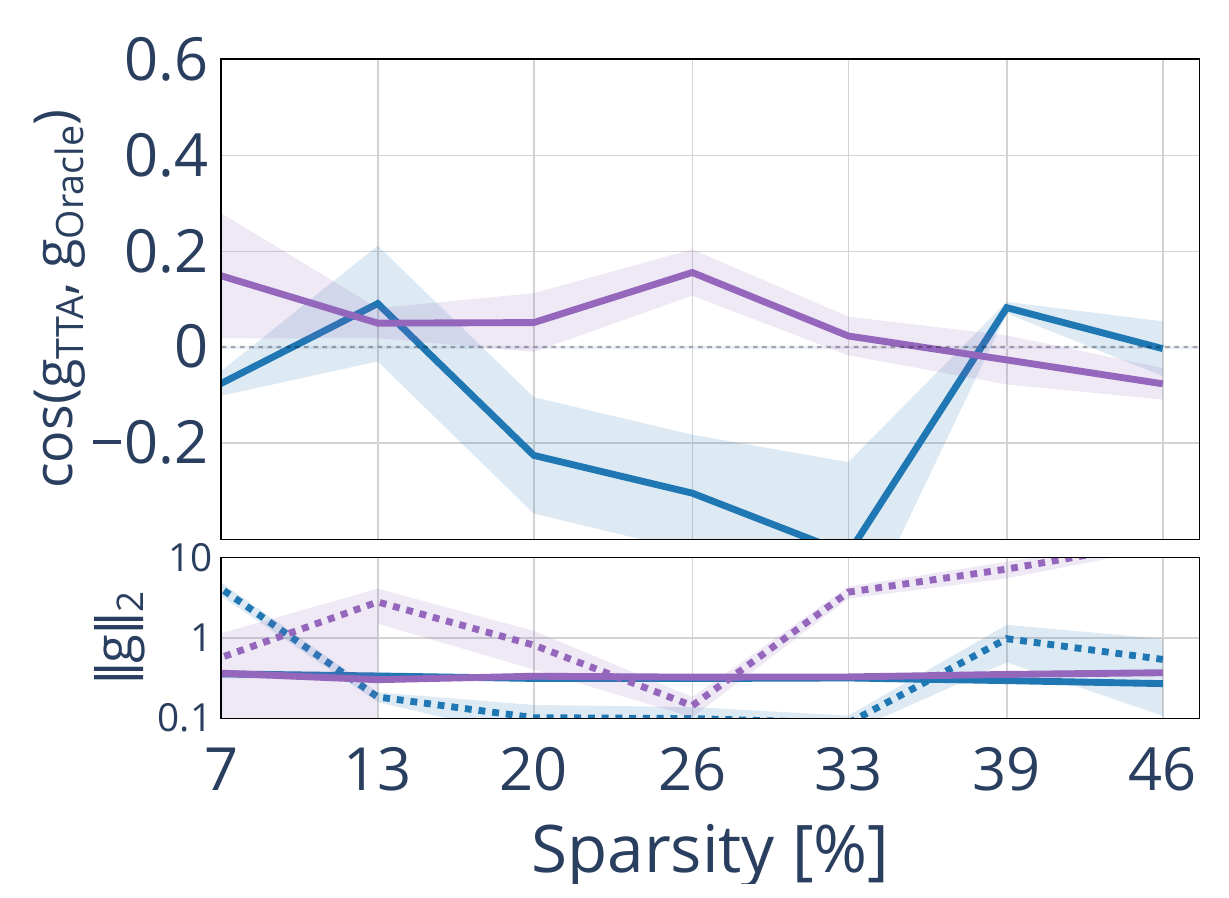}\caption{ViT-Base / ImageNet-C.}\label{fig:cs_digital_free_vit}\end{subfigure}

\caption{\textbf{Gradient alignment restricted to the \emph{Digital} corruption family} (Contrast, Elastic, Pixelate, JPEG). Same visualization scheme as Fig.~\ref{fig:cosine_similarity}; $\pm \sigma$ band across the four Digital corruptions.}
\label{fig:cs_digital}
\end{figure*}

\begin{figure*}[h]

\centering
\begin{subfigure}[t]{0.33\textwidth}
\centering
\includegraphics[width=\textwidth]{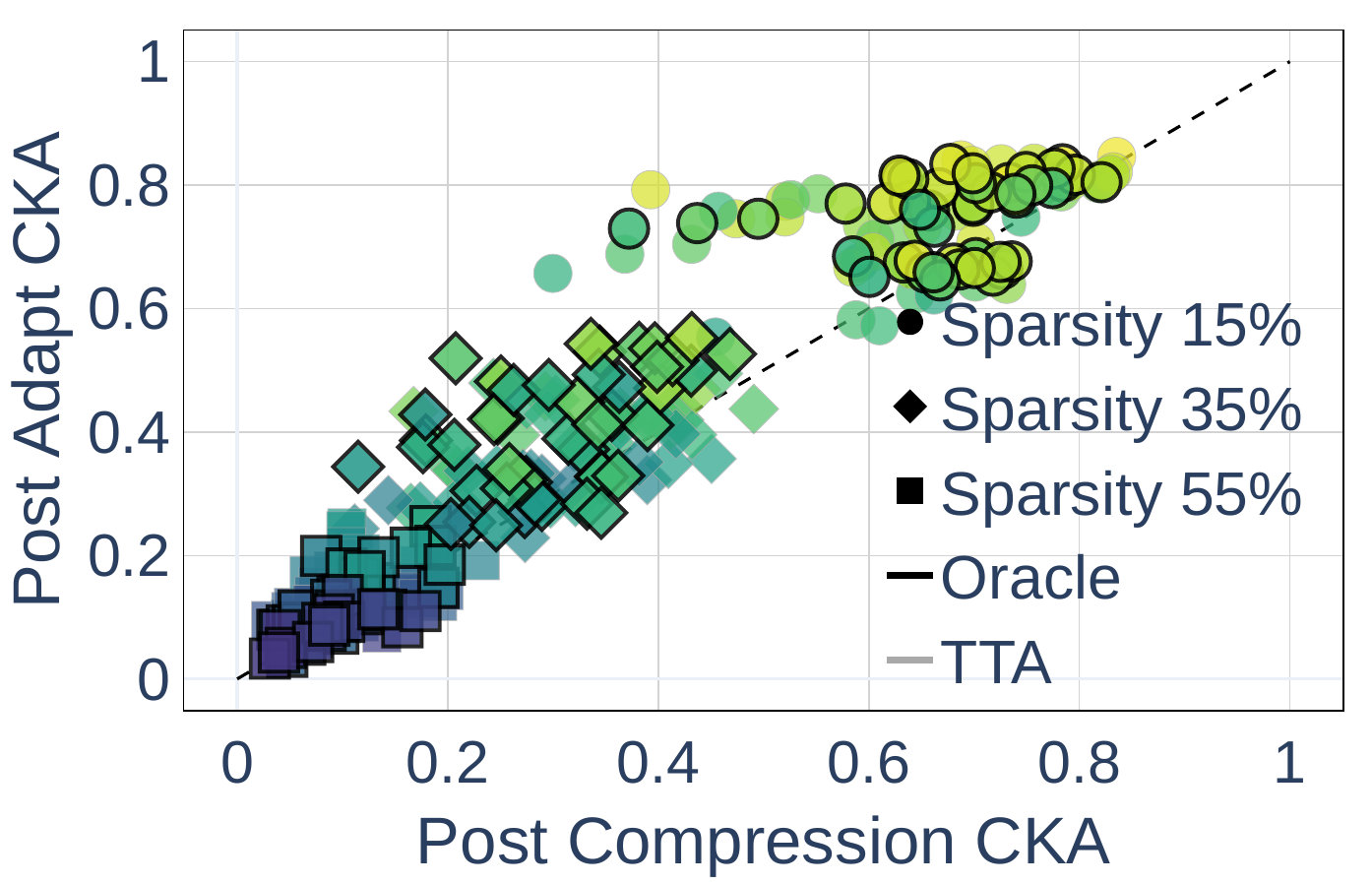}
\caption{ResNet-18 on CIFAR-10-C}
\label{fig:cka_data_cifar10}
\end{subfigure}
\hfill
\begin{subfigure}[t]{0.33\textwidth}
\centering
\includegraphics[width=\textwidth]{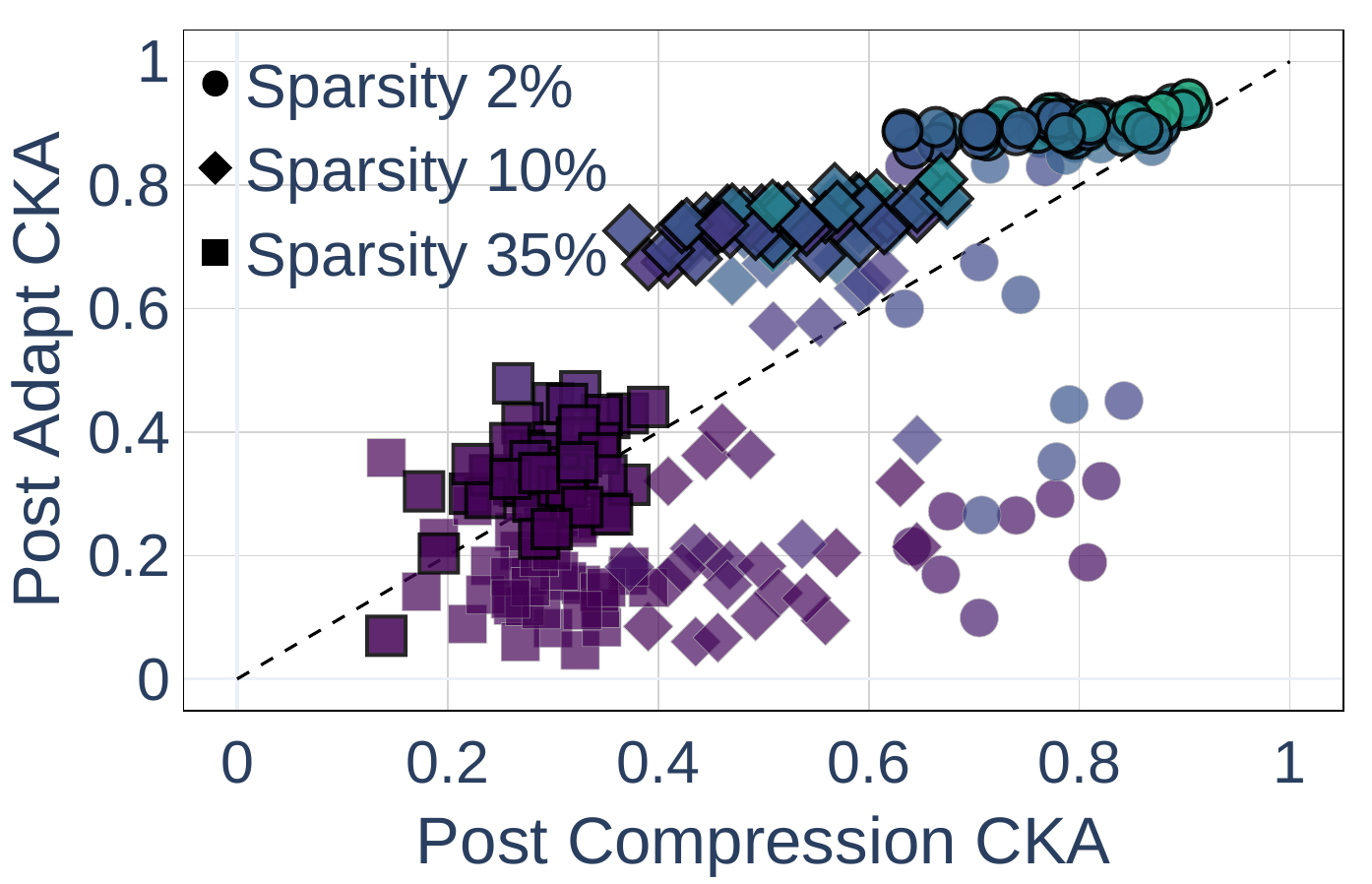}
\caption{ResNet-18 on ImageNet-C}
\label{fig:cka_data_imagenet}
\end{subfigure}
\hfill
\begin{subfigure}[t]{0.33\textwidth}
\centering
\includegraphics[width=\textwidth]{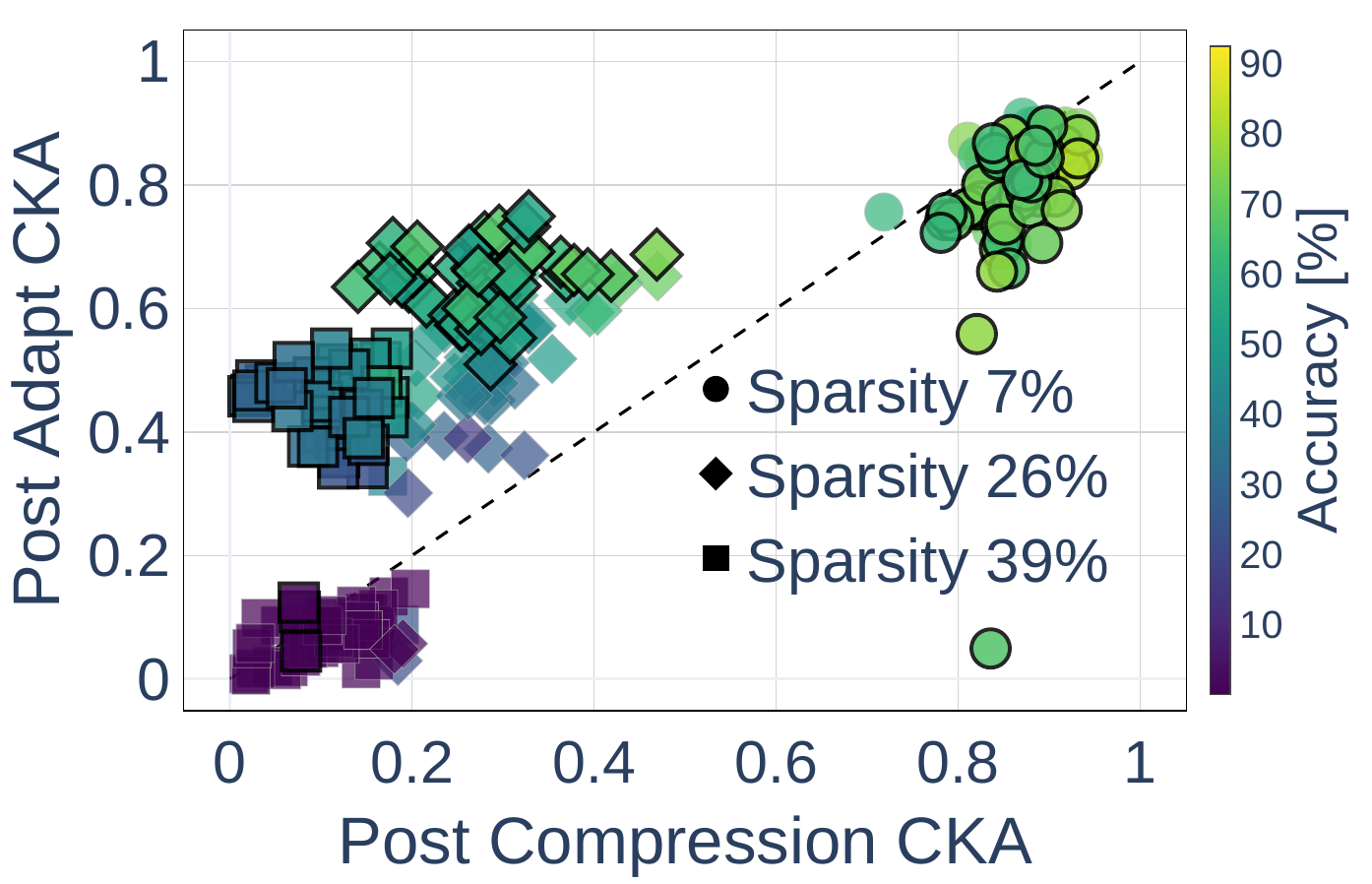}
\caption{ViT-Base on ImageNet-C}
\label{fig:cka_data_vit}
\end{subfigure}

\vspace{0.3em}

\begin{subfigure}[t]{0.33\textwidth}
\centering
\includegraphics[width=\textwidth]{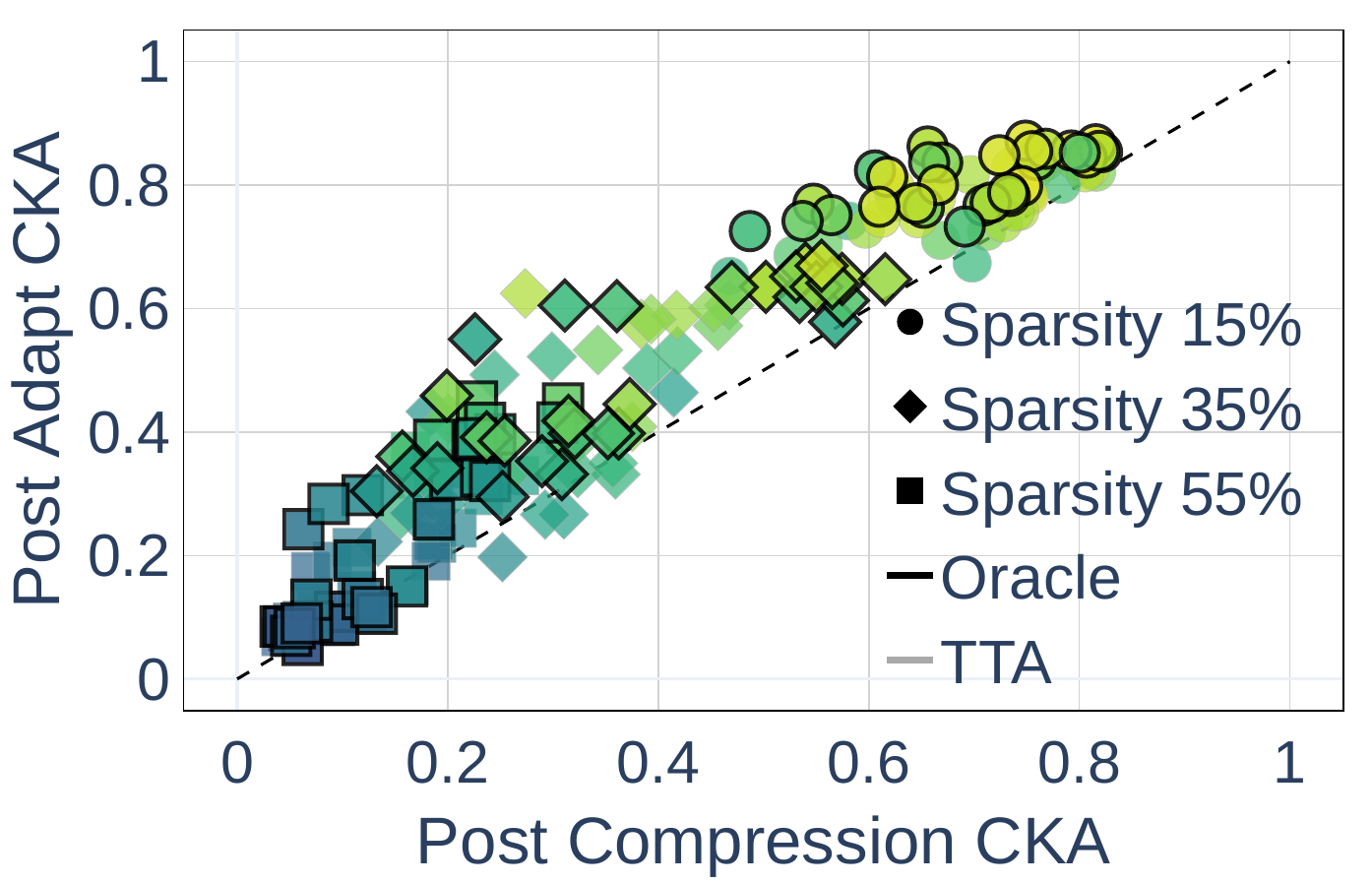}
\caption{ResNet-18 on CIFAR-10-C}
\label{fig:cka_free_cifar10}
\end{subfigure}
\hfill
\begin{subfigure}[t]{0.33\textwidth}
\centering
\includegraphics[width=\textwidth]{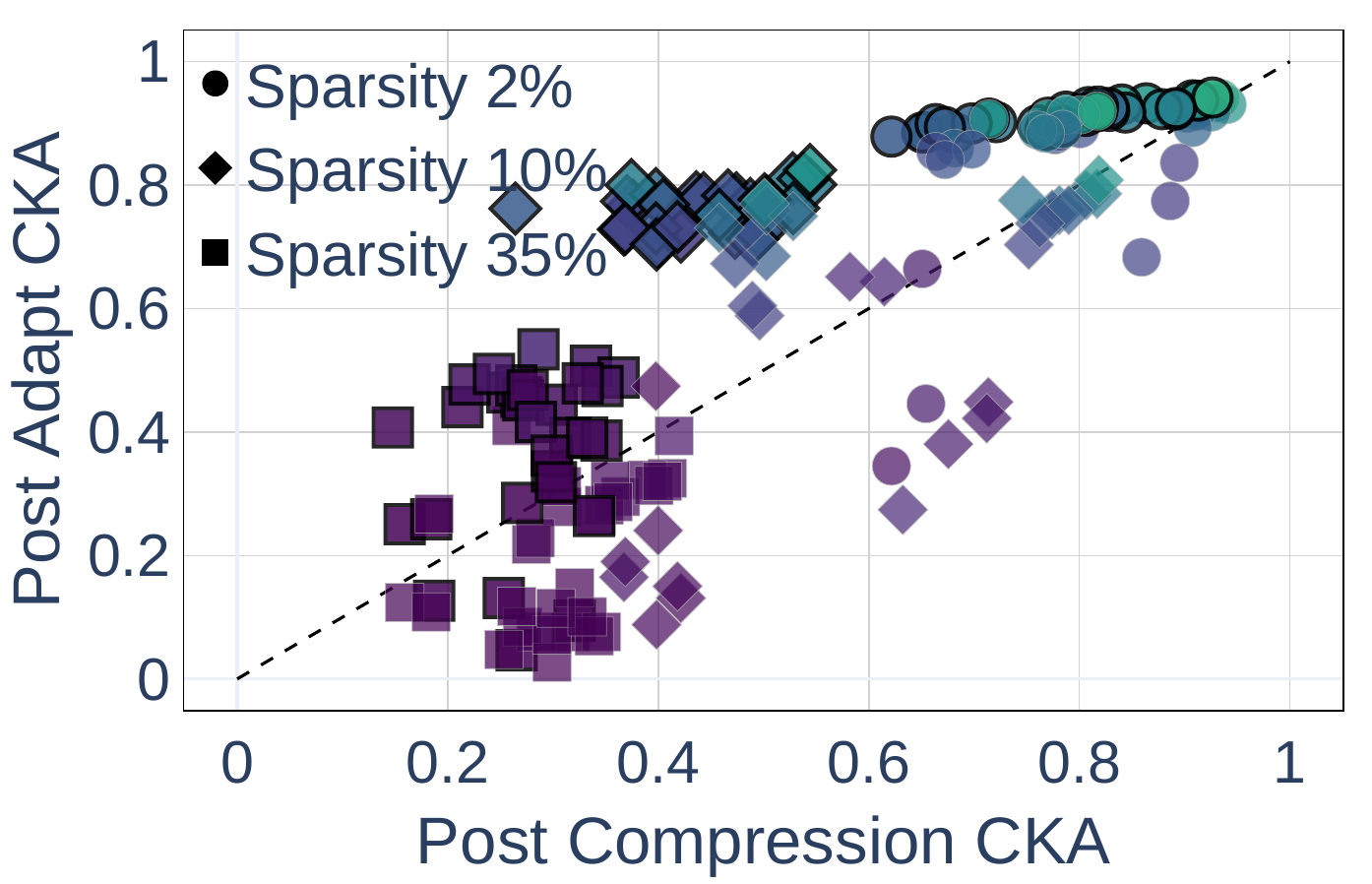}
\caption{ResNet-18 on ImageNet-C}
\label{fig:cka_free_imagenet}
\end{subfigure}
\hfill
\begin{subfigure}[t]{0.33\textwidth}
\centering
\includegraphics[width=\textwidth]{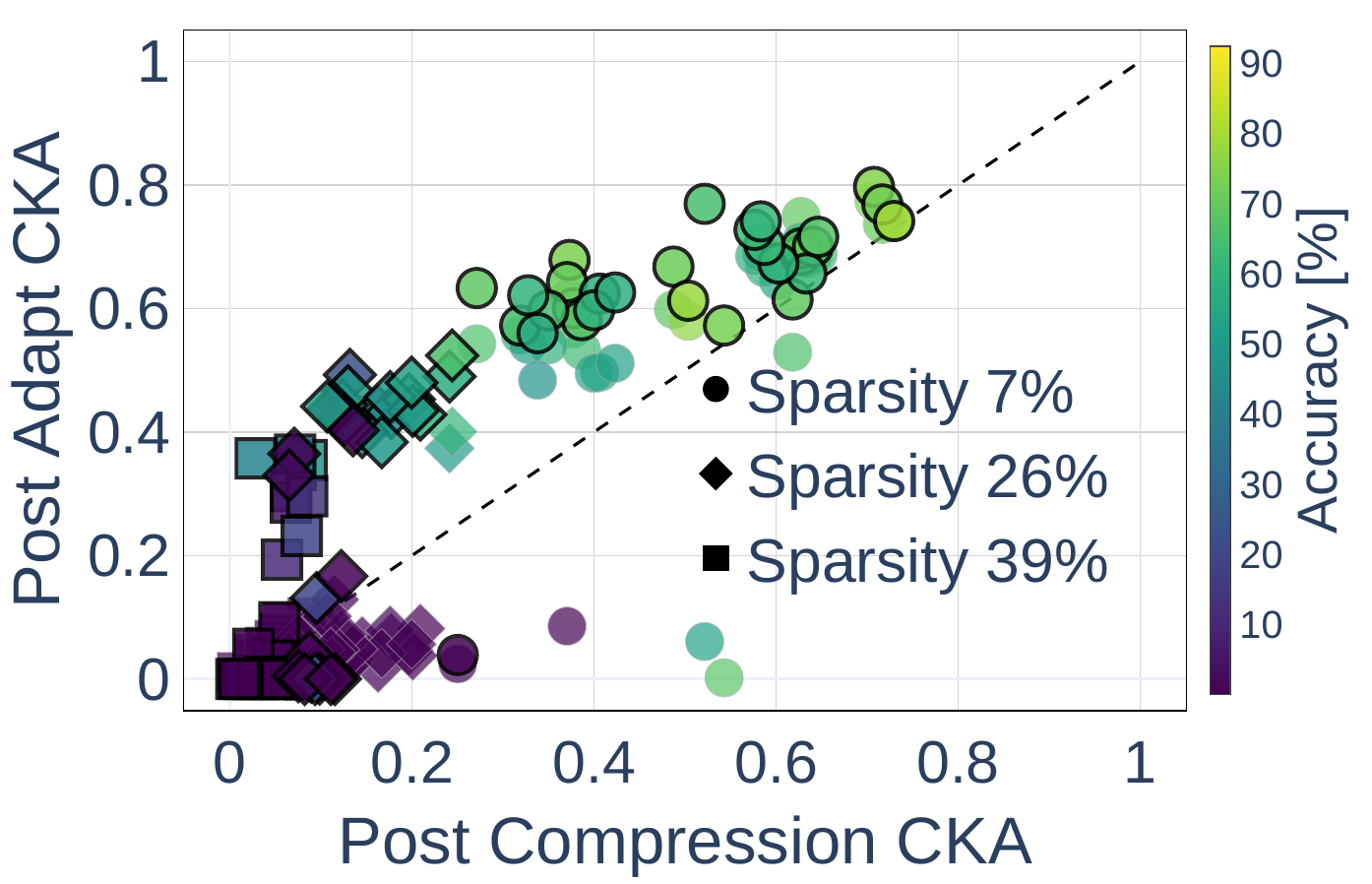}
\caption{ViT-Base on ImageNet-C}
\label{fig:cka_free_vit}
\end{subfigure}

\caption{\textbf{Test-time adaptation (TTA) degrades representations more than supervised fine-tuning (Oracle), with the gap depending on the compression method.}}
\label{fig:cka_main_appendix}
\end{figure*}

\begin{figure*}[h]
\centering
\begin{subfigure}[t]{0.33\textwidth}
\centering
\includegraphics[width=\textwidth]{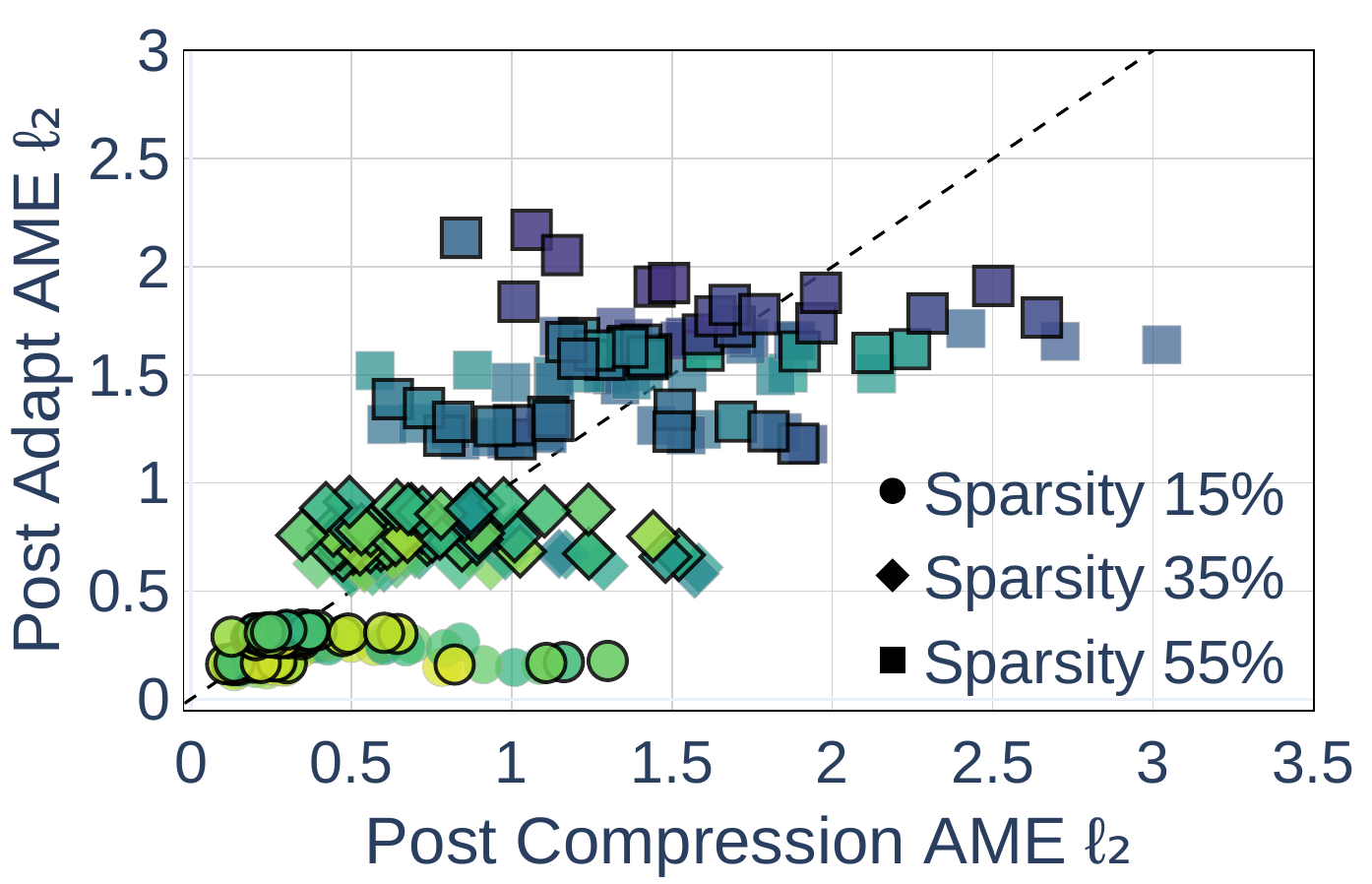}
\caption{ResNet-18 on CIFAR-10-C}
\label{fig:ame_data_cifar10}
\end{subfigure}
\hfill
\begin{subfigure}[t]{0.33\textwidth}
\centering
\includegraphics[width=\textwidth]{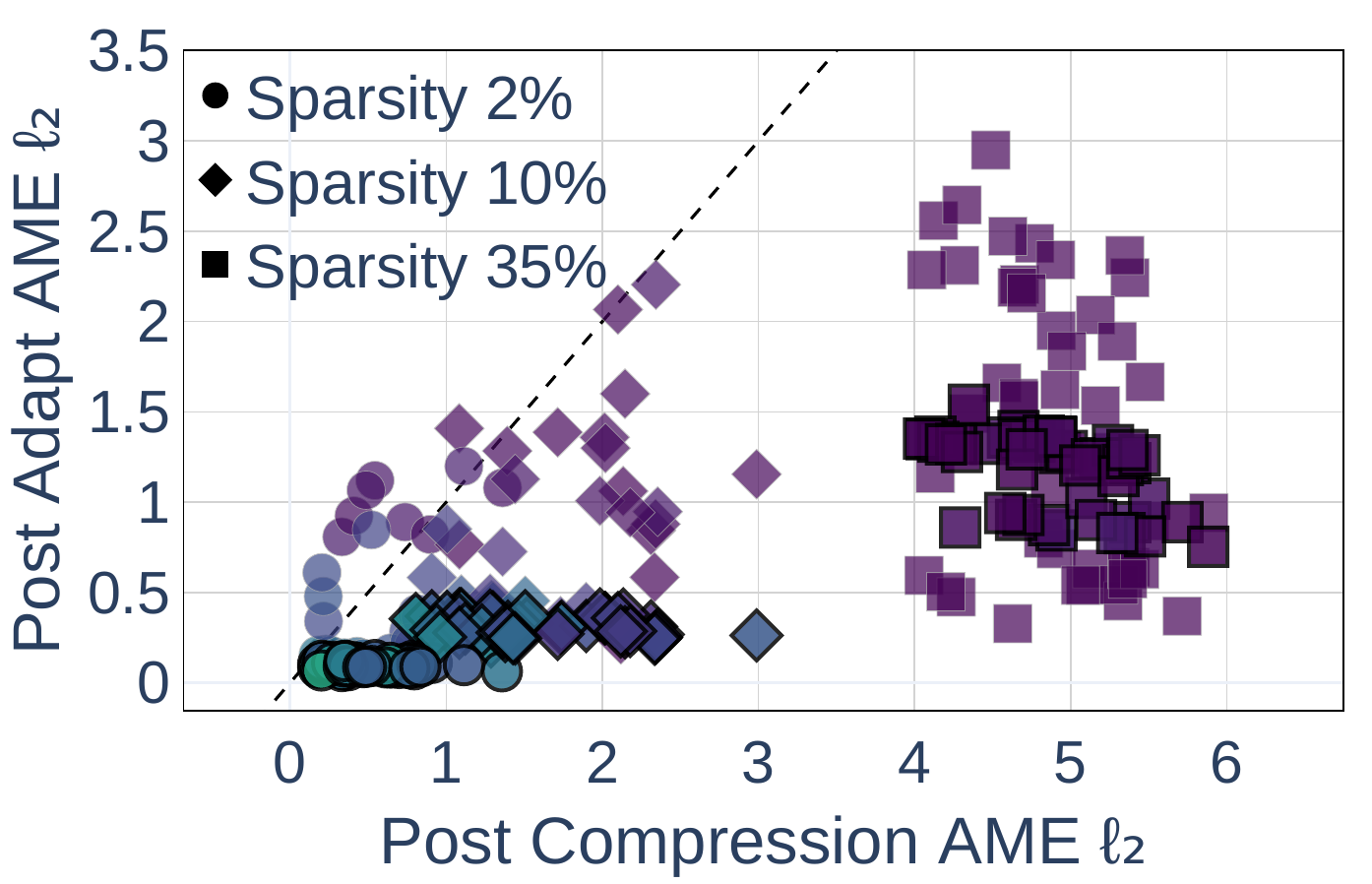}
\caption{ResNet-18 on ImageNet-C}
\label{fig:ame_data_imagenet}
\end{subfigure}
\hfill
\begin{subfigure}[t]{0.33\textwidth}
\centering
\includegraphics[width=\textwidth]{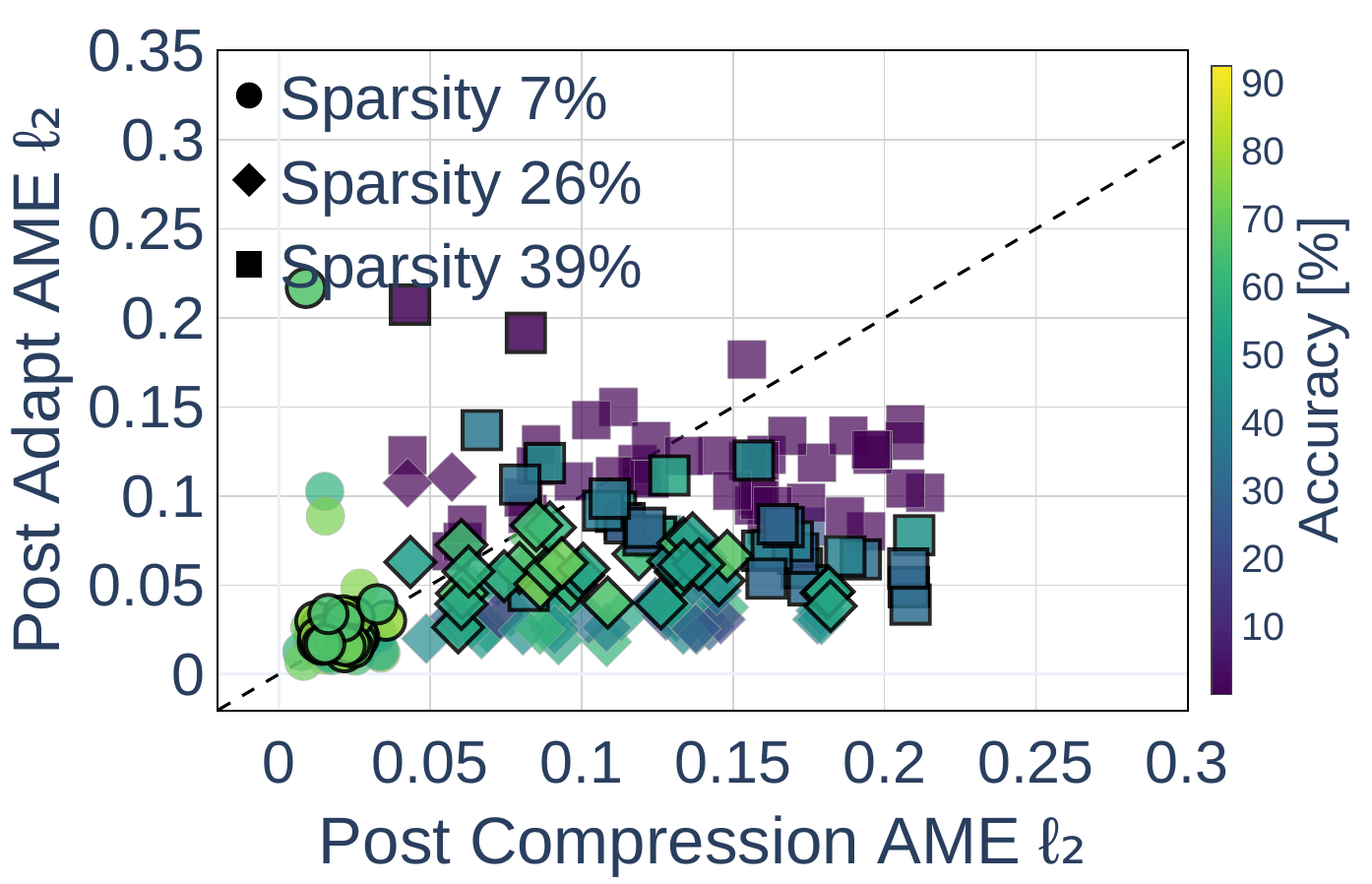}
\caption{ViT-Base on ImageNet-C}
\label{fig:ame_data_vit}
\end{subfigure}

\vspace{0.3em}

\begin{subfigure}[t]{0.33\textwidth}
\centering
\includegraphics[width=\textwidth]{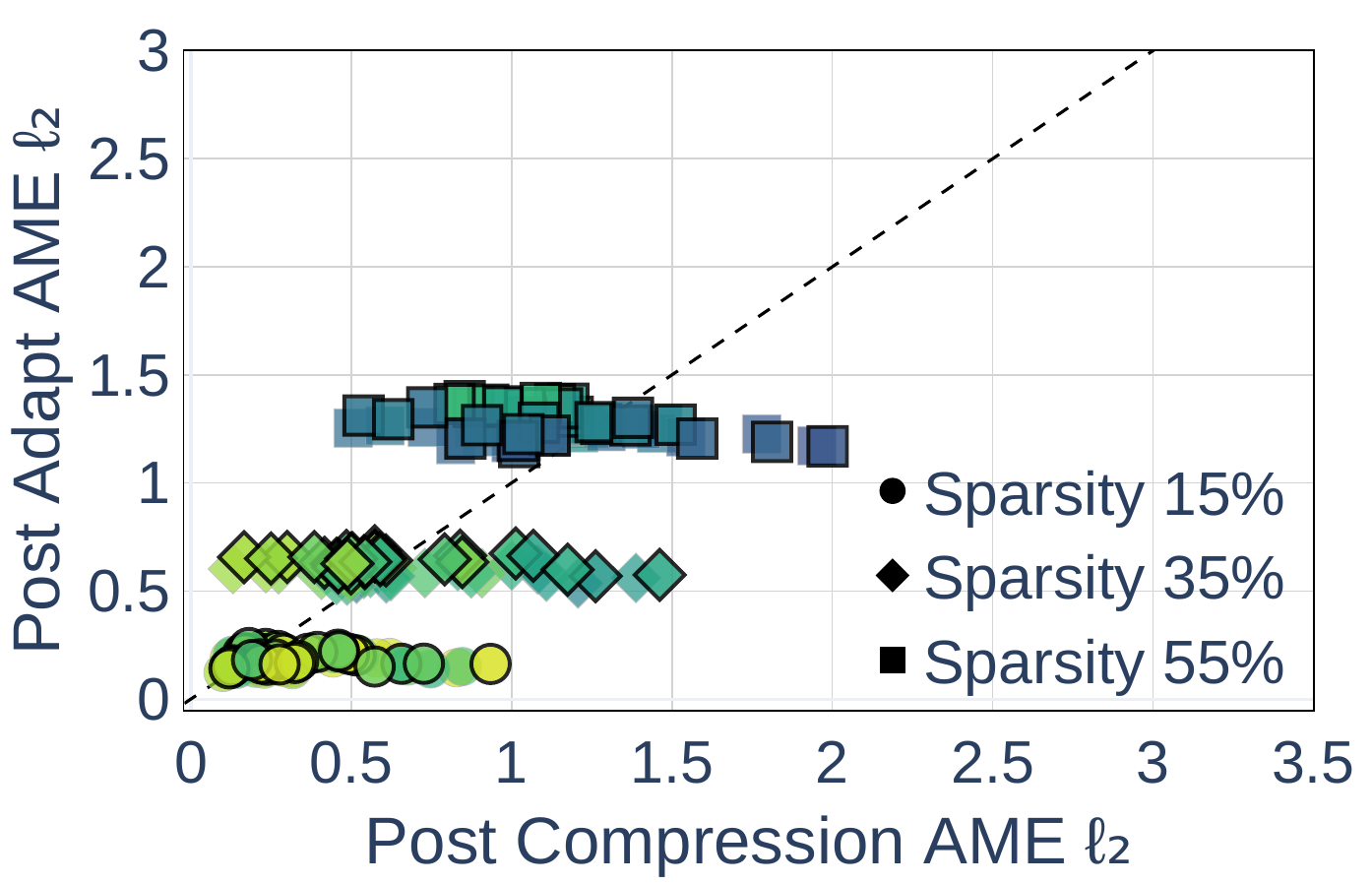}
\caption{ResNet-18 on CIFAR-10-C}
\label{fig:ame_free_cifar10}
\end{subfigure}
\hfill
\begin{subfigure}[t]{0.33\textwidth}
\centering
\includegraphics[width=\textwidth]{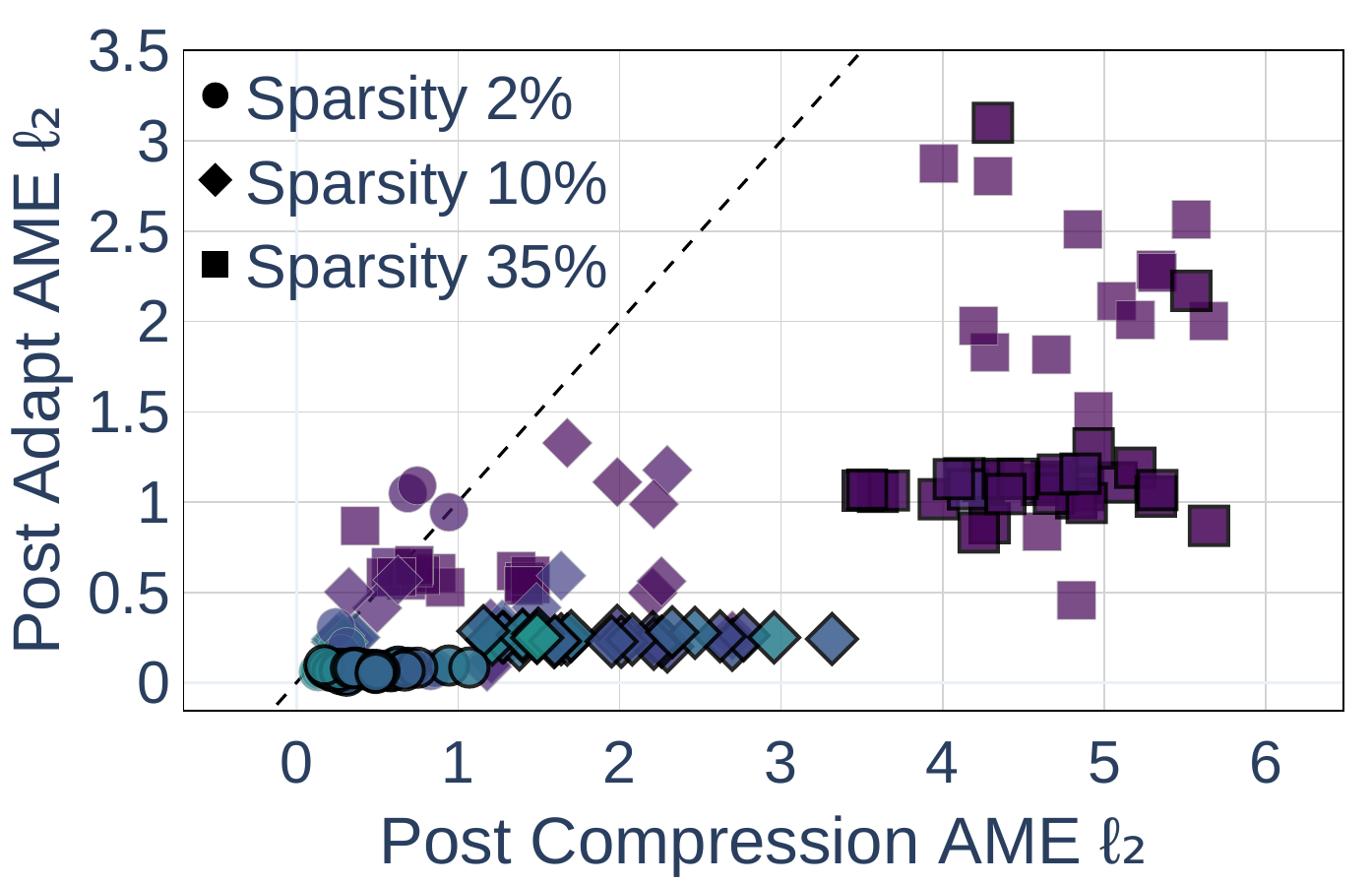}
\caption{ResNet-18 on ImageNet-C}
\label{fig:ame_free_imagenet}
\end{subfigure}
\hfill
\begin{subfigure}[t]{0.33\textwidth}
\centering
\includegraphics[width=\textwidth]{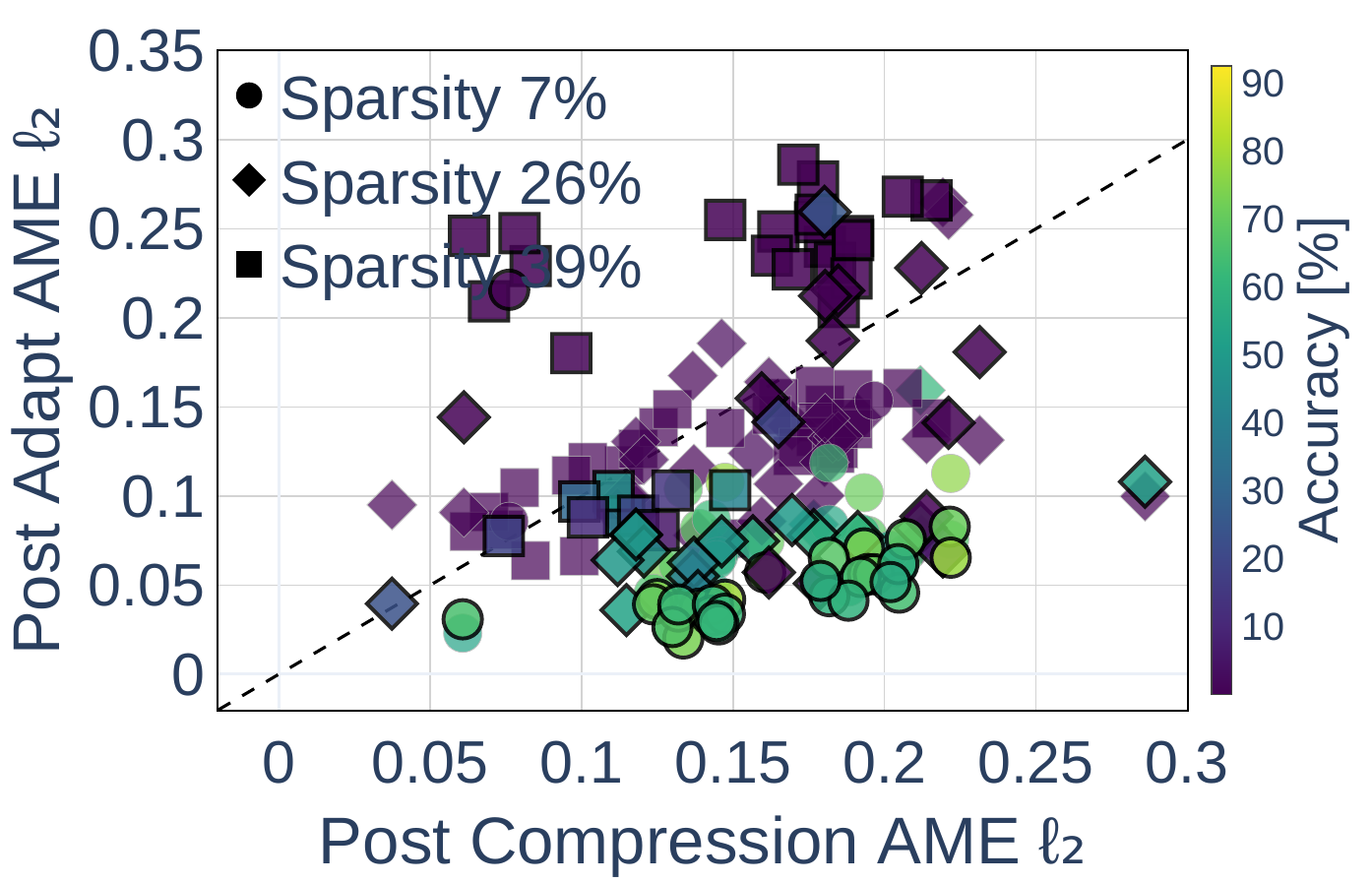}
\caption{ViT-Base on ImageNet-C}
\label{fig:ame_free_vit}
\end{subfigure}
\caption{\textbf{Test-time adaptation (TTA) only partially restores output expressivity, with recovery strongly dependent on the compression method.}}
\label{fig:ame_main_methods_colors}
\end{figure*}

Fig.~\ref{fig:gap_scatter} decomposes the averaged Oracle--\tta adaptation gap observed in Fig.~\ref{fig:hero} into individual corruptions. In Fig.~\ref{fig:cka_main_appendix} and Fig.~\ref{fig:ame_main_methods_colors}, marker shape encodes sparsity, color encodes post-adaptation accuracy; thick black edge denotes Oracle, gray edge denotes \tta. Points above the identity line indicate representational recovery after adaptation in Fig.~\ref{fig:cka_main_appendix}, in Fig.~\ref{fig:ame_main_methods_colors} points \emph{below} the identity line indicate entropy-distance adaptation recovery. Across all figures, \emph{Top row} (a--c): data-dependent pruning criteria (Wanda, Taylor, OBD). \emph{Bottom row} (d--f): data-free compression criteria (Folding, Mag-$\ell_2$).

\textbf{Multi-seed evaluation.}
We train three checkpoints per (architecture, dataset, pre-training paradigm) with different seeds.  Fig.~\ref{fig:cka_dense_prune_tta_sev3} reports the multi-seed analysis of \cka for severity 3 corruptions, supporting our previous severity 5 findings. Shaded bands of Fig.~\ref{fig:hero_multiseed} are the across-seed standard deviation of the per-seed corruption-averaged accuracy. These checkpoints follow a different pre-training recipe than the single checkpoint of Fig.~\ref{fig:hero}, shifting the \emph{absolute} post-adaptation accuracy of Fig.~\ref{fig:hero_multiseed} relative to Fig.~\ref{fig:hero}. Nevertheless, the compression-induced Oracle-\tta gap and its widening trend persist under both recipes, indicating that the effect reflects structured compression rather than a particular pre-training run.

\begin{figure*}[h]
\centering
\begin{subfigure}[t]{0.32\textwidth}\centering\includegraphics[width=\textwidth]{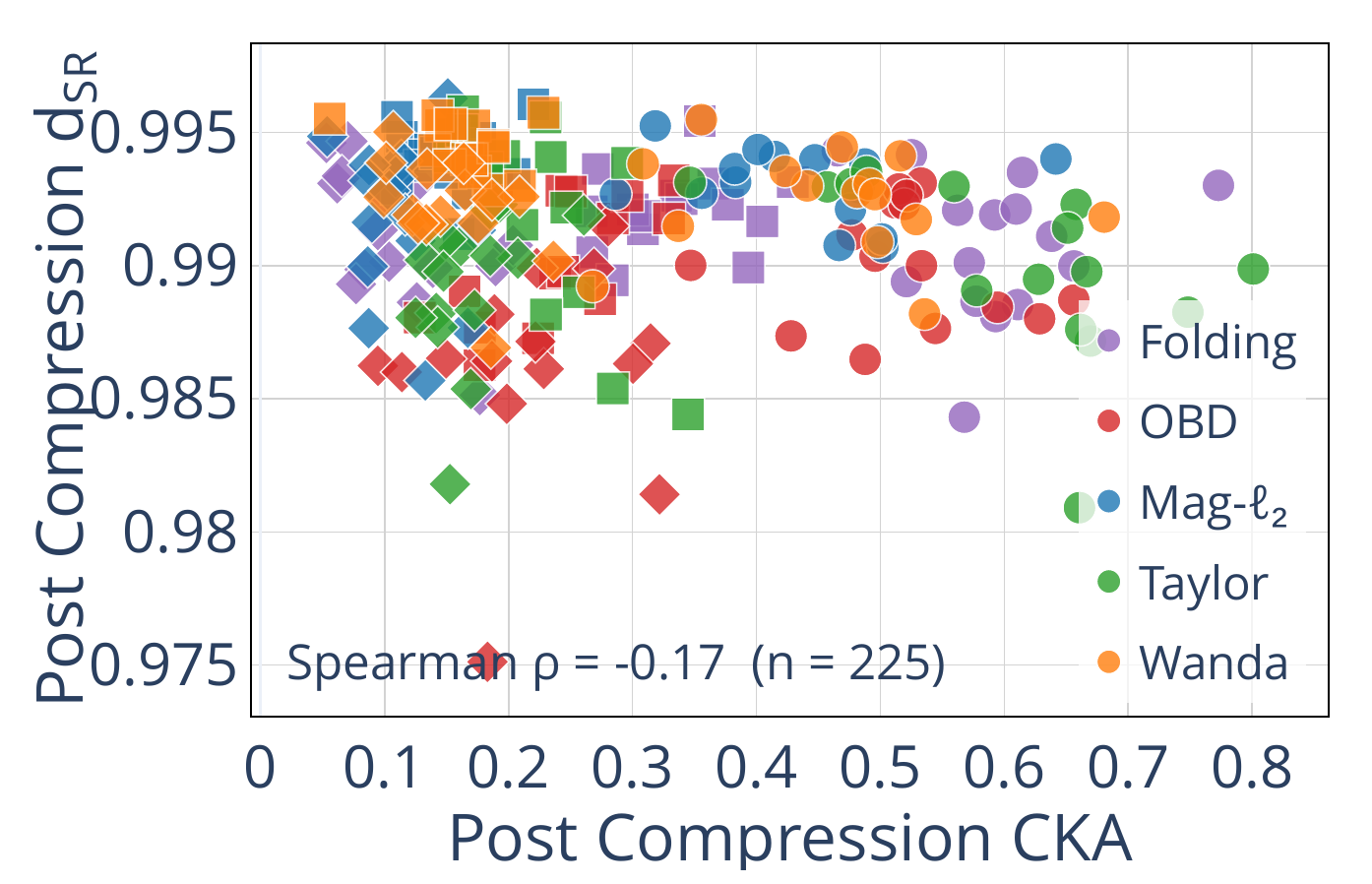}\caption{ResNet-18 / CIFAR-10-C.}\label{fig:msa_cifar10}\end{subfigure}
\hfill
\begin{subfigure}[t]{0.32\textwidth}\centering\includegraphics[width=\textwidth]{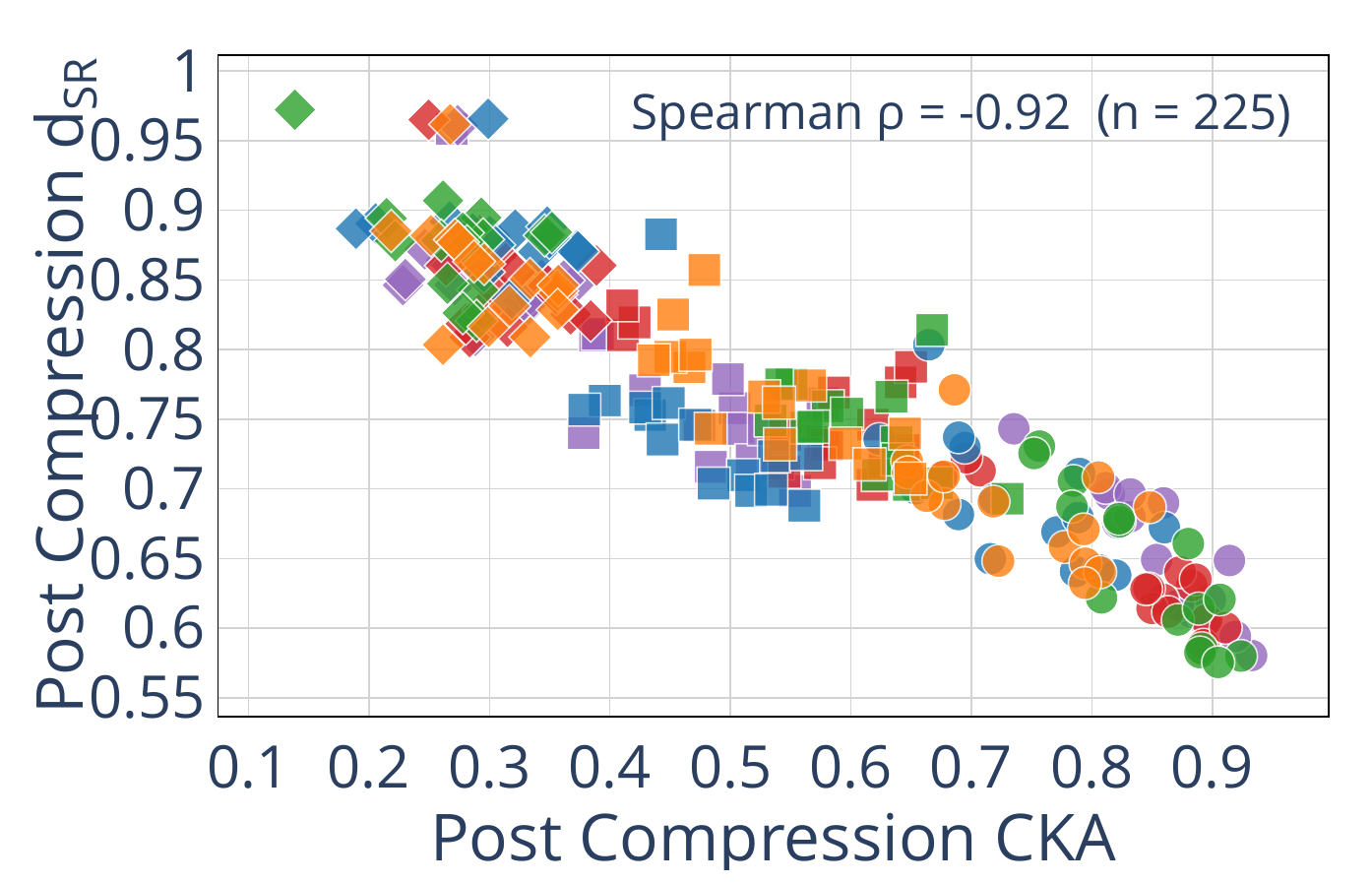}\caption{ResNet-18 / ImageNet-C.}\label{fig:msa_imagenet}\end{subfigure}
\hfill
\begin{subfigure}[t]{0.32\textwidth}\centering\includegraphics[width=\textwidth]{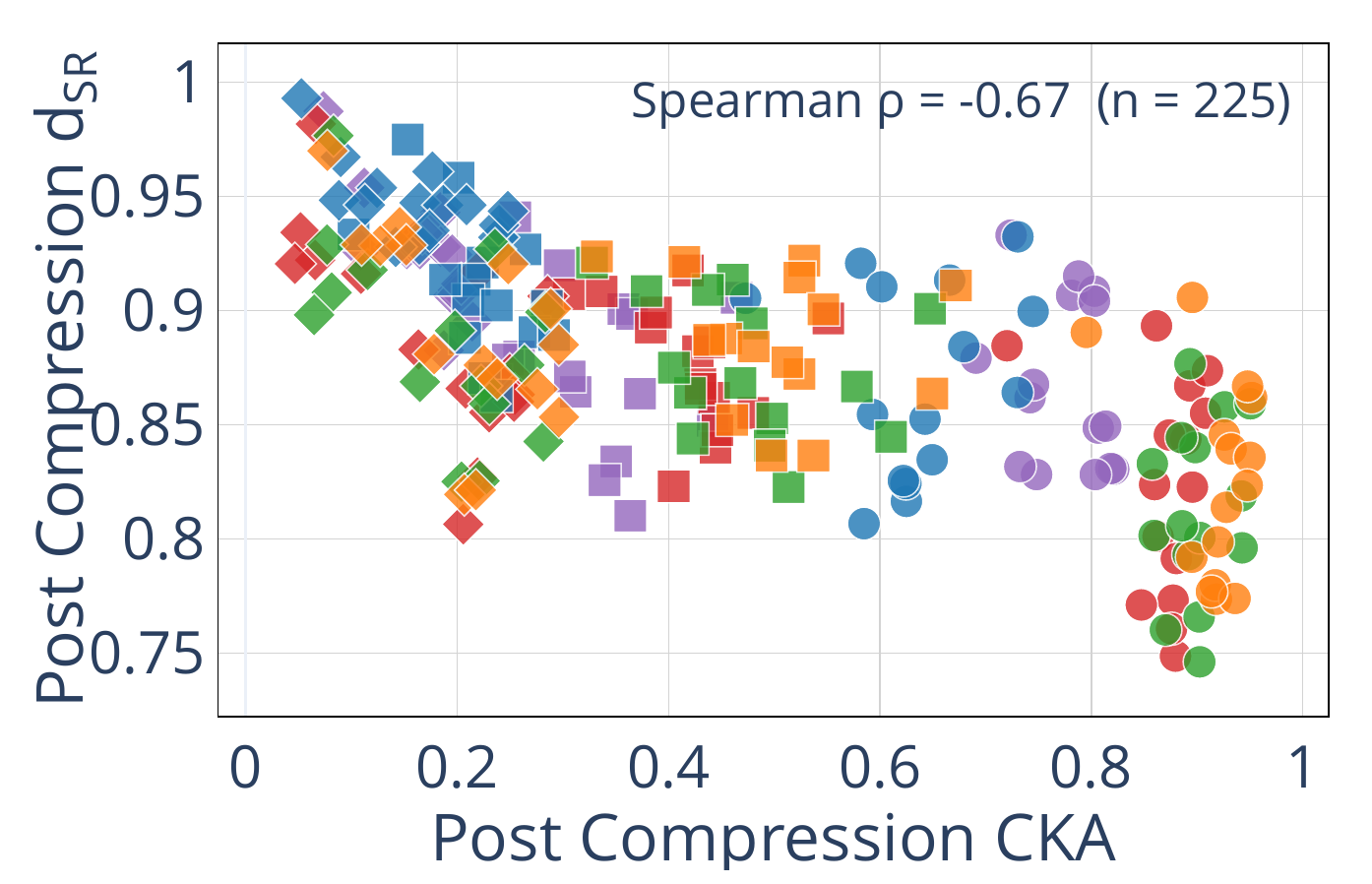}\caption{ViT-Base / ImageNet-C.}\label{fig:msa_vit}\end{subfigure}

\caption{\textbf{Geometry-aware similarity agrees with worst-layer \cka representation metric on ImageNet-C.} Each point is one (criterion, sparsity, corruption) cell at severity 5 post-compression pre-adaptation ($n$ equals to the number of total samples considered per panel; colors denote the compression criterion). The $x$-axis is the worst-layer \cka between the dense and the compressed model, the $y$-axis is the spectral-ratio pseudo-distance $d_{\mathrm{SR}}$~\citep{caycogajic2026geometry} computed on the same batches; since $d_{\mathrm{SR}}$ is a distance, agreement corresponds to negative rank correlation.}
\label{fig:msa_original}
\end{figure*}

\begin{figure*}[h]
\centering
\begin{subfigure}[t]{0.32\textwidth}\centering\includegraphics[width=\textwidth]{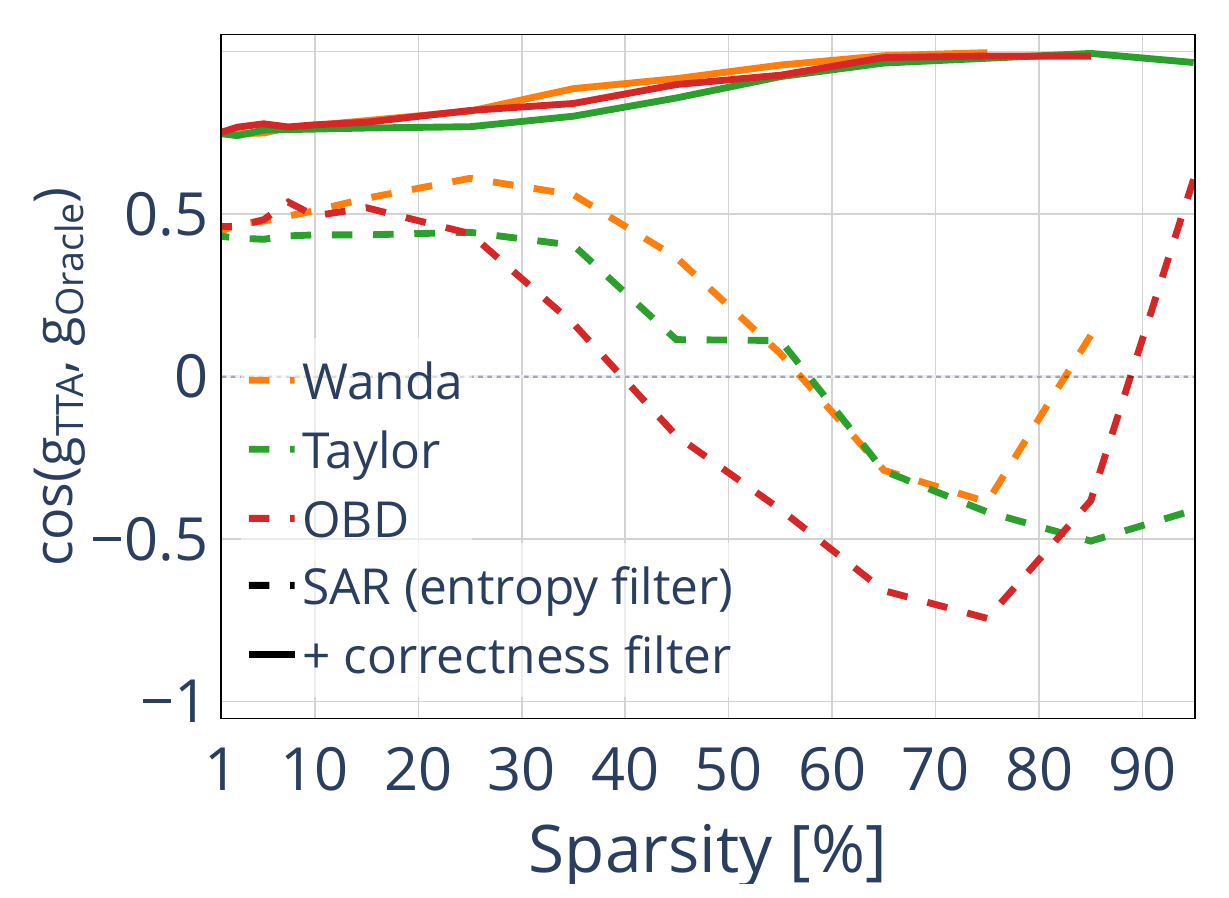}\caption{ResNet-18 / CIFAR-10-C.}\label{fig:abl_data_cifar10}\end{subfigure}
\hspace{0.04\textwidth}
\begin{subfigure}[t]{0.32\textwidth}\centering\includegraphics[width=\textwidth]{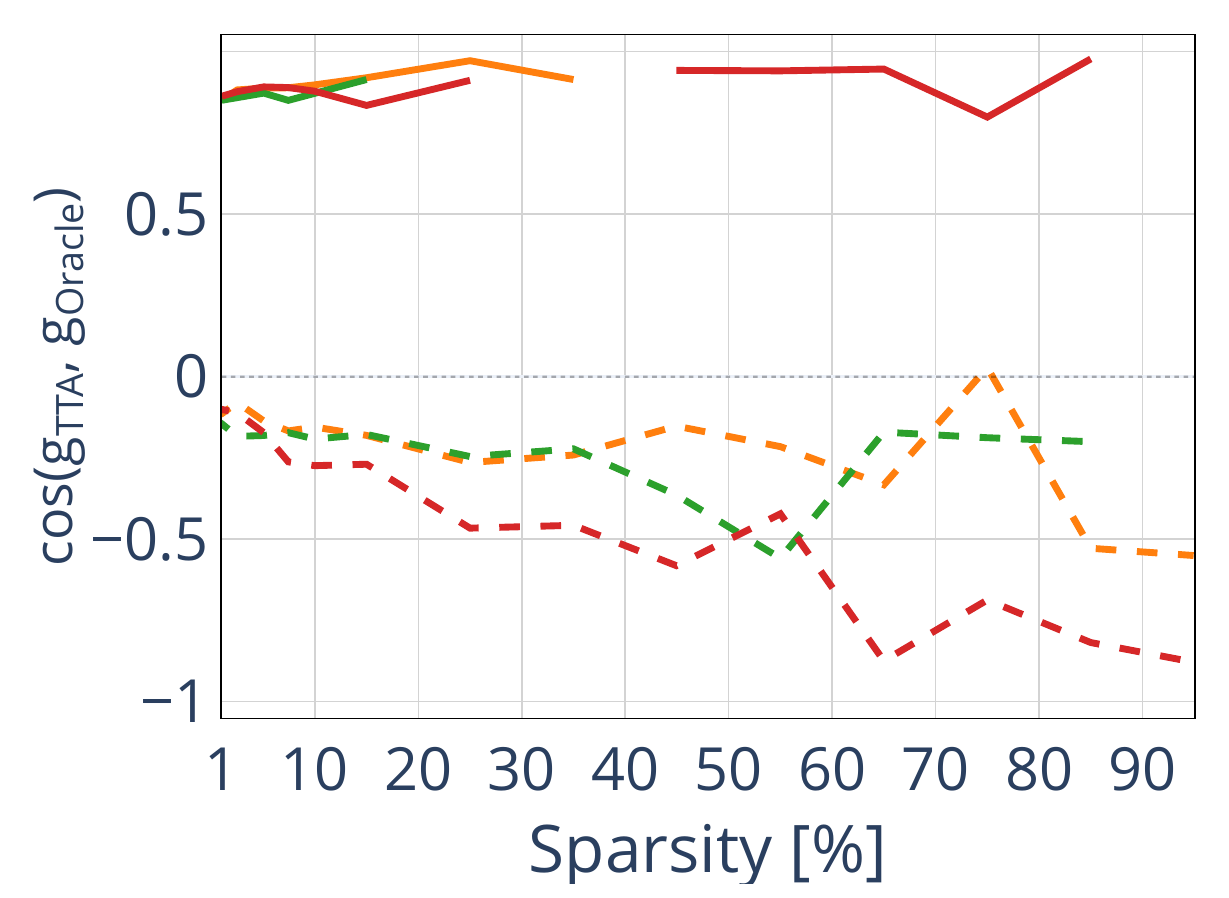}\caption{ResNet-18 / ImageNet-C.}\label{fig:abl_data_imagenet}\end{subfigure}
\vspace{0.3em}

\begin{subfigure}[t]{0.32\textwidth}\centering\includegraphics[width=\textwidth]{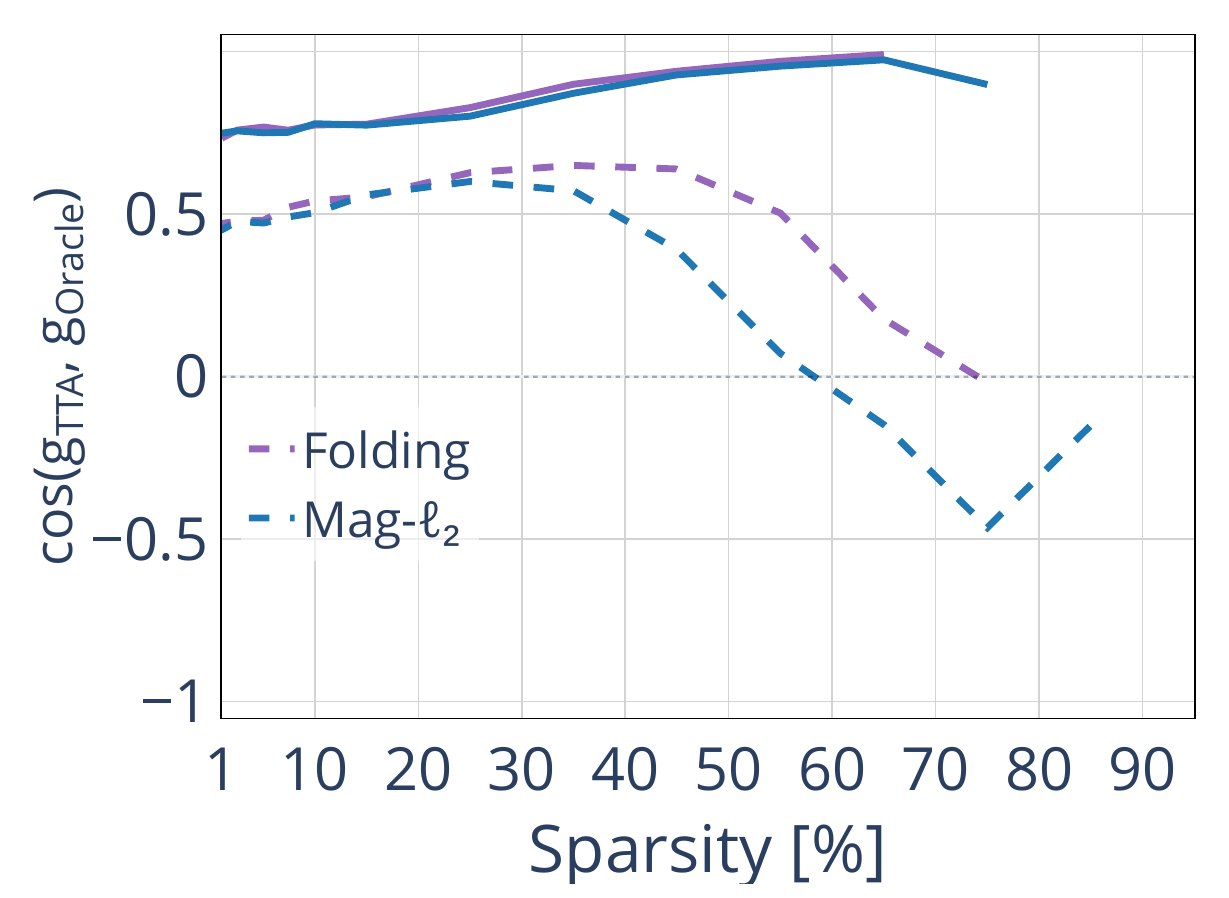}\caption{ResNet-18 / CIFAR-10-C.}\label{fig:abl_free_cifar10}\end{subfigure}
\hspace{0.04\textwidth}
\begin{subfigure}[t]{0.32\textwidth}\centering\includegraphics[width=\textwidth]{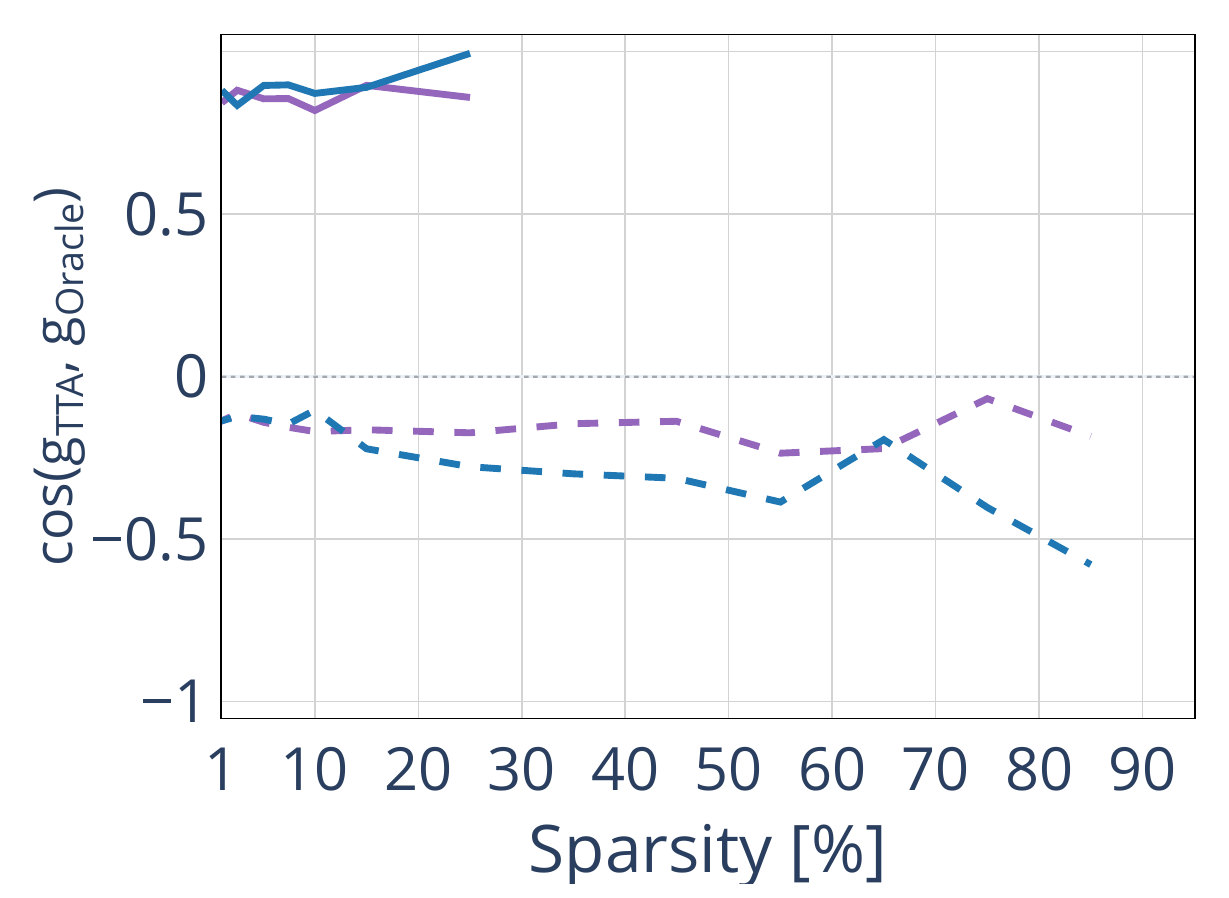}\caption{ResNet-18 / ImageNet-C.}\label{fig:abl_free_imagenet}\end{subfigure}

\caption{\textbf{Removing the confident-but-wrong samples that pass SAR's reliability filters recovers gradient alignment.} We report $\cos(\mathbf{g}_{\text{\tta}},\mathbf{g}_{\text{Oracle}})$ at $\phi_0$ (Eq.~\ref{eq:grad_cosine}), averaged across all corruptions, with colors as in Fig.~\ref{fig:cosine_similarity}. Dashed lines reproduce the cosine-similarity curves of Fig.~\ref{fig:cosine_similarity} unchanged; solid lines restrict the reliable set to correctly classified samples and evaluate the Oracle-SAR  gradient over the same set. Excluding the confident-wrong samples restores the alignment to $\approx{+}0.9$ in every considered scenario.}
\label{fig:ablation_recovery}
\end{figure*}

\begin{figure*}[t]
\centering
\begin{subfigure}[t]{0.32\textwidth}
\centering
\includegraphics[width=\textwidth]{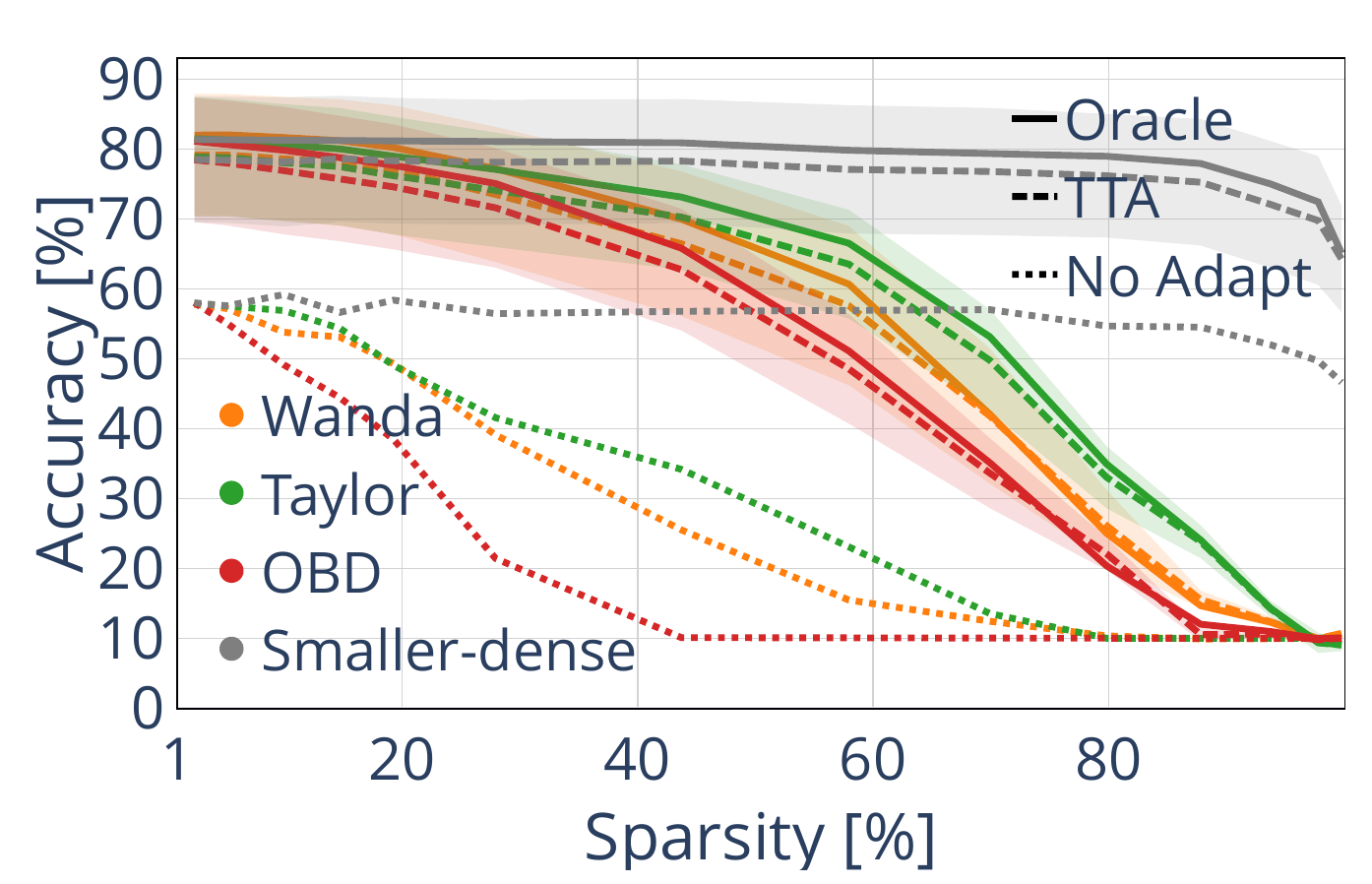}
\caption{ResNet-18 / CIFAR-10-C}
\label{fig:scale_cifar10}
\end{subfigure}
\hfill
\begin{subfigure}[t]{0.32\textwidth}
\centering
\includegraphics[width=\textwidth]{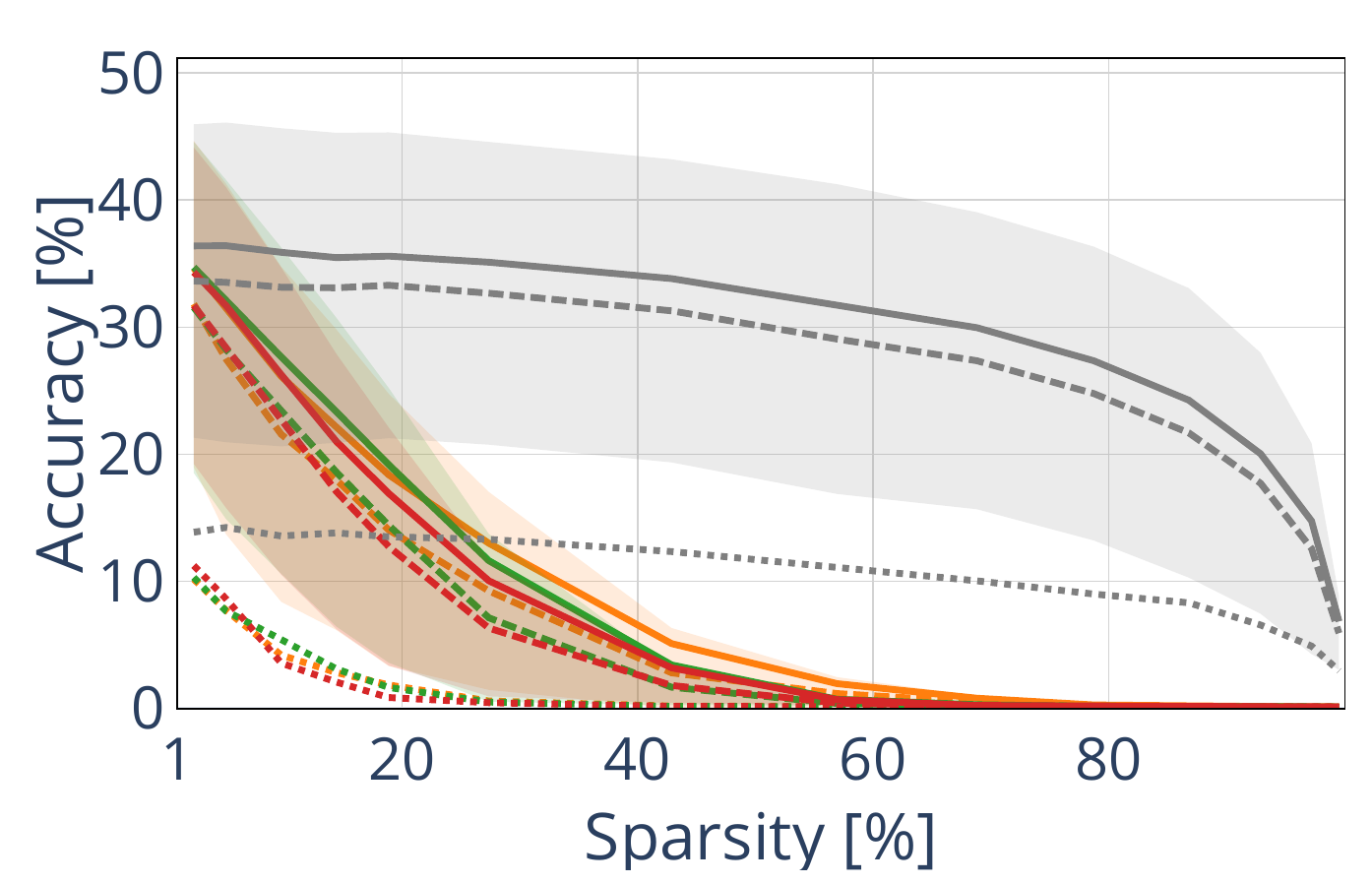}
\caption{ResNet-18 / ImageNet-C}
\label{fig:scale_imagenet}
\end{subfigure}
\hfill
\begin{subfigure}[t]{0.32\textwidth}
\centering
\includegraphics[width=\textwidth]{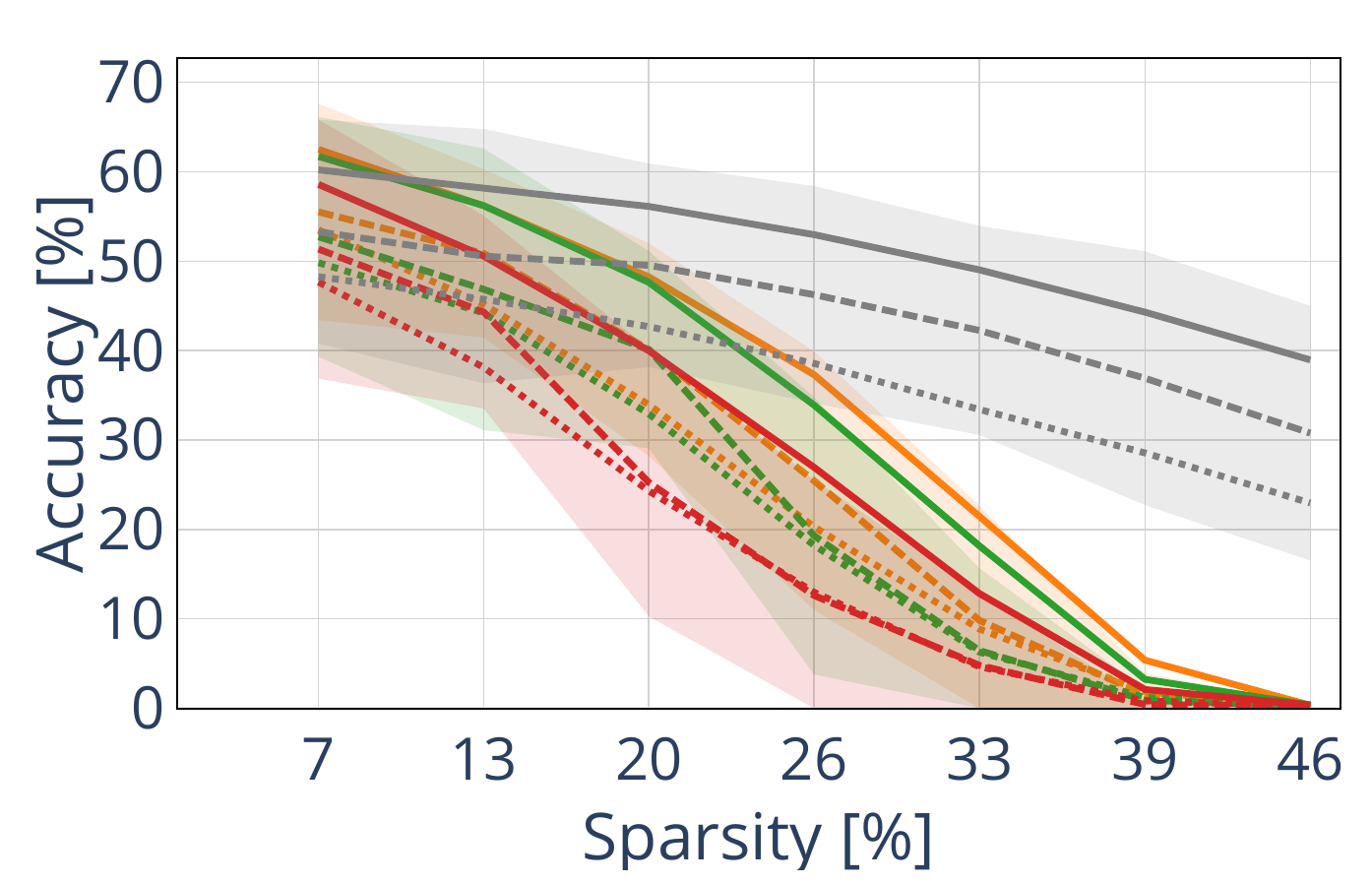}
\caption{ViT-Base / ImageNet-C}
\label{fig:scale_vit_single}
\end{subfigure}

\vspace{0.6em}
\begin{subfigure}[t]{0.32\textwidth}
\centering
\includegraphics[width=\textwidth]{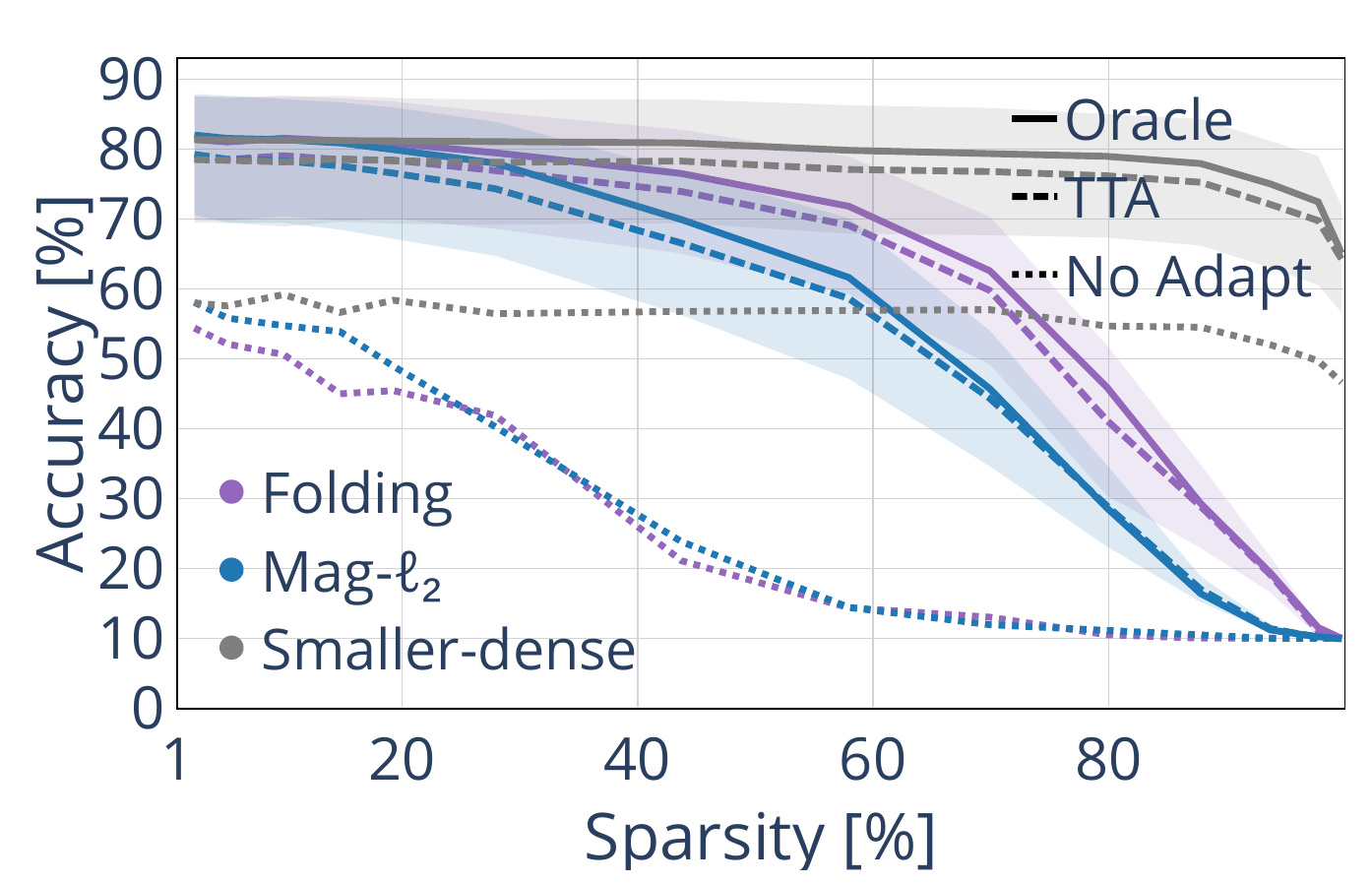}
\caption{ResNet-18 / CIFAR-10-C}
\label{fig:scale_df_cifar10}
\end{subfigure}
\hfill
\begin{subfigure}[t]{0.32\textwidth}
\centering
\includegraphics[width=\textwidth]{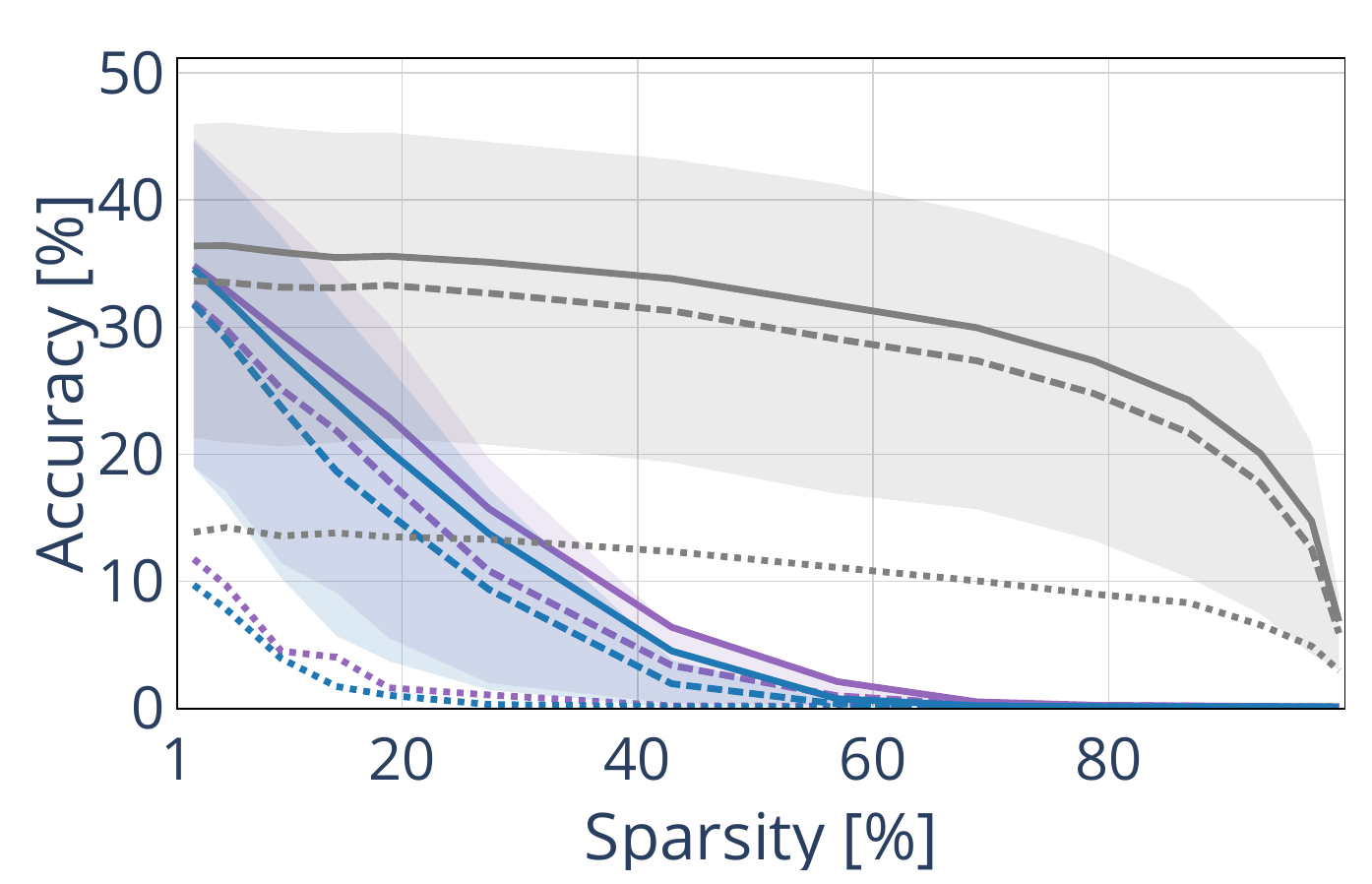}
\caption{ResNet-18 / ImageNet-C}
\label{fig:scale_df_imagenet}
\end{subfigure}
\hfill
\begin{subfigure}[t]{0.32\textwidth}
\centering
\includegraphics[width=\textwidth]{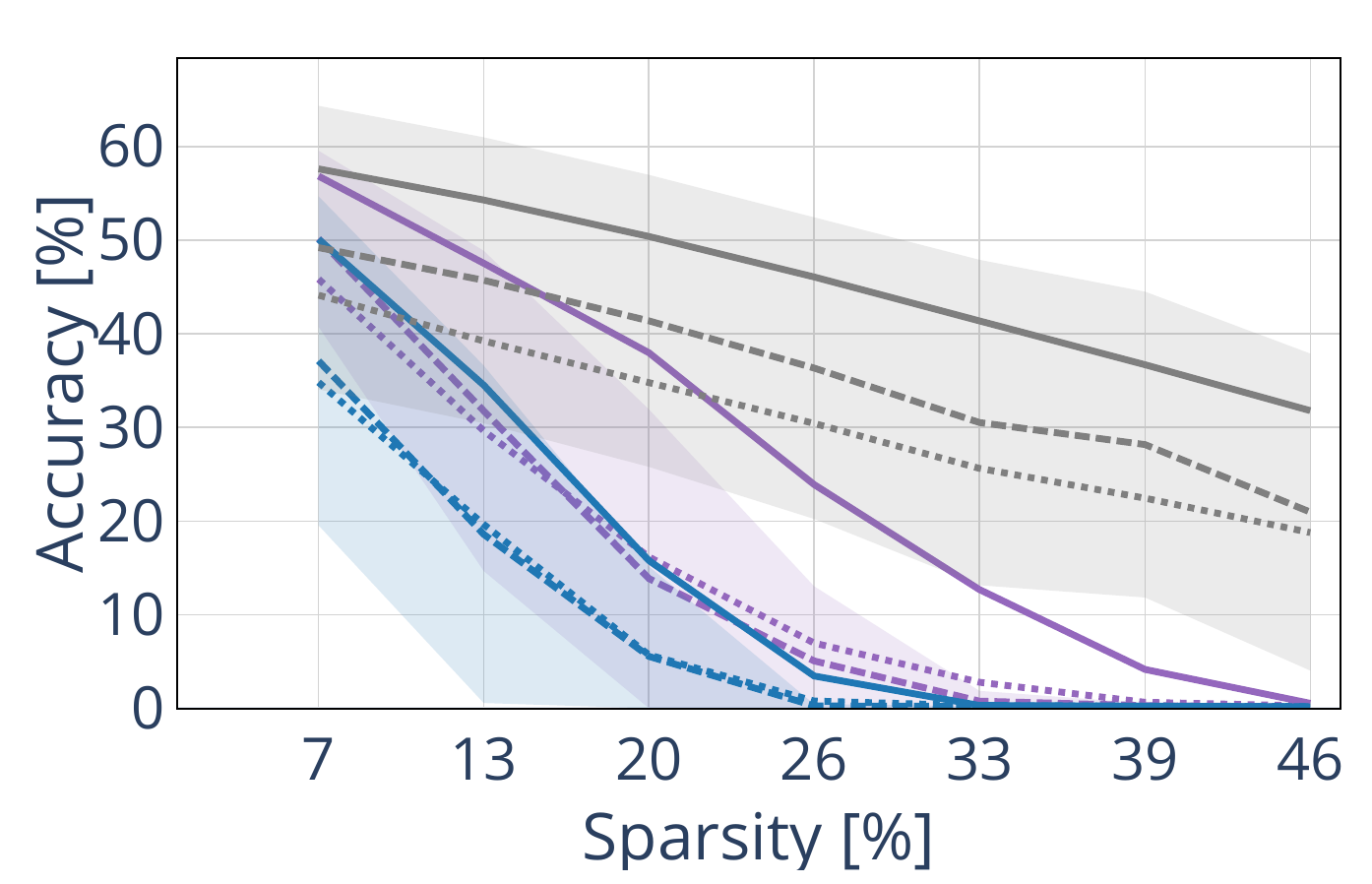}
\caption{ViT-Base / ImageNet-C}
\label{fig:scale_df_vit_single}
\end{subfigure}

\caption{\textbf{Test-time adaptation collapses with increasing compression for the pruned models, unlike the smaller-dense models of equal parameter budget trained from scratch.} Same layout as Fig.~\ref{fig:hero} using SAR and Oracle-SAR on all the displayed architectures. The gray curves show the method-agnostic \emph{smaller-dense} \tta reference. At low sparsity, \tta on the compressed models remains competitive with the references; as compression increases, \tta and its Oracle both collapse on every criterion, while the smaller-dense models retain adaptation gain. Together, these observations suggest that the loss of adaptability is induced by the structured compression rather than by the reduced model size. }
\label{fig:scale_study}
\end{figure*}

\end{document}